\documentclass[aap]{imsart}
\usepackage[T1]{fontenc}
\usepackage{textcomp}
\usepackage{newunicodechar}
\newunicodechar{§}{\S}
\newunicodechar{·}{\textperiodcentered{}}
\newunicodechar{½}{\ensuremath{\tfrac12}}
\newunicodechar{–}{--}
\newunicodechar{…}{\dots{}}
\newunicodechar{é}{\'e}
\newunicodechar{á}{\'a}
\newunicodechar{à}{\`a}
\newunicodechar{è}{\`e}
\newunicodechar{â}{\^a}
\newunicodechar{ö}{\"o}
\newunicodechar{ä}{\"a}
\newunicodechar{ó}{\'o}
\newunicodechar{ú}{\'u}
\newunicodechar{É}{\'E}
\newunicodechar{ø}{\o{}}
\newunicodechar{ā}{\=a}
\newunicodechar{ć}{\'c}
\newunicodechar{č}{\v c}
\newunicodechar{ł}{\l{}}
\newunicodechar{ő}{\H o}
\newunicodechar{′}{\ensuremath{{}^{\prime}}}
\newunicodechar{″}{\ensuremath{{}^{\prime\prime}}}
\newunicodechar{⁺}{\ensuremath{{}^{+}}}
\newunicodechar{⁻}{\ensuremath{{}^{-}}}
\newunicodechar{ℓ}{\ensuremath{\ell}}
\newunicodechar{∩}{\ensuremath{\cap}}
\newunicodechar{→}{\ensuremath{\to}}
\newunicodechar{←}{\ensuremath{\leftarrow}}
\newunicodechar{↔}{\ensuremath{\leftrightarrow}}
\newunicodechar{⇒}{\ensuremath{\Rightarrow}}
\newunicodechar{↦}{\ensuremath{\mapsto}}
\newunicodechar{⌈}{\ensuremath{\lceil}}
\newunicodechar{⌉}{\ensuremath{\rceil}}
\newunicodechar{⌊}{\ensuremath{\lfloor}}
\newunicodechar{⌋}{\ensuremath{\rfloor}}
\newunicodechar{−}{\ensuremath{-}}
\newunicodechar{≥}{\ensuremath{\ge}}
\newunicodechar{≤}{\ensuremath{\le}}
\newunicodechar{≈}{\ensuremath{\approx}}
\newunicodechar{≠}{\ensuremath{\neq}}
\newunicodechar{×}{\ensuremath{\times}}
\newunicodechar{±}{\ensuremath{\pm}}
\newunicodechar{∖}{\ensuremath{\setminus}}
\newunicodechar{∈}{\ensuremath{\in}}
\newunicodechar{∪}{\ensuremath{\cup}}
\newunicodechar{∅}{\ensuremath{\emptyset}}
\newunicodechar{⋯}{\ensuremath{\cdots}}
\newunicodechar{Σ}{\ensuremath{\Sigma}}
\newunicodechar{σ}{\ensuremath{\sigma}}
\newunicodechar{₁}{\textsubscript{1}}
\newunicodechar{₂}{\textsubscript{2}}
\newunicodechar{₃}{\textsubscript{3}}
\newunicodechar{ₙ}{\textsubscript{n}}
\newunicodechar{✓}{\ensuremath{\checkmark}}
\newunicodechar{✗}{\ensuremath{\times}}
\newunicodechar{✅}{}
\newunicodechar{⬜}{}
\newunicodechar{🟨}{}
\newunicodechar{🟩}{}
\newunicodechar{🔒}{}
\newunicodechar{⚑}{}
\usepackage{amsmath,amssymb,amsthm,mathtools,bm}
\usepackage{graphicx,booktabs,longtable,array,enumitem,calc}
\usepackage[colorlinks,citecolor=blue,urlcolor=blue,linkcolor=blue]{hyperref}
\graphicspath{{figs/}}
\setkeys{Gin}{width=0.95\linewidth,height=0.75\textheight,keepaspectratio}

\usepackage{etoolbox}
\usepackage{adjustbox}
\usepackage[htt]{hyphenat}
\usepackage[htt]{hyphenat}
\AtBeginEnvironment{longtable}{\footnotesize\setlength{\tabcolsep}{4pt}}
\AtBeginEnvironment{tabular}{\small\setlength{\tabcolsep}{4pt}}
\AtBeginEnvironment{verbatim}{\footnotesize}
\makeatletter\def\journal@name{}\makeatother
\startlocaldefs

\newcommand{\rsaSpan}{s}
\newcommand{\rsaN}{N}
\newcommand{\rsan}{n}

\newcommand{\rsaX}[1]{X_{#1}}
\newcommand{\rsaXord}[1]{X_{(#1)}}

\newcommand{\rsaMc}[1]{M_{#1, c}^{\infty}}

\providecommand{\pandocbounded}[1]{\adjustbox{max width=\linewidth}{#1}}

\providecommand{\tightlist}{\setlength{\itemsep}{0pt}\setlength{\parskip}{0pt}}
\theoremstyle{plain}

\theoremstyle{definition}

\newcounter{thmcnt}[section]
\renewcommand{\thethmcnt}{\thesection.\arabic{thmcnt}}
\numberwithin{equation}{section}
\endlocaldefs
\begin{document}
\begin{frontmatter}
\title{Exact finite-length theory of Uniform Car Parking:
Spatial laws, Absorption, and Aggregates}
\runtitle{Finite-length theory of the UCP}
\begin{aug}
\author{\fnms{Ganesh P.} \snm{Kumar}}
\address{pkganeshpk [AT] gmail [DOT] com}
\end{aug}
\begin{abstract}
The uniform car-parking process is the one-dimensional random sequential adsorption of unit cars on a segment of finite length $s$: cars arrive at uniformly random positions and park wherever they fit, until no gap admits another. This paper develops the exact finite-$s$ theory. The joint density of the parked positions is resolved into jamming cells, on each of which it is a rational function, and evaluated by a subset recursion in $O(2^n n)$ operations; the marginal and gap order statistics are obtained as hyperlogarithms whose weight is fixed by the number of coordinates integrated out; and the absorption count and the aggregate quantities are treated through the integral equation descending from R\'enyi.
\end{abstract}
\end{frontmatter}

\begin{center}
\fbox{\parbox{0.88\linewidth}{\small\textbf{Note on AI use.}\enspace
Large language models (Claude, Anthropic) were used throughout this work:
for the calculations, for the proofs, and for writing the paper, under the
author's direction and review. The formalised statements are additionally
machine-checked in Lean~4.}}
\end{center}

\setcounter{tocdepth}{2}
\setcounter{secnumdepth}{3}
\tableofcontents

\section{Introduction}
\label{sec:introduction}

\subsubsection{Background}

A common multi-robot deployment problem is the placement of a team of identical robots along an unknown
one-dimensional boundary using only minimal local sensing: each robot, on arrival, commits to a position
that does not overlap any robot already in place, and the team's collective behaviour is determined by the
sequence of these local choices. Stochastic boundary coverage and related deployment strategies
(Kumar and Berman 2014; Pavlic, Wilson, Kumar and Berman 2015; Kumar and Berman 2017; Kumar 2016; Kumar and Berman 2016)
formalise this setting. The underlying mathematical object, sequential deposition of unit-length objects on
a bounded substrate under a hard-overlap constraint, appears in statistical mechanics as \emph{random
sequential adsorption} (RSA) (Evans 1993; Krapivsky, Redner and Ben-Naim 2010) and in applied probability as Rényi's \emph{car
parking problem} (R\textquotesingle enyi 1958; Dvoretzky and Robbins 1964).

The car-parking version of this model has been the most thoroughly studied case: deposited objects are
unit-length intervals on a bounded segment \([0, \rsaSpan]\), attachment is sequential and uniform on the
unoccupied region, and the only inter-object constraint is non-overlap. We adopt Rényi's terminology
of \emph{cars} for deposited objects and \emph{segment} for the substrate; throughout the paper we use \textbf{UCP},
reserving the broader term RSA for the citing context in statistical mechanics.

This paper revisits the uniform car-parking process (UCP) on a segment of finite length \(\rsaSpan\), in two
regimes: a fixed number of attached cars \(\rsan\) (joint order statistics, gap distributions), and the random
jamming count \(\rsaN\) (PMF on a fixed segment). The principal contributions are four, each exact at finite \(\rsaSpan\).

\begin{enumerate}
\def\labelenumi{\arabic{enumi}.}
\tightlist
\item
  \textbf{The joint law of the car positions.} On each region of its support the joint density of the order
  statistics \(\rsaXord{1}<\cdots<\rsaXord{\rsan}\) is a single rational function, the sum over arrival
  orders of the reciprocal free lengths available to the cars as they attach; the regions are indexed by the
  jamming bitstrings, and the free lengths are affine in the gaps with coefficients in \(\{-1,0,1\}\).
  The order-statistics section states this form, counts the regions in closed form, and grades the
  transcendental weight of every marginal obtained from it.
\item
  \textbf{An enumeration of the support, and a separation between evaluating the density and writing it down.}
  The regions carrying positive mass are generated by a recurrence over sub-intervals rather than by
  testing candidate sign patterns for feasibility. One recursion then serves two purposes according to the
  semiring its accumulator carries. Over the reals the free lengths are determined by the set of cars
  already placed, so grouping the temporal orders by their last-placed car evaluates the density at a point in
  \(O(2^{\rsan}\rsan)\) operations rather than \(\rsan!\). Whether that can be improved to a polynomial
  algorithm is open: a further collapse to intervals would need the free lengths to be determined by an
  interval rather than by the whole placed set, and they are not. Over the field of rational functions the
  piecewise closed form has one piece per feasible jamming bitstring, and at the jamming-typical density
  that count grows like \(2^{H(2-1/m)\,\rsan}\) with \(H\) the binary entropy and \(m\) the R\textquotesingle enyi constant, so
  no sub-exponential symbolic form exists. Evaluating this density is therefore far cheaper than writing it down: the gap between \(2^{\rsan}\rsan\) arithmetic operations and a symbolic form with \(2^{\Theta(\rsan)}\) distinct pieces is the separation the section records.
\item
  \textbf{A representability theory for the aggregates.} The first car makes the process regenerate in the
  segment length, and an aggregate satisfies a finite integral equation exactly when it is compatible
  with that regeneration, possibly after augmenting the state by a finite mark. The aggregate section
  develops the three resulting regimes --- a linear Volterra equation for additive functionals with its
  resolvent, a quadratic Dyson--Schwinger equation for the count and the largest gap, and triangular
  hierarchies for order-statistic marginals and multi-point densities --- together with the boundary at
  which no finite equation exists.
\item
  \textbf{Limit theorems for the spatial functionals at jamming.} The asymptotics section establishes a
  central limit theorem for bulk-gap sums and its density-general extension, a sub-Gaussian concentration
  bound, a Gaussian fluctuation law for the median position, a Weibull law for the largest gap with an
  explicit constant, and a functional limit theorem for the empirical gap measure.
\end{enumerate}

\textbf{Mechanised verification.} A companion Lean 4 development (mathlib, Lean \texttt{v4.\allowbreak 33.\allowbreak 0-\allowbreak rc1}) machine-checks a
combinatorial and measure-theoretic core of what follows: the jamming-count bounds, the affineness of the
free lengths in the gap coordinates, and the order-statistic symmetrisation for a dependent parent. It
contains no \texttt{sorry}, and it is deliberately partial: the sequential-uniform process measure and the
polylogarithmic weight theory are carried as explicit hypotheses rather than constructed, so the development
certifies the combinatorial layer and not the analytic results drawn from it.

Rényi's integral equation for the mean count is the one exact result of this kind in the prior literature.
The finite-\(\rsaSpan\) statements above appear neither in the random-sequential-adsorption literature nor in
adjacent order-statistics or applied-probability work; obstacles and open cases are stated where they arise,
and the results that rest on numerical rather than symbolic verification are labelled as such.

\subsubsection{Significance and the elusiveness of exact results}

The uniform car-parking process (UCP) is the model defined above: on a segment \([0, \rsaSpan] \subset
\mathbb{R}\) of length \(\rsaSpan > 0\), unit-length cars attach sequentially with left-endpoint position
\(\rsaX{k}\) at step \(k\) drawn uniformly on the unoccupied portion of \([0, \rsaSpan - 1]\) that admits a new
unit interval without overlap (formal definition in the process section). The process is a canonical model
of \emph{irreversible spatial competition} under a hard exclusion constraint: each car commits to a position the
moment it attaches, without relaxation, diffusion, or re-arrangement. The minimality is sharp in three
respects: UCP is the lowest spatial dimension in which the hard-core constraint carries geometric content;
the deposited object is the simplest convex body in one dimension (a unit interval, no shape or orientation
parameter); and the attachment density is uniform on its support and conditionally Markovian on the current
free region.

Despite this minimality, \emph{exact} finite-boundary results have been elusive for nearly seven decades.
Rényi's 1958 paper (R\textquotesingle enyi 1958) reduces the mean jamming density to a linear Volterra integral equation
that has no elementary solution; the limiting constant \(\rsaMc{1} \approx 0.7475979\) is itself an improper
integral with no closed form, and no exact analogue exists for the variance, higher moments, or the joint
spatial law of the cars at jamming. The closed-form joint order-statistic density of \(\rsan\) uniformly
attached hard rods does not appear in the statistical-mechanics literature (Evans 1993; Krapivsky, Redner and Ben-Naim 2010),
the order-statistics literature (David and Nagaraja 2003), or the dedicated coverage literature
(Kumar and Berman 2014; Pavlic, Wilson, Kumar and Berman 2015). Even the distribution of the random jamming count \(\rsaN\) is not
known exactly at finite \(\rsaSpan\); only its mean and asymptotic variance constant are characterised.

Three structural features explain why exact computations have resisted: \textbf{(i)} the conditional-uniform
attachment rule makes the joint density depend on the temporal sequence of placements, so the symbolic
representation has \(\rsan!\) permutation summands; \textbf{(ii)} the hard-core constraint partitions configuration
space into a piecewise polytopal structure whose cells must be enumerated combinatorially; and \textbf{(iii)} the
underlying Volterra-style integral equations couple solutions across all sub-interval lengths in
\([0, \rsaSpan]\), ruling out lifted closed-form solutions. This paper attacks each feature directly --- a
permutation-sum representation organised by attachment-type codewords, an enumeration of the polytopal
pieces, and an explicit Volterra cascade for the moment hierarchy --- and so produces the first exact,
finite-\(\rsaSpan\) statements which extant literature could characterise only asymptotically.

\paragraph{Prior work, positioned by object and regime}

Prior results on the parking process sort along two axes: the \emph{object} computed --- the jamming count, a single gap, or the full joint law --- and the \emph{regime} of validity --- asymptotic in the segment length, or exact at finite \((\rsaSpan, \rsan)\). The count column is the settled one: Rényi (1958) fixed its limiting mean; Dvoretzky and Robbins (1964) its variance and central limit theorem; Mannion (1976) its higher moments; Penrose and Yukich (2002) extend the count CLT to packing in every dimension. Exact distributional statements stop at one gap at a time: Coffman, Flatto and Jelenković (2000) obtain the limiting law of a single vacant interval, and Gerin (2015) a single-site marginal of the Page--Rényi process. Exact finite-\((\rsaSpan,\rsan)\) joint laws exist only for the non-sequential baselines, uniform and Tonks-equilibrium placement (Kumar and Berman 2014; Kumar 2016). The exact full joint law at finite \((\rsaSpan, \rsan)\) --- the upper-right corner of the map below --- is the region this paper fills.

\pandocbounded{\includegraphics[keepaspectratio]{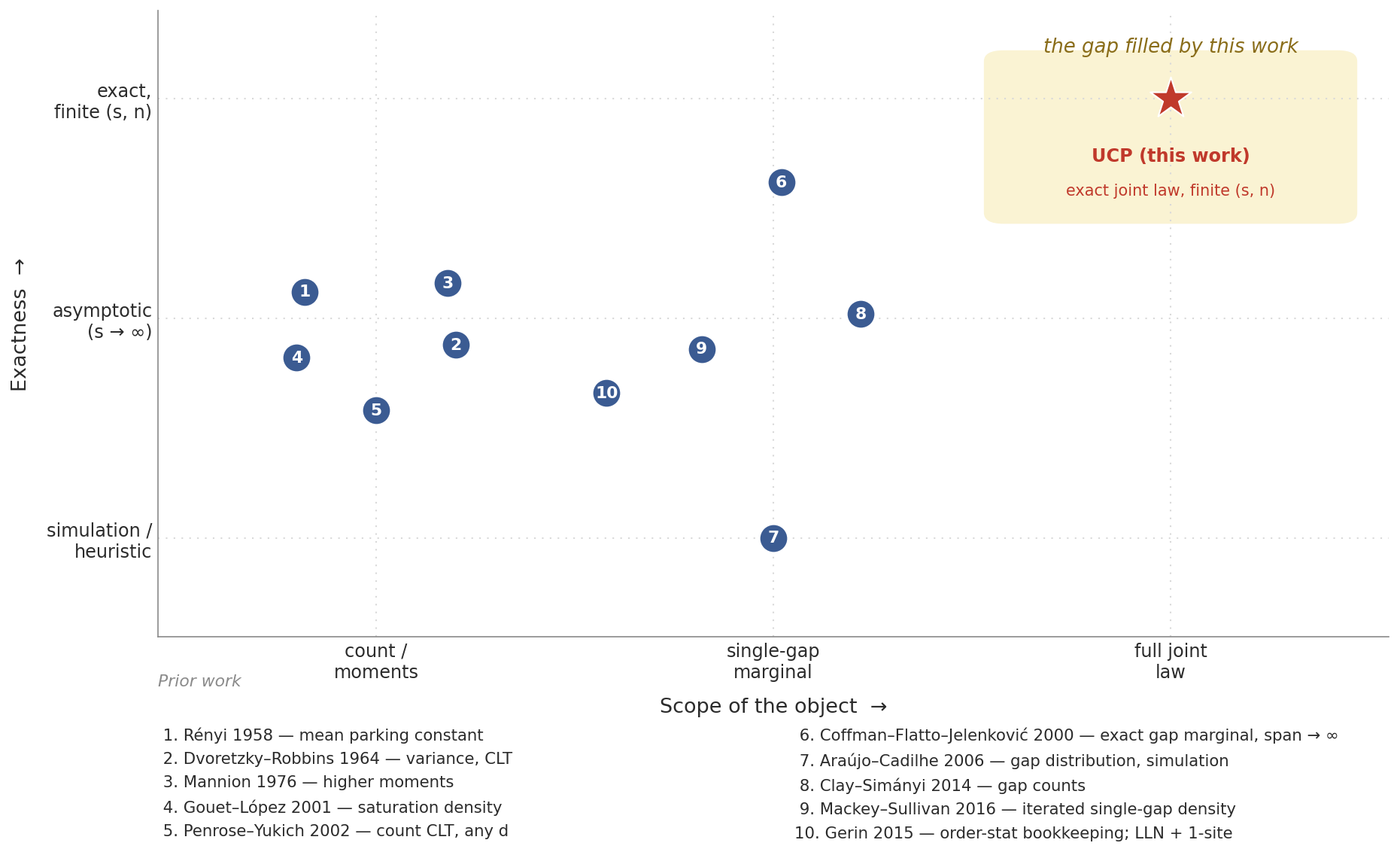}}

\subsubsection{Further examples of the process}

While the car-parking metaphor is a convenient anchor, the sequential placement of identical hard objects on
a one-dimensional substrate recurs across domains:

\begin{itemize}
\tightlist
\item
  \textbf{Multi-robot boundary coverage.} Mobile robots entering a one-dimensional corridor and attaching at
  random positions under a hard non-overlap constraint were analysed in
  (Kumar 2016; Kumar and Berman 2014) as a stochastic boundary-coverage process; the UCP is the
  geometric kernel of the steady-state coverage distribution and the motivating application here.
\item
  \textbf{Polymer adsorption.} Linear polymers of fixed length irreversibly adsorb onto a one-dimensional
  substrate, with sites chosen uniformly and overlapping attempts discarded (Gonz\textquotesingle alez, Hemmer and H\o{}ye 1974).
\item
  \textbf{Hard-rod statistical mechanics.} Tonks's partition function (Tonks 1936) for one-dimensional gases
  of hard elastic spheres is the equilibrium analogue; the UCP is its sequential, non-equilibrium counterpart.
\item
  \textbf{Protein binding on DNA.} Ligands of fixed footprint bind irreversibly at uniform random positions along
  a strand without overlap; the McGhee--von Hippel model's irreversible limit reduces to one-dimensional RSA
  (McGhee and von Hippel 1974).
\item
  \textbf{Nucleosome positioning.} Histone octamers occupy linker-separated intervals along chromatin, a
  sequential hard-core deposition identical to RSA (Kornberg and Stryer 1988).
\item
  \textbf{Vehicular traffic and pedestrian queueing.} Drivers selecting parking slots and pedestrians spacing on
  a queue line have been modelled via one-dimensional RSA (Krapivsky, Redner and Ben-Naim 2010).
\end{itemize}

\section{Summary of results}

\label{sec:summary-of-results}

This section lists what the paper establishes and the evidence carried for each item. The four principal contributions are stated in \ref{sec:introduction}; the table below resolves them into individual results, separating those that are proved, those that are machine-checked, and those that are conjectured with a named obstacle.

The tables use the paper's notation before the sections that construct it, so we name the objects here and leave the constructions where they belong. The process places unit cars sequentially on a segment of length \(s\) (\ref{sec:background-ucp}). Its \textbf{absorption time} \(N_{\mathrm{abs}}(s)\) is the number of cars parked when no gap admits another, \(p_n(s)\) is the probability that this number is \(n\), and \(P_s\) collects those probabilities as a polynomial (\ref{sec:absorption-time}). For a fixed number of cars, \(f_{\mathrm{UCP}}\) is the joint density of their positions and \(\ell_k\) is the total free length available to the \(k\)-th arrival, the two objects the density is built from (\ref{sec:joint-order-stats}).

We use four labels, and they are used strictly. \textbf{Machine-checked} means a Lean 4 declaration against Mathlib whose axiom footprint is exactly \(\{\texttt{propext},\ \texttt{Classical.\allowbreak choice},\ \texttt{Quot.\allowbreak sound}\}\), so no \texttt{sorry} and no external assumption enters. \textbf{Proved} means a proof given here in the usual mathematical register, not formalised. \textbf{Conditional} means proved from a hypothesis that is stated and not discharged. \textbf{Conjectured} means unproved, and we name the obstacle that blocks each one.

\subsubsection{Principal results}

{\footnotesize\begin{longtable}[]{@{}
  >{\raggedright\arraybackslash}p{\dimexpr 0.2500\linewidth-2\tabcolsep\relax}
  >{\raggedright\arraybackslash}p{\dimexpr 0.2500\linewidth-2\tabcolsep\relax}
  >{\raggedright\arraybackslash}p{\dimexpr 0.2500\linewidth-2\tabcolsep\relax}
  >{\raggedright\arraybackslash}p{\dimexpr 0.2500\linewidth-2\tabcolsep\relax}@{}}
\toprule\noalign{}
\begin{minipage}[b]{\linewidth}\raggedright
\#
\end{minipage} & \begin{minipage}[b]{\linewidth}\raggedright
Result
\end{minipage} & \begin{minipage}[b]{\linewidth}\raggedright
Section
\end{minipage} & \begin{minipage}[b]{\linewidth}\raggedright
Evidence
\end{minipage} \\
\midrule\noalign{}
\endhead
\bottomrule\noalign{}
\endlastfoot
1 & The joint density of the positions is the temporal-order sum \(f_{\mathrm{UCP}}=\sum_{\sigma}\prod_k 1/\ell_k^{\sigma}\), one rational function on each region & \ref{sec:joint-order-stats} & machine-checked \\
2 & Every free length is affine in the gaps, with coefficients in \(\{-1,0,1\}\) & \ref{sec:joint-order-stats} & machine-checked \\
3 & The regions are indexed by jamming bitstrings, with a closed-form feasibility criterion and count & \ref{sec:order-stats-cells} & machine-checked \\
4 & On a region the density is a single rational function, with poles only on free-length hyperplanes & \ref{sec:order-stats-cells} & machine-checked \\
5 & The parking arrangement is linearly reducible, so every marginal is a hyperlogarithm of weight at most \(n-1\) & \ref{sec:joint-order-stats} & machine-checked \\
6 & The gap map carries \(f_{\mathrm{UCP}}\): it is affine with unit Jacobian & \ref{sec:gap-order-stats} & machine-checked \\
7 & \(\int_{J_n}f_{\mathrm{UCP}}=\Pr[N_{\mathrm{abs}}=n]\), and these masses sum to one & \ref{sec:background-ucp} & machine-checked \\
8 & Held--Karp evaluation: grouping temporal orders by the last-placed car evaluates the density in \(O(2^n n)\) operations rather than \(n!\) & \ref{sec:order-stats-cells} & recursion machine-checked; the cost bound is not formalised, as no cost model is available \\
9 & The symbolic form has one piece per feasible bitstring, a count growing like \(2^{H(2-1/m)n}\), so no sub-exponential closed form exists & \ref{sec:properties-order-stats} & proved \\
10 & Realizability of a jamming bitstring is decidable in polynomial time & \ref{sec:order-stats-cells} & proved \\
11 & Log-convexity of \(f_{\mathrm{UCP}}\) holds on each region and fails globally, with an exact counterexample at a wall & \ref{sec:properties-order-stats} & machine-checked (both halves) \\
12 & The absorption time is stochastically non-decreasing in the segment length & \ref{sec:absorption-time} & machine-checked, conditional on the splitting law \\
13 & The number of modes of the absorption pmf is at most \(\lceil m/2\rceil\) for a support of \(m\) points & \ref{sec:absorption-time} & machine-checked \\
14 & The absorption pmf admits plateaux, so its mode need not be unique & \ref{sec:absorption-time} & machine-checked, conditional on the splitting law \\
15 & The jamming graph is a linear forest, which reduces every graph invariant to a functional of the jamming bitstring & \ref{sec:process-related} & proved \\
16 & Independence, matching and domination numbers in closed form, with König's identity & \ref{sec:process-related} & machine-checked \\
17 & The void probability, the probability generating functional and the factorial-moment densities in closed form & \ref{sec:process-related} & machine-checked \\
18 & The largest gap obeys a Weibull law, and the extreme-value class flips between the jammed and partial ensembles & \ref{sec:asymptotic-properties} & machine-checked \\
19 & Aggregates satisfy a finite integral equation exactly when they are compatible with regeneration: a linear Volterra equation for additive functionals, a quadratic equation for the count & \ref{sec:aggregate-representations} & proved \\
\end{longtable}}

\subsubsection{What is open, and why}

The open problems concentrate at two places, and each is a single obstacle rather than a list.

{\footnotesize\begin{longtable}[]{@{}
  >{\raggedright\arraybackslash}p{\dimexpr 0.3333\linewidth-2\tabcolsep\relax}
  >{\raggedright\arraybackslash}p{\dimexpr 0.3333\linewidth-2\tabcolsep\relax}
  >{\raggedright\arraybackslash}p{\dimexpr 0.3333\linewidth-2\tabcolsep\relax}@{}}
\toprule\noalign{}
\begin{minipage}[b]{\linewidth}\raggedright
Question
\end{minipage} & \begin{minipage}[b]{\linewidth}\raggedright
Section
\end{minipage} & \begin{minipage}[b]{\linewidth}\raggedright
Obstacle, and the step that would remove it
\end{minipage} \\
\midrule\noalign{}
\endhead
\bottomrule\noalign{}
\endlastfoot
Real-rootedness of the count generating polynomial \(P_s\), hence log-concavity and unimodality of \(p_n\) & \ref{sec:absorption-time} & Real-rootedness is not preserved by the continuous mixture over the split position, and the splitting family has non-constant degree, so the common-interlacer route is unavailable. The remaining candidate is to establish real-rootedness of the symmetric self-convolution \(\int_0^{s-1}P_xP_{s-1-x}\,\mathrm dx\) by a tree-structured aggregation. \\
The likelihood-ratio order in the segment length & \ref{sec:absorption-time} & The same mixture: the order is preserved by independent sums and not by mixtures. Real-rootedness would give it, so this problem is subsumed by the row above. \\
Whether the density can be evaluated in polynomial time, and whether the associated counting problems are \(\#\mathrm P\)-hard & \ref{sec:order-stats-cells} & Open in both directions. A collapse of the subset recursion to intervals would need the free lengths to be interval-determined, and they are not. Settling either direction needs new work: a polynomial algorithm exploiting the geometry of the free lengths, or a gadget reproducing the temporal-order sum from a known hard problem. \\
The asymptotic two-point gap density inside the present framework & \ref{sec:asymptotic-properties} & Its kernel is an exponential period, outside the algebra of multiple polylogarithms in which the finite-length theory sits. The step is to work in that transcendence class directly, targeting the double-Laplace form of the kernel rather than a polylogarithmic one. \\
\end{longtable}}

The Tonks process carries the same unit rods under the Gibbs measure, so its density is flat on the hard-core support; its count polynomial is real-rooted, and every shape property left open above holds for it (\ref{sec:absorption-time}).

\subsubsection{Machine verification}

The formalisation is described in \ref{sec:implementation-details}. Two properties of it are worth stating here, because they bound how much of the paper the reader must take on trust.

First, every result imported from the literature is carried as an explicit hypothesis on the statements that use it, never as an \texttt{axiom}, so each theorem exhibits its own assumptions. Every declaration's axiom footprint is audited, and a single genuine axiom remains in the development, isolated in one module.

Second, verification is reported at the granularity of the individual result rather than the section, which is why the table above carries an evidence column instead of a blanket claim. Where a result is only partly formalised, we state which part.

\section{Problem statement}

\label{sec:problem-statement}

\emph{Spatial statistics of one-dimensional sequential absorption --- the predicates in full
generality, the problem areas confined to unit-length cars.}

\subsubsection{Scope and conventions}

This section fixes the objects, predicates, and problem areas that the rest of the paper works with.
Two conventions organize what follows.

\textbf{General predicates, unit-car problems.} The objects and predicates of §§1--4 are stated for cars
of arbitrary size, the most general one-dimensional hard-core adsorption. The problem areas of §6
then specialize to \textbf{unit-length cars}: the classical \textbf{uniform car-parking process (UCP)},
one-dimensional random sequential adsorption with unit cars, and the object of Rényi's parking
problem. The general framework fixes the language and situates the unit case; the heterogeneous and
non-uniform extensions it admits are recorded but not pursued here.

\textbf{Each area carries both results and open questions.} Every problem area in §6 has an established
core and an open frontier. The finite-\((s,n)\) joint law (G1) is the corner that was missing from the
prior literature; it, like the others, carries results in hand alongside questions that remain open.

\subsubsection{1. Setting and primitives}

\textbf{D1 (Ambient).} The cars are placed on a one-dimensional \textbf{ambient} \(\mathcal{A}\) of extent
\(s > 0\): either the \textbf{segment} \(\Lambda := [0, s]\) of length \(s\), or the \textbf{circle}
\(\mathbb{T}_s := \mathbb{R}/s\mathbb{Z}\) of circumference \(s\). The segment has two \textbf{endpoints}
\(\partial\Lambda = \{0, s\}\); the circle has none. Unless stated otherwise \(\mathcal{A} = \Lambda\).

\textbf{D2 (Cars and sizes).} A \textbf{car} of size \(\ell \ge 0\) at position \(x\) has \textbf{footprint}
\([x, x + \ell] \subseteq \mathcal{A}\). A configuration of \(n\) cars carries a position vector
\(\mathbf{x} = (x_1, \ldots, x_n)\) and a size vector \(\boldsymbol{\ell} = (\ell_1, \ldots, \ell_n)\),
whose entries may be fixed constants or marks drawn from a size distribution. A \textbf{unit car} has size
one; unit cars are the object of the problem areas in §6.

\textbf{D3 (Endpoints).} The two endpoints \(\{0, s\}\) of the segment are degenerate zero-length cars; they
participate in the hard-core constraint on the same footing as cars. The circle has no endpoints,
the structural difference exploited by the integral-equation analysis of §6.

\textbf{D4 (Hard-core constraint; the predicate \(\texttt{is\_\allowbreak admissible}\)).} A set of \(k\) cars is
\textbf{admissible} when their footprints are pairwise disjoint and contained in the ambient:

\[\resizebox{\ifdim\width>\linewidth\linewidth\else\width\fi}{!}{$\displaystyle \texttt{is\_\allowbreak admissible}(\mathbf{x}, \boldsymbol{\ell})
\;\Longleftrightarrow\;
\Bigl(\,\bigwedge_{1 \le i < j \le k} [x_i, x_i{+}\ell_i] \cap [x_j, x_j{+}\ell_j] = \varnothing\,\Bigr)
\;\wedge\;
\Bigl(\,\bigwedge_{i=1}^{k} [x_i, x_i{+}\ell_i] \subseteq \mathcal{A}\,\Bigr).$}\]

Admissibility is the hard-core (non-overlap) exclusion.

\textbf{D5 (Configuration).} A \textbf{configuration} is an admissible set of cars, recorded by
\((\mathbf{x}, \boldsymbol{\ell})\). Its \textbf{occupancy} is the number of cars \(k\).

\subsubsection{2. Configuration space, indices, and jamming}

\textbf{D6 (Spatial and temporal indices).} The \(k\) cars carry two indexings. Sorting the positions gives
the \textbf{order statistics} \(x_{(1)} < \cdots < x_{(k)}\); the rank \(i\) is the \textbf{spatial index}. The
attachment process (D9) places the cars in an temporal order \(X_1, \ldots, X_k\); the arrival rank is
the \textbf{temporal index}. A permutation \(\sigma\) relates the two, \(x_{(i)} = X_{\sigma(i)}\), and is
the \textbf{temporal order} of the configuration. Figure 1 shows both indexings on one instance.

\textbf{D7 (Gaps, spacing, and the jamming vector).} With ordered positions \(x_{(1)} < \cdots < x_{(k)}\),
the \textbf{gaps} \(g_0, g_1, \ldots, g_k\) are the free lengths between consecutive footprints, with \(g_0,
g_k\) the two \textbf{end gaps} against the endpoints; on the segment \(\sum_{m=0}^{k} g_m = s - \sum_i
\ell_i\). The \textbf{spacing} of consecutive cars is \(d_i = x_{(i+1)} - x_{(i)} = \ell + g_i\). A gap
\textbf{admits} a car of size \(\ell\) when its length is at least \(\ell\). For unit cars, the \textbf{jamming
vector} records which gaps are capped:

\[\resizebox{\ifdim\width>\linewidth\linewidth\else\width\fi}{!}{$\displaystyle \mathbf{b} \in \{0,1\}^{k+1}, \quad b_i = \mathbf{1}[\, g_i < 1 \,], \qquad
j \;=\; \mathbf{1}^{\!\top}\mathbf{b} \;=\; \sum_{m=0}^{k} b_m ,$}\]

so \(j\) is the \textbf{number of jammed (capped) gaps}, the count that drives the jamming fraction of §5.

\textbf{D8 (Hard-core polytope).} For a common size \(\ell_0\) the ordered positions of \(k\) cars range
over the \textbf{order-statistic simplex}

\[\resizebox{\ifdim\width>\linewidth\linewidth\else\width\fi}{!}{$\displaystyle \mathsf S_k(s; \ell_0) \;=\; \bigl\{\, 0 \le x_{(1)},\;\; x_{(m+1)} \ge x_{(m)} + \ell_0,\;\;
x_{(k)} \le s - \ell_0 \,\bigr\},$}\]

whose facets are the hyperplanes on which a gap equals \(\ell_0\). Under heterogeneous sizes the domain
becomes a union of such polytopes over the size-to-rank assignments. The problem areas adopt
\(\ell_0 = 1\), on which \(\mathsf S_n(s; 1)\) is the domain of every finite-\((s,n)\) spatial law in §6.

\pandocbounded{\includegraphics[keepaspectratio]{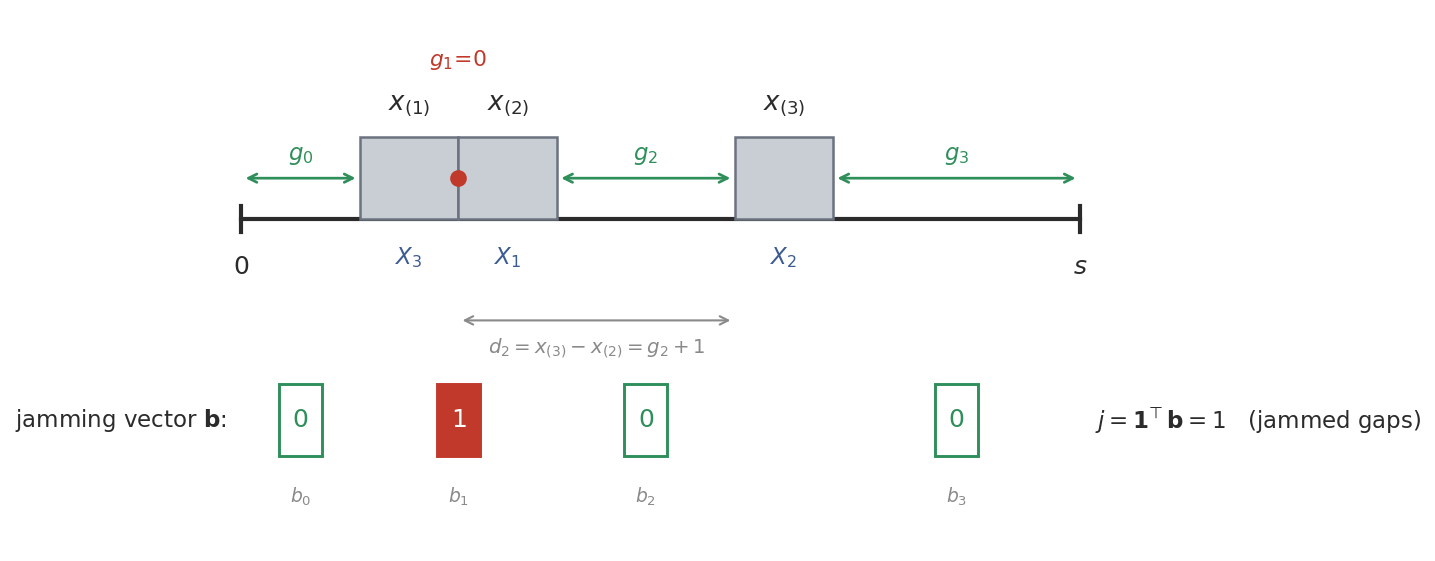}}

\textbf{Figure 1.} A partial configuration of three unit cars on a segment of length \(s\).
\emph{Above} each car: the spatial index (order statistic) \(x_{(i)}\); \emph{below}: the temporal
(attachment) index \(X_j\), so the temporal order here is \(\sigma = (x_{(1)},x_{(2)},x_{(3)}) =
(X_3,X_1,X_2)\). Gaps \(g_i\) are green when open (\(\ge 1\)) and red when capped (\(<1\)); here \(g_1 = 0\)
is the single jammed gap. The spacing \(d_2 = x_{(3)}-x_{(2)} = g_2 + 1\). The jamming vector is
\(\mathbf{b} = (0,1,0,0)\) with \(j = \mathbf{1}^{\!\top}\mathbf{b} = 1\); the configuration is
\emph{not} jammed, since three gaps still admit a car. Reproduces the paper's
fig:segment-spatial-temporal, extended with \(\mathbf{b}\).

\subsubsection{3. Dynamics and absorption}

\textbf{D9 (Attachment process).} The \textbf{attachment process} places cars sequentially: each new car is
drawn uniformly over the currently admissible positions (the \textbf{free region}), independently of the
past given the current configuration. The process is the one-dimensional case of \textbf{random sequential
adsorption}, and its placements are irreversible: a car, once placed, never moves. The framework
admits a general placement intensity in place of the uniform law; the uniform intensity is the one
studied here.

\textbf{D10 (Attachment density).} After \(k\) attachments the ordered positions have a joint density on the
order-statistic simplex, the \textbf{attachment density} \(f^{(k)}_s\). Its finite-\((s,n)\) form for unit
cars is the paper's central result and the object of problem area G1.

\textbf{D11 (Jamming and absorption; the predicate \(\texttt{is\_\allowbreak jammed}\)).} A configuration is \textbf{jammed}
when no gap admits a further car --- equivalently, when the jamming vector of D7 is all ones:

\[\resizebox{\ifdim\width>\linewidth\linewidth\else\width\fi}{!}{$\displaystyle \texttt{is\_\allowbreak jammed}(\mathbf{x}) \;\Longleftrightarrow\; \mathbf{b} = \mathbf{1}
\;\Longleftrightarrow\; j = k + 1 \;\Longleftrightarrow\; \max_{0 \le m \le k} g_m < 1 .$}\]

\(\texttt{is\_\allowbreak jammed}\) is the absorbing condition of the attachment process: the process runs until it
reaches a jammed configuration and then stops. The \textbf{jamming count} \(N = N(s)\) is the occupancy at
absorption. The jammed configurations are the absorbing states of the attachment Markov process, so
the count law, the time to absorption, and the quasi-stationary behaviour are the questions of
problem area G2.

\textbf{D12 (Two ensembles).} Two distinct laws share the same geometry:

\begin{itemize}
\tightlist
\item
  the \textbf{partial ensemble} --- the configuration after a \emph{fixed} number \(n\) of attachments,
  supported on the hard-core polytope \(\mathsf S_n(s; 1)\);
\item
  the \textbf{jammed ensemble} --- the configuration \emph{conditioned to be jammed} with \(N = n\),
  supported on the sub-region \(\{\texttt{is\_\allowbreak jammed}\} \subseteq \mathsf S_n(s; 1)\).
\end{itemize}

For unit cars they coincide only when \(s < n + 1\); elsewhere they are genuinely different, and a
spatial statistic must name which ensemble it refers to.

\subsubsection{4. Specialization: homogeneity and unit cars}

The predicates of §§1--3 hold for arbitrary sizes. Two further predicates cut the general
framework down to the object the problem areas study.

\textbf{D13 (The predicate \(\texttt{is\_\allowbreak homogeneous}\)).} A size vector is \textbf{homogeneous} when all cars
share one size:

\[\resizebox{\ifdim\width>\linewidth\linewidth\else\width\fi}{!}{$\displaystyle \texttt{is\_\allowbreak homogeneous}(\boldsymbol{\ell}) \;\Longleftrightarrow\;
\ell_1 = \ell_2 = \cdots = \ell_n \;=:\; \ell_0 .$}\]

Rescaling the ambient by \(1/\ell_0\) carries the homogeneous case to unit cars on an ambient of extent
\(s/\ell_0\), so homogeneity loses no generality beyond the single common size.

\textbf{D14 (The predicate \(\texttt{unit\_\allowbreak length}\)).} A size vector is \textbf{unit} when it is homogeneous with
common size one:

\[\resizebox{\ifdim\width>\linewidth\linewidth\else\width\fi}{!}{$\displaystyle \texttt{unit\_\allowbreak length}(\boldsymbol{\ell}) \;\Longleftrightarrow\;
\texttt{is\_\allowbreak homogeneous}(\boldsymbol{\ell}) \;\wedge\; \ell_0 = 1 .$}\]

This is the unit car of D2. The attachment process of D9 on unit cars is the \textbf{uniform car-parking
process (UCP)} --- one-dimensional random sequential adsorption with unit cars, the classical
object of Rényi's parking problem, and the process every problem area below studies. The chain
\(\texttt{unit\_\allowbreak length} \Rightarrow \texttt{is\_\allowbreak homogeneous} \Rightarrow (\text{general})\) runs from
the studied case outward to the framework.

\textbf{Out of scope.} The predicates admit two extensions the problem areas do not pursue: heterogeneous
sizes (a size distribution over cars) and a non-uniform placement intensity. Both are recorded so the
framework is stated once in full generality; the problem areas are confined to unit cars under the
uniform law, since heterogeneity would take the analysis too far afield.

\subsubsection{5. The regime plane for asymptotics}

The asymptotic and absorption goals are organized on a plane of two scaling axes, along which \(s\) and
\(n\) grow together. All of §§5--7 fix unit cars.

\textbf{D15 (Coverage and jamming fraction).} Fix a scaling of \((s, n)\) with \(s \to \infty\).

\begin{itemize}
\tightlist
\item
  The \textbf{coverage} \(\theta = n/s \in [0, 1]\) is the occupied fraction of the segment. Random
  sequential adsorption fills toward the jamming coverage \(\theta^{\*} = m \approx 0.7476\), the
  Rényi parking constant.
\item
  The \textbf{jamming fraction} \(\varphi = j/n \in [0, 1]\) is the share of gaps that are capped, with the
  jammed-gap count \(j = \mathbf{1}^{\!\top}\mathbf{b}\) of D7. The value \(\varphi = 1\) is the fully
  jammed boundary (\(\mathbf{b} = \mathbf{1}\)).
\end{itemize}

The two axes are linked, not free: a fully capped configuration (\(\varphi = 1\)) forces mean gap below
one, hence \(\theta \gtrsim \tfrac12\); small coverage forces \(\varphi < 1\). A typical adsorption
trajectory runs from \((\theta, \varphi) = (0, 0)\) to the corner \((m, 1)\). The plane carries four
regimes --- \textbf{sparse} (\(\theta \to 0\)), \textbf{partial} (\(\theta \in (0, m)\), \(\varphi < 1\)),
\textbf{jammed} (\(\theta = m\), \(\varphi = 1\)), and \textbf{super-dense} (\(\theta \in (m, 1]\)) --- and the
line \(\varphi = 1\) is a phase boundary at which the extreme-value class of the largest gap changes.
Each asymptotic result in §6 is a statement about a region of this plane, and which regions admit an
integral-equation treatment is itself an open question (G6).

\subsubsection{6. Open problem areas}

The areas below fix unit cars. Each carries an established core and an open frontier; the detail of
the frontier is deliberately left open.

\textbf{G1. Joint density laws at finite \((s, n)\).} The exact per-cell attachment density \(f^{(n)}_s\) and
everything read from it --- marginals, joint and marginal distribution functions, quantiles, mode
structure, mixed moments, log-convexity. \emph{Core:} the density and its principal functionals are in
hand for unit cars. \emph{Open:} extension to the jammed ensemble and to the circle.

\textbf{G2. Absorption and the jamming count.} The law of the jamming count \(N(s)\) as the absorption level
of the attachment process: probability mass function, log-concavity and real-rootedness, exact facet
laws at small \(n\), mean and variance and their convergence to the Rényi constants, and the
time-to-absorption and quasi-stationary reading of the same process. \emph{Core:} exact small-\(n\) laws and
the mean/variance asymptotics. \emph{Open:} real-rootedness in general and the full law of \(N_{\mathrm{abs}}\).
Located on the coverage axis of the regime plane.

\textbf{G3. Gap and spatial order statistics.} The joint law of the gaps and of the ordered positions:
gap-order-statistic marginals and means, the spacing law, the largest gap as a jamming certificate,
dependence structure and reflection symmetry, and the extreme-value behaviour whose class flips
across the \(\varphi = 1\) boundary. \emph{Core:} the finite joint gap density and its marginals. \emph{Open:} the
asymptotic extreme-value theory across the regime plane.

\textbf{G4. Asymptotics and limit theorems.} Central limit theorems for bulk-gap functionals, Gaussian
behaviour of the sample median, sub-Gaussian concentration, Berry--Esseen rates, the empirical
gap law, and stabilization. \emph{Core:} the bulk-gap CLT and concentration. \emph{Open:} the super-dense strip
and atypical jamming fractions, each a marked region of the plane.

\textbf{G5. Point-process and random-geometric-graph structure.} The jammed configuration as a point
process and as a random geometric graph: reduction of Penrose graph invariants and Haenggi
descriptors to the finite density, Palm and Mecke factorization, the structure factor and the
hyperuniformity question, and factorial-moment densities. \emph{Core:} the finite-\((s,n)\) reductions to the
density. \emph{Open:} the hyperuniformity classification and the asymptotic graph limits.

\textbf{G6. Aggregate and integral-equation analysis.} Integral-equation representations of the process:
Rényi-style representability of aggregate functionals, the order-statistic integral-equation
hierarchy, the closure dichotomy (which additive functionals close into a single equation), and
moment cascades. \emph{Core:} the count and moment integral equations and the closure dichotomy. \emph{Open:}
which regimes and constraints admit an integral equation at all, against those needing the BBGKY/BBV
hierarchy. The Hopf-algebraic machinery underpinning these representations is a \emph{method} for this
area, introduced in the preliminaries rather than posed as a goal.

\textbf{G7. Generalization axes (within unit length).} Three directions keep the unit-car object while
relaxing another feature, each recovering the studied case at one end: \textbf{reversibility} (irreversible
adsorption, then the reversible Tonks equilibrium, then hybrid interpolations); \textbf{ambient} (segment,
then circle); and \textbf{dimension} (the line, then higher-dimensional adsorption of unit balls). The size
axis of D12 --- heterogeneous cars --- and a non-uniform placement intensity are deliberately
excluded, as noted in §4.

\section{Preliminaries}
\label{sec:preliminaries}

\subsection{Background: the UCP as a process}

\label{sec:background-ucp}

\emph{The uniform car-parking process (UCP) as an absorbing Markov chain --- the attachment kernel,
the all-jam absorbing state, the \(N=n\) conditioning, and the cited large-\(s\) stabilisation tool
--- developed once for the sections that specialise it.}

\subsubsection{Scope}

The problem statement fixed the objects; this section develops the process theory the later sections
use: the attachment dynamics and absorption (§§1--2), the strong Markov factorisation (§3), the two
ensembles (§4), and the cited large-\(s\) stabilisation tool (§5). The configuration-space geometry and
the Tonks control live in Section \ref{sec:background-order-statistics}. Section 6 tabulates which aspect each later
section uses.

\subsubsection{1. The attachment chain}

\refstepcounter{thmcnt}\label{def:attachment-chain}\textbf{Definition\nobreakspace{}\thethmcnt{} (attachment chain).} Cars are unit intervals placed on \([0, s]\) one at a time. Write
\(\Pi_k = \{X_1, \dots, X_k\}\) for the set of left endpoints after \(k\) placements, with order statistics
\(X_{(1)} < \cdots < X_{(k)}\), and \(\widehat F_k = \{\,x \in [0, s-1] : [x, x+1]\text{ overlaps no car of }\Pi_k\,\}\)
for the \textbf{free region}. The \textbf{attachment kernel} places \(X_{k+1}\) uniformly on \(\widehat F_k\) and sets
\(\Pi_{k+1} = \Pi_k \cup \{X_{k+1}\}\), from \(\Pi_0 = \varnothing\). The sequence \((\Pi_k)_{k \ge 0}\) is the
\textbf{attachment chain}, and the process it defines is the \textbf{UCP}. Placements are irreversible; the UCP
is the one-dimensional random sequential adsorption process (R\textquotesingle enyi 1958).

\refstepcounter{thmcnt}\label{prop:markov-property}\textbf{Proposition\nobreakspace{}\thethmcnt{} (Markov property).} \((\Pi_k)_{k \ge 0}\) is a Markov chain.

\emph{Proof.} The free region \(\widehat F_k\) is a deterministic function of \(\Pi_k\), and the kernel draws
\(X_{k+1}\) from the uniform law on \(\widehat F_k\). Hence the conditional law of \(\Pi_{k+1}\) given
\(\Pi_0, \dots, \Pi_k\) depends only on \(\Pi_k\). \(\quad\blacksquare\)

\pandocbounded{\includegraphics[keepaspectratio]{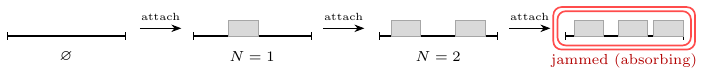}}

\textbf{Figure 1.} The attachment chain: each step draws a car uniformly on the free region, until the
configuration jams. The jammed state (double frame) is absorbing.

\subsubsection{2. Absorption at jamming}

\refstepcounter{thmcnt}\label{def:jammed-configuration}\textbf{Definition\nobreakspace{}\thethmcnt{} (jammed configuration).} A configuration is \textbf{jammed} when \(\widehat F_k = \varnothing\),
that is, when no admissible left endpoint remains. Writing \(g_0, \dots, g_k\) for the \(k+1\) gaps (the
components of \([0, s]\) not covered by a car), the configuration is jammed if and only if
\(\max_i g_i < 1\).

The jammed configurations are the \textbf{absorbing states} of the attachment chain, since from a jammed
state the free region is empty and no transition is defined. Each car occupies unit length, so at most
\(\lfloor s \rfloor\) disjoint cars fit and the chain reaches a jammed state after finitely many steps
almost surely. The \textbf{jamming count} \(N = N(s)\) is the number of cars at absorption; read on the step
index it is the \textbf{absorption time} of the chain.

\refstepcounter{thmcnt}\label{prop:feasible-range}\textbf{Proposition\nobreakspace{}\thethmcnt{} (feasible range).} The jamming count satisfies
\[\resizebox{\ifdim\width>\linewidth\linewidth\else\width\fi}{!}{$\displaystyle \Bigl\lfloor \tfrac{s+1}{2} \Bigr\rfloor \;\le\; N(s) \;\le\; \lfloor s \rfloor .$}\]

\emph{Proof.} The \(N\) cars are disjoint unit intervals in \([0, s]\), so \(N \le s\) and hence
\(N \le \lfloor s \rfloor\). For the lower bound, the \(N\) cars leave \(N+1\) gaps of total length \(s - N\).
At a jammed configuration every gap has length below one (Definition \ref{def:jammed-configuration}), so
\(s - N = \sum_{i=0}^{N} g_i < N + 1\). Therefore \(s - 1 < 2N\), and since \(N\) is an integer,
\(N \ge \lfloor (s+1)/2 \rfloor\). \(\quad\blacksquare\)

\subsubsection{3. The strong Markov property and sub-interval factorisation}

The attachment chain restarts as a fresh UCP after any placement, independently on the sub-intervals a
placement creates. The arrivals landing in a given gap occur at random indices, so the restart is taken at
optional times rather than at fixed ones; those times are \(\mathbb N\)-valued, and the strong Markov property
for a time-homogeneous chain at an optional time taking countably many values is Kallenberg (2021),
Proposition 11.9.

\refstepcounter{thmcnt}\label{prop:sub-interval-factorisation}\textbf{Proposition\nobreakspace{}\thethmcnt{} (sub-interval factorisation).} Condition on \(\Pi_k\). Each gap \([a_i, b_i]\) with
\(b_i - a_i \ge 1\) still admits cars, and the placements landing in it form an independent UCP on
\([a_i, b_i]\); gaps shorter than one admit no further car. Consequently the conditional law of
\((\Pi_m)_{m > k}\) factors into one UCP per admitting gap.

\emph{Proof.} Given \(\Pi_k\), the free region is the disjoint union
\(\widehat F_k = \bigcup_{i :\, b_i - a_i \ge 1} [a_i,\, b_i - 1]\) over the admitting gaps. The kernel draws
each subsequent left endpoint uniformly on the current free region, and a car placed in one gap alters
only that gap's admissible set. The restriction of the kernel to \([a_i, b_i]\) is therefore the UCP
kernel on \([a_i, b_i]\) and is independent of the restrictions to the other gaps. Let \(\tau^i_1<\tau^i_2<\cdots\)
be the successive indices at which an arrival lands in \([a_i,b_i]\); each is an optional time for the chain,
so the strong Markov property applies at it and the subsequence \((\Pi_{\tau^i_m})_m\) restricted to that gap
is a UCP on \([a_i,b_i]\), independent across \(i\). \(\quad\blacksquare\)

Conditioning on the first arrival and averaging is \textbf{first-step analysis} for this chain: an aggregate
evaluated on the whole configuration is written as its first-car contribution plus the same aggregate on the
two sub-segments, and taking expectations turns that identity into an equation. The integral equations of the
aggregate section are exactly those equations, and their solution operator is the resolvent of the attachment
kernel; which of them close in the segment length alone is settled there.

The factorisation yields the recursions for \textbf{count-level} aggregates. Splitting on the first
placement and averaging over its position gives the Rényi split-convolution integral equation for
\(N\) and its moments: the count is additive across the split, so the two sides contribute
independently and the interleaving of their arrivals is irrelevant to the total.

It does \textbf{not} yield the subset recursion for the joint order-statistic density, and that recursion
is not derived from it. A first-placement split cannot carry the density's weight: the \(k\)-th
arriving car divides by the free length of the \emph{whole} configuration, which after a split is the sum
of the two sides' free lengths, and \(1/(\ell_{\mathrm{left}}+\ell_{\mathrm{right}})\) does not
factor. What Proposition \ref{prop:sub-interval-factorisation} supplies is independence of the gap \textbf{subsequences in their own time
index} --- enough for the count, whose value is independent of the interleaving, and not enough for
a fixed-\(n\) joint density, whose weights are not. The subset recursion is obtained instead by peeling the
\textbf{last} car and conditioning on the placed \emph{set}, for which the free length is determined; that
argument is independent of this section.

\subsubsection{4. The partial and jammed ensembles}

\refstepcounter{thmcnt}\label{def:partial-and-jammed-ensembles}\textbf{Definition\nobreakspace{}\thethmcnt{} (partial and jammed ensembles).} Fix \(n\) in the feasible range of Proposition \ref{prop:feasible-range}. The
\textbf{partial ensemble}, \(\Pi_n\) conditioned on \(\{n \le N\}\), is the step-\(n\) marginal of the chain,
supported on the hard-core polytope \(\mathsf S_n\) (\ref{sec:background-order-statistics}, Definition \ref{def:hard-core-polytope}). The \textbf{jammed ensemble}, \(\Pi_n\) conditioned on
\(\{N = n\}\), is fully jammed with \(n\) cars, supported on the jammed sub-region \(J_n\) (\ref{sec:background-order-statistics}, Definition \ref{def:gap-simplex}).

\refstepcounter{thmcnt}\label{prop:coincidence-of-the-ensembles}\textbf{Proposition\nobreakspace{}\thethmcnt{} (coincidence of the ensembles).} The partial and jammed ensembles coincide if
\(s < n + 1\) and differ otherwise.

\emph{Proof.} With \(n\) cars placed, the total free length is \(s - n\). If \(s < n + 1\) then \(s - n < 1\), so
every gap is shorter than one and every \(n\)-car configuration is already jammed; conditioning on
\(\{N = n\}\) is vacuous and the two ensembles coincide. If \(s \ge n + 1\) then \(s - n \ge 1\), and a
configuration with a single gap of length at least one is unjammed yet carries positive mass under the
partial ensemble and none under the jammed ensemble, so the two differ. \(\quad\blacksquare\)

We therefore state every spatial statistic below for a named ensemble.

\refstepcounter{thmcnt}\label{prop:count-as-jammed-region-mass}\textbf{Proposition\nobreakspace{}\thethmcnt{} (the count as jammed-region mass).} For \(n\) in the feasible range, the law of the
configuration after \(n\) attachments (a sub-probability measure of total mass \(\Pr[N\ge n]\)) assigns the
jammed sub-region \(J_n\) the mass \(\Pr[N(s)=n]\). Equivalently, for any density \(f^{(n)}\) of that
configuration, \(\int_{J_n} f^{(n)} = \Pr[N(s)=n]\).

\emph{Proof.} Deleting any car from a jammed \(n\)-configuration merges its two neighbouring gaps with the
car's unit length into a single gap of length \(g_{i-1}+1+g_i\ge1\); every proper \((n{-}1)\)-subconfiguration
therefore has a gap that admits a car and is unjammed. The attachment chain cannot absorb before the
\(n\)-th car and absorbs exactly when the \(n\)-th makes every gap below one, so \(\{N{=}n\}\) is the event that
the configuration after \(n\) attachments lies in \(J_n\); its probability is that configuration's \(J_n\)-mass.
\(\quad\blacksquare\)

\subsubsection{5. Gap and position coordinates}

The configuration of \(n\) parked cars on \([0,s]\) carries two coordinate systems, and every later
section moves between them. The positions are the order statistics
\(X_{(1)}<\cdots<X_{(n)}\) of the left edges; the gaps are the \(n{+}1\) free lengths
\[\resizebox{\ifdim\width>\linewidth\linewidth\else\width\fi}{!}{$\displaystyle G_0=X_{(1)},\qquad G_i=X_{(i+1)}-X_{(i)}-1\ \ (1\le i\le n-1),\qquad G_n=(s-1)-X_{(n)} .$}\]

\refstepcounter{thmcnt}\label{prop:gap-position-coordinates}\textbf{Proposition\nobreakspace{}\thethmcnt{} (gap--position coordinates).} Write \(\mathbf G=(G_0,\dots,G_n)^{\mathsf T}\) and
\(\mathbf X=(X_{(1)},\dots,X_{(n)})^{\mathsf T}\). Then \(\mathbf X=L\mathbf G+\mathbf c\), where \(L\) is
the \(n\times(n{+}1)\) matrix \(L_{ik}=\mathbf 1\{k\le i-1\}\) and \(c_i=i-1\).

\begin{enumerate}
\def\labelenumi{\arabic{enumi}.}
\tightlist
\item
  \(L\) has rank \(n\) and is \textbf{not} injective: the gap vector carries one redundant coordinate, fixed
  by the conservation law \(\sum_{k=0}^{n}G_k=s-n\).
\item
  On that hyperplane, eliminating \(G_n\) leaves the square map
  \(L'\colon(G_0,\dots,G_{n-1})\mapsto\mathbf X\) with \(L'_{ik}=\mathbf 1\{k\le i\}\), the
  lower-triangular matrix of ones. Then \(\det L'=1\), and
  \[\resizebox{\ifdim\width>\linewidth\linewidth\else\width\fi}{!}{$\displaystyle (L')^{-1}=\begin{pmatrix}1&&&\\-1&1&&\\&\ddots&\ddots&\\&&-1&1\end{pmatrix},$}\]
  the lower-bidiagonal difference operator, which is the display above read backwards.
\item
  Consequently \(\mathbf G\mapsto\mathbf X\) is an affine bijection from the gap simplex
  \(\mathcal G_n=\{\mathbf G\ge\mathbf 0:\sum_k G_k=s-n\}\) onto the position simplex \(\mathsf S_n\),
  it is unimodular, and Lebesgue measure is preserved.
\end{enumerate}

\emph{Proof.} Summing the display telescopes to \(X_{(i)}=(i-1)+\sum_{k<i}G_k\), which is the stated affine
form. \(L'\) is triangular with unit diagonal, so \(\det L'=1\) and \(L'\) is invertible over \(\mathbb Z\);
inverting a triangular system of ones gives the difference operator. A unimodular affine map has
Jacobian \(1\). \(\qquad\square\)

\textbf{Why it matters here.} Because the Jacobian is \(1\), a density written in gap coordinates and the
same density written in positions are the \emph{same function of the configuration}: no factor is picked
up in either direction. Later sections therefore state a law once, in whichever coordinate makes the
constraints linear, and read it in the other without comment. The gaps make the jamming conditions
coordinate-wise (\(G_i<1\)); the positions make the ordering coordinate-wise
(\(X_{(i+1)}\ge X_{(i)}+1\)).

\textbf{Two objects that are not this one.} The \emph{gap order statistics} \(G_{(1)}\le\cdots\le G_{(n+1)}\)
sort the gaps and are studied in \ref{sec:gap-order-stats}; sorting is not affine and does not preserve
the correspondence above. And \(L\) itself is not invertible as a map on \(\mathbb R^{n+1}\) --- only its
restriction to the conservation hyperplane is.

\subsubsection{\texorpdfstring{6. The stabilisation tool for the large-\(s\) count limit}{6. The stabilisation tool for the large-s count limit}}

As \(s \to \infty\) the saturated count \(N(s)\) has a limit governed by the stabilisation method, cited
here and scoped to the saturation regime. The paper's spatial limit laws --- the bulk-gap and extreme-gap
limits, which differ between the partial and jammed ensembles --- are proved in the asymptotics section
against the spacing baseline of \ref{sec:background-order-statistics} §2, not here.

\textbf{Stabilisation (Penrose--Yukich 2002).} For the saturated (jammed) count \(N(s)\), the packing
functional is a sum of local contributions with almost-surely finite range of dependence. The law of
large numbers \(N(s)/s \to m\), with \(m \approx 0.7476\) the R\textquotesingle enyi parking constant, is R\textquotesingle enyi's (1958)
one-dimensional result; Penrose and Yukich (2002) prove, in every dimension by stabilisation, the
accompanying central limit theorem --- the asymptotic normality of the saturated count \(N(s)\). This is a
saturation-regime statement: it governs the jammed count, not the partial ensemble.

\subsubsection{7. Each later section studies one aspect of the chain}

{\footnotesize\begin{longtable}[]{@{}
  >{\raggedright\arraybackslash}p{\dimexpr 0.5000\linewidth-2\tabcolsep\relax}
  >{\raggedright\arraybackslash}p{\dimexpr 0.5000\linewidth-2\tabcolsep\relax}@{}}
\toprule\noalign{}
\begin{minipage}[b]{\linewidth}\raggedright
Section
\end{minipage} & \begin{minipage}[b]{\linewidth}\raggedright
Aspect of the attachment chain
\end{minipage} \\
\midrule\noalign{}
\endhead
\bottomrule\noalign{}
\endlastfoot
order statistics & the density on the hard-core polytope \(\mathsf S_n\); \textbf{jammed} \(= N{=}n\) conditioning, \textbf{partial} \(=\) step-\(n\) marginal \\
absorption time & the absorption level / time \(N(s)\) and its law \\
aggregate / count & the law and moments of \(N(s)\), via the strong-Markov split-convolution (R\textquotesingle enyi IE) \\
point process / RGG & the absorbed configuration as a point process and random geometric graph \\
asymptotics & the large-\(s\) limit --- stabilisation, scoped to saturation (\S5) \\
\end{longtable}}

Definitions and predicates (car, segment, endpoint, gap, \texttt{is\_\allowbreak jammed}, the two ensembles) are those of
the problem statement; this section adds the Markov, absorption, factorisation, and (cited, scoped) limit theory; the configuration-space geometry and the Tonks control are in Section \ref{sec:background-order-statistics}.

\subsection{Background: order statistics and the configuration spaces}

\label{sec:background-order-statistics}

\emph{The order-statistics toolbox and the configuration-space geometry the spatial sections use ---
the simplices and the Tonks transform, the iid uniform baseline, the Tonks (jointly uniform)
equilibrium control, and the order statistics of a dependent parent.}

\subsubsection{Scope}

The car positions of the UCP are order statistics, so this is the ambient language for the joint
density and the gaps. The section fixes the configuration-space geometry the spatial sections
integrate over (§1), records the two reference laws the UCP is measured against --- the iid
uniform baseline (§2) and the Tonks equilibrium control (§3) --- and states the two classical
tools the dependent theory needs: and the order-statistic density of a dependent parent (§4).

\subsubsection{1. Order statistics, spacings, and the configuration spaces}

Let \(X_1, \ldots, X_n\) be real random variables (the \textbf{parent} sample). Their \textbf{order statistics}
\(X_{(1)} \le \cdots \le X_{(n)}\) are the sorted values; the rank \(i\) is the \textbf{spatial index}, and the
consecutive differences \(X_{(i)} - X_{(i-1)}\) are the \textbf{spacings}. For a continuous parent the
inequalities are strict almost surely. The ordered positions of \(n\) points range over one of four
affinely related polytopes, fixed here as the coordinate framework the spatial sections use.

\refstepcounter{thmcnt}\label{def:order-statistic-simplex}\textbf{Definition\nobreakspace{}\thethmcnt{} (order-statistic simplex).} The order statistics of \(n\) points on an interval of
length \(\ell\) occupy
\[\resizebox{\ifdim\width>\linewidth\linewidth\else\width\fi}{!}{$\displaystyle \Lambda_n(\ell) \;=\; \{\, \mathbf u \in \mathbb R^n :\ 0 \le u_1 \le \cdots \le u_n \le \ell \,\},
\qquad \operatorname{Vol}\Lambda_n(\ell) \;=\; \ell^{\,n}/n!.$}\]
With no interaction between the points, \(\Lambda_n(s)\) is the support of the iid uniform baseline (§2).

\refstepcounter{thmcnt}\label{def:hard-core-polytope}\textbf{Definition\nobreakspace{}\thethmcnt{} (hard-core polytope).} For \(n\) \textbf{unit} cars the non-overlap constraint confines the
ordered left endpoints to
\[\resizebox{\ifdim\width>\linewidth\linewidth\else\width\fi}{!}{$\displaystyle \mathsf S_n \;=\; \{\, \mathbf x :\ 0 \le x_{(1)} \ \wedge\ x_{(i+1)} - x_{(i)} \ge 1 \ (1 \le i < n)
\ \wedge\ x_{(n)} \le s - 1 \,\},$}\]
the \(n-1\) interior inequalities being the hard-core constraints and the two outer ones the endpoints.
\(\mathsf S_n\) is the support both of the UCP and of the Tonks control (§3); the plain simplex
\(\Lambda_n(s)\) is \emph{not} their support.

\refstepcounter{thmcnt}\label{prop:tonks-transform}\textbf{Proposition\nobreakspace{}\thethmcnt{} (Tonks transform).} The per-coordinate shift \(z_i = x_{(i)} - (i-1)\) is a
volume-preserving affine bijection of \(\mathsf S_n\) onto the \textbf{Tonks simplex}
\(\mathsf T_n = \Lambda_n(s-n)\), carrying each hard-core inequality \(x_{(i+1)} - x_{(i)} \ge 1\) to the
plain ordering \(z_{i+1} \ge z_i\). Hence \(\operatorname{Vol}\mathsf S_n = (s-n)^n/n!\).

\emph{Proof.} The map is a translation in each coordinate, so its Jacobian is \(1\). Substituting
\(z_i = x_{(i)} - (i-1)\) sends \(x_{(i+1)} - x_{(i)} \ge 1\) to \(z_{i+1} \ge z_i\) and the endpoint
constraints to \(0 \le z_1\) and \(z_n \le s-n\); the image is \(\Lambda_n(s-n)\), of volume \((s-n)^n/n!\) by
Definition \ref{def:order-statistic-simplex}. \(\quad\blacksquare\)

\refstepcounter{thmcnt}\label{def:gap-simplex}\textbf{Definition\nobreakspace{}\thethmcnt{} (gap simplex).} In the gap coordinates \(g_i\) (the free lengths between consecutive
footprints, with \(g_0, g_n\) the two end gaps), the \(n\)-car configurations occupy the \textbf{gap simplex}
\[\resizebox{\ifdim\width>\linewidth\linewidth\else\width\fi}{!}{$\displaystyle \mathcal G_n \;=\; \{\, \mathbf g \ge 0 :\ \textstyle\sum_{i=0}^n g_i = s - n \,\},$}\]
affinely equivalent to \(\mathsf T_n\); the \textbf{jammed sub-region} is
\(J_n = \mathcal G_n \cap \{g_i < 1 \ \forall i\}\). Throughout, \(\Lambda_n\) (baseline), \(\mathsf S_n\)
(hard-core support), \(\mathsf T_n\) (Tonks image), and \(\mathcal G_n\) (gaps) are the canonical names for
these spaces; the spatial sections integrate over \(\mathsf S_n\) by pulling back along the transform to
\(\mathsf T_n\), or work directly in \(\mathcal G_n\).

\subsubsection{2. The iid uniform baseline}

\refstepcounter{thmcnt}\label{thm:iid-uniform-order-statistics}\textbf{Theorem\nobreakspace{}\thethmcnt{}.1 (iid uniform order statistics; David \& Nagaraja 2003, §2.2).} Let \(U_1, \ldots, U_n\)
be independent and uniform on \([0, s]\). Their order statistics have the constant joint density
\[\resizebox{\ifdim\width>\linewidth\linewidth\else\width\fi}{!}{$\displaystyle f_{U_{(1)}, \ldots, U_{(n)}}(u_1, \ldots, u_n) \;=\; \frac{n!}{s^{\,n}} \qquad\text{on } \Lambda_n(s);$}\]
the normalised spacings \(\bigl(U_{(i)} - U_{(i-1)}\bigr)/s\) are \(\mathrm{Dirichlet}(1, \ldots, 1)\)
(uniform on the unit simplex), and \(U_{(i)}/s \sim \mathrm{Beta}(i, n-i+1)\).

\refstepcounter{thmcnt}\label{cor:exponential-representation-of-spacings}\textbf{Corollary\nobreakspace{}\thethmcnt{}.1a (exponential representation of the spacings; Pyke 1965).} With
\(E_1, \ldots, E_{n+1}\) iid \(\mathrm{Exp}(1)\) and \(T = \sum_j E_j\), the uniform spacings satisfy
\(\bigl(U_{(1)}, U_{(2)}-U_{(1)}, \ldots, s - U_{(n)}\bigr) \stackrel{d}{=} s\,(E_1, \ldots, E_{n+1})/T\).
This carries the smallest and largest spacing to functionals of an iid exponential vector, the classical
route to the extreme spacings of a uniform sample.

This flat law is the \textbf{baseline} the UCP is read against. It lives on the full simplex \(\Lambda_n(s)\),
where the points do not interact; the UCP places its mass on the smaller hard-core support \(\mathsf S_n\).
Whether the flat density survives the hard-core and jamming constraints, or is deformed by them, is the
question the order-statistics section resolves; the exponential representation above is the baseline
against which the gap-order-statistic and extreme-gap sections measure the jammed spacings.

\subsubsection{3. The Tonks equilibrium control}

The equilibrium counterpart of the UCP --- the control the results sections measure the UCP against
--- is the Tonks process, the hard-rod gas of Tonks (1936) in equilibrium rather than filled
sequentially.

\refstepcounter{thmcnt}\label{def:tonks-process-the-jointly-uniform-hard-rod}\textbf{Definition\nobreakspace{}\thethmcnt{} (Tonks process; the jointly uniform hard-rod law).} The \textbf{Tonks process} \(\mathbb H_n\),
written \(\mathbb H\) for the hard-rod gas it models, places \(n\)
unit cars on \([0, s]\) \emph{jointly uniformly} subject to non-overlap: the ordered positions are uniform on
the hard-core polytope \(\mathsf S_n\), with the constant density
\[\resizebox{\ifdim\width>\linewidth\linewidth\else\width\fi}{!}{$\displaystyle f_{\mathbb H}(\mathbf x) \;=\; \frac{n!}{(s-n)^n} \qquad\text{on } \mathsf S_n,$}\]
normalised by Proposition \ref{prop:tonks-transform}.

\refstepcounter{thmcnt}\label{prop:tonks-order-statistics-are-transformed}\textbf{Proposition\nobreakspace{}\thethmcnt{} (Tonks order statistics are transformed uniform).} Under the Tonks transform
(Proposition \ref{prop:tonks-transform}) the Tonks order statistics are those of \(n\) iid uniform points on \([0, s-n]\).
Consequently every uniform order-statistic law of Theorem \ref{thm:iid-uniform-order-statistics} transports to Tonks by
\(z_i = x_{(i)} - (i-1)\): the Tonks spacings are \(\mathrm{Dirichlet}(1, \ldots, 1)\) on \([0, s-n]\) and
\(z_{(i)}/(s-n) \sim \mathrm{Beta}(i, n-i+1)\).

\emph{Proof.} \(f_{\mathbb H}\) is uniform on \(\mathsf S_n\) (Definition \ref{def:tonks-process-the-jointly-uniform-hard-rod}), and Proposition \ref{prop:tonks-transform} carries it,
volume-preservingly, to the uniform law on \(\mathsf T_n = \Lambda_n(s-n)\) --- the order statistics of
\(n\) iid uniform points on \([0, s-n]\); the Dirichlet and Beta laws are the facts of Theorem \ref{thm:iid-uniform-order-statistics} read on
the reduced interval. \(\quad\blacksquare\)

Tonks and the UCP share the support \(\mathsf S_n\) but not the law: Tonks is uniform there, so its order
statistics reduce to the baseline of §2 on a shorter interval. Whether the sequential UCP is likewise
uniform, or tilts the mass toward the tight (jamming) configurations, is the contrast the
order-statistics section settles.

\subsubsection{4. Order statistics of a dependent parent}

For \textbf{independent} parents the order-statistic density is a permanent: if \(X_1, \ldots, X_n\) are
independent with densities \(f_1, \ldots, f_n\), then
\[\resizebox{\ifdim\width>\linewidth\linewidth\else\width\fi}{!}{$\displaystyle f_{X_{(1)}, \ldots, X_{(n)}}(x_1 < \cdots < x_n) \;=\; \operatorname{per}\bigl[\, f_j(x_i) \,\bigr]
\;=\; \sum_{\pi \in \mathfrak S_n} \prod_{i=1}^{n} f_{\pi(i)}(x_i)$}\]
(Vaughan \& Venables 1972; Bapat \& Beg 1989; David \& Nagaraja 2003, §5.2). The permanent form
requires the coordinates to be \textbf{independent} --- the iid case collapses it to one term, the
non-identical case keeps the permanent.

The UCP gaps are \textbf{not} independent: the total length is conserved, \(\sum_{i} g_i = s - n\). No permanent
form applies to them; the order-statistic density of a general, possibly dependent, parent is instead
the full symmetrization.

\refstepcounter{thmcnt}\label{thm:dependent-parent-order-statistics}\textbf{Theorem\nobreakspace{}\thethmcnt{}.2 (order statistics of a dependent parent; David \& Nagaraja 2003, §6.1).} Let
\(\mathbf Z = (Z_1, \ldots, Z_m)\) be absolutely continuous with joint density \(f_{\mathbf Z}\). Its order
statistics have density
\[\resizebox{\ifdim\width>\linewidth\linewidth\else\width\fi}{!}{$\displaystyle f_{Z_{(1)}, \ldots, Z_{(m)}}(y_1 \le \cdots \le y_m) \;=\; \sum_{\pi \in \mathfrak S_m}
f_{\mathbf Z}\bigl(y_{\pi(1)}, \ldots, y_{\pi(m)}\bigr).$}\]

\emph{Proof.} The sort map \(\mathbf z \mapsto (z_{(1)}, \ldots, z_{(m)})\) is \(m!\)-to-one. The open sort cell
\(\{z_{\pi(1)} < \cdots < z_{\pi(m)}\}\) is carried onto the ordered wedge by the coordinate permutation
\(\pi\), a measure-preserving bijection; pushing \(f_{\mathbf Z}\) forward on each cell and summing over the
\(m!\) cells gives the density. \(\quad\blacksquare\)

When \(f_{\mathbf Z}\) factors, the sum reduces to the permanent above; for the dependent UCP gaps it does
not factor, so this symmetrization --- not a permanent --- is the tool the order-statistics section applies.
Whether its \(n!\)-term sum collapses to a tractable form is that section's question.

\pandocbounded{\includegraphics[keepaspectratio]{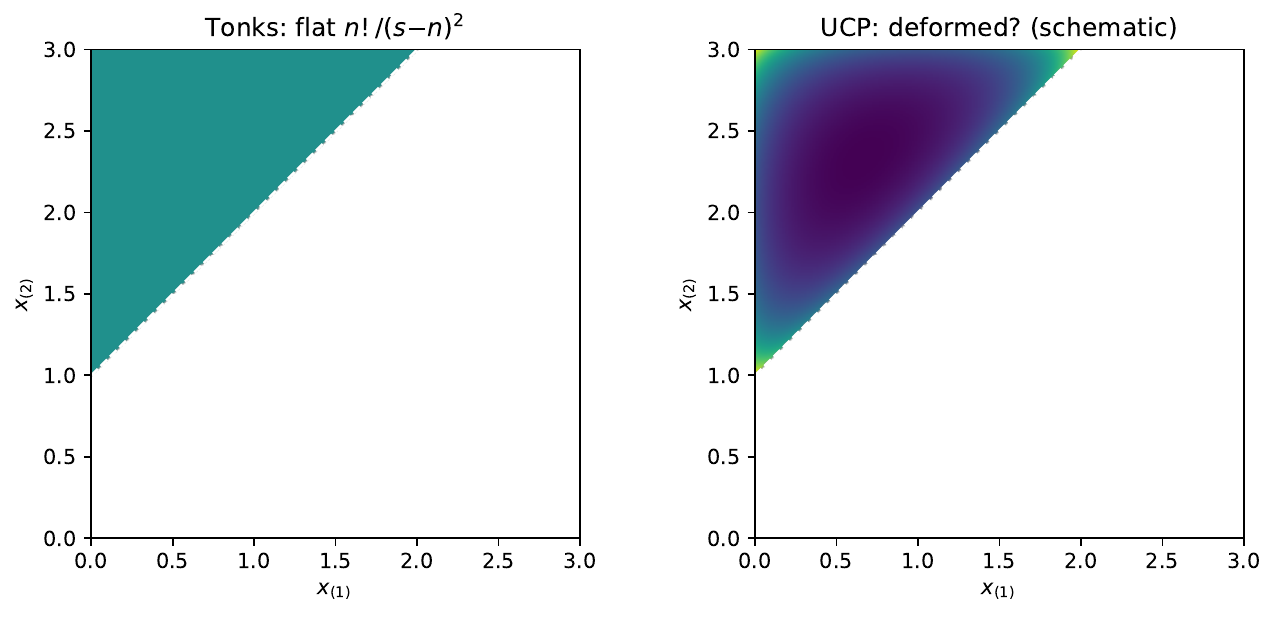}}

\textbf{Figure 1.} Order statistics of two unit cars on a segment of length \(s = 4\), over the hard-core
simplex \(\mathsf S_2 = \{0 \le x_{(1)},\ x_{(1)} + 1 \le x_{(2)} \le 3\}\) (Definition \ref{def:hard-core-polytope}). \emph{Left:}
the Tonks equilibrium density is flat, \(n!/(s-n)^n\) (Definition \ref{def:tonks-process-the-jointly-uniform-hard-rod}). \emph{Right (schematic):} the UCP
density on the same support --- flat versus deformed is the contrast the order-statistics section
resolves.

\subsubsection{Prior art for the gap and order-statistic laws}

Four strands of earlier work touch the densities computed here. The interior-gap density of the
saturated process (Rényi 1958; Dvoretzky--Robbins 1964; Coffman--Flatto--Jelenković 2000) is
recovered as the \(s\to\infty\) limit of the finite-\(s\) gap marginals of \ref{sec:order-stats-cells}, and the
Bonnier--Boyer--Viot (1994) pair correlation, with its super-exponential decay, is the two-gap
marginal of the cell density at jamming. The Bapat--Beg permanent representation of an
order-statistic marginal (David--Nagaraja 2003) does not apply here: it presumes independent
parents, whereas the free lengths couple every coordinate to all the others, and that coupling is
what forces a decomposition into cells in place of a product-form permanent. The general
David--Nagaraja formula for order statistics of dependent parents does apply, and it reproduces \(f_{\mathrm{UCP}}\) term by term. The remaining comparable results
(Coffman--Flatto--Jelenković 1995 on the vacant interval; Araújo--Cadilhe 2006; Clay--Simányi 2014;
Gerin 2015; Mackey--Sullivan 2016) are univariate, single-gap, or simulation-based. The joint
density is known in the form of an average over temporal orders of the reciprocal free-length
product (Talbot, Tarjus, Van Tassel and Viot 2000, Eq. (22); the same kernel appears as a Janossy
density in van Lieshout 2006), but it is stated there formally and in arbitrary dimension; what is
supplied here is its resolution in one dimension at finite \((s,n)\). The nearest exact statement is
Iwasa--Fukuda (2008), whose saturation-count law is the Tonks equilibrium count rather than the
sequential one.

The construction that these results bound is carried out in \ref{sec:joint-order-stats}; the piece count and the symbolic evaluations are in \ref{sec:order-stats-cells}.

\subsection{Background: the Hopf-algebraic machinery and the parking integrals}

\label{sec:background-hopf}

\emph{The classical algebra the aggregate and order-statistics sections invoke --- the parking integrals
and the shuffle Hopf algebra they live in, the Aomoto--Goncharov period theorem and Goncharov's
symbol, and the combinatorial Hopf algebra FQSym with its Boolean-lattice antipode.}

\subsubsection{Scope}

The aggregate and order-statistics sections reduce every moment and density to one family of integrals and analyse them with classical Hopf-algebraic machinery. This section states that machinery as the tools those sections invoke: the parking integrals (§1), the shuffle Hopf algebra and coradical raising (§2), the Aomoto--Goncharov period theorem (§3), Goncharov's coaction and symbol (§4), FQSym and the Boolean-lattice antipode (§5), and the bridge \(\Phi\) (§6). The configuration-space geometry these integrals run over is that of \ref{sec:background-order-statistics}, and the first-car split that generates them is the strong Markov factorisation of \ref{sec:background-ucp}; this section cites across and does not duplicate them. The parking-specific theorems the program proves with these tools --- linear reducibility and the weight grading --- are stated and proved in the aggregate sections; here we supply only what they rest on.

\subsubsection{1. The parking integrals}

The parking integrals are the analytic object the rest of this section grades. Fix a jammed configuration of \(n\) unit cars with gap coordinates \(\mathbf g=(g_0,\dots,g_n)\), ranging over the jammed sub-region \(J_n\) of the gap simplex (\ref{sec:background-order-statistics}, Definition \ref{def:gap-simplex}). Let \(\sigma\) range over the \(n!\) temporal orders of the cars, and let \(A_k\) denote the set of cars already placed when the \(k\)-th attaches. A \textbf{jamming-bitstring cell} is a region of \(J_n\) on which each gap has a fixed sign relative to \(1\) (the bitstring records, per gap, whether \(g_j<1\)). On one such cell the joint density of a fixed temporal order is a product of reciprocal \textbf{free lengths},
\[\resizebox{\ifdim\width>\linewidth\linewidth\else\width\fi}{!}{$\displaystyle g_\sigma=\prod_{k}\frac{1}{\ell_k^\sigma},\qquad
\ell_k^\sigma(\mathbf g)=(s-|A_k|)-\sum_{j}\!\min(g_j,1),$}\]
so \(\ell_k^\sigma\) is the free space available to car \(k\); within one cell each \(\min(g_j,1)\) resolves to either \(g_j\) or \(1\), so every \(\ell_k^\sigma\) is \textbf{affine in \((\mathbf g,s)\) with coefficients in \(\{0,\pm1\}\) and integer constant term}.

\refstepcounter{thmcnt}\label{def:parking-integral}\textbf{Definition\nobreakspace{}\thethmcnt{} (parking integral).} A \textbf{parking integral} is any integral of such a density over \(J_n\) (or over a coordinate slice of it), together with the one-dimensional \textbf{convolutions}
\[\resizebox{\ifdim\width>\linewidth\linewidth\else\width\fi}{!}{$\displaystyle \frac{1}{s-1}\int_0^{s-1} M_b(x)\,M_c(s-1-x)\,dx$}\]
of the count-moment functions \(M_b(x)=\mathbb E[N(x)^b]\), which assemble the moments from lower ones.

These are exactly \textbf{Aomoto polylogarithms}: iterated integrals of \(d\log(\text{affine})\) forms over a polytope (Aomoto 1982; Goncharov, \emph{Aomoto polylogarithms and the coproduct}).

\refstepcounter{thmcnt}\label{def:weight-and-the-integer-delay-alphabet}\textbf{Definition\nobreakspace{}\thethmcnt{} (weight and the integer delay alphabet).} The \textbf{weight} of such an integral is the number of integrations, equivalently the polylogarithm order (\(\log\) at weight \(1\), \(\mathrm{Li}_2\) at weight \(2\), and so on). The \textbf{integer delay alphabet} is the set of one-forms \(\{\,ds/(s-k):k\in\mathbb Z\,\}\); the unit car length forces every breakpoint to an integer, so this is the only alphabet that appears. We write \(\mathcal H_{\le w}\) for the \(\mathbb Q[s]\)-module of hyperlogarithms of weight at most \(w\) over this alphabet, together with the weight-\(\le w\) constants (\(\pi^2\) at weight \(2\), \(\zeta_3\) at weight \(3\)).

A representative weight-\(0\) element is \(3s-5\); a weight-\(1\) element is \(7s-17-4\log(s-2)\); weight-\(2\) elements include \(\log(s-2)\log(s-3)\), \(\mathrm{Li}_2(3-s)\), and \(\pi^2\).

\subsubsection{2. The shuffle Hopf algebra and coradical raising}

Iterated integrals over a fixed alphabet form a \textbf{shuffle Hopf algebra}: the product of two of them is the sum over shuffles of their letter words, the coproduct is deconcatenation, and the coradical filtration coincides with the weight. The single fact we use downstream is \textbf{coradical raising}: one integration appends one letter and therefore raises the weight by exactly one. This is the abstract content of the informal statement that ``each interval crossing raises the polylog order by one.'' A separate consequence of the shuffle structure, used in the period theorem (§3), is that a product of two forms of weights \(w_1\) and \(w_2\) has weight \(w_1+w_2\) and no more.

\subsubsection{3. The period theorem: linearly reducible Aomoto polylogarithms are weight-graded}

The analytic input the aggregate and order-statistics sections draw from this machinery is a period theorem. An \textbf{Aomoto polylogarithm} is an iterated integral of \(d\log(\text{affine})\) forms over a polytope in \(\mathbb R^m\); the parking integrals of §1 are of this type, since \(g_\sigma=\prod_k 1/\ell_k^\sigma\) has only the affine singularities \(\{\ell_k^\sigma=0\}\).

\refstepcounter{thmcnt}\label{thm:aomotogoncharov-period-theorem-affine-case}\textbf{Theorem\nobreakspace{}\thethmcnt{} (Aomoto--Goncharov period theorem, \(d\log\)-affine case).} Let \(I=\int_P\omega\) be an Aomoto polylogarithm in \(m\) integration variables whose singular hyperplane arrangement is \textbf{linearly reducible} (Brown). Then \(I\) is a multiple polylogarithm of \textbf{weight equal to \(m\)}, the number of integrations, over the alphabet of the arrangement's affine letters; no algebraic (square-root) or elliptic letters arise.

\emph{Classical justification.} Each integration appends one \(d\log\) letter and raises the weight by exactly one (coradical raising, §2). Linear reducibility furnishes an elimination order under which Brown's polynomial reduction keeps every letter affine, so no irreducible quadric --- hence no square root --- can appear (Brown 2009; Panzer 2015); weight \(=\) number of integrations is then the Chen--Aomoto--Goncharov grading (Chen 1977; Aomoto 1982; Goncharov 2005).

The theorem is a \textbf{conditional tool}: it grades an integral by weight \emph{provided} the arrangement is linearly reducible, which is not automatic --- a generic arrangement produces algebraic or elliptic letters and is not so graded. Whether the parking arrangement is reducible, and hence whether this grading may be invoked, is a parking-specific question discharged in the aggregate section.

\subsubsection{4. The coaction and the symbol}

Goncharov's coaction, and its maximal iteration the \textbf{symbol}, send a weight-\(w\) hyperlogarithm to a tensor of its \(d\log\) letters and reduce an identity between such functions to linear algebra on tensors. The symbol's use as evidence is limited by its \textbf{kernel}, which annihilates \(\zeta_2\), \(i\pi\), and all products of lower-weight constants. Consequently \(\mathcal S(R)=0\) reduces the identity \(R=0\) to a lower-weight constant but does not establish it; closing that gap needs a separate step, either a numerical evaluation (which counts as a verification) or an exact functional equation (which counts as a proof).

\subsubsection{5. FQSym and the antipode of the Boolean lattice}

The analytic Hopf algebra of §§1--4 is matched by a combinatorial Hopf algebra on permutations, whose product is the shifted shuffle and whose antipode performs the inclusion--exclusion over subsets used in §6.1. Its structure is recorded in §5.1 and that antipode in §5.2.

\paragraph{5.1 FQSym (Malvenuto--Reutenauer 1995)}

\(\mathbf{FQSym}=\bigoplus_{n\ge0}\operatorname{span}\{F_\sigma:\sigma\in\mathfrak S_n\}\) has a basis indexed by \textbf{all permutations}, which one reads as the maximal chains of the Boolean lattice \(B_n\). The product is the shifted shuffle of one-line words, so \(F_1\cdot F_1=F_{12}+F_{21}\) and \(F_{12}\cdot F_1=F_{123}+F_{132}+F_{312}\). The coproduct deconcatenates and standardizes, so \(\Delta F_{132}=1\otimes F_{132}+F_1\otimes F_{21}+F_{12}\otimes F_1+F_{132}\otimes1\). The algebra is non-commutative, non-cocommutative, and self-dual; it is the free lift of \(\mathrm{QSym}\) and surjects onto the Loday--Ronco algebra of planar binary trees (Loday--Ronco 1998). Both displayed identities are the two operations applied to short words: the product ranges over the shuffles of \(12\) with the shifted letter \(3\), and the coproduct over the four cut points of \(132\), each factor standardized (Malvenuto and Reutenauer 1995).

\paragraph{5.2 The antipode}

\refstepcounter{thmcnt}\label{def:antipode}\textbf{Definition\nobreakspace{}\thethmcnt{} (antipode).} In a bialgebra with Sweedler coproduct \(\Delta(x)=\sum x_{(1)}\otimes x_{(2)}\), the \textbf{antipode} \(S\) is the map satisfying \(\sum S(x_{(1)})\,x_{(2)}=\varepsilon(x)\,1=\sum x_{(1)}\,S(x_{(2)})\); equivalently it is the convolution inverse of the identity, \(S*\mathrm{id}=\varepsilon(\cdot)\,1=\mathrm{id}*S\). It generalizes group inversion, since \(S(h)=h^{-1}\) for a group element \(h\).

\refstepcounter{thmcnt}\label{thm:boolean-antipode-is-mobius-inversion}\textbf{Theorem\nobreakspace{}\thethmcnt{} (the Boolean antipode is Möbius inversion; Joni--Rota 1979).} On the Boolean lattice \(B_n\), with its incidence Hopf algebra, the antipode is the Möbius function \(\mu(A,T)=(-1)^{|T\setminus A|}\), and the identity \(\mu*\zeta=\delta\) is inclusion--exclusion over subsets.

\subsubsection{\texorpdfstring{6. The bridge \(\Phi\)}{6. The bridge \textbackslash Phi}}

The analytic Hopf algebra of §§1--4 and the combinatorial one of §5 are distinct objects, joined by a bridge \(\Phi\). The bridge is the Picard iteration of the parking Volterra recursion, which coincides with the Chen iterated-integral series of the cell density. Under \(\Phi\) the first-car recursion is the FQSym coproduct. The multiplicative structure does not transport: the weight map \(F_\sigma\mapsto\prod_k 1/\ell_k^\sigma\) is \textbf{not} a character on \(\mathbf{FQSym}\), refuted by Counterexample \texttt{cex:\allowbreak bridge-\allowbreak not-\allowbreak a-\allowbreak character} of \ref{sec:joint-order-stats}, where the reason is also given. So FQSym indexes the temporal orders and matches the first-car recursion to the coproduct, and supplies no product rule for the weights; the period theorem (Theorem \ref{thm:aomotogoncharov-period-theorem-affine-case}) and the Boolean antipode (Theorem \ref{thm:boolean-antipode-is-mobius-inversion}) are the established statements this section supplies. The Chen series of §1 and the algebra \(\mathbf{FQSym}\) of §5 are different objects that \(\Phi\) relates.

\paragraph{6.1 The antipode passes between the two decompositions of the joint density}

The order-statistics section derives two decompositions of the joint density, trading algebraic structure against computational cost; one pairs with the shuffle algebra of §§1--4, the other with the antipode of §5.2.

\begin{itemize}
\tightlist
\item
  \textbf{Temporal-order form.} \(f_{\mathrm{UCP}}=\sum_{\sigma\in\mathfrak S_n}\prod_k 1/\ell_k^\sigma\), the symmetrization over the \(n!\) temporal orders; it carries the temporal-order algebra and has \(n!\) terms.
\item
  \textbf{Cell form.} The geometric bit-cell decomposition, on which a Held--Karp subset dynamic program collapses the \(n!\) sum to \(O(2^n n)\), at the cost of cutting across the temporal-order algebra. (An \emph{interval} dynamic program would collapse it further, but the free lengths are not interval-determined, so no such collapse is available --- see \ref{sec:properties-order-stats} §7.)
\end{itemize}

The passage between them is not the antipode. The free length depends on the \emph{set} of cars already
placed and not on their order, so temporal orders that share a placed set share a subproblem; the
\(n!\)-term sum collapses to \(O(2^n)\) by the unsigned recursion
\(W(S)=\sum_{a\in S}W(S\setminus a)/L(S\setminus a)\), which is memoisation. A subset dynamic program of
that kind carries no signs and is not an inclusion--exclusion: Held--Karp and Ryser are both \(O(2^n)\)
over subsets but are different algorithms, and only the latter is Möbius inversion. Ryser's formula
needs a matrix, one factor per row depending on a single index, whereas the parking weight \(1/L(S_k)\)
depends on the whole placed set --- the same set-dependence that blocks a product-form permanent (\ref{sec:properties-order-stats}). A signed sum over subsets does not evaluate the
density: the
temporal-order sum and the dynamic program agree exactly in rational arithmetic while the signed sum
disagrees in sign and magnitude. The bridge \(\Phi\) and the integral-equation machinery pair with the
temporal-order form; the cell form is handled by the set recursion. The antipode of §5.2 is quoted here
as the Boolean-lattice fact it is, and is not applied to the density. A permutation-invariant functional converts between the two by \(\sum_i h(X_i)=\sum_i h(X_{(i)})\); an order-dependent functional has no such identity and must be symmetrized.

\subsubsection{7. The algebraic picture of an arriving car}

An arriving car does two things at once. It adds one more object to the configuration, and it cuts
the region it lands in into two sub-segments that fill independently of each other. A product and a
compatible coproduct are the defining data of a Hopf algebra, and the sequential density, one
uniform factor per car, is a character on it. This section fixes that dictionary, names the three
Hopf algebras the process lives in, and computes the character where it can be computed in closed
form: its image under the symbol map is a single staircase word, which pins the transcendental
weight of the fully marginalized density at every \(k\).

\textbf{Algebraic vocabulary.} The terms below are used throughout the section; each is given with a
minimal example and with its meaning for the parking process.

\begin{itemize}
\tightlist
\item
  \textbf{Algebra, a product} \(m\colon A\otimes A\to A\): multiply two things into one. \emph{Words under concatenation:} \(ab\cdot cd=abcd\). \emph{Here:} place these cars, then those.
\item
  \textbf{Coproduct, splitting} \(\Delta\colon A\to A\otimes A\): one element \(\mapsto\) the sum of all (left part)\(\otimes\)(right part) cuts. \emph{Deconcatenation:} \(\Delta(abc)=\varnothing\otimes abc+a\otimes bc+ab\otimes c+abc\otimes\varnothing\). A space with a coproduct is a \textbf{coalgebra}. For iterated integrals: split the path into (first stretch)\(\otimes\)(rest).
\item
  \textbf{Character, a multiplicative measurement} \(\chi\colon A\to\) numbers/functions, \(\chi(xy)=\chi(x)\chi(y)\). \emph{Evaluate a polynomial at a point.} \emph{Here:} the \textbf{temporal-order density} \(\prod_k 1/\ell_k\), which turns ``combine configurations'' into ``multiply densities''.
\item
  \textbf{Coaction, a coproduct into another space} \(\rho\colon M\to A\otimes M\): decompose \(M\) with coefficients in \(A\). \emph{Here:} weight-\(w\) polylogs coact (Goncharov) with lower-weight ones, \(\Delta\,\mathrm{Li}_n=(\text{derivative pieces})\otimes(\text{remainder})\); ``peel one layer,'' \(w\to(w{-}1)+1\).
\item
  \textbf{Hopf algebra}, product \(+\) coproduct (compatible) \(+\) \textbf{antipode} \(S\) (an algebraic inverse/reversal). Our three: words/shuffle \(H_{\sqcup\!\sqcup}\), rooted-trees/grafting \(H_{\mathrm{CK}}\) (grafting \(B_+\) \(=\) ``one arriving car''), polylogs/symbol \(H_{\mathrm{MPL}}\).
\item
  \textbf{Symbol, the coaction run to the bottom}: keep splitting until every slot is weight 1 (a bare \(\log\) \(=\) a ``letter''), \(\mathrm{Li}_w\mapsto \text{letter}\otimes\cdots\otimes\text{letter}\) (\(w\) slots). The atomized fingerprint; constants/\(i\pi\)/products die under full splitting. This is what collapsed \(M(s)\) to the staircase.
\item
  \textbf{Weight (grading)}, product \emph{adds} weight, coaction moves \emph{one} unit. Rational \(=0\), \(\log=1\), \(\mathrm{Li}_2=2,\dots\). \emph{Here:} joint density \(=0\) (bottom), \(M(s)\) on \([k,k{+}1)\) \(=k{-}2\) (top).
\end{itemize}

\textbf{In one line.} The product puts configurations together (insert a car); the coproduct and coaction
take them apart (marginalize); the character is the density, which turns ``together'' into ``multiply'';
and the symbol is the coaction run to the bottom. References are collected at the end of the
section.

\paragraph{7.1 The three Hopf algebras an arriving car generates}

One arriving car generates three linked Hopf algebras, and those three are the algebra behind the
subset recursion, the weight ladder, and the higher-weight reductions.

\paragraph{1. What a car does, algebraically}

A car landing at \(X\) in a host interval \(I=[a,b]\) makes two \textbf{dual} moves:
- \textbf{graft and split (coproduct side):} it becomes a node and cuts \(I\) into two sub-segments \(I_L=[a,X]\) and \(I_R=[X{+}1,b]\) that fill independently of each other, by the strong Markov property of the process;
- \textbf{interleave (product side):} the global temporal order is a \textbf{shuffle} of the two sub-segments' orders.

Splitting is a \textbf{coproduct}; interleaving is the \textbf{shuffle product} \(\sqcup\!\sqcup\).

\paragraph{2. The three Hopf algebras}

\textbf{(a) Shuffle algebra of attachment orders \(H_{\sqcup\!\sqcup}\).} Words \(=\) temporal orders; product \(=\) shuffle (interleavings, with binomial multiplicities); coproduct \(=\) deconcatenation. Graded by \(n\), commutative, connected \(\Rightarrow\) Hopf. The \(n!\) orders on a region are the shuffle \(\sqcup\!\sqcup\) of the sub-segments' orders.

\textbf{(b) Connes--Kreimer rooted-tree algebra \(H_{CK}\).} The \textbf{attachment tree} (each car a node, two child sub-segments) lives here; the car is the grafting operator \(B_+\), which roots onto the forest of sub-segments, and the coproduct is the sum over admissible cuts. Conditioning on the first car placed is exactly that cut.

\textbf{(c) Goncharov--Brown MPL algebra \(H_{\mathrm{MPL}}\) (the symbol / coaction).} Densities and marginals are multiple polylogarithms; iterated integrals carry the shuffle product, the \textbf{weight grading}, and the \textbf{coproduct \(\Delta\)} (symbol/coaction) that breaks a weight-\(w\) MPL into lower-weight tensors.

The \textbf{density is a character} \(\chi:H_{CK}\to H_{\mathrm{MPL}}\), sending a tree to its sub-segment integral; a car raises weight by one through the integration \(\int(\cdot)\,d\log(u-1)\), which is a coalgebra map, and that is why marginalization depth equals weight.

\paragraph{3. The objects of this section, restated in that language}

\begin{itemize}
\item
  The interval recursion \(\;D[i,j]=\sum_{c}\dfrac{1}{\ell_{[i,j]}(c)}\binom{j-i}{c-i}\,D[i,c{-}1]\,D[c{+}1,j]\;\) \textbf{is} a character recursion: \(D[i,c{-}1]\otimes D[c{+}1,j]\) is the coproduct, \(\binom{j-i}{c-i}\) is the shuffle count, \(\sum_c\) is \(B_+\). Being a character means precisely that values \textbf{factor through the coproduct}.
\item
  \textbf{That is why it does not compute \(f_{\mathrm{UCP}}\).} The free length an arriving car divides by is the total measure of the free region left by the whole placed set; after a split it is \(\ell_{\mathrm{left}}+\ell_{\mathrm{right}}\), and \(1/(\ell_{\mathrm{left}}+\ell_{\mathrm{right}})\) does not factor. So \(f_{\mathrm{UCP}}\) is not a character value for this coproduct, and the interval recursion, which computes one, computes something else. Numerically, already at \(n=2\): \(s=8\), \(\mathbf x=(2,5)\) gives \(1/14\) against the true \(2/35\) (\ref{sec:properties-order-stats} §7). The Hopf reading is therefore an \textbf{obstruction theorem}, not a speedup: it says exactly which functionals the sub-segment coproduct can carry, and the density is not among them.
\item
  What \emph{does} compute \(f_{\mathrm{UCP}}\) is the \textbf{last-car subset recursion} on the Boolean lattice, at \(O(2^n n)\): it groups temporal orders by their final element, for which the free length \emph{is} determined by the set already present. That is the \(\zeta\)-side chain sum of \(B_n\), not a character recursion for the sub-segment coproduct, which is why it escapes the obstruction, and why it costs \(2^n\) rather than \(\mathrm{poly}(n)\).
\item
  The relabelling by \(\sigma\), together with the reflection of the segment, is the antipode-level symmetry of \(H_{\sqcup\!\sqcup}\).
\end{itemize}

\paragraph{4. What the structure buys computationally}

\begin{enumerate}
\def\labelenumi{\arabic{enumi}.}
\tightlist
\item
  \textbf{Factorization, and its limit:} the coproduct \emph{is} the sub-segment recursion, so any functional that factors through it is cheap per cell. The joint density does not factor through it (above), so the cheap route is closed for the density and open for the aggregates, which are additive and do factor.
\item
  \textbf{Weight grading for free:} integration is a coalgebra map, so the rational \(\to\) log \(\to\) dilog \(\to\cdots\) ladder is the grading, no case analysis.
\item
  \textbf{Identities become linear algebra:} \(\Delta\) (the symbol) turns polylog functional equations into \emph{linear} relations on lower-weight tensors. This is exactly the missing tool for the open \([5,6),[6,7)\) clean forms: the spurious \(i\pi\), \texttt{delta}, and \texttt{signum(Im\textbackslash{},s)} terms met in the aggregate-moment section lie in the \textbf{kernel of the symbol} and vanish modulo the coaction, and the fibration basis of a hyperlogarithm integrator is one realization of that coaction.
\item
  \textbf{Structural cross-checks:} a character into a \emph{graded} Hopf algebra makes identities like \(M(s)=\sum_n n\Pr[N{=}n]\) and the moment cascade structural, not coincidental.
\end{enumerate}

\paragraph{5. Scope}

The shuffle structure of temporal orders and the polylogarithm Hopf algebra are standard (Goncharov, Brown), and the Connes--Kreimer reading of the attachment tree is exact. We compute the character on the ladder trees that carry the fully marginalized facet, where the symbol proves the staircase form. Writing the character explicitly on a general attachment tree, and with it the higher-weight reductions for the multivariable marginals, remains open.

\paragraph{7.2 The dictionary: temporal-order density, joint density, character, shuffle, coaction}

{\footnotesize\begin{longtable}[]{@{}
  >{\raggedright\arraybackslash}p{\dimexpr 0.5000\linewidth-2\tabcolsep\relax}
  >{\raggedright\arraybackslash}p{\dimexpr 0.5000\linewidth-2\tabcolsep\relax}@{}}
\toprule\noalign{}
\begin{minipage}[b]{\linewidth}\raggedright
probabilistic object
\end{minipage} & \begin{minipage}[b]{\linewidth}\raggedright
algebraic object
\end{minipage} \\
\midrule\noalign{}
\endhead
\bottomrule\noalign{}
\endlastfoot
one arriving car & grafting \(B_+\) in \(H_{\mathrm{CK}}\), depth \(+1\) \\
temporal order \(\sigma\) (position against arrival time) & a \textbf{word} in \(H_{\sqcup\!\sqcup}\) (ladder tree in \(H_{\mathrm{CK}}\)) \\
temporal-order density \(\prod_k 1/\ell_k^{(\sigma)}\) & the \textbf{character \(\chi\)} on that word (comultiplicative for deconcatenation) \\
sorting \(=\) sum over temporal orders & the \textbf{shuffle} \(\sqcup\!\sqcup\) (symmetrize over \(S_n\)) \\
order-stat joint density \(f_{\mathrm{UCP}}\) & \(\chi(e_1\sqcup\!\sqcup\cdots\sqcup\!\sqcup e_n)=\sum_\sigma \chi(\mathrm{word}_\sigma)\) \\
marginalize one coordinate & one \textbf{coaction \(\Delta\)}, weight \(+1\) \\
jamming bitstring \(\mathbf b\) (region \(\mathcal R_{\mathbf b}\)) & a \textbf{grading of the target \(H_{\mathrm{MPL}}\)}: which letters \((s{-}j)\) the realization uses \\
\end{longtable}}

\textbf{The joint density.} The temporal-order density is the character \(\chi\): multiplicative along the
sequential coproduct, which is the statement that each car is uniform on the remaining free space.
The order-statistic density is its shuffle-symmetrization
\(f_{\mathrm{UCP}}=\chi(e_1\sqcup\!\!\sqcup\cdots\sqcup\!\!\sqcup e_n)\), and it sits at the bottom of the
weight filtration, rational of weight \(0\), with no polylogarithmic content of its own. The \(n!\)
shuffle terms are the reason the subset recursion of §3 exists: the shuffle is evaluated by the
coproduct recursion rather than by expanding \(n!\) words.

\textbf{Permutation against bitstring.} The character has two sides:
- the \textbf{permutation \(\sigma\) acts on the source} \(H_{\sqcup\!\sqcup}\) (reorder the word); \(\chi\) is \emph{not} shuffle-invariant, \(\ell_k\) depends on order, so the joint density is the projection onto the symmetric part;
- the \textbf{bitstring \(\mathbf b\) grades the target} \(H_{\mathrm{MPL}}\) (the alphabet: which gaps are \(<1\)); it is \emph{not} a permutation.

The character \(\chi\) \textbf{intertwines} them: \(\sigma\) permutes inputs, \(\mathbf b\) selects the output grade, marginalization climbs the weight via \(\Delta\). The fully-marginalized top is \(M(s)\), where the coaction collapses to the single staircase symbol, joint density (weight 0) and \(M(s)\) (weight \(k{-}2\)) are the bottom and top of one filtration.

\subsubsection{Glossary}

\begin{itemize}
\tightlist
\item
  \textbf{grading} --- a decomposition \(\mathcal H=\bigoplus_w\mathcal H_w\) under which the product satisfies \(\mathcal H_{w_1}\mathcal H_{w_2}\subseteq\mathcal H_{w_1+w_2}\).
\item
  \textbf{weight} --- the word length of an iterated integral, equal to the number of integrations and to the polylogarithm order.
\item
  \textbf{coradical raising} --- the fact that one integration appends one letter and raises the weight by one.
\item
  \textbf{Aomoto polylogarithm} --- an iterated integral of \(d\log(\text{affine})\) forms over a polytope; the parking integrals of Definition \ref{def:parking-integral} are of this type.
\item
  \textbf{linear reducibility} --- the property that Brown's polynomial reduction stays linear for some elimination order, which guarantees the integral is a hyperlogarithm.
\item
  \textbf{\(\mathcal H_{\le w}\)} --- the \(\mathbb Q[s]\)-module of hyperlogarithms of weight at most \(w\) over the integer delay alphabet, with the weight-\(\le w\) constants adjoined.
\item
  \textbf{antipode} --- the convolution inverse of the identity in a Hopf algebra; on a poset, the Möbius function.
\item
  \textbf{period} --- an integral of an algebraic function over an algebraic domain in the sense of Kontsevich--Zagier; the multiple polylogarithms of the period theorem (Theorem \ref{thm:aomotogoncharov-period-theorem-affine-case}) are periods.
\end{itemize}

\subsection{Background: integral equations and their extensions}

\label{sec:background-aggregate}

\emph{Rényi's parking integral equation and the moment and correlation cascades that grew from it ---
the second-kind Volterra theory, the Dyson--Schwinger structure, and the transcendence tools
that \ref{sec:aggregate-representations} applies to finite length.}

\subsubsection{Scope}

The aggregate theory descends from a single equation. Rényi (1958) wrote an integral equation for the
mean number of cars a segment holds at saturation --- the jammed count \(N(s)\) of \ref{sec:background-ucp}; later
work extended it along two lines --- a \textbf{moment cascade} (the variance and higher cumulants of the
count) and a \textbf{correlation cascade} (the spatial \(k\)-point densities) --- and supplied the analytic
theory for such equations. This section develops that historical line and lays its groundwork: Rényi's
equation and the second-kind Volterra theory that formalises it (§1), the moment cascade and its
Dyson--Schwinger form (§2), the correlation cascade and its thermodynamic limit (§3), and the
transcendence classes that grade the solutions (§4). The Hopf machinery is in \ref{sec:background-hopf}, the
process and its strong-Markov split in \ref{sec:background-ucp}, and the order-statistic geometry in
\ref{sec:background-order-statistics}; this section cites across and does not duplicate them. No result of the
aggregate section is stated here --- §5 tabulates which tool each subsection applies.

\subsubsection{1. Rényi's integral equation and the second-kind Volterra framework}

The founding object is Rényi's equation for the mean saturated count \(N=N(s)\).

\refstepcounter{thmcnt}\label{thm:parking-integral-equation-renyi-1958}\textbf{Theorem\nobreakspace{}\thethmcnt{} (the parking integral equation; Rényi 1958).} Let \(M(s)=\mathbb E[N(s)]\) be the mean
number of unit cars a segment \([0,s]\) holds at saturation. Then \(M(s)=0\) for \(s<1\), and
\[\resizebox{\ifdim\width>\linewidth\linewidth\else\width\fi}{!}{$\displaystyle M(s)=1+\frac{2}{s-1}\int_0^{s-1}M(u)\,\mathrm du\qquad(s\ge1).$}\]
Its solution satisfies \(M(s)/s\to m\) as \(s\to\infty\), with
\[\resizebox{\ifdim\width>\linewidth\linewidth\else\width\fi}{!}{$\displaystyle m=\int_0^\infty\exp\!\Big(-2\int_0^t\frac{1-e^{-v}}{v}\,\mathrm dv\Big)\,\mathrm dt\approx0.7476,$}\]
the \textbf{Rényi parking constant}.

The equation is a \emph{second-kind linear Volterra equation with unit delay}: the unknown appears both alone and
under an integral whose upper limit trails the argument by one. The general theory of such equations supplies
existence, uniqueness, and a solution operator.

\refstepcounter{thmcnt}\label{thm:existence-uniqueness-and-the-resolvent}\textbf{Theorem\nobreakspace{}\thethmcnt{} (existence, uniqueness, and the resolvent; Gripenberg--Londen--Staffans 1990, Ch.~9;
Brunner 2004, Ch.~2).} A second-kind linear Volterra equation
\[\resizebox{\ifdim\width>\linewidth\linewidth\else\width\fi}{!}{$\displaystyle \varphi(x)=f(x)+\int_0^x K(x,u)\,\varphi(u)\,\mathrm du,$}\]
with \(f\) and the kernel \(K\) locally integrable, has a unique locally integrable solution
\[\resizebox{\ifdim\width>\linewidth\linewidth\else\width\fi}{!}{$\displaystyle \varphi(x)=f(x)+\int_0^x R(x,u)\,f(u)\,\mathrm du,\qquad R=\sum_{r\ge1}K_r,$}\]
where the iterated kernels are \(K_1=K\) and \(K_{r}(x,u)=\int_u^x K(x,t)\,K_{r-1}(t,u)\,\mathrm dt\). The
Volterra operator is quasi-nilpotent, so the \textbf{resolvent} series \(R\) converges on every bounded interval.

This is the machinery the additivity and order-statistic (integral-equation) subsections invoke (§5);
the resolvent of the specific parking kernel is constructed there, not here.

The parking recursion sits beside a broader class of processes, and the contrast locates it. Splitting the
segment at the first car fragments it into two sub-intervals that fill independently, which is the shape of a
\textbf{self-similar fragmentation} in the sense of Bertoin (2006). The parking process is not one. In a
self-similar fragmentation a fragment of size \(m\) splits into pieces \(m x_i\) whose relative sizes are drawn
from a dislocation measure that does not depend on \(m\); here the pieces are \(U/s\) and \((s-1-U)/s\) with \(U\)
uniform on \([0,s-1]\), so the relative law is supported on \([0,(s-1)/s]\) and retains mass \((s-1)/s\), both of
which move with \(s\). A unit car is consumed at every split, and a car length is an absolute quantity, so no
rescaling makes the law scale-invariant. The departure is exactly \(1/s\) in total variation, so the process is
\textbf{asymptotically} self-similar and the finite-\(s\) theory below is precisely the correction. Subdividing an \(n\)-simplex
by random points and recursing on the sub-simplices would give a higher-dimensional Rényi-type
equation --- an extension of the one-dimensional theory beyond this section's scope.

\subsubsection{2. The moment cascade}

Rényi's mean was the first term of a cascade in the moments of the count.

Dvoretzky and Robbins (1964) carried the analysis to the variance, showing it extensive,
\(\operatorname{Var}N(s)=C_2\,s+O(1)\) with the asymptotic constant \(C_2\approx0.0382\); the accompanying
central limit theorem for the count is developed in \ref{sec:background-ucp} §5. Mannion (1964) gave the
first-car split recursion in its standard form and the Tauberian extraction of the higher cumulant-rate
constants \(C_k=\lim_{s\to\infty}\kappa_k(s)/s\), with \(\kappa_k\) the \(k\)-th cumulant.

The cascade has an algebraic source. A combinatorial \textbf{Dyson--Schwinger equation} --- a fixed-point
equation for a generating function under a grafting operator --- has graded coordinates that form a
\textbf{Faà di Bruno subalgebra} of the Hopf algebra of planar trees (Foissy 2008; \ref{sec:background-hopf}); this
is the algebraic reason such a moment cascade is triangular. The section applies the correspondence to the
count, obtaining the triangular moment system; the concrete equation, the per-interval forms, and the
constants \(C_k\) are computed there.

\subsubsection{3. The correlation cascade and its thermodynamic limit}

The spatial analogue of the moment cascade is the hierarchy of correlation functions.

\refstepcounter{thmcnt}\label{def:factorial-moment-densities-daleyvere-jones}\textbf{Definition\nobreakspace{}\thethmcnt{} (factorial-moment densities; Daley--Vere-Jones 2003, §5.4).} For a point process
with points \(X_1,\dots,X_N\), the \(k\)-point factorial-moment (correlation) density is
\[\resizebox{\ifdim\width>\linewidth\linewidth\else\width\fi}{!}{$\displaystyle \rho_k(y_1,\dots,y_k)=\mathbb E\Big[\sum_{i_1\neq\cdots\neq i_k}\ \prod_{l=1}^{k}\delta(X_{i_l}-y_l)\Big],$}\]
the intensity of ordered \(k\)-tuples of distinct points. The family \(\rho_1,\rho_2,\dots\) is the correlation
cascade.

For the saturated parking process the infinite-volume pair correlation \(g_2\) (the limit of \(\rho_2\)) was computed in closed form by
Bonnier, Boyer and Viot (1994). Penrose (2001) established the thermodynamic limit itself: as the segment
grows the saturated configuration converges to a stationary point process, whose correlations do not factorise
--- the limit is a \textbf{non-renewal} process, whose gaps stay correlated at all separations.

The stationary correlation functions obey a \textbf{BBGKY hierarchy}, coupling the \(k\)-point density to the
\((k+1)\)-point one (Evans 1993, §III, for the random-sequential-adsorption case). Unlike the finite-length
recursion, this hierarchy does not truncate beyond the pair, so no density \(g_k\) with \(k\ge3\) is representable
by a finite integral equation; \(g_2\) is the last order with a known closed form. The finite-length
correlation densities, representable by an integral equation via the split, are the section's subject.

\subsubsection{4. Transcendence classes of the cascade solutions}

The cascades are graded by transcendence class, and the classifying tools are classical.

\textbf{Hyperlogarithms} are the iterated integrals over a pole \(\mathrm dt/(t-1)\); an integral stays in this
class when the underlying hyperplane arrangement is \emph{linearly reducible}, Brown's condition (Brown 2009; the
period theorem of \ref{sec:background-hopf}). Whether a count-weighted sum \(\sum_n a_n\,\Pr[N=n]\) admits a finite
equation is a question about the sequence \((a_n)\): it is \textbf{P-recursive} --- a linear recurrence with
polynomial coefficients --- exactly when its generating function is holonomic (Stanley 1980; Zeilberger
1990).
Beyond the algebraic-integrand periods that hyperlogarithms realise lie the \textbf{exponential periods} of
Kontsevich and Zagier (2001), periods of the exponential of an algebraic function.

The section places each cascade solution, finite-length and thermodynamic-limit alike, into one of these
classes; the classification is its result, and the classes are the tools it uses.

\subsubsection{5. Each aggregate subsection continues the cascade}

{\footnotesize\begin{longtable}[]{@{}
  >{\raggedright\arraybackslash}p{\dimexpr 0.3333\linewidth-2\tabcolsep\relax}
  >{\raggedright\arraybackslash}p{\dimexpr 0.3333\linewidth-2\tabcolsep\relax}
  >{\raggedright\arraybackslash}p{\dimexpr 0.3333\linewidth-2\tabcolsep\relax}@{}}
\toprule\noalign{}
\begin{minipage}[b]{\linewidth}\raggedright
aggregate subsection
\end{minipage} & \begin{minipage}[b]{\linewidth}\raggedright
continues
\end{minipage} & \begin{minipage}[b]{\linewidth}\raggedright
tool laid here
\end{minipage} \\
\midrule\noalign{}
\endhead
\bottomrule\noalign{}
\endlastfoot
representations & the split-convolution dictionary & the Volterra / Dyson--Schwinger dichotomy (§1--2) + Hopf (\ref{sec:background-hopf}) \\
additivity & Rényi's mean equation & the second-kind Volterra resolvent (§1) \\
special-cases & the moment cascade & the Dyson--Schwinger / Faà di Bruno structure (§2) \\
order statistics (integral-equation route) & the rank marginals and the joint density & the Volterra hierarchy + linear reducibility (§1, §4) \\
boundary & the correlation cascade & correlation densities (§3) + transcendence classes (§4) \\
\end{longtable}}

The founding equation and the two cascades are Rényi's, Dvoretzky--Robbins's, Mannion's,
Bonnier--Boyer--Viot's, and Penrose's; this section adds the second-kind Volterra theory, the
Dyson--Schwinger reading, and the transcendence classification, and cites the Hopf machinery, the process,
and the order-statistic geometry from the sibling backgrounds.

\section{Order statistics}
\label{sec:order-statistics}

\subsection{\texorpdfstring{Joint order statistics of \(\mathbb U_n\)}{Joint order statistics of \textbackslash mathbb U\_n}}

\label{sec:joint-order-stats}

We will now compute the joint pdf \(f_{\mathrm{UCP}}\) of UCP stopped at \(n\) attachments,
\(\mathbb U_n\). The main difficulty here is that all \(n!\) temporal orders contribute to one spatial
configuration \eqref{eq:temporal-law}, so the order statistics \(X_{(\cdot)}\) are not Markovian, which
forces us to construct their density with techniques distinct from those in the literature
(David--Nagaraja 2003, where the recursion proceeds one rank at a time).

While each temporal order on its own \emph{is} Markov, contributing a product of nearest-neighbour
factors, it is the mixture over orders that breaks the property.

\textbf{Example (\(n=3\), \(s=12\)).} At \(\mathbf x=(2,\tfrac72,5)\), on the cell where consecutive cars clip,
a single temporal order gives \(\partial^2\log g_\sigma/\partial x_1\partial x_3=0\), while the
mixture gives
\[\resizebox{\ifdim\width>\linewidth\linewidth\else\width\fi}{!}{$\displaystyle \frac{\partial^2\log f_{\mathrm{UCP}}}{\partial x_1\,\partial x_3}=\frac{784}{416025}\neq0 .$}\]

\paragraph{The mapping from temporal to spatial pdf}

The temporal left edges \(X_1,\dots,X_n\) and the free length \(\ell_k\) are those of
\ref{sec:problem-statement} and \ref{sec:background-ucp}. Only one abbreviation is new: write
\(\ell_k^\sigma=\ell_k(x_{\sigma(1)},\dots,x_{\sigma(k-1)})\) for the free length seen by the \(k\)-th
car when the cars arrive in the order \(\sigma\).

\refstepcounter{thmcnt}\label{thm:temporal-sum}\textbf{Theorem\nobreakspace{}\thethmcnt{} (the temporal sum).} For every configuration \(\mathbf x\in\mathsf S_n\),
\[\resizebox{\ifdim\width>\linewidth\linewidth\else\width\fi}{!}{$\displaystyle f_{\mathrm{UCP}}(x_1,\dots,x_n)\;=\;\sum_{\sigma\in\mathfrak S_n}\ \prod_{k=1}^{n}\frac1{\ell_k^{\sigma}} .$}\]

\emph{Proof (disintegration of the temporal law).} By the attachment kernel of \ref{sec:background-ucp},
conditionally on \(X_1,\dots,X_{k-1}\) the point \(X_k\) is uniform on the free set \(F_{k-1}\), so
\begin{equation}\label{eq:temporal-law}
p\bigl(X_k\mid X_1,\dots,X_{k-1}\bigr)=\frac{\mathbf 1[X_k\in F_{k-1}]}{\ell_k},
\end{equation}
which by the chain rule becomes
\(p(X_1,\dots,X_n)=\prod_{k=1}^{n}\mathbf 1[X_k\in F_{k-1}]\,\ell_k^{-1}\).

The temporal vectors split into regions on which the bookkeeping is constant. For
\(\sigma\in\mathfrak S_n\) let the \textbf{chamber}
\[\resizebox{\ifdim\width>\linewidth\linewidth\else\width\fi}{!}{$\displaystyle C_\sigma=\{\,\mathbf X:\ X_{\sigma(1)}<\cdots<X_{\sigma(n)}\,\}$}\]
be the set of temporal vectors whose \(k\)-th car has spatial rank \(\sigma(k)\); the \(n!\) chambers partition temporal space. On \(C_\sigma\)
every indicator is \(1\), because a car that arrives can only land where it fits. Each free set
\(F_{k-1}\) is a union of intervals whose endpoints are affine in the cars already placed, so
\(\ell_k=\ell_k^\sigma\) is one fixed affine form throughout the chamber, and the temporal density on
\(C_\sigma\) is exactly \(\prod_k 1/\ell_k^\sigma\).

Sorting maps \(C_\sigma\) onto \(\mathsf S_n\) by the coordinate permutation \(\sigma^{-1}\). A coordinate
permutation is affine with unit Jacobian, so it preserves Lebesgue measure and the density carries
over unchanged. The chambers are disjoint while their images all fill \(\mathsf S_n\), so a sorted
configuration receives one contribution per chamber and the contributions add. \(\qquad\blacksquare\)

\textbf{Prior-art check.} The identity is an instance of
Theorem \ref{thm:dependent-parent-order-statistics} of \ref{sec:background-order-statistics}. Taking the
parent there to be the temporal vector \(\mathbf X\), whose density is the chain-rule product above,
the \(\pi\)-th summand of that theorem becomes \(\prod_k 1/\ell_k^{\pi}\) and the symmetrisation is the
display.

\textbf{Weight map}

Call \(w_\sigma=\prod_{k=1}^{n}1/\ell_k^\sigma\) the \textbf{weight} of the temporal order \(\sigma\), so that
Theorem \ref{thm:temporal-sum} reads \(f_{\mathrm{UCP}}=\sum_\sigma w_\sigma\), and let the \textbf{weight map}
\(\Phi\) send the basis element \(F_\sigma\) of \(\mathbf{FQSym}\) to \(w_\sigma\). It is natural to ask whether
the temporal indexing carries the algebra of \ref{sec:background-hopf} §5, that is whether \(\Phi\) respects
the product.
\(\mathbf{FQSym}\) conveys the combinatorics of the temporal orders exactly: its basis elements
\(F_\sigma\) \emph{are} the orders, the shifted shuffle is their interleaving, and the coproduct is the
first-car recursion.However, it does not encode the \emph{weights} \(\prod_k 1/\ell_k^\sigma\) attached to each order. As such, the weights analytic quantity that cannot be encoded by FQSym. So the weight map
\(F_\sigma\mapsto\prod_k 1/\ell_k^\sigma\) is not multiplicative.

\textbf{Counterexample (the weight map is not a character of \(\mathbf{FQSym}\)).} Take \(s=20\) and \(n=2\),
with cars at \(x_1=2\) and \(x_2=12\), so that \(\ell(\varnothing)=19\) and
\(\ell(\{x_1\})=\ell(\{x_2\})=17\). Write \(F_1\) and \(F_2\) for the two one-car orders; their shifted
shuffle is \(F_1\cdot F_2=F_{12}+F_{21}\), the two ways of interleaving them. Then
\[\resizebox{\ifdim\width>\linewidth\linewidth\else\width\fi}{!}{$\displaystyle \Phi(F_1\cdot F_2)=\underbrace{\tfrac1{19}\cdot\tfrac1{17}}_{F_{12}}
+\underbrace{\tfrac1{19}\cdot\tfrac1{17}}_{F_{21}}=\frac{2}{323},
\qquad
\Phi(F_1)\,\Phi(F_2)=\frac1{19}\cdot\frac1{19}=\frac1{361},$}\]
and \(\tfrac{2}{323}\neq\tfrac1{361}\). \(\qquad\blacksquare\)

\paragraph{\texorpdfstring{Efficient computation of \(f_{\mathrm{UCP}}\)}{Efficient computation of f\_\{\textbackslash mathrm\{UCP\}\}}}

Evaluating \(f_{\mathrm{UCP}}\) from Theorem \ref{thm:temporal-sum} at a single configuration costs \(O(n!)\) arithmetic
operations. Since the free length \(\ell_k^\sigma\) depends only on the \emph{set} of
cars placed before step \(k\) and not on their temporal order, it is inefficient to apply \ref{thm:temporal-sum} directly.

Two routes exploit that redundancy, and they answer different questions. Grouping the orders by
their last-placed car collapses the sum onto subsets; that gives a fast evaluation at one
configuration and is the subject of \ref{sec:order-stats-cells}. It leaves the density as a single
expression with the case analysis still inside it, so it yields no closed form.

The route taken here splits the \emph{domain} instead. Partitioning the polytope \(\mathsf S_n\) so that
each piece carries one rational function with no case analysis left inside is what makes a symbolic
form available at all, and it is what the marginals are later integrated over. The partition is
dictated by the density itself.

\paragraph{\texorpdfstring{\(f_{\mathrm{UCP}}\) is not \(C^1\) on \(\mathsf S_n\)}{f\_\{\textbackslash mathrm\{UCP\}\} is not C\^{}1 on \textbackslash mathsf S\_n}}

The free length is \(\ell(S)=(s-\lvert S\rvert)-\sum_j\min(g_j,1)\), and \(\min(g_j,1)\) is continuous
but not differentiable at \(g_j=1\). The density inherits exactly that regularity: it is continuous
across a \textbf{jamming wall} \(\{g_j=1\}\), where an interior gap equals one car length, and its
derivative jumps there.

\textbf{Running Example (\(n=3\), \(s=13/2\)).} Fix the leftmost car at \(x_{(1)}=\tfrac12\) and the rightmost one \(x_{(3)}=\tfrac{11}2\). The central car \$x\_\{(2)\} creates two interior gaps \(g_1=x_{(2)}-x_{(1)}-1\) and \(g_2=x_{(3)}-x_{(2)}-1\), which
intersect \(1\) (what is the significance of this one?) at \(x_{(2)}=\tfrac52\) and \(x_{(2)}=\tfrac72\). The density \(f_{\mathrm{UCP}}\) is continuous at both crossings
and has a corner at each.

\begin{figure}
\centering
\pandocbounded{\includegraphics[keepaspectratio,alt={The joint density across two jamming walls: continuous, not differentiable}]{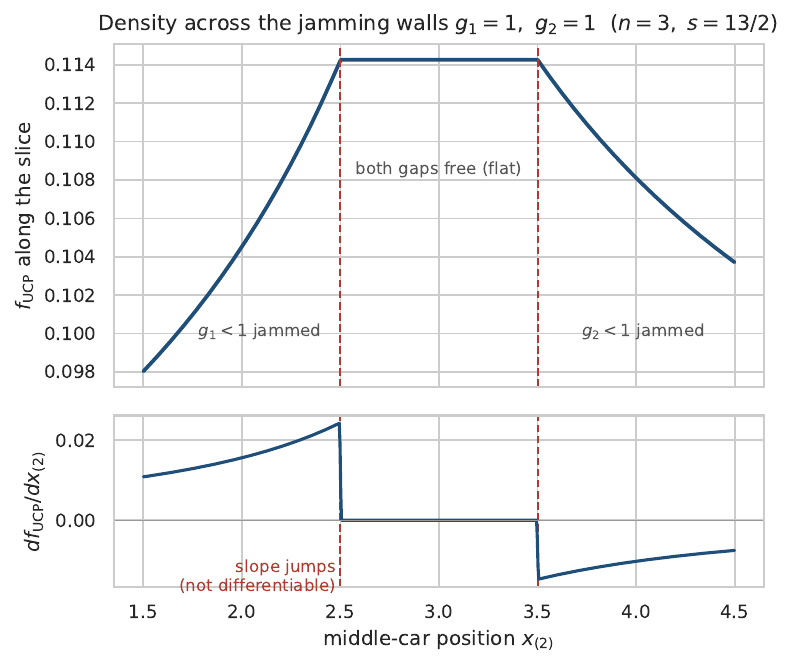}}
\caption{The joint density across two jamming walls: continuous, not differentiable}
\end{figure}

The jamming hyperplanes are therefore the natural cut locus. Slicing \(\mathsf S_n\) along every hyperplane
\(\{g_j=1\}\) leaves regions on which each \(\min(g_j,1)\) resolves to one branch, so every \(\ell_k\) is
a single affine form and the density is one rational function. Those regions are the \textbf{cells}, and
the branch choices that label them are the jamming bitstrings of \ref{sec:order-stats-cells}.

\subsection{\texorpdfstring{Cells and algorithms: computing \(f_{\mathrm{UCP}}\)}{Cells and algorithms: computing f\_\{\textbackslash mathrm\{UCP\}\}}}

\label{sec:order-stats-cells}

\ref{sec:joint-order-stats} shows that the density has a corner at every jamming wall, so the
walls are the natural cut locus. This section carries that out: it partitions the polytope,
gives the algorithm that evaluates the density on a cell, and reports the running example.

\subsubsection{The recursion as a dynamic program}

The \(n!\) sum of \ref{sec:joint-order-stats} is evaluated here in \(O(2^n n)\) operations. The saving comes
from a single structural fact, and it is worth naming the thing that fails first. A car does not
choose its gap independently of the others: it divides by the free length summed over \emph{all} gaps, so
the two sides of a split compete for every later arrival. The density therefore does not decompose
across a split, and no tree recursion over sub-segments computes it. What does decompose is the
index set, and the recursion below is accordingly a subset recursion on \(\{1,\dots,n\}\).

\paragraph{The recursion}

Write \([n]=\{1,\dots,n\}\) for the index set of the cars and, for \(S\subseteq[n]\), write
\(\mathfrak S(S)\) for the set of linear orders of \(S\), the temporal orders in which those cars could
have arrived. The sum defining \(f_{\mathrm{UCP}}\) runs over \(\mathfrak S([n])\) and so has \(n!\) terms.
The step that replaces it by a sum over the \(2^n\) subsets is that the free length is a function of a
set.

\refstepcounter{thmcnt}\label{prop:set-determined-free-length}\textbf{Proposition\nobreakspace{}\thethmcnt{} (set-determined free length).} Fix sorted positions \(\mathbf x\). For
\(\sigma\in\mathfrak S([n])\), the free length \(\ell_k^\sigma\) available to the car placed at step \(k\)
depends only on the set \(S=\{\sigma_1,\dots,\sigma_{k-1}\}\) of cars already placed, not on the order
in which they were placed.

\emph{Proof.} By definition \(\ell_k^\sigma\) is the Lebesgue measure of
\(\{p\in[0,s{-}1]:[p,p{+}1]\ \text{meets}\ [x_j,x_j{+}1]\ \text{for no}\ j\in S\}\). The set being
measured is cut out by a condition quantified over the members of \(S\), so permuting them leaves it,
and hence its measure, unchanged. \(\square\)

Write \(\ell(S)\) for that common value. Group the orders of \(S\) by the car that comes \textbf{last}: every
\(\sigma\in\mathfrak S(S)\) ending at \(i\) is an order \(\sigma'\in\mathfrak S(S\setminus\{i\})\) followed
by \(i\), and along it the weight splits as
\[\resizebox{\ifdim\width>\linewidth\linewidth\else\width\fi}{!}{$\displaystyle \prod_{k=1}^{|S|}\frac1{\ell_k^{\sigma}}
=\Big(\prod_{k=1}^{|S|-1}\frac1{\ell_k^{\sigma'}}\Big)\cdot\frac1{\ell(S\setminus\{i\})}.$}\]
The last factor is the free length \(i\) arrives into. By the Proposition it depends on
\(S\setminus\{i\}\) alone and not on \(\sigma'\), so it is constant across the group and leaves the sum:
\[\resizebox{\ifdim\width>\linewidth\linewidth\else\width\fi}{!}{$\displaystyle \sum_{\substack{\sigma\in\mathfrak S(S)\\ \sigma\ \text{ends at}\ i}}\ \prod_k\frac1{\ell_k^\sigma}
=\frac1{\ell(S\setminus\{i\})}\sum_{\sigma'\in\mathfrak S(S\setminus\{i\})}\prod_k\frac1{\ell_k^{\sigma'}}.$}\]
Each order of \(S\) ends at exactly one car, so summing over the \(|S|\) choices of \(i\) counts every
order once. Writing \(D[S]\) for the inner sum over \(\mathfrak S(S)\), the right-hand side is
\(D[S\setminus\{i\}]/\ell(S\setminus\{i\})\) and the recursion closes.

\refstepcounter{thmcnt}\label{thm:heldkarp-evaluation}\textbf{Theorem\nobreakspace{}\thethmcnt{} DP (Held--Karp evaluation).} For sorted positions \(\mathbf x\) in \(\mathsf S_n\),
\[\resizebox{\ifdim\width>\linewidth\linewidth\else\width\fi}{!}{$\displaystyle D[\varnothing]=1,\qquad
D[S]=\sum_{i\in S}\frac{D[S\setminus\{i\}]}{\ell(S\setminus\{i\})},\qquad
f_{\mathrm{UCP}}(\mathbf x)=D\big[[n]\big],$}\]
and the table is filled in \(O(2^n n)\) arithmetic operations.

\emph{Proof.} The displayed grouping is the recursion, and \(D[[n]]=\sum_{\sigma\in\mathfrak S([n])}\prod_k
1/\ell_k^\sigma=f_{\mathrm{UCP}}(\mathbf x)\) by the definition of \(D\). For the count: the table has
one entry per subset, so \(2^n\) entries, and the entry for \(S\) costs \(|S|\le n\) terms, one per choice
of last car. The total \(\sum_{S\subseteq[n]}|S|=n\,2^{n-1}\) is bounded by \(n\,2^n\). \(\square\)

Both the \(n\,2^n\) bound and the exact value \(n\,2^{n-1}\) are machine-checked in Lean. Both bound the cost of \emph{this} algorithm; neither is a
lower bound on the problem. Whether evaluating \(f_{\mathrm{UCP}}\) admits a \(\mathrm{poly}(n)\)
algorithm, and whether it is \(\#P\)-hard, are both open.

\paragraph{The algorithm}

Executed as a single pass over the subsets in increasing order of size, the recursion above
becomes the following procedure. It fills a table \(W\) indexed by the subsets of \([n]\), each entry
accumulating one term per choice of last-placed car.

\begin{verbatim}
Algorithm DENSITY(x_1 < ... < x_n, span s)
  W[∅] ← 1
  for S over subsets of {1,…,n} by increasing |S|, S ≠ ∅:
      W[S] ← 0
      for i in S:                # i = last car placed
          # ℓ = TOTAL free length left by S∖{i}: the sum
          # over ALL its gaps, not the one holding x_i.
          (p_1,…,p_r) ← sorted {x_j : j ∈ S∖{i}}
          ℓ ← (p_1−1)_+ + (s−p_r−2)_+
                        + Σ_m (p_{m+1}−p_m−2)_+
          W[S] ← W[S] + W[S∖{i}] · (1 / ℓ)
  return W[{1,…,n}]        # ℓ ← s−1 when S∖{i} = ∅
\end{verbatim}

Correctness is Theorem DP: the loop computes \(W[S]=D[S]\) by induction on \(|S|\), since the entries it
reads have strictly smaller size, and it returns \(W[[n]]=f_{\mathrm{UCP}}(\mathbf x)\). The
accumulator only adds and multiplies, taking each \(1/\ell(S)\) as an input, so the same pass runs over
any commutative semiring once those reciprocals are supplied; over \(\mathbb Q\) it returns the density
in exact rationals, which is how the tables of the running example are computed.

\textbf{Example (\(n=3\)).} The six temporal orders of \(\{1,2,3\}\) are grouped by last car into three pairs,
so the table holds eight entries against the six products of the direct sum, and the two agree in
exact rationals. The agreement is checked to \(n\le8\) in the running example.

\subsubsection{The cell partition and the geometry of a cell}

The rational form of the density changes exactly where a gap crosses length one, because a gap
shorter than a car can host no further car and therefore stops contributing to every later free
length. Which gaps are short is thus the combinatorial datum that selects the formula, and
configuration space splits into the regions on which that datum is constant. This section describes
those regions. Gap coordinates make each one a slice of a box by a hyperplane, the walls being
facets of the unit cube; the H-description below gives their facets and vertices, and the
half-integrality of the vertices controls how many pieces a single-gap marginal is cut into.

\paragraph{From gap coordinates to order statistics}

The cells are defined by the gaps, the density is wanted in the order statistics, and the two are
interchangeable. Writing \(g_0=X_{(1)}\), \(g_i=X_{(i+1)}-X_{(i)}-1\) and \(g_n=s-X_{(n)}-1\) gives
\(\mathbf X=L\mathbf G+\mathbf c\), the affine map established in \ref{sec:background-ucp}, whose
reduced matrix is lower-triangular with unit diagonal and therefore unimodular. Its determinant is
\(\pm1\), so densities and volumes carry across unchanged:
\[\resizebox{\ifdim\width>\linewidth\linewidth\else\width\fi}{!}{$\displaystyle f_{X_{(1)},\dots,X_{(n)}}(\mathbf x)=f_{G_0,\dots,G_n}(\mathbf g(\mathbf x)).$}\]
Every statement below may therefore be read in whichever coordinates are convenient. The cells are
described in the gaps, where a wall is the single condition \(\{G_i=1\}\); the density is reported in
the order statistics.

\paragraph{Enumerating the cells}

A cell is indexed by its \textbf{jamming bitstring} \(\mathbf b\in\{0,1\}^{n+1}\), with \(b_i=1\) when the
gap \(g_i\) is shorter than a car. Since \(\sum_i g_i=s-n\) and every gap with \(b_i=0\) has \(g_i\ge1\),
the bitstring describes a non-empty region exactly when the unjammed gaps fit:
\[\resizebox{\ifdim\width>\linewidth\linewidth\else\width\fi}{!}{$\displaystyle \#\{i:b_i=0\}\le s-n\qquad\Longleftrightarrow\qquad\#\{i:b_i=1\}\ge 2n+1-s.$}\]
Feasibility depends only on how many gaps are jammed, not on which, so the cells may be enumerated
by running over all \(2^{n+1}\) bitstrings and keeping those of sufficient weight.

\begin{verbatim}
Algorithm CELLS(n, span s)
  j0 ← max(0, ⌈2n + 1 − s⌉)   # fewest jammed gaps a cell can have
  for b in {0,1}^(n+1):
      if popcount(b) ≥ j0:
          emit (b, R_b)
  # R_b = {g ≥ 0 : Σ g_i = s−n, and g_i < 1 exactly when b_i = 1}
\end{verbatim}

The count follows by summing the binomial coefficients that survive the test,
\[\resizebox{\ifdim\width>\linewidth\linewidth\else\width\fi}{!}{$\displaystyle \lvert\mathcal B(n,s)\rvert=\sum_{j\ge\lceil 2n+1-s\rceil}\binom{n+1}{j},$}\]
which is \(15,16,7,1\) at \(n=3,4,5,6\) and \(s=13/2\). Pairing \texttt{CELLS} with \texttt{DENSITY}, evaluated at any
interior point of \(\mathcal R_{\mathbf b}\), gives the density on every cell at a cost of
\(O(2^n n)\) per cell.

\paragraph{A general cell, and the cell where the density is uniform}

The bitstring splits the gaps into two kinds, and the split is geometric. The wall \(\{G_i=1\}\) is a
facet of the unit cube in the gap coordinates, so \(b_i=1\) places the cell on the interior side of
that facet and \(b_i=0\) on the exterior side. The jammed sub-region \(J_n\), where every \(b_i=1\), is
therefore the one cell lying interior to all \(n+1\) facets, the slice of the open unit cube by the
conservation hyperplane; every other cell lies exterior to at least one of them. A general cell mixes
the two: some gaps are shorter than a car and some are not.

\textbf{Example (a mixed cell).} Take \(n=3\) and \(s=\tfrac{15}2\), so \(\sum_i g_i=s-n=\tfrac92\), and the
cell \(\mathbf b=(1,0,1,0)\), in which \(g_0\) and \(g_2\) are jammed and \(g_1\) and \(g_3\) are not. At the
gaps \((\tfrac12,\tfrac32,\tfrac12,2)\) the density is \(\tfrac{116}{1755}\), and at
\((\tfrac58,\tfrac{11}8,\tfrac58,\tfrac{15}8)\), another point of the same cell, it is
\(\tfrac{2432}{34983}\). The density varies across the cell: it is a non-constant rational function
there, as it is on every cell with at least one jammed gap.

The exception is the cell with no jammed gap at all.

\refstepcounter{thmcnt}\label{prop:unjammed-cell-uniform}\textbf{Proposition\nobreakspace{}\thethmcnt{} (the fully unjammed cell carries the uniform law).} The cell \(\mathbf b=\mathbf 0\) is
non-empty exactly when \(s\ge 2n+1\), and on it
\[\resizebox{\ifdim\width>\linewidth\linewidth\else\width\fi}{!}{$\displaystyle f_{\mathrm{UCP}}(\mathbf x)=\frac{n!}{\prod_{k=0}^{n-1}\bigl((s-1)-2k\bigr)},$}\]
a constant. The law restricted to that cell is uniform.

\emph{Proof.} Every gap satisfies \(g_i\ge1\), so no exclusion zone of a placed car is clipped by another
car or by an endpoint, and the free length left by a set \(S\) is \(\ell(S)=(s-1)-2\lvert S\rvert\),
depending on \(S\) only through its size. Every temporal order then contributes the same product
\(\prod_{k=0}^{n-1}\bigl((s-1)-2k\bigr)^{-1}\), and by the recursion there are \(n!\) of them. Non-emptiness
is the feasibility criterion with all \(n+1\) gaps unjammed: \(n+1\le s-n\). \(\square\)

\textbf{Example.} At \(n=3\) the cell exists once \(s\ge7\). At \(s=\tfrac{15}2\) the density on it is
\(3!/[(s{-}1)(s{-}3)(s{-}5)]=\tfrac{16}{195}\), and two different points of the cell return that same
value. At the running example \(s=\tfrac{13}2\) the cell is empty, since \(\tfrac{13}2<7\), which is why
no plateau of full dimension appears there.

\paragraph{\texorpdfstring{The density changes rational form across \(\{G_i=1\}\), and nowhere else}{The density changes rational form across \textbackslash\{G\_i=1\textbackslash\}, and nowhere else}}

Every free length has the form \(\ell(S) = (s-m) - \sum_i \min(G_i,1)\), and \(\min(G_i,1)\) has a kink at
\(G_i=1\): it equals \(G_i\) on a short gap (\(b_i=1\)) and \(1\) on a long one (\(b_i=0\)). The density
assembled from these free lengths is therefore piecewise rational with break loci exactly the
hyperplanes \(\{G_i=1\}\), and the bit \(b_i=\mathbf 1[G_i<1]\) names the active branch. The bitstring
partition is thus the coarsest one on which \(f_{\mathrm{UCP}}\) is a single rational function, which is
what keeps the marginal integrals within one rational integrand per region. The \(n=3\) case exhibits it: across \(g_1=1\) the density switches from \(6/((s-1)(s-3)(s-5))\)
(no \(g_1\)) to a \(g_1\)-dependent form: \textbf{one continuous density, two rational pieces}, agreeing
exactly on \(g_1=1\) and differing off it.

\paragraph{\texorpdfstring{Density on the jamming polytope \(J_n\)}{Density on the jamming polytope J\_n}}

\refstepcounter{thmcnt}\label{thm:jamming-polytope-density}\textbf{Theorem\nobreakspace{}\thethmcnt{}.} In the gap coordinates of \ref{def:gap-simplex} the configurations occupy the gap simplex
\(\mathcal G_n\), and the jammed sub-region \(J_n=\mathcal G_n\cap\{g_i<1\ \forall i\}\) is the slice of
the open unit hypercube by the conservation hyperplane; for \(s<n+1\) it is all of \(\mathcal G_n\). On \(J_n\)
the density is the single weight-\(0\) rational function
\[\resizebox{\ifdim\width>\linewidth\linewidth\else\width\fi}{!}{$\displaystyle f_{\mathrm{UCP}}=\sum_{\sigma\in S_n}\prod_{k=1}^n \frac{1}{\ell_k^\sigma},$}\]
all \(n!\) temporal orders are feasible at every interior point of the polytope, on \(J_n\) and on every other cell alike: the cars are pairwise at least a car length apart, so the free region left by any prefix contains a positive-measure neighbourhood of every unplaced car. It is
\textbf{positive and pole-free on \(\operatorname{int}J_n\)}, every pole hyperplane is a boundary block
\(\{\sum_{i\in\text{block}}g_i=0\}\) or lies outside \(J_n\) (each \(\ell_k^\sigma>0\) along every feasible
order, since all \(n!\) orders reach the jammed config), and it is the \textbf{size-biased counterpart of the
constant Tonks equilibrium density} \(f_{\mathbb H}=n!/(s-n)^n\). Consequently
\[\resizebox{\ifdim\width>\linewidth\linewidth\else\width\fi}{!}{$\displaystyle \int_{J_n} f_{\mathrm{UCP}}=\Pr[N=n]\ \text{(true RSA)},\qquad
  \int_{J_n} f_{\mathbb H}=\text{the Iwasa–Fukuda / Tonks value}$}\], the two differ, and that difference is the correction from the equilibrium law to the sequential one.

\textbf{Smallest instance \((n=2)\).} \(f_{\mathrm{UCP}}|_{J_2}=\dfrac{1}{s-1}\Big(\dfrac{1}{g_0+g_1}+\dfrac{1}{g_1+g_2}\Big)\), the two temporal orders written out, symmetric under \(g_0\leftrightarrow g_2\).

\textbf{Running example \((s=13/2,\ n=4)\).} \(f_{\mathrm{UCP}}|_{J_4}\) is a weight-0 rational summing all
\(4!=24\) temporal orders (assembled over \(2^4=16\) subsets), non-constant, with \(14\) pole hyperplanes all
off \(\operatorname{int}J_4\). Over \(J_4=\{g\in(0,1)^5:\sum g_i=5/2\}\),
\[\resizebox{\ifdim\width>\linewidth\linewidth\else\width\fi}{!}{$\displaystyle \int_{J_4} f_{\mathrm{UCP}}=0.406=\Pr[N{=}4]\ \text{(RSA)},\qquad
  \int_{J_4} f_{\mathbb H}=0.368\ \text{(Iwasa–Fukuda / Tonks)}$}\], a wider RSA\(-\)Tonks gap than at \(n=2\). The subset recursion computes the density and both integrals.

\paragraph{Schlegel diagrams of the cell partition}

\begin{figure}
\centering
\pandocbounded{\includegraphics[keepaspectratio,alt={cell partition and Schlegel diagram}]{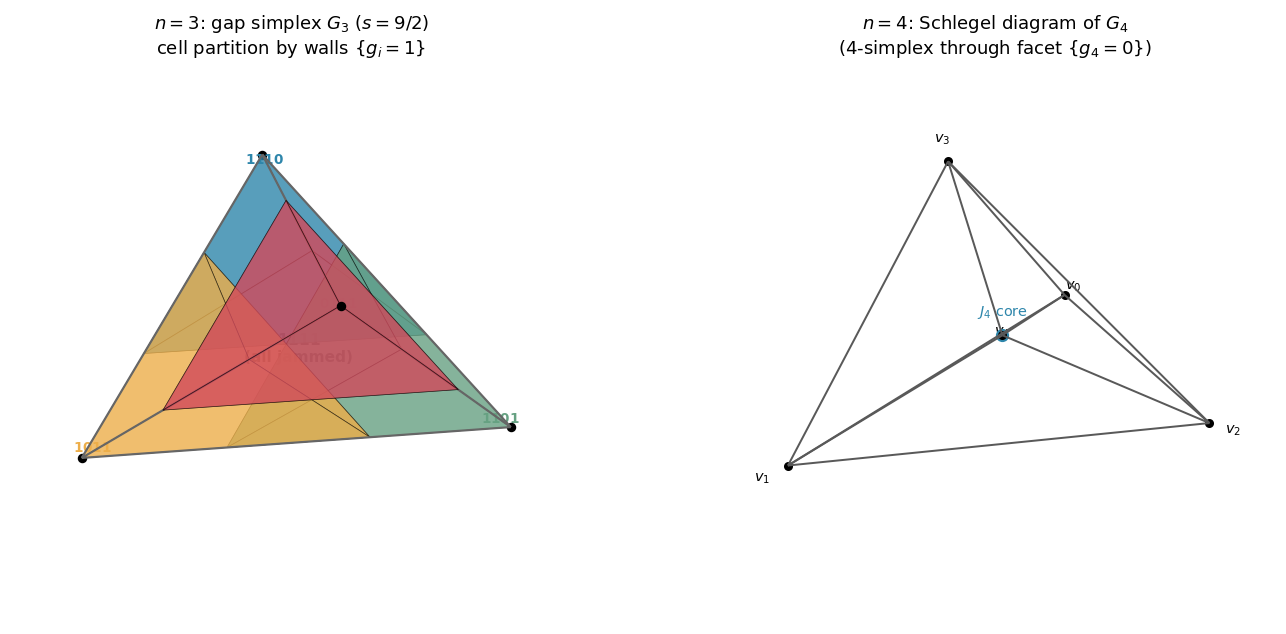}}
\caption{cell partition and Schlegel diagram}
\end{figure}

\textbf{Left (\(n=3\), \(s=9/2\)).} The gap simplex \(\mathcal G_3=\{g\ge0,\ \sum g_i=3/2\}\) is a tetrahedron; the four walls \(\{g_i=1\}\) (equivalently \(\lambda_i=2/3\) in barycentric coordinates) cut off four corner simplices, the cells with one \emph{unjammed} gap (\(b_i{=}0\): bitstrings \(\mathbf{0111},\mathbf{1011},\mathbf{1101},\mathbf{1110}\)), leaving the central all-jammed cell \(\mathbf{1111}\), the jamming polytope \(J_3\). This is the H-description of the cell, with \(\tau=s-2n-1+w\): each corner is a small simplex and the centre is \(J_3\).

\textbf{Right (\(n=4\)).} The gap simplex \(\mathcal G_4\) is a \(4\)-simplex; its Schlegel diagram (perspective projection through the facet \(\{g_4=0\}\)) renders the \(4\)-polytope in three dimensions as a star of tetrahedra with the jamming core \(J_4\) at the centre.

\paragraph{The number of pieces of a single-gap density}

Every cell is cut out by facets of the form \(\{G_i=1\}\) together with the hard-core and endpoint
constraints, all of them \(\pm1\)-affine, so each vertex solves \(n\) of these equations with
half-integer coordinates. That locates the breakpoints of a single-gap marginal before it is
computed: the formula can change only where the slicing hyperplane crosses a vertex.

\refstepcounter{thmcnt}\label{thm:order-stats-theorem-33-1}\textbf{Theorem\nobreakspace{}\thethmcnt{} C (pieces of a single-gap density).} The single-gap density \(g\mapsto f_{G_i}(g)\) is piecewise with \(O(s)\) pieces; every breakpoint lies at a half-integer (an integer when \(s\in\mathbb Z\)).

\textbf{Proof.} The slice \(\{G_i=g\}\), the hard-core spacings \(t_j-t_{j-1}\ge1\), the endpoint bounds \(t_0\ge0,\ t_{n-1}\le s-1\), and the gap thresholds \(g_j\lessgtr1\) are all constraints on the sorted positions \(\mathbf t\) of the form \(t_j\lessgtr c\) or \(t_j-t_{j'}\lessgtr c\), a \textbf{difference system}, whose coefficient matrix is the incidence matrix of a digraph and is therefore \textbf{totally unimodular} (a network matrix; Ghouila-Houri criterion). Hence every vertex of every sliced polytope has coordinates in the additive group generated by the right-hand sides \(\{0,\ s-1,\ \mathbb Z,\ g\}\). The combinatorial type, and therefore the density formula, changes only where two vertex coordinates coincide, i.e.~at values of \(g\) solving an integer-coefficient equation with right-hand side in \(\{0,\,s-1,\,\mathbb Z\}\); for \(s\in\tfrac12\mathbb Z\) these are half-integers. As \(g\) ranges over \([0,s-n]\) there are \(O(s)\) such values. \(\square\)

\textbf{Count.} The distinct-form breakpoints are the at most \(2(s-n)\) half-integers in \([0,s-n]\), and a curvature scan of \(f_{G_0}\) at \(s=9/2\) confirms the breakpoint at \(g=1\). The finer tallies \(11,7,4,1\) at \(s-n=\tfrac72,\tfrac52,\tfrac32,\tfrac12\) count bitstring-by-interval sub-expressions rather than distinct formula pieces; at \(n{=}4\) the sweep of the running example finds five pieces, in agreement with the half-integer enumeration.

\refstepcounter{thmcnt}\label{rem:order-stats-remark-33-2}\textbf{Remark\nobreakspace{}\thethmcnt{}.} The breakpoints observed near \(g=\tfrac12\) at \(s=13/2\) are genuine half-integer breakpoints, a consequence of \(s\) itself being a half-integer.

\paragraph{The density on the hard-core polytope}

At \(n=2\) and \(s=\tfrac{13}2\) the polytope of the gap coordinates above is two-dimensional, so the
density can be drawn whole. The walls \(\{G_i=1\}\) cut it into the cells of the partition, and the
density is a different rational function on each.

\begin{figure}
\centering
\pandocbounded{\includegraphics[keepaspectratio,alt={The UCP joint density at n=2, s=13/2; dotted lines are the cell walls \textbackslash\{G\_i=1\textbackslash\}}]{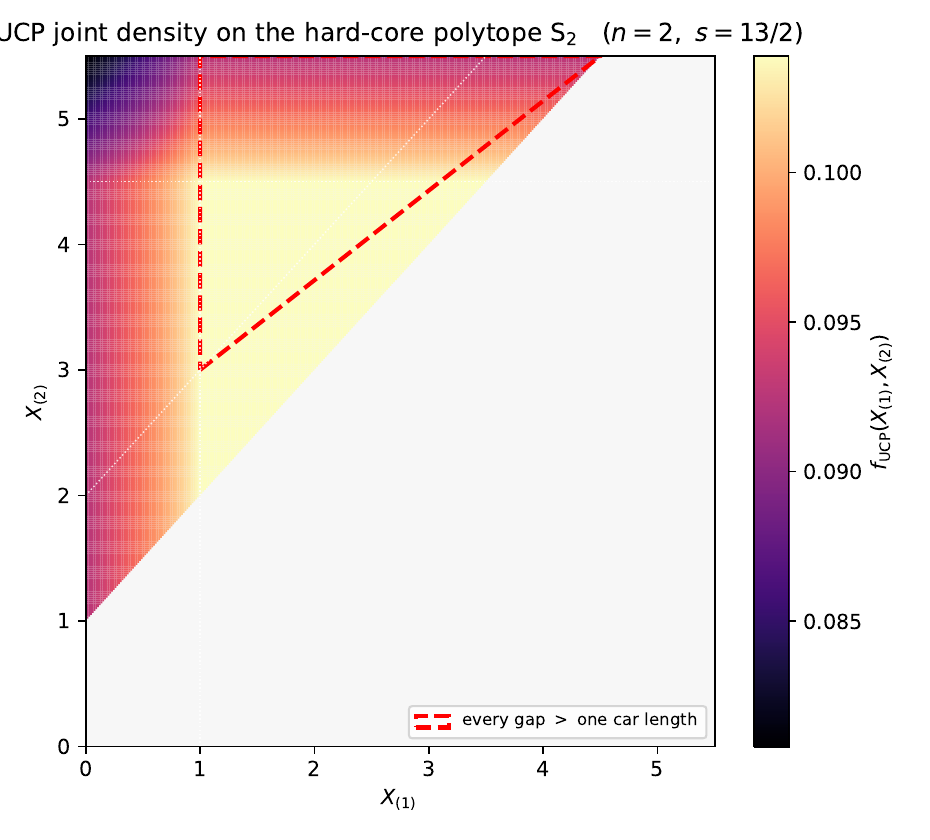}}
\caption{The UCP joint density at \(n=2\), \(s=13/2\); dotted lines are the cell walls \(\{G_i=1\}\)}
\end{figure}

On the cell where every gap exceeds a car length the surface is flat, at height
\(2!/[(s{-}1)(s{-}3)]=\tfrac8{77}\). The plateau is not special to \(n=2\): where no exclusion zone is
clipped, every free length is \(\ell(S)=(s-1)-2\lvert S\rvert\) and so depends on \(S\) only through
its size, all \(n!\) temporal orders carry the same product, and the density takes the constant value
\(n!\big/\prod_{k=0}^{n-1}\bigl((s-1)-2k\bigr)\). The surface falls away from the plateau toward
the boundaries where a gap closes, and the corners appear across those walls.

\subsubsection{\texorpdfstring{The running example: the size of the computation at one \((s,n)\)}{The running example: the size of the computation at one (s,n)}}

A cell is a jamming-bitstring region \(\mathcal R_{\mathbf b}\); the \(n!\) attachment words lying over it are
an internal device, and their kernels sum to one rational function per cell. The table records how
many cells there are, how many are distinct up to reflection, and how many reciprocal-affine
products the density on one of them carries.

{\footnotesize\begin{longtable}[]{@{}
  >{\raggedright\arraybackslash}p{\dimexpr 0.2500\linewidth-2\tabcolsep\relax}
  >{\raggedright\arraybackslash}p{\dimexpr 0.2500\linewidth-2\tabcolsep\relax}
  >{\raggedright\arraybackslash}p{\dimexpr 0.2500\linewidth-2\tabcolsep\relax}
  >{\raggedright\arraybackslash}p{\dimexpr 0.2500\linewidth-2\tabcolsep\relax}@{}}
\toprule\noalign{}
\begin{minipage}[b]{\linewidth}\raggedright
\end{minipage} & \begin{minipage}[b]{\linewidth}\raggedright
general \((s,n)\)
\end{minipage} & \begin{minipage}[b]{\linewidth}\raggedright
\(s=13/2\): \(\,n=3,4,5,6\)
\end{minipage} & \begin{minipage}[b]{\linewidth}\raggedright
in the literature?
\end{minipage} \\
\midrule\noalign{}
\endhead
\bottomrule\noalign{}
\endlastfoot
\textbf{1. \# cells} & \(\lvert\mathcal B(n,s)\rvert=\displaystyle\sum_{k=w_{\min}}^{n+1}\binom{n+1}{k}-\mathbf 1\{s>2n+1\},\ \ w_{\min}=\max\{0,\lceil 2n{+}1{-}s\rceil\}\) & \(15,\,16,\,7,\,1\) & no prior source found (the jamming-cell partition) \\
\textbf{2. qualitatively distinct} & \(\lceil\lvert\mathcal B(n,s)\rvert/2\rceil\) (reflection of the segment; the density is rational on every cell, of weight \(0\)) & \(8,\,8,\,4,\,1\) & reflection is classical; the partition is not \\
\textbf{3. min terms / cell} & \(n!\): every temporal order is feasible at an interior point & \(6,\,24,\,120,\,720\) & no prior source found \\
\textbf{4. max terms / cell} & exactly \(n!\) reciprocal-products: \(f\big|_{\mathcal R_{\mathbf b}}=\sum_{\sigma\in\mathfrak S_n}\prod_{k=1}^{n}1/\ell_k^{(\sigma)}\) & \(6,\,24,\,120,\,720\) & \textbf{prior art} (the order-sum \emph{form}) \\
\end{longtable}}

\textbf{General/exact: 1, 2, 4.} A cell's density is \(\sum_{\sigma}\prod_{k}1/\ell_k^{(\sigma)}\), at most \(n!\) products, the bound being attained only when no two temporal orders give the same product (measured distinct products: \(6\) at \(n{=}3\), \(22\) of the \(24\) orders giving distinct products at \(n{=}4\)). \textbf{\#3 min} has no closed form. \emph{(Combined over a common denominator the expanded numerator is larger, \(3\)--\(19\) at \(n{=}3\), \(63\)--\(378\) at \(n{=}4\), and not fit from two points.)}

\paragraph{Prior art for the cell decomposition}

The order-sum form of the density is classical; the finite-\((s,n)\) resolution into cells is not.

\begin{itemize}
\tightlist
\item
  \textbf{The form is known.} Writing a cell's density as the order-sum
  \(f=\sum_\sigma\prod_k 1/\ell_k\), hence the bound of \(n!\) reciprocal-products, goes back to
  Widom (1966) and the cooperative sequential adsorption literature surveyed by Talbot; Gerin

  \begin{enumerate}
  \def\labelenumi{(\arabic{enumi})}
  \setcounter{enumi}{2013}
  \tightlist
  \item
    likewise indexes by a permutation. This form is cited here, not claimed.
  \end{enumerate}
\item
  \textbf{The targets studied are different.} The literature treats the count \(N(s)\) and its moments or
  limits (Rényi 1958; Gouet--López 2001; Gerin 2014), the single-gap marginal and vacant-interval
  law (Coffman--Flatto--Jelenković 2000; Araújo--Cadilhe 2006; Mackey--Sullivan 2016; Clay--Simányi
  2014), or the asymptotic regime \(n,s\to\infty\).
\item
  \textbf{What is new here.} The exact finite-\((s,n)\) \emph{joint} order-statistic density resolved into
  jamming cells, the cell count \(\lvert\mathcal B(n,s)\rvert\), the partition itself, and the
  per-cell rational structure. The nearest prior work is Gerin (2014), which indexes by
  permutation but computes marginals only, and Araújo--Cadilhe (2006), whose saturated-gap law is
  the closest object.
\end{itemize}

The claim is therefore the cell count, the partition, and the finite-\((s,n)\) joint resolution; the
order-sum form itself belongs to Widom and the adsorption literature.

\paragraph{\texorpdfstring{Computation summary at \(n = 3,\dots,6\) and \(s = 13/2\)}{Computation summary at n = 3,\textbackslash dots,6 and s = 13/2}}

\textbf{Words, cells, integrals.} Reflection gives \(G_{n-i}=G_i\), so only \(\lceil(n+1)/2\rceil\) gaps need be
computed; the integral counts are Duffy entries summed over all \(g\)-pieces.

{\footnotesize\begin{longtable}[]{@{}
  >{\raggedright\arraybackslash}p{\dimexpr 0.1250\linewidth-2\tabcolsep\relax}
  >{\raggedright\arraybackslash}p{\dimexpr 0.1250\linewidth-2\tabcolsep\relax}
  >{\raggedright\arraybackslash}p{\dimexpr 0.1250\linewidth-2\tabcolsep\relax}
  >{\raggedright\arraybackslash}p{\dimexpr 0.1250\linewidth-2\tabcolsep\relax}
  >{\raggedright\arraybackslash}p{\dimexpr 0.1250\linewidth-2\tabcolsep\relax}
  >{\raggedright\arraybackslash}p{\dimexpr 0.1250\linewidth-2\tabcolsep\relax}
  >{\raggedright\arraybackslash}p{\dimexpr 0.1250\linewidth-2\tabcolsep\relax}
  >{\raggedright\arraybackslash}p{\dimexpr 0.1250\linewidth-2\tabcolsep\relax}@{}}
\toprule\noalign{}
\begin{minipage}[b]{\linewidth}\raggedright
\end{minipage} & \begin{minipage}[b]{\linewidth}\raggedright
attachment words
\end{minipage} & \begin{minipage}[b]{\linewidth}\raggedright
cells (bitstrings)
\end{minipage} & \begin{minipage}[b]{\linewidth}\raggedright
collapse
\end{minipage} & \begin{minipage}[b]{\linewidth}\raggedright
gaps computed
\end{minipage} & \begin{minipage}[b]{\linewidth}\raggedright
\(g\)-pieces / gap
\end{minipage} & \begin{minipage}[b]{\linewidth}\raggedright
integrals / gap
\end{minipage} & \begin{minipage}[b]{\linewidth}\raggedright
max / cell·piece
\end{minipage} \\
\midrule\noalign{}
\endhead
\bottomrule\noalign{}
\endlastfoot
\textbf{n=3} & 90 & 15 & \(90=3!\cdot15\) & \(G_0,G_1\) (→ \(G_2,G_3\) refl.) & 11 & \textbf{588} & 24 \\
\textbf{n=4} & 384 & 16 & \(4!\cdot16\) & \(G_0,G_1,G_2\) (refl.) & 5 & \textbf{≈3600} & 288 \\
\end{longtable}}

Totals actually integrated: n=3 → \(588{+}588 = 1176\); n=4 → \(3672{+}3552{+}3648 = 10{,}872\) weight-\((n{-}1)\) integrals. (The maximum per cell and piece is \(n!\) attachment words times the number of simplices in that cell's slice: \(6\times4\) at \(n=3\), \(24\times12\) at \(n=4\).)

\textbf{Form of the formulas.}
- \textbf{Cell count} \(|\mathcal B|\): an exact integer closed form, the tail sum of binomials \(\sum_{k=w_{\min}}^{n+1}\binom{n+1}{k}-\mathbf 1\{s>2n+1\}\).
- \textbf{Gap density} \(f_{G_i}^{(\mathbf b)}(g)\): \emph{piecewise} in \(g\) (breakpoints at half-integers); each piece is a \(\mathbb Q(g)\)-linear combination of \textbf{multiple polylogarithms of weight \(\le n-1\)}, with arguments \emph{affine in \(g\)}:
- at \(n=3\), weight at most \(2\): rational multiples of \(\log\), together with dilogarithms;
- at \(n=4\), weight at most \(3\), so hyperlogarithms as far as weight three.

\textbf{Terms per formula} (leaf size of the symbolic output):
- n=3 forms per attachment word: about \textbf{160 to 1700} leaves; a typical hand-reduced piece is short, e.g.~one \(G_0\) region on \(g\in(1,\tfrac32)\) is \(-\tfrac4{77}(g{+}1)\log 385+\tfrac4{77}(g{+}1)\log(154g{+}154)-\tfrac4{77}g+\tfrac6{77}\) (≈4 terms).
- n=4 per-cell \emph{accumulated} forms: large sums, up to \textbf{288 terms} for the dominant cell \texttt{01111}; correct, but pathological under numeric evaluation as one object, so kept termwise.

\textbf{Validation.} At \(n=3\) the per-bitstring sums match the exact polygon integrator to six or seven digits with no evaluation failures. At \(n=4\) a termwise Monte-Carlo integration of the integrand matches the numeric backend on every gap: \(0.2198\) against \(0.2197\) for \(G_0\), \(0.1615\) against \(0.1616\) for \(G_1\), and \(0.0252\) against \(0.0252\) for \(G_2\).

\subsection{Marginal order statistics}

\label{sec:marginal-order-stats}

\ref{sec:order-stats-cells} gives the joint density as a rational function on each cell. A marginal is
obtained by integrating that rational function over the coordinates to be discarded. This section
records what the integration costs and what it produces: the class of function is fixed in advance by
a weight count, the answers at \(n=4\) are explicit, and the point at which the computation stops is
reached well before the theory does.

\paragraph{Linear reducibility, and the field the letters live in}

The weight bound of the next subsection presumes that each integration can be carried out in closed
form at all. What guarantees this is a property of the singular locus, and in this problem it holds as
a theorem rather than as an assumption.

\refstepcounter{thmcnt}\label{lem:reducibility}\textbf{Lemma\nobreakspace{}\thethmcnt{} (linear reducibility of the parking integrals).} Every free length \(\ell_k^\sigma\) and every
facet of every cell is affine in the gap coordinates, with each coefficient in \(\{-1,0,+1\}\) and
constant term in \(\mathbb Z+\mathbb Z s\). Consequently the singular locus of \(f_{\mathrm{UCP}}\) is a
rational hyperplane arrangement, and Brown's polynomial reduction terminates in every variable order.
The parking integrals are therefore linearly reducible, and each marginal is a hyperlogarithm whose
letters are affine forms over \(\mathbb Q\) in the remaining coordinates and in \(s\).

\emph{Proof.} For the first claim, \(\ell(S)\) is a sum of clipped gaps and so is a signed sum of the \(g_i\)
with an integer constant; no product of two coordinates occurs, which is the one-dimensional hard-core
feature. The facets \(\{G_i=1\}\) and the endpoint constraints are of the same shape.

For the reduction, Brown eliminates one variable \(v\) at a time and asks that the polynomials met along
the way stay factorizable into forms linear in the next variable. Here every form is \(\alpha v+\beta\)
with \(\alpha\in\{-1,0,+1\}\) and \(\beta\) affine in the remaining coordinates. Eliminating \(v\) produces
two kinds of polynomial: the endpoints \(-\beta/\alpha\), which are affine because \(\alpha\) is a unit;
and the crossings, whose resultant is
\[\resizebox{\ifdim\width>\linewidth\linewidth\else\width\fi}{!}{$\displaystyle \operatorname{Res}_v(\alpha_1v+\beta_1,\ \alpha_2v+\beta_2)=\alpha_1\beta_2-\alpha_2\beta_1,$}\]
an integer combination of affine forms and hence affine. So the defining property is preserved by one
elimination step, and by induction by all of them. The arrangement is finite, so the process
terminates, and it does so whatever order the variables are eliminated in. Linear reducibility is
Brown's criterion for the iterated integrals to stay within the hyperlogarithms, which gives the last
claim. \(\square\)

Two consequences are worth naming. No irreducible quadratic can appear in the reduction, so no
elliptic period arises anywhere in the finite theory: the letters never leave the rational function
field \(\mathbb Q(g_{i_1},\dots,g_{i_d},s)\). And because the criterion is met in \emph{every} variable
order, the order of integration may be chosen for convenience, which is what makes peeling one
coordinate at a time legitimate.

The general tool being invoked, Brown's criterion and the hyperlogarithm class it controls, is stated
in \ref{sec:background-aggregate}; what is proved here is that this problem satisfies its hypothesis.

\paragraph{The marginalization theorem}

The gaps satisfy \(\sum_{i=0}^n g_i = s-n\), so the gap simplex \(\mathcal G_n\) of \ref{def:gap-simplex} is
\(n\)-dimensional even though it carries \(n+1\) coordinates. Keeping \(d\) of them therefore calls for
\(n-d\) integrations.

\refstepcounter{thmcnt}\label{thm:marginal-weight}\textbf{Theorem\nobreakspace{}\thethmcnt{} (weight of a marginal).} Fix a cell \(\mathcal R_{\mathbf b}\) and keep \(d\) of the \(n+1\)
gaps. The marginal of \(f_{\mathrm{UCP}}\) over \(\mathcal R_{\mathbf b}\) is a hyperlogarithm of weight
at most \(n-d\) in the kept variables, with letters that are affine forms in those variables. The bound
does not depend on \(n\) except through \(n-d\), and it is attained.

\emph{Proof.} On \(\mathcal R_{\mathbf b}\) the density is rational with poles on the facet hyperplanes,
which is weight \(0\). Each of the \(n-d\) integrations is one coaction step: integrating a rational
function of one variable against affine denominators produces a \(\mathrm{d}\log\) primitive, raising
the weight by exactly one and appending one letter, and integrating a weight-\(w\) hyperlogarithm
similarly yields weight \(w+1\). After \(n-d\) steps the weight is at most \(n-d\). The letters are the
images of the facet forms, which are affine in the kept variables because the facets of a cell are
\(\pm1\)-affine. \(\square\)

Two corollaries fix the shape of everything below. A single-gap density keeps \(d=1\) and so has weight
at most \(n-1\): logarithms at \(n=2\), dilogarithms at \(n=3\), weight \(3\) at \(n=4\). A marginal that keeps
almost everything is cheap, since \(n-d\) is small whatever \(n\) is. By \ref{thm:order-stats-theorem-33-1}
each marginal is also piecewise, with breakpoints at the half-integers.

\paragraph{\texorpdfstring{Running example: the single-gap densities at \(n=4\)}{Running example: the single-gap densities at n=4}}

At \(s=\tfrac{13}2\) and \(n=4\) the single-gap densities \(f_{G_i}(g)\) were computed exactly, piece by
piece, with \(g\) carried as a symbol rather than sampled. The pieces are the half-integer intervals
predicted by \ref{thm:order-stats-theorem-33-1}, and on each the answer is a sum over the cells that meet
the slice \(\{G_i=g\}\).

{\footnotesize\begin{longtable}[]{@{}llll@{}}
\toprule\noalign{}
gap & piece & cells meeting the slice & largest term count \\
\midrule\noalign{}
\endhead
\bottomrule\noalign{}
\endlastfoot
\(G_0\) & \([1,\tfrac32]\) & 5 & 39,840 \\
\(G_0\) & \([\tfrac32,2]\) & 1 & 149 \\
\(G_1\) & \([0,\tfrac12]\) & 11 & 50,210 \\
\(G_1\) & \([1,\tfrac32]\) & 5 & 60,253 \\
\(G_2\) & \([1,\tfrac32]\) & 5 & 55,324 \\
\(G_2\) & \([\tfrac32,2]\) & 1 & 471 \\
\end{longtable}}

The forms are hyperlogarithms in \(g\) of weight at most \(3\), matching \ref{thm:marginal-weight} at \(d=1\),
\(n=4\), and the bound is attained: the deepest iterated integrals carry three letters.

\textbf{Example (one cell's contribution).} On \(\mathbf b=01011\) over \(g\in[1,\tfrac32]\) the density is a
combination of \(74\) terms, built from logarithms and from weight-\(2\) and weight-\(3\) hyperlogarithms
whose arguments are affine in \(g\), among them \(2(g+2)/(2g-3)\) and \((2g-5)/(2g-3)\). The letters are
exactly the facet forms of that cell, as \ref{lem:reducibility} requires.

\textbf{Example (both marginals at \(n=3\), \(s=\tfrac{13}2\)).} At \(n=3\) the slice fixing one coordinate is
two-dimensional, so both marginals can be drawn directly from the joint density without any of the
machinery above. The first position \(X_{(1)}\) is a marginal order statistic; the interior gap \(G_1\)
is a marginal gap. Since \(G_0=X_{(1)}\) identically, the leading gap would repeat the left panel, and
the interior gap is shown instead.

\begin{figure}
\centering
\pandocbounded{\includegraphics[keepaspectratio,alt={Marginals of the joint density at n=3, s=13/2: the first position f\_\{X\_\{(1)\}\} and the interior gap f\_\{G\_1\}, with the half-integer breakpoints marked}]{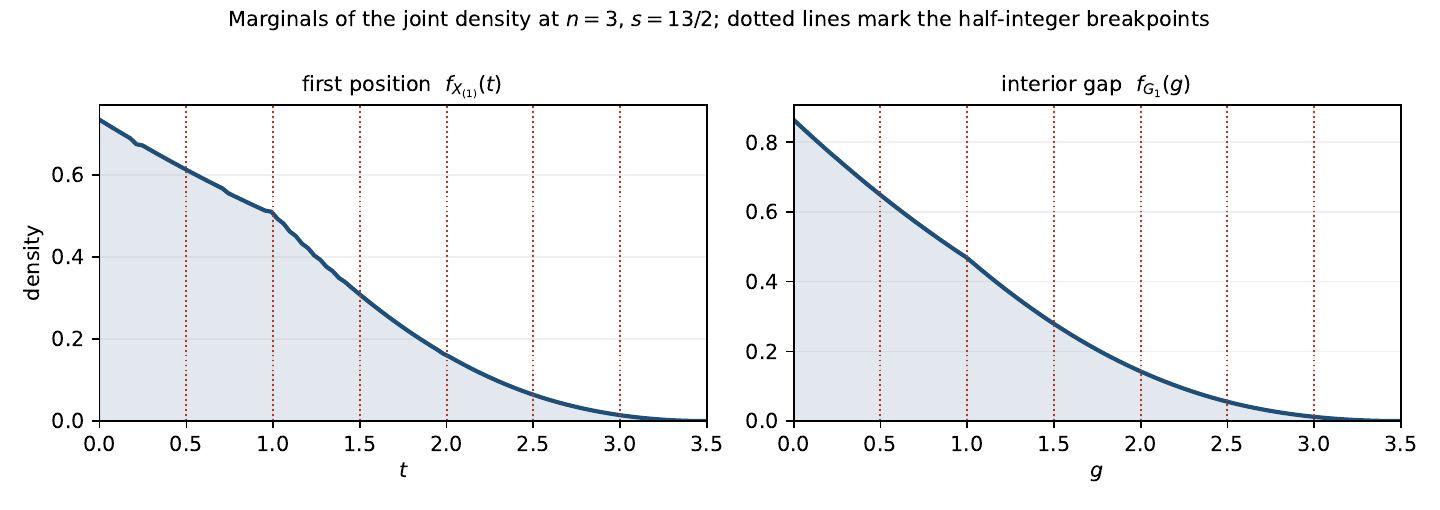}}
\caption{Marginals of the joint density at \(n=3\), \(s=13/2\): the first position \(f_{X_{(1)}}\) and the interior gap \(f_{G_1}\), with the half-integer breakpoints marked}
\end{figure}

Both curves are piecewise, with the kinks falling on the half-integers of
\ref{thm:order-stats-theorem-33-1}; the change of form at \(t=1\) and at \(g=1\) is where the slice crosses a
jamming wall. Both decrease to zero at \(s-n=\tfrac72\), where the remaining cars no longer fit. At
\(n=3\) the weight bound of \ref{thm:marginal-weight} reads \(n-1=2\), so each curve is a combination of
logarithms and dilogarithms, and neither is a polynomial. The two integrate to the same mass, as
they must, being marginals of one density.

\paragraph{Computational limitations}

The integration, not the theory, is what stops. Three limits appeared, and each is recorded rather
than worked around.

\begin{itemize}
\tightlist
\item
  \textbf{Coverage.} Of the fourteen \((\text{gap},\text{piece})\) tasks at \(n=4\), eleven completed and
  three did not: two failed on a floating-point simplification of a symbolic wall position and one
  exceeded its time budget. At \(n=5\) six of eight completed. The position marginals \(f_{X_{(i)}}\)
  are harder, four of thirteen pieces completing, because the slice \(\{X_{(i)}=t\}\) is oblique to
  the facets and meets more cells.
\item
  \textbf{Expression size.} Term counts grow with the number of cells meeting the slice, reaching
  \(60\,253\) on one piece at \(n=4\) and \(73\,423\) for a position marginal. These are raw forms before
  reduction; their size, not their weight, is what makes them impractical to present or to evaluate
  term by term.
\item
  \textbf{Branch selection.} Blind indefinite integration of a rational kernel can select the wrong
  branch of a dilogarithm; on one \(n=3\) word this returned \(-3.58\) against a reference value of
  \(0.012\). Carrying \(g\) symbolically through the integration, so that it enters the arguments of the
  answer rather than its numerical value, is what avoids this.
\end{itemize}

The per-cell forms computed at \(n=4\) and \(n=5\) are archived as plain text and are re-evaluated
independently of the program that produced them, so the results quoted above do not depend on any
one implementation.

None of these limits is a statement about the complexity of the problem, which is open in both
directions. They bound the present pipeline.

\paragraph{The link to the gap marginals}

The gap marginals and the position marginals are two readings of one object. By the affine map
\(\mathbf X=L\mathbf G+\mathbf c\) of \ref{sec:background-ucp}, which is unimodular, the joint densities
agree coordinate for coordinate; the marginals do not, because marginalizing does not commute with an
affine map that mixes coordinates. Each position \(X_{(i)}=\sum_{k<i}g_k+(i-1)\) is a \emph{sum} of gaps, so
a position marginal is the pushforward of the joint gap density along that sum, and its slice
\(\{X_{(i)}=t\}\) is a hyperplane oblique to the facets rather than a coordinate slice.

That obliqueness is the whole difference in cost. A gap slice constrains one coordinate, so it meets
only the cells whose bitstring is compatible with that coordinate; a position slice constrains a sum
of coordinates, so it meets every cell whose gaps can reach the given total. The second set is the
larger, and it is why the position sweep is the one that stalls.

\textbf{Example (\(n=4\), \(s=\tfrac{13}2\)).} The gap slice \(\{G_0=g\}\) on \([1,\tfrac32]\) meets five cells;
the position slice \(\{X_{(1)}=t\}\) on \([2,\tfrac52]\) meets twelve. Eleven of the fourteen gap pieces
completed, against four of the thirteen position pieces.

\ref{sec:gap-order-stats} develops the sorted gaps \(g_{(1)}\le\cdots\le g_{(n+1)}\) from the same joint
density.

\subsection{Gap order statistics}

\label{sec:gap-order-stats}

The largest free interval left by a jammed configuration determines whether another car can attach, so
the sorted gaps \(g_{(1)}\le\cdots\le g_{(n+1)}\) carry the jamming certificate directly. In this section
we derive their joint law from the position density \(f_{\mathrm{UCP}}\), compute the moments and the
marginal laws of the individual order statistics, and bound the transcendental weight of every marginal.

\subsubsection{Prior art for the single-gap marginal and its order statistics}

\textbf{Widom (1966)} obtained the single-gap (typical-spacing) marginal of one-dimensional random sequential adsorption, deriving the gap-length distribution on the infinite line at every density up to jamming by the kinetic empty-interval method. The gap density \(G(\ell,t)\) evolves at arrival time \(t\) by
\[\resizebox{\ifdim\width>\linewidth\linewidth\else\width\fi}{!}{$\displaystyle \partial_t G(\ell,t)=-(\ell-1)_+\,G(\ell,t)+2\!\int_{\ell+1}^\infty G(\ell',t)\,d\ell',$}\]
with loss when a car lands in a gap \(\ge1\) and gain when a longer gap is split. The jamming limit \(t\to\infty\) gives the saturated gap density \(f_G\) in the exponential-period \(\varphi\)-kernel form, with contact behaviour \(f_G(\ell)\sim -e^{-2\gamma}\ln\ell\) as \(\ell\to0\) (\textbf{Bonnier, Boyer \& Viot 1994}).

The combinatorial gap marginal of this section coincides with the Widom--Bonnier--Boyer--Viot result. The saturated single-gap density reproduces the \(\varphi\)-kernel form, with endpoint \(f_G(1^-)=0.4217\) and contact coefficient \(0.843\) (\ref{sec:asymptotic-properties} §A5). The two derivations are independent, Widom's kinetic time-integral and the finite-\(n\) order-statistic construction here, and agree in the jamming limit.

What the sections below add is the law of the \textbf{gap order statistics}: the densities of the smallest, median, and largest gaps \(g_{(1)},g_{(2)},\dots\) (Theorem D), their boundary/interior asymmetry (non-exchangeability), and per-cell log-convexity. The kinetic single-gap treatment does not address the order statistics of the gap set.

\subsubsection{Construction of the order statistics from the joint gap law}

Condition on \(N(s)=n\). The \(n{+}1\) gaps carry the same joint density \(f_{\mathrm{UCP}}\) as the positions
(Lemma \ref{lem:gaps-carry}, unit Jacobian) but are \textbf{dependent and non-exchangeable} (Lemma \ref{lem:non-exchangeability}). In this section we build the order statistics \(g_{(1)}\le\cdots\le g_{(n+1)}\) from that joint law. One route serves both these and the position order statistics: the last-car subset dynamic program of \ref{sec:order-stats-cells} supplies each density value, and the David--Nagaraja symmetrization and marginalization of the sections below run on top of it unchanged.

\begin{enumerate}
\def\labelenumi{\arabic{enumi}.}
\tightlist
\item
  \textbf{Non-exchangeability (the gap law).} The gap vector's distributional symmetry is exactly \(\mathbb Z_2\)
  (segment reflection): \(g_i\stackrel{d}{=}g_{n-i}\) and no more, even distinct-depth interior gaps differ.
  The order statistics sort a \emph{dependent, non-identically-distributed} family.
\item
  \textbf{Joint order-statistic density (the David--Nagaraja derivation).} The \textbf{David--Nagaraja dependency theorem} gives the
  exact joint law \(f_{g_{()}}=\sum_{\pi}f_{\mathrm{UCP}}(\pi\cdot\mathbf y)\) on the ordered jammed simplex, weight-0 rational, with a complexity analysis.
\item
  \textbf{Moments (the means).} \(\mathbb E[G_{(k)}]\) (smallest / median / largest gap) as integrals of the joint law.
\item
  \textbf{Marginal order statistics (the marginal densities).} Single-order-statistic densities (median \(g_{(2)}\)) as marginals
  of the David--Nagaraja derivation.
\item
  \textbf{Contrast (the distribution contrast).} The gap distribution set beside Tonks, Poisson and Mat\textquotesingle ern.
\end{enumerate}

\textbf{Running examples.} \(s=\tfrac52,\,n=2\) (three gaps summing to \(\tfrac12\)) and the primary
\(s=\tfrac{13}{2},\,n=4\) (five gaps summing to \(\tfrac52\)), shared with the companions.

\subsubsection{The two objects this section adds}

The coordinates are those already fixed elsewhere and are not restated. The gaps
\(G_0=X_{(1)}\), \(G_i=X_{(i+1)}-X_{(i)}-1\) and \(G_n=(s-1)-X_{(n)}\), the conservation law
\(\sum_{i=0}^n G_i=s-n\), and the unimodular affine map \(\mathbf X=L\mathbf G+\mathbf c\) between gaps
and positions are \ref{prop:gap-position-coordinates} of \ref{sec:background-ucp}. The gap simplex
\(\mathcal G_n\) and its jammed sub-region \(J_n=\mathcal G_n\cap\{g_i<1\ \forall i\}\) are
\ref{def:gap-simplex} of \ref{sec:background-order-statistics}. The joint density \(f_{\mathrm{UCP}}\) is
\ref{sec:joint-order-stats}, and by \ref{lem:gaps-carry} below it is the same function of \(\mathbf g\) as of
\(\mathbf x\). One convention is local to this section: the two walls are written as sentinels
\(x_{(0)}=0\) and \(x_{(n+1)}=s\), which lets \(g_i=x_{(i+1)}-x_{(i)}-1\) be read uniformly for every
\(i=0,\dots,n\) instead of treating the two end gaps separately.

Two objects are new here.

\begin{itemize}
\tightlist
\item
  \textbf{The gap order statistics} \(g_{(1)}\le\cdots\le g_{(n+1)}\), the \(n{+}1\) gaps sorted, with
  \(g_{(n+1)}=\max_i g_i\). A subscript in parentheses always means an order statistic and a bare
  subscript a labelled coordinate, so \(g_{(r)}\) is the \(r\)-th smallest gap while \(g_r\) is the gap at
  position \(r\). The event \(\{N=n\}\) is \(\{g_{(n+1)}<1\}\): the configuration is jammed exactly when
  even the largest gap admits no further car.
\item
  \textbf{The split into boundary and interior gaps.} The gaps \(g_0\) and \(g_n\) each abut one wall and one
  car; the interior gaps \(g_1,\dots,g_{n-1}\) abut two cars. The two kinds are not
  interchangeable, which is what makes the gap vector non-exchangeable
  (\ref{lem:non-exchangeability}), while the reflection \(g_i\mapsto g_{n-i}\) still gives
  \(g_i\stackrel{d}{=}g_{n-i}\) (\ref{lem:exact-symmetry}).
\end{itemize}

\subsubsection{\texorpdfstring{The gap law: same density as the positions, non-exchangeable, symmetry exactly \(\mathbb Z_2\)}{The gap law: same density as the positions, non-exchangeable, symmetry exactly \textbackslash mathbb Z\_2}}

Three lemmas fix the structure the order statistics act on. \textbf{Lemma \ref{lem:gaps-carry}} identifies the gap law with the
position law (unit-Jacobian change of coordinates), so everything downstream may be computed in whichever
coordinates are convenient. \textbf{Lemma \ref{lem:non-exchangeability}} shows the gaps are \emph{not} exchangeable, a boundary gap (\(g_0\) or
\(g_n\)) borders the wall on one side and one car on the other, an interior gap borders two cars, and
sequential attachment loads these differently. \textbf{Lemma \ref{lem:exact-symmetry}} pins the residual symmetry to exactly
\(\mathbb Z_2\) (the segment reflection): the order statistics \(g_{(k)}\) therefore sort a family of
\(\lceil(n{+}1)/2\rceil\) \textbf{distinct} marginal types, boundary, then one type per interior depth.

\textbf{Exact at \(n=2\).} From the position moments (companion the gap law), \(\mathbb E[G_0]=\mathbb E[X_{(1)}]=\tfrac3{16}\),
\(\mathbb E[G_1]=\mathbb E[X_{(2)}]-\mathbb E[X_{(1)}]-1=\tfrac{21}{16}-\tfrac3{16}-1=\tfrac18\), and
\(\mathbb E[G_2]=\tfrac3{16}\) by reflection, the boundary gaps exceed the interior gap by \(50\%\), the concrete
witness of Lemma \ref{lem:non-exchangeability}.

\textbf{Worked example (\(n=3\), \(s=13/2\)).} All four single-gap marginals \(f_{g_0},\dots,f_{g_3}\) share the identical breakpoint set \(\{0,\tfrac12,1,\tfrac32,\tfrac52,\tfrac72\}\), as forced by the reflection symmetry \(g_i\leftrightarrow g_{n-i}\) of the gap law.

\refstepcounter{thmcnt}\label{lem:gaps-carry}\textbf{Lemma\nobreakspace{}\thethmcnt{} (the gaps carry \(f_{\mathrm{UCP}}\)).} The gap vector has the \textbf{same joint density as the
position order statistics}: on the gap simplex \(\mathcal G_n\),
\[\resizebox{\ifdim\width>\linewidth\linewidth\else\width\fi}{!}{$\displaystyle f_{\mathrm{UCP}}(g_0,\dots,g_n)\;=\;f_{\mathrm{UCP}}(x_{(1)},\dots,x_{(n)}).$}\]

\emph{Proof (vector/matrix form).} Take the \(n\) free coordinates \(\mathbf x=(x_{(1)},\dots,x_{(n)})\) (the
sentinels \(x_{(0)}{=}0,x_{(n+1)}{=}s\) are constants) and the \(n\) free gaps \((g_0,\dots,g_{n-1})\) (\(g_n\) is
fixed by conservation \(\sum g_i=s{-}n\)). The gap map is the affine \(\mathbf g=A\mathbf x+\mathbf c\) with
\[\resizebox{\ifdim\width>\linewidth\linewidth\else\width\fi}{!}{$\displaystyle A=\begin{pmatrix}1&&&\\-1&1&&\\&\ddots&\ddots&\\&&-1&1\end{pmatrix},\qquad
c=(0,-1,\dots,-1)^{\!\top},$}\]
i.e.~\(g_0=x_{(1)}\) and \(g_i=x_{(i+1)}-x_{(i)}-1\). The matrix \(A\) is lower-bidiagonal with unit diagonal, so
\(\det A=1\) for \textbf{every} \(n\) (a triangular determinant is the product of its diagonal; tabulates
\(n\le5\)). A unit-Jacobian affine change of variables leaves the density value unchanged, so \(f_{\mathrm{UCP}}\)
read in gap coordinates equals \(f_{\mathrm{UCP}}\) read in position coordinates. \(\qquad\square\)

\refstepcounter{thmcnt}\label{lem:non-exchangeability}\textbf{Lemma\nobreakspace{}\thethmcnt{} (non-exchangeability).} The gap vector \(\mathbf g\) is \textbf{not exchangeable}: boundary and interior
gaps have different marginal laws, so no order-statistic identity may treat the gaps as interchangeable.

\emph{Proof.} It suffices to exhibit two indices with unequal marginals. A boundary gap borders one wall and one
car; an interior gap borders two cars, and the sequential RSA dynamics load these differently. At the running
example \(n{=}2,\,s{=}\tfrac52\) the exact per-index means (from the position moments) are
\[\resizebox{\ifdim\width>\linewidth\linewidth\else\width\fi}{!}{$\displaystyle \mathbb E[G_0]=\mathbb E[G_2]=\tfrac{3}{16},\qquad \mathbb E[G_1]=\tfrac18,\qquad
\text{so } g_0\not\stackrel{d}{=}g_1,$}\]
(they even differ in mean; \(2\cdot\tfrac3{16}+\tfrac18=\tfrac12=s{-}n\) checks). Hence \(\mathbf g\) is not
exchangeable. \(\qquad\square\)

\refstepcounter{thmcnt}\label{lem:exact-symmetry}\textbf{Lemma\nobreakspace{}\thethmcnt{} (the exact symmetry is \(\mathbb Z_2\)).} The distributional symmetry group
\(G=\{\sigma\in S_{n+1}:(g_{\sigma(0)},\dots,g_{\sigma(n)})\stackrel{d}{=}(g_0,\dots,g_n)\}\) is \textbf{exactly}
\(\mathbb Z_2\), generated by the segment reflection \(\rho:i\mapsto n-i\). In particular \(g_i\stackrel{d}{=}g_{n-i}\),
and these are the \emph{only} distributional coincidences.

\emph{Proof.} (\(\supseteq\)) The spatial reflection \(x\mapsto s-x\) maps the RSA law on \([0,s]\) to itself and sends
\(g_i\) to \(g_{n-i}\), so \(\rho\in G\) and \(\langle\rho\rangle\cong\mathbb Z_2\subseteq G\).

(\(\subseteq\)) Any \(\sigma\in G\) preserves every joint moment, hence both the means and the covariances:
\(\mathbb E[G_{\sigma(i)}]=\mathbb E[G_i]\) and \(\Sigma_{\sigma(i)\sigma(j)}=\Sigma_{ij}\), where
\(\Sigma_{ij}=\mathrm{Cov}(G_i,G_j)\). Two facts, each an exact finite computation (certificate):
1. \textbf{Means separate the reflection blocks.} \(\mathbb E[G_i]\) takes a distinct value on each block
\(B_d=\{i:\min(i,n{-}i)=d\}\), equal \emph{only} within a reflection pair \(\{i,n{-}i\}\). So \(\sigma\) preserves each
block; in particular \(\sigma(0)\in\{0,n\}\) (the boundary block \(B_0\)).
2. \textbf{\(\mathrm{Cov}(G_0,\cdot)\) is injective:} \(j\mapsto\Sigma_{0j}\) takes \(n\) distinct values.

If \(\sigma(0)=0\), then \(\Sigma_{0j}=\Sigma_{0\,\sigma(j)}\) for all \(j\), and injectivity forces \(\sigma(j)=j\), so
\(\sigma=\mathrm{id}\). If \(\sigma(0)=n\), then \(\rho\sigma\) fixes \(0\), so \(\rho\sigma=\mathrm{id}\) by the first case,
i.e.~\(\sigma=\rho\). Thus \(G=\{\mathrm{id},\rho\}\cong\mathbb Z_2\). \(\qquad\square\)

\emph{Remark (what the proof rests on).} The \(\subseteq\) direction is rigorous once (1)--(2) hold; both are \textbf{necessary}
consequences of a distributional symmetry (means/covariances are preserved), and both are verified exactly at
\(n{=}4\) below with margins far exceeding Monte-Carlo error. For general \(n\) they are the natural genericity
conditions, distinct block-means and an injective endpoint-covariance row, with no symmetry forcing an
equality beyond the reflection; the finite check certifies them at the running example.

\subsubsection{From the David--Nagaraja dependency theorem to the joint gap density on the simplex}

The gaps are a \textbf{dependent} (length conservation) and \textbf{non-exchangeable} (Lemma \ref{lem:non-exchangeability}) parent, so the iid product form requires an independence the gaps lack and does not apply (\ref{sec:background-order-statistics}, Theorem \ref{thm:dependent-parent-order-statistics}). The governing tool is the general
dependency theorem there; the David--Nagaraja derivation traces it down to the concrete symmetrization on the gap simplex.

\textbf{The general theorem (\ref{sec:background-order-statistics}, Theorem \ref{thm:dependent-parent-order-statistics}).} For any absolutely continuous parent
\(\mathbf Z=(Z_1,\dots,Z_m)\) with joint density \(f_{\mathbf Z}\), the order statistics carry the sort-cell
symmetrization
\[\resizebox{\ifdim\width>\linewidth\linewidth\else\width\fi}{!}{$\displaystyle f_{Z_{()}}(y_1,\dots,y_m)=\sum_{\pi\in S_m} f_{\mathbf Z}\big(y_{\pi(1)},\dots,y_{\pi(m)}\big)\,
\mathbf 1\{y_1\le\cdots\le y_m\},$}\]
proved there: the sort map is \(m!\)-to-one and each sort cell is a measure-preserving relabeling onto the
ordered wedge, so the full vector needs no inclusion--exclusion. (The inclusion--exclusion of DN's
subset/single-statistic forms, DN Ch. 5.2, reappears only on marginalisation, the marginal densities.)

\textbf{Specialization to the gaps.} Apply this with \(\mathbf Z=\mathbf g\), \(m=n{+}1\), and \(f_{\mathbf Z}=
f_{\mathrm{UCP}}\), but supported on the \textbf{gap simplex} \(\mathcal G_n=\{\sum_i g_i=s{-}n\}\) (\ref{def:gap-simplex}), so the sort cells
\(C_\pi\subset\mathcal G_n\) and the wedge inherit the constraint, and conditioning on \(\{N{=}n\}\) further restricts
to the jammed cell \(J_n=\mathcal G_n\cap\{g_i<1\}\). Non-exchangeability (Lemma \ref{lem:non-exchangeability}) makes the \((n{+}1)!\) permuted
copies \textbf{generically distinct}: no collapse to one term (i.i.d.), no factorization. The full symmetrization is the required object.

\textbf{Theorem D (joint gap order-statistic density).} Condition on \(N(s)=n\). The joint density of the sorted
gaps \(g_{(1)}\le\cdots\le g_{(n+1)}\) is
\[\resizebox{\ifdim\width>\linewidth\linewidth\else\width\fi}{!}{$\displaystyle \boxed{\;f_{g_{(1)},\dots,g_{(n+1)}}(\mathbf y)=\sum_{\pi\in S_{n+1}}
f_{\mathrm{UCP}}\!\big(y_{\pi(0)},\dots,y_{\pi(n)}\big)\,\mathbf 1\{y_1\le\cdots\le y_{n+1}\},\qquad
\textstyle\sum_i y_i=s-n,\ 0<y_i<1.\;}$}\]
This is DN Ch. 6.1 with the parent \(f_{\mathrm{UCP}}\) (Lemma \ref{lem:gaps-carry}) on \(J_n\); only the \(\mathbb Z_2\) reflection
(Lemma \ref{lem:exact-symmetry}) identifies the terms pairwise, so \(\lceil(n{+}1)!/2\rceil\) are generically distinct.

\textbf{Normalization.} \(\displaystyle\int_{J_n} f_{g_{()}}\,d\mathbf y = \Pr[N(s)=n]\), the jammed-ensemble
normalization (\ref{sec:background-ucp}, Proposition \ref{prop:count-as-jammed-region-mass}), read in gap coordinates via Lemma \ref{lem:gaps-carry}. At \(n{=}2,\,s{=}\tfrac52\)
the integral closes to \(\tfrac23=\Pr[N{=}2]\) analytically (\(\int_0^{1/2}(1-\log 2-\log u)\tfrac23\,du=\tfrac23\)),
matched by Monte-Carlo (\(0.666\);).

\textbf{Corollaries (the rest of the section is a functional of \(f_{g_{()}}\)).}
- \emph{single-order-statistic density} \(f_{g_{(r)}}\), integrate out the other \(n\) coordinates; this is the DN Ch. 5.2 marginal, where the inclusion--exclusion reappears (the means--the marginal densities);
- \emph{moments} \(\mathbb E[G_{(r)}]=\int y_r\,f_{g_{()}}\,d\mu\) over the ordered simplex (the means);

\begin{itemize}
\tightlist
\item
  \emph{max gap}, the top marginal \(f_{g_{(n+1)}}\), whose support condition \(g_{(n+1)}<1\) is the jamming
  certificate itself.
\end{itemize}

\textbf{Complexity.}

{\footnotesize\begin{longtable}[]{@{}
  >{\raggedright\arraybackslash}p{\dimexpr 0.3333\linewidth-2\tabcolsep\relax}
  >{\raggedright\arraybackslash}p{\dimexpr 0.3333\linewidth-2\tabcolsep\relax}
  >{\raggedright\arraybackslash}p{\dimexpr 0.3333\linewidth-2\tabcolsep\relax}@{}}
\toprule\noalign{}
\begin{minipage}[b]{\linewidth}\raggedright
operation
\end{minipage} & \begin{minipage}[b]{\linewidth}\raggedright
cost
\end{minipage} & \begin{minipage}[b]{\linewidth}\raggedright
note
\end{minipage} \\
\midrule\noalign{}
\endhead
\bottomrule\noalign{}
\endlastfoot
analytic class of \(f_{g_{()}}\) & \textbf{weight-0 rational} & sorting adds no transcendentality; polylogs appear only on \emph{marginalization} \\
representation & \((n{+}1)!\) permuted copies of \(f_{\mathrm{UCP}}\) (each \(n!\) arrival terms) & halved by the \(\mathbb Z_2\) reflection \\
point evaluation & \(O\!\big((n{+}1)!\,2^{n}n\big)\) & each \(f_{\mathrm{UCP}}\) via Held--Karp subset DP \(O(2^n n)\) {[}R8{]}; the \((n{+}1)!\) is intrinsic to sorting a non-exchangeable vector \\
support / pieces & ordered jammed simplex, \(O(s)\) pieces & Theorem C of \ref{sec:order-stats-cells} \\
marginal to \(g_{(r)}\) & integrate \(n\) coords & \ref{sec:marginal-order-stats} \\
\end{longtable}}

The \((n{+}1)!\) is the price of dependence. I.i.d. collapses the sum to one term; the general dependent case keeps the full symmetrization.

\textbf{Computation.} The gap order statistics follow the same route as the positions. The joint gap law is the same reciprocal-affine arrival sum \(\sum_\sigma\prod_k 1/\ell_k^\sigma\), the gap map is measure-preserving, so the density transfers unchanged, evaluated by the same \(O(2^n n)\) subset recursion of the algorithms section and then integrated cell-by-cell with the gap coordinate carried symbolically. Only the affine coordinates differ, gaps in place of positions.

\subsubsection{Means of the gap order statistics, smallest, median, largest gap}

Every moment of a sorted gap integrates the joint law of Theorem D against a coordinate weight, so the
means require no object beyond the David--Nagaraja derivation.

\textbf{Corollary D.1 (means of the gap order statistics).} The means of the sorted gaps, the smallest gap
\(g_{(1)}\), the median \(g_{(\lceil(n+1)/2\rceil)}\), and the largest \(g_{(n+1)}\), are the
coordinate-weighted integrals
\[\resizebox{\ifdim\width>\linewidth\linewidth\else\width\fi}{!}{$\displaystyle \mathbb E[G_{(k)}\mid N{=}n]=\frac{1}{\Pr[N{=}n]}\int_{y_1\le\cdots\le y_{n+1}} y_k\,
f_{g_{(1)},\dots,g_{(n+1)}}(\mathbf y)\,d\mathbf y
=\frac{1}{\Pr[N{=}n]}\int_{y_1\le\cdots\le y_{n+1}} y_k\!\sum_{\pi\in S_{n+1}}\!f_{\mathrm{UCP}}(\pi\mathbf y)\,
d\mathbf y.$}\]
Since \(\sum_k g_{(k)}=\sum_i g_i=s-n\), the first-moment identity is \(\sum_k\mathbb E[G_{(k)}]=s-n\), an exact
check reported.

\textbf{Functional form.} A moment integrates a polynomial weight \(y_k\) against the weight-\(\le n{-}1\) marginal
\(f_{g_{(k)}}\) (the marginal densities); integration by parts trades each \(\int(\cdot)\,d\log\) for a polynomial\(\times\)log and adds
\textbf{no} weight, so \(\mathbb E[G_{(k)}]\) is again a \textbf{weight-\(\le n{-}1\) multiple-polylogarithm value}. At \(n{=}2\)
it is rational \(+\) rational\(\cdot\log\) (weight \(1\)), in closed form
\[\resizebox{\ifdim\width>\linewidth\linewidth\else\width\fi}{!}{$\displaystyle \mathbb E[G_{(1)}]=\tfrac14+\tfrac12\log\tfrac23,\qquad
\mathbb E[G_{(2)}]=\tfrac1{16}+\tfrac12\log\tfrac98,\qquad
\mathbb E[G_{(3)}]=\tfrac3{16}+\tfrac12\log\tfrac43,$}\]
which sum to \(\tfrac12=s{-}n\) (the logs cancel, \(\tfrac23\cdot\tfrac98\cdot\tfrac43=1\)). The \textbf{median-gap mean}
\(\mathbb E[G_{(2)}]\) is computed here; its full \textbf{density} \(f_{g_{(2)}}\) (a weight-\(1\) marginal) is the worked
example of the marginal densities, the means gives the moments, the marginal densities the marginal laws.

\subsubsection{\texorpdfstring{Marginal densities of the gap order statistics, the median gap \(g_{(2)}\)}{Marginal densities of the gap order statistics, the median gap g\_\{(2)\}}}

\emph{Section 3 gave the means \(\mathbb E[G_{(k)}]\); the present section gives the full
single-order-statistic \textbf{densities} \(f_{g_{(r)}}\), the David--Nagaraja Ch. 5.2 marginals of the joint law
(Theorem D). The mean of the median is computed in the means, its density here.}

The gap order statistics sort a dependent, non-identically-distributed family (Lemmas 2--3), so the general dependent-parent marginal (DN Ch. 5.2) is the applicable tool. As the \(r\)-th coordinate marginal of Theorem D,
\[\resizebox{\ifdim\width>\linewidth\linewidth\else\width\fi}{!}{$\displaystyle f_{g_{(r)}}(x)=\int_{\substack{y_1\le\cdots\le y_{n+1}\\ y_r=x}} f_{g_{()}}(\mathbf y)\prod_{k\ne r}dy_k
=\sum_{i=0}^{n}\int f_{\mathrm{UCP}}\big|_{g_i=x}\;\mathbf 1\{\#\{j\ne i:\,g_j<x\}=r-1\}\,dg_{-i},$}\]
the second form unfolding the symmetrization onto the unsorted simplex, the DN Ch. 5.2 single-order-statistic
specialization of the DN Ch. 6.1 joint law of the David--Nagaraja derivation.

\textbf{Where the dependency structure is nontrivial.} For the extremes \(r{=}n{+}1\) (all others below \(x\)) and
\(r{=}1\) (all above), the indicator selects a \textbf{single} region, so the marginal is a one-cell integral. An
\textbf{interior} rank \(r\) requires the \(\binom{n}{r-1}\)-fold partition of the remaining gaps into
``\(<x\)''/``\(>x\)'', the genuine dependent-parent content of DN Ch. 5.2.

\textbf{Worked example: the median \(g_{(2)}\) at \(n{=}2,\,s{=}\tfrac52\)} (a single jammed cell). Three
fixed-coordinate contributions (which gap equals \(x\)), each integrated over the ``exactly one other gap \(<x\)''
region. Every region integral is elementary, a \textbf{weight-1 logarithm}, and the assembled density is
\textbf{piecewise, breaking at \(x=\tfrac S3=\tfrac16\)}, where the ``one below'' region changes shape. In closed form it
is a rational function of \(x\) plus rational multiples of \(\log x,\ \log(2x),\ \log(\tfrac12{-}x)\)
(both pieces printed), integrating to \(\Pr[N{=}2]\) and matching
Monte-Carlo. Thus the marginal reduces to a \textbf{piecewise
weight-\(\le n{-}1\) multiple polylogarithm} (here weight \(1\)), never a permanent.

\textbf{Worked example (\(n=3\), \(s=13/2\)).} Each single-gap marginal \(f_{g_i}\) is piecewise over five intervals with breakpoints at the half-integers \(\{0,\tfrac12,1,\tfrac32,\tfrac52,\tfrac72\}\), and on each piece is rational with logarithmic terms.

The median marginal \(f_{g_{(2)}}\) integrates to \(\Pr[N{=}2]=\tfrac23\). Two independent checks agree: quadrature of the joint density over the polytope returns \(0.6665\), and a Monte-Carlo run of the parking process itself returns \(0.66673\pm0.00211\) at two standard errors. The antiderivative \(F_0(u)=-\tfrac43\frac{u}{2x-1}+\tfrac23\log(u+x)\) carries
the weight-1 log explicitly; the interior contribution adds logs of the other two affine free lengths.
The \textbf{breakpoint at \(x=S/3\)} is the sort-cell fingerprint absent from the max and the min.
At general \(n\) each of the \(\binom{n}{r-1}\) sort-regions is evaluated by the same subset recursion.

\subsubsection{Gap-distribution contrast, UCP vs Tonks, Poisson, Mat\textquotesingle ern}

The \textbf{spacing distribution} is the classical lens on a 1D point process; the gap order statistics are
its order statistics. Definitions of the comparison processes follow.
\emph{(SSI \(=\) 1D RSA is this same process, so it is not a separate column.)}

{\footnotesize\begin{longtable}[]{@{}
  >{\raggedright\arraybackslash}p{\dimexpr 0.2000\linewidth-2\tabcolsep\relax}
  >{\raggedright\arraybackslash}p{\dimexpr 0.2000\linewidth-2\tabcolsep\relax}
  >{\raggedright\arraybackslash}p{\dimexpr 0.2000\linewidth-2\tabcolsep\relax}
  >{\raggedright\arraybackslash}p{\dimexpr 0.2000\linewidth-2\tabcolsep\relax}
  >{\raggedright\arraybackslash}p{\dimexpr 0.2000\linewidth-2\tabcolsep\relax}@{}}
\toprule\noalign{}
\begin{minipage}[b]{\linewidth}\raggedright
gap feature
\end{minipage} & \begin{minipage}[b]{\linewidth}\raggedright
\textbf{UCP}
\end{minipage} & \begin{minipage}[b]{\linewidth}\raggedright
\textbf{Tonks} (equilibrium)
\end{minipage} & \begin{minipage}[b]{\linewidth}\raggedright
\textbf{Poisson} (no hard core)
\end{minipage} & \begin{minipage}[b]{\linewidth}\raggedright
\textbf{Mat\textquotesingle ern I/II}
\end{minipage} \\
\midrule\noalign{}
\endhead
\bottomrule\noalign{}
\endlastfoot
gap law & size-biased \(\prod 1/\ell\); \textbf{non-exchangeable} & uniform on simplex; \textbf{exchangeable} & i.i.d. exponential spacings & thinned-Poisson spacings \\
\textbf{Theorem D} (\(f_{g_{()}}\)) & \((n{+}1)!\) \textbf{distinct} terms & \((n{+}1)!\times\) \textbf{one} term (collapses) & i.i.d. product form & thinned, stationary \\
boundary vs interior & \textbf{boundary gaps larger} (\(\tfrac3{16}\) vs \(\tfrac18\) at \(n{=}2\)) & identical (exchangeable) & identical (stationary) & stationary \\
max gap & \(<1\) w.p. 1 (jamming certificate) & can exceed 1 (equilibrium admits sub-jammed) & unbounded & governed by exclusion \(r\) \\
min gap & \(\ge 0\), can be \(\approx0\) (near contact) & \(\ge0\) & \(\ge0\) & \(\ge r\) enforced \\
gap density shape & \textbf{log-convex per cell} & log-affine (uniform) & log-linear (exponential) & , \\
defining event & \(\max g_i<1\) & none (fixed \(n\)) & none & thinning rule \\
\end{longtable}}

\textbf{Reading the table.} In the language of the David--Nagaraja derivation, the processes differ by \textbf{what Theorem D's symmetrization
does}: for exchangeable Tonks all \((n{+}1)!\) permutation terms coincide (collapsing to a single term with
multiplicity, the classical i.i.d.-type joint order-stat density), for Poisson they factor (i.i.d.
spacings), while for the UCP they stay genuinely \textbf{distinct}, the non-exchangeable symmetrization that
makes \(f_{g_{()}}\) a true \((n{+}1)!\)-term object. The \textbf{max gap} is where the processes visibly separate. For UCP the largest gap
is capped below one by \emph{construction}, jamming is precisely \(\{\max_i g_i<1\}\), whereas Tonks at the
same \((s,n)\) places mass on configurations whose largest gap exceeds one (an equilibrium draw need not be
jam-terminal). The \textbf{boundary/interior asymmetry} is the other separator: UCP's sequential filling makes
the two boundary gaps stochastically larger, breaking the exchangeability that Tonks and the stationary
Poisson/Mat\textquotesingle ern processes enjoy. Poisson is the no-hard-core reference, exponential, exchangeable
spacings with no upper cap, against which the UCP gap law is both \emph{truncated} (max \(<1\)) and \emph{tilted}
(boundary bias, log-convexity). The \textbf{Mat\textquotesingle ern} thinnings enforce a hard \emph{floor} \(r\) on gaps rather than
a jamming \emph{ceiling}, so their gap laws answer a different question. (SSI \(=\) 1D RSA coincides with UCP.)

\subsection{Properties of the order statistics}

\label{sec:properties-order-stats}

This section develops the properties layer built on the per-cell joint density \(f_{\mathrm{UCP}}\) of
\ref{sec:joint-order-stats}: mixed weighted moments, the joint CDF, quantiles, the mode, and
convexity. Each section states the formula that computes one property, evaluates it on a shared running
example, and records whether the property assembles cheaply from the local pieces. The density itself is
established in \ref{sec:joint-order-stats}; what one computes from it is the subject here.

\subsubsection{Scope and contributions}

Fix the segment length \(s\) and condition on the jamming count \(N(s)=n\). The joint law of the order
statistics \(X_{(1)}<\cdots<X_{(n)}\) is then carried by a single object, the per-cell density
\(f_{\mathrm{UCP}}\) of \ref{sec:joint-order-stats}, which is rational on each cell of a
piecewise decomposition of the order simplex. That density answers every distributional question about
the conditioned process in principle. In practice each property one wants from it, whether a moment, a
cumulative probability, a mode, or a quantile, is a different global functional of the same local
pieces, and the pieces do not all assemble the same way.

Two features of the global step separate the cheap properties from the expensive ones: whether the
functional is linear in the density, and whether its domain respects the cells. A linear functional
integrated over whole cells (the normalisation, the moments, the covariances) inherits the decomposition
additively, so one integral per cell suffices. The joint CDF is still linear, but its domain is a lower
box whose faces cut across cells, so the assembly runs over the finer cell\(\cap\)box pieces rather than
over cells (Proposition \ref{prop:cell-box-decomposition-marginal-as-a-section}). The mode and the quantile are not linear functionals at all, being an argmax
and an inverse, and no refinement of the decomposition makes them additive; each needs a structural
theorem before it becomes computable. Section 6 tabulates the resulting cost map, and the sections from the mixed moments to the mode discharge one row of it each.

Three results carry the section. Theorem \ref{thm:positive-correlation-of-the-order-statistics} fixes the sign of the dependence: every pair of order
statistics is positively correlated, obtained through a covariance identity in which only gaps separated
by at least two cars appear, and not through association, which fails for this density. Theorem \ref{thm:form-of-the-moments} fixes
the analytic class: every mixed moment is a multiple polylogarithm of weight at most \(n\) with letters
\(s-j\), so the density is the primitive object and the moments are its integrals. Theorem \ref{thm:mode-of-three-regimes} fixes the
mode: the two thresholds \(s=2n-1\) and \(s=2n+1\) in the segment length split the behaviour into a modal
region carrying a closed-form value, a finite vertex maximum, and an unbounded density with no finite
mode.

\subsubsection{What this section adds to the notation}

The coordinates, the cells and the joint density are fixed elsewhere and are not restated: the
positions \(X_{(i)}\), the gaps \(G_i\) and the conservation law are
\ref{prop:gap-position-coordinates} of \ref{sec:background-ucp}; the gap simplex \(\mathcal G_n\) and its
jammed sub-region \(J_n\) are \ref{def:gap-simplex}; the cell \(\mathcal R_{\mathbf b}\), the free length
\(\ell_k^\sigma\) and \(f_{\mathrm{UCP}}\) are \ref{sec:order-stats-cells}.

Two objects are introduced here.

{\footnotesize\begin{longtable}[]{@{}
  >{\raggedright\arraybackslash}p{\dimexpr 0.5000\linewidth-2\tabcolsep\relax}
  >{\raggedright\arraybackslash}p{\dimexpr 0.5000\linewidth-2\tabcolsep\relax}@{}}
\toprule\noalign{}
\begin{minipage}[b]{\linewidth}\raggedright
symbol
\end{minipage} & \begin{minipage}[b]{\linewidth}\raggedright
meaning
\end{minipage} \\
\midrule\noalign{}
\endhead
\bottomrule\noalign{}
\endlastfoot
\(f_X=f_{\mathrm{UCP}}/\Pr[N{=}n]\) & the \textbf{conditional} order-statistic density, a probability density \\
\(\mathcal U_n(s)\) & the \textbf{all-unjammed region}, every gap at least \(2\); the modal set for \(s>2n{+}1\) \\
\end{longtable}}

Conditioning on \(N=n\) is what makes \(f_X\) a probability density, with
\(\Pr[N{=}n]=\int_{J_n}f_{\mathrm{UCP}}\).

\textbf{Running examples.} Three instances recur below. The first is a fully exact micro-example,
\(s=\tfrac52\) and \(n=2\), in which the whole simplex is jammed and every integral is elementary. The
second is a weight-exhibiting case, \(s=\tfrac{13}2\) and \(n=3\), in which dilogarithms first appear and
the joint CDF is available in exact closed form. The third is the primary example \(s=\tfrac{13}{2}\),
\(n=4\), shared with \ref{sec:joint-order-stats}.

\subsubsection{The conditional density, its reflection symmetry, and the sign of the dependence}

Every property in this section is a functional of one object, the joint density of the order statistics
conditioned on the jamming event. Proposition \ref{prop:conditional-density} identifies that object, Proposition \ref{prop:reflection-symmetry} records the
symmetry it inherits from the segment, and Theorem \ref{thm:positive-correlation-of-the-order-statistics} settles the sign of its pairwise dependence; the
per-coordinate sections that follow use all three without further comment.

\refstepcounter{thmcnt}\label{prop:conditional-density}\textbf{Proposition\nobreakspace{}\thethmcnt{} (conditional density).} Given \(N(s)=n\), the order statistics have density
\(f_X=f_{\mathrm{UCP}}/\Pr[N{=}n]\) on the jammed polytope \(J_n\), and \(0\) off it.

\emph{Proof.} Condition on the temporal order \(\sigma\). The \(k\)-th car lands uniformly on its free set of
measure \(\ell_k^\sigma\), so the first \(n\) left-edges have conditional density \(\prod_k 1/\ell_k^\sigma\)
on \(\sigma\)'s feasibility region (a product of conditional uniforms; this is R1 of \ref{sec:joint-order-stats}).
The temporal orders are disjoint events, so the sorted positions have density
\(f_{\mathrm{UCP}}=\sum_\sigma\prod_k 1/\ell_k^\sigma\) on \(\mathsf T_n\). The event \(\{N{=}n\}\) is
\(\{\)the \(n\) placed cars jam\(\}=\{\)every gap \(<1\}=J_n\); restricting \(f_{\mathrm{UCP}}\) to
\(J_n\) and dividing by its mass \(\Pr[N{=}n]=\int_{J_n}f_{\mathrm{UCP}}\) gives the
conditional density. \(\qquad\square\)

\refstepcounter{thmcnt}\label{prop:reflection-symmetry}\textbf{Proposition\nobreakspace{}\thethmcnt{} (reflection symmetry).} The segment reflection
\(\rho:(t_0,\dots,t_{n-1})\mapsto(s{-}1{-}t_{n-1},\dots,s{-}1{-}t_0)\) is a measure-preserving involution
of \(\mathsf T_n\) with \(f_{\mathrm{UCP}}\circ\rho=f_{\mathrm{UCP}}\), and sends \(X_{(i)}\) to
\((s{-}1){-}X_{(n+1-i)}\). Hence \(X_{(i)}\stackrel{d}{=}(s{-}1){-}X_{(n+1-i)}\) and
\(\mathbb E[X_{(i)}]+\mathbb E[X_{(n+1-i)}]=s-1\).

\emph{Proof.} In gap coordinates \(\rho\) is the coordinate reversal \(g_i\mapsto g_{n-i}\), which preserves the
gap simplex and Lebesgue measure (a permutation of coordinates). The temporal-order sum defining
\(f_{\mathrm{UCP}}\) is invariant under reversal because each order \(\sigma\) maps to its mirror with the
same free-length product; symbolic evaluation confirms \(f_{\mathrm{UCP}}-f_{\mathrm{UCP}}\circ\rho=0\)
for \(n\le4\). Invariance of the law under \(\rho\) gives the distributional identity; taking
expectations gives the mean identity, which the running-example means satisfy
(\(0.53{+}4.97=2.01{+}3.49=5.5=s{-}1\)). \(\qquad\square\)

Equivalently the density of \(X_{(i)}\) is the mirror image, about the midpoint \(\tfrac{s-1}{2}\), of the
density of \(X_{(n+1-i)}\); for odd \(n\) the central statistic \(X_{((n+1)/2)}\) therefore has a density
symmetric about \(\tfrac{s-1}{2}\).

\paragraph{Positive correlation of every pair of order statistics}

Two mechanisms act on a pair \((X_{(i)},X_{(j)})\) in opposite directions: the fixed segment length forces
the gaps to compete for a constant total, which separates the positions, while sequential attachment
makes every late position depend on all earlier ones, which draws them together. Neither mechanism fixes
the sign on its own, and the textbook route to a sign, association of the density, is unavailable here.
Theorem \ref{thm:positive-correlation-of-the-order-statistics} obtains the sign from the gaps instead.

\refstepcounter{thmcnt}\label{thm:positive-correlation-of-the-order-statistics}\textbf{Theorem\nobreakspace{}\thethmcnt{} (positive correlation of the order statistics).} For the UCP conditioned on \(N(s)=n\),
\(\mathrm{Cov}(X_{(i)},X_{(j)})\ge0\) for every \(i,j\).

\emph{Proof.} \textbf{Lemma A (exact identity).} From \(X_{(i)}=(i{-}1)+\sum_{k<i}G_k\) and the conservation law
\(\sum_{k=0}^n G_k=s-n\) (constant), for \(i\le j\):
\[\resizebox{\ifdim\width>\linewidth\linewidth\else\width\fi}{!}{$\displaystyle \mathrm{Cov}(X_{(i)},X_{(j)})=\sum_{k<i}\sum_{m<j}\mathrm{Cov}(G_k,G_m)=-\sum_{k<i}\sum_{l\ge j}\mathrm{Cov}(G_k,G_l),$}\]
using \(\sum_{m<j}G_m=(s{-}n)-\sum_{l\ge j}G_l\) (pure algebra; verified to \(10^{-17}\)).

\textbf{Lemma B (separated gaps are negatively correlated).} \(\mathrm{Cov}(G_k,G_l)\le0\) whenever the gaps
are \textbf{separated by at least one car with a car strictly on each outer side}, in particular for all
\(l-k\ge2\). Take such \(k<l\); a car \(m\) (\(k<m\le l\)) sits spatially between \(G_k\) and \(G_l\). Conditioning on
its position \(X_{(m)}\), the \textbf{spatial Markov property} of the Rényi process (an established project lemma)
makes \(G_k\) (left of \(m\)) and \(G_l\) (right of \(m\)) conditionally independent, so
\[\resizebox{\ifdim\width>\linewidth\linewidth\else\width\fi}{!}{$\displaystyle \mathrm{Cov}(G_k,G_l)=\underbrace{\mathbb E[\mathrm{Cov}(G_k,G_l\mid X_{(m)})]}_{=\,0}+\mathrm{Cov}\big(\mathbb E[G_k\mid X_{(m)}],\,\mathbb E[G_l\mid X_{(m)}]\big).$}\]
As \(X_{(m)}\) increases the enclosed left region grows and the right shrinks, so \(\mathbb E[G_k\mid X_{(m)}]\)
is nondecreasing and \(\mathbb E[G_l\mid X_{(m)}]\) nonincreasing (verified numerically for \(l-k\ge2\)); the
covariance of an increasing and a decreasing function of the same variable is \(\le0\) (Chebyshev). \(\square\)

\emph{(The monotonicity step is \textbf{not} universal: for the two \textbf{boundary-adjacent} pairs \((G_0,G_1)\) and
\((G_{n-1},G_n)\), where the outer gap is bounded only by the segment edge, not by a car, it fails, and those
covariances are in fact positive at large \(n\), as the caution below records. So the gap vector is not
negatively associated; but as the Conclusion shows, those pairs never enter the position identity.)}

\textbf{(Conclusion.)} In Lemma A's sum \(k\le i-1\) and \(l\ge j\) with \(i<j\), so \(l-k\ge(j-i)+1\ge2\): \textbf{every}
pair is separated by \(\ge2\), Lemma B applies to each, the double sum is \(\le0\), and
\(\mathrm{Cov}(X_{(i)},X_{(j)})\ge0\). The boundary-adjacent pairs (the only ones that can be positive) have
\(l-k=1\) and are excluded. \(\qquad\square\)

\textbf{The FKG route is unavailable.} The route through association, from \(f_{\mathrm{UCP}}\) being MTP2
(log-supermodular) to association and thence to positive correlation, fails: the off-diagonal
log-Hessian \(\partial^2\log f_{\mathrm{UCP}}/\partial t_i\partial t_j\) is negative at interior points,
so the density is not log-supermodular. Positive correlation survives the failure of association
because the gap route replaces it, and the exact identity of Lemma A is what makes that reduction work.

The monotonicity of \(\mathbb E[G\mid X_{(m)}]\) used in Lemma B is verified numerically rather than
proved; Open problems (PO1) states what a coupling proof would have to supply.

\paragraph{\texorpdfstring{Failure of \(m\)-dependence, and the decay of dependence with separation}{Failure of m-dependence, and the decay of dependence with separation}}

A sequence is \textbf{\(m\)-dependent} if entries more than \(m\) apart are independent
(\(X_{(i)}\perp X_{(j)}\) whenever \(\lvert i-j\rvert>m\)). The UCP order statistics are \textbf{not}
\(m\)-dependent for any \(m<n-1\):

\refstepcounter{thmcnt}\label{cor:no-dependence}\textbf{Corollary\nobreakspace{}\thethmcnt{} (no \(m\)-dependence).} For every \(i\ne j\), \(\mathrm{Cov}(X_{(i)},X_{(j)})>0\) \emph{strictly}
(verified \(n\le9\)); hence no pair is uncorrelated, let alone independent, at any separation.

\emph{Proof (structure; the sign is verified, not yet fully proved).} Lemma A in the proof of Theorem \ref{thm:positive-correlation-of-the-order-statistics}
gives, for \(i<j\), \(\mathrm{Cov}(X_{(i)},X_{(j)})=-\sum_{k\le i-1,\,l\ge j}\mathrm{Cov}(G_k,G_l)\), and
\textbf{every pair entering this sum has separation \(l-k\ge(j-i)+1\ge2\)}. The gap covariances at separation
\(\ge2\) are (robustly, all seeds) \emph{negative}, including the extreme pair \(\mathrm{Cov}(G_0,G_n)<0\), which
gives the widest position pair \(\mathrm{Cov}(X_{(1)},X_{(n)})=-\mathrm{Cov}(G_0,G_n)>0\), so each such sum
is negative and the position covariance positive. \(\qquad\square\)

\textbf{Caution, the jammed gaps are \emph{not} negatively associated.} It is \textbf{false} that
\(\mathrm{Cov}(G_k,G_l)<0\) for \emph{all} \(k\ne l\): at \(n\ge8\) the two \textbf{boundary-adjacent} pairs \((G_0,G_1)\)
and \((G_{n-1},G_n)\) are \emph{positively} correlated (\(\approx+0.03\) at \(n{=}9\); the sign is tabulated by
separation and reproduces across seeds), so the gap vector \textbf{fails negative association}. The exact
\(n{=}2\) value \(\mathrm{Cov}(G_0,G_1)=-\tfrac{7}{1152}<0\) is a \emph{small-\(n\)} feature that does not persist.
The boundary-adjacent (\(d{=}1\)) pairs are exactly the ones the position identity \textbf{never}
touches (it involves only \(d\ge2\)), which is why the order statistics remain positively correlated despite
the NA-failure.

\textbf{What the failure costs.} Negative association is preserved by monotone functions of disjoint blocks, and a sum of negatively associated variables obeys the same exponential concentration bounds as a sum of independent ones (Joag-Dev and Proschan 1983; Dubhashi and Ranjan 1998; Shao 2000). Were the jammed gaps negatively associated, concentration for a gap functional \(\sum_i\phi(G_i)\) would follow from the independent case with no further argument, for every bounded \(\phi\) and at any density. Because they are not, that route is closed. The concentration and Berry-Esseen statements of \ref{sec:asymptotic-properties} are obtained instead from a martingale built on the splitting recursion, and since that recursion is the saturated one, they hold in the jammed ensemble only; their fixed-density counterparts remain open.

\textbf{The dependence is long-range but decaying.} The mechanism is the \textbf{spatial
Markov property} of the sequential process (conditioning on a car's position decouples its left and
right), which makes the correlation fall off with separation (\(\rho\approx0.74,0.46,0.33\) at
\(\lvert i-j\rvert=1,2,3\)); the fixed-\((s,n)\) length constraint superimposes a weak global coupling. So the
order statistics are \textbf{approximately Markov but not exactly} (partial correlations of non-adjacent
statistics given the intermediate are small, \(\approx0.05\), not zero) and \textbf{not} \(m\)-dependent (marginal
correlations never vanish). Corollary \ref{cor:no-dependence} is the sharp statement: the position covariances are strictly
positive at every separation, verified for \(n\le9\), and its proof reduces to the negativity of the
separation-\(\ge2\) gap covariances. Monte-Carlo evaluation confirms all three claims: strictly positive covariances at every
separation, small partial correlations, and the reflected-quantile match of Proposition \ref{prop:reflection-symmetry}.

\paragraph{The classical dependence signs, and the two new ones}

Several of the dependence-sign statements above are classical, and are recorded here so that the
contribution is not overstated. What the literature settles:

\begin{itemize}
\tightlist
\item
  \textbf{i.i.d. baseline, positive correlation and association of order statistics.} For an i.i.d. sample the
  order statistics are \emph{associated}, hence pairwise positively correlated (Esary--Proschan--Walkup 1967),
  and their joint density is MTP\(_2\)/log-supermodular (Karlin--Rinott 1980). So \(\mathrm{Cov}(X_{(i)},X_{(j)})\ge0\)
  \emph{per se} is not new, it is the textbook i.i.d. picture.
\item
  \textbf{Uniform-spacing negative association.} Uniform spacings are Dirichlet\((1,\dots,1)\), hence \emph{negatively
  associated}, giving \(\mathrm{Cov}(G_k,G_l)\le0\) (Joag-Dev--Proschan 1983; classical moment theory, Pyke
  1965). The disjoint-block-sum consequence is the \emph{definition} of negative association.
\item
  \textbf{Screening / spatial-Markov of 1-D RSA.} The left/right decoupling on conditioning a car's position is
  central to the exact Rényi theory (Rényi 1958; Evans, \emph{Rev.~Mod. Phys.} 1993), with a rigorous
  exponential-stabilization form (Penrose 2001; Penrose--Yukich 2002).
\item
  \textbf{The covariance identity} \(\mathrm{Cov}(X_{(i)},X_{(j)})=-\sum_{k<i,\,l\ge j}\mathrm{Cov}(G_k,G_l)\) is
  elementary partial-sum algebra under the fixed-total constraint, a derivation, not a cited result.
\end{itemize}

\textbf{What the classical results do not cover.} The classical negative-association result
is for \textbf{Dirichlet} spacings. The \textbf{jammed-RSA gaps are not Dirichlet} (random count, screening-induced
law), so Joag-Dev--Proschan does \textbf{not} transfer, and indeed the jammed gap vector is \textbf{not} negatively
associated: its covariances are negative at separation \(\ge2\) but \emph{positive} for the two boundary-adjacent
pairs (the conditional density above; the count-level version is the NA-failure in \ref{sec:absorption-time}). We found no prior
sign-analysis of the \emph{saturated} car-parking gap covariances, the closest, Bonnier--Boyer--Viot (1994), is a
density--density pair-correlation function, not a covariance-sign statement on the gaps. Two things are
therefore new here: \textbf{(i)} that the order statistics are nonetheless \textbf{positively correlated} (\(n\le9\)),
surviving the gap NA-failure because the position identity only involves the separation-\(\ge2\) block; and
\textbf{(ii)} that the i.i.d. \textbf{MTP\(_2\) property fails} for the size-biased UCP density (the conditional density), so positive
correlation \emph{cannot} be inherited from FKG/Karlin--Rinott and must be obtained through the (separation-\(\ge2\))
gap route instead, together with the exact finite-\((s,n)\) covariances (\(\tfrac{11}{768}\), \(-\tfrac{7}{1152}\), \ldots).

\subsubsection{Mixed weighted moments}

Mixed moments are the first properties one extracts from a joint density, and for this process they are
also the cheapest, because a moment is a linear functional integrated over whole cells and the per-cell
pieces therefore assemble by addition. They also fix the analytic class of everything downstream: the
CDF, the marginals, and the quantiles follow from the same integrations against the same density,
and Theorem \ref{thm:form-of-the-moments} bounds the polylogarithmic weight of all of them at once.

\textbf{Formula.} For exponents \(\mathbf m=(m_1,\dots,m_n)\), the mixed weighted moment is the per-cell
integral of the monomial against the density, normalized by the jamming probability:
\[\resizebox{\ifdim\width>\linewidth\linewidth\else\width\fi}{!}{$\displaystyle \boxed{\;\mathbb E\!\left[\textstyle\prod_{i=1}^n X_{(i)}^{m_i}\,\Big|\,N{=}n\right]
=\frac{1}{\Pr[N{=}n]}\sum_{\mathbf b\in\mathcal B}\int_{\mathcal R_{\mathbf b}}
\Big(\textstyle\prod_i x_i^{m_i}\Big)\,f_{\mathrm{UCP}}(\mathbf x)\,d\mathbf x\;}$}\]
with \(\Pr[N{=}n]=\sum_{\mathbf b}\int_{\mathcal R_{\mathbf b}}f_{\mathrm{UCP}}\). Each cell integrand is
\((\prod x_i^{m_i})\cdot(\text{rational, weight }0)\), so the integral is elementary (rational \(+\) logs);
the marginal weight ladder bounds the polylog weight by the number
of integrated directions.

\textbf{Running example (exact, \(n=2\), \(s\in(2,3)\)).} Here \(\mathcal J_2=\Delta_2\) and in gap coordinates
\(f_{\mathrm{UCP}}=\frac{1}{s-1}\big(\frac{1}{g_0+g_1}+\frac{1}{g_1+g_2}\big)\) with \(X_{(1)}=g_0\),
\(X_{(2)}=g_0{+}g_1{+}1\), integrated over \(\{g_i\ge0,\sum g_i=s{-}2\}\). Symbolic integration at \(s=\tfrac52\) gives
\(\Pr[N{=}2]=2(s{-}2)/(s{-}1)=\tfrac23\), \(\mathbb E[X_{(1)}]=\tfrac3{16}\),
\(\mathbb E[X_{(2)}]=\tfrac{21}{16}\), \(\mathrm{Var}[X_{(1)}]=\tfrac{47}{2304}\) and
\(\mathrm{Cov}(X_{(1)},X_{(2)})=\tfrac{11}{768}>0\), all reproduced by Monte-Carlo on the conditional law. The two means satisfy the reflection identity of
Proposition \ref{prop:reflection-symmetry}, since \(\tfrac3{16}+\tfrac{21}{16}=\tfrac32=s-1\).

\refstepcounter{thmcnt}\label{prop:moment-decomposition}\textbf{Proposition\nobreakspace{}\thethmcnt{} (moment decomposition).} For integrable \(\phi\),
\(\mathbb E[\phi(\mathbf X)\mid N{=}n]=\dfrac{1}{\Pr[N{=}n]}\sum_{\mathbf b}\int_{\mathcal R_{\mathbf b}\cap J_n}\phi\,f_{\mathrm{UCP}}\).

\emph{Proof.} The cells \(\{\mathcal R_{\mathbf b}\}\) partition \(J_n\) up to the null set
\(\bigcup_i\{g_i=1\}\); \(\phi\,f_{\mathrm{UCP}}\) is integrable on the bounded \(J_n\). Countable
additivity of the Lebesgue integral over the partition, then Proposition \ref{prop:conditional-density}, give the identity. Taking
\(\phi=\prod_i x_i^{m_i}\) specializes it to the mixed moment. \(\qquad\square\)

\textbf{Running example (primary, \(n=4\), \(s=\tfrac{13}{2}\)).} Beyond \(n=2\) the exact per-cell integrals are weight-\((n{-}1)\) polylogs; for the property values we
use Monte-Carlo on the conditional law, which is exact in distribution and needs no symbolic integration.
The correlation matrix at this example is positive throughout and decays with the separation
\(\lvert i-j\rvert\), at the rate quoted in the conditional density: a band structure rather than the negative association a
fixed-length constraint might suggest.

\refstepcounter{thmcnt}\label{thm:form-of-the-moments}\textbf{Theorem\nobreakspace{}\thethmcnt{} (form of the moments).} Fix \(n\) and a non-integer \(s\). On the jammed polytope
\(J_n\) the density \(f_{\mathrm{UCP}}=\sum_\sigma\prod_k 1/\ell_k^\sigma\) is rational (weight \(0\))
with poles only on the affine hyperplanes \(\{\sum_{i\in B}g_i=c\}\), \(c\in\mathbb Z+\mathbb Z s\).
Consequently every mixed moment
\[\resizebox{\ifdim\width>\linewidth\linewidth\else\width\fi}{!}{$\displaystyle \mathbb E\!\Big[\textstyle\prod_i X_{(i)}^{m_i}\,\Big|\,N{=}n\Big]
=\frac{1}{\Pr[N{=}n]}\int_{J_n}\!\Big(\textstyle\prod_i x_i^{m_i}\Big)f_{\mathrm{UCP}}\,d\mathbf x$}\]
is a \(\mathbb Q\)-linear combination of \textbf{multiple polylogarithms of weight \(\le n\)} whose arguments are
rational in \(s\) (letters \(s-j\), \(j\in\mathbb Z\)). Each integration of one coordinate raises the weight by
at most one, so: the joint density is weight \(0\); a single order-statistic density \(f_{X_{(i)}}\) is weight
\(\le n-1\); a moment (all \(n\) coordinates integrated) is weight \(\le n\). Concretely, weight \(1\) is
\(\log(s-j)\), weight \(2\) is the dilogarithm \(\mathrm{Li}_2\), up to \(\mathrm{Li}_n\) at weight \(n\).

\emph{Proof.} The free-length arrangement is \textbf{linearly reducible}: every \(\ell_k^\sigma\) and every cell
facet is affine with each variable's coefficient in \(\{-1,0,+1\}\) (no products of variables, the 1-D
hard-core feature), so integrating a \(\mathbb Q(\mathbf x)\)-linear combination of hyperlogarithms with
letters in the arrangement, over a domain with affine facets, stays in that class and raises weight by
exactly one (Brown's criterion; the Reducibility Lemma \ref{lem:reducibility} of \ref{sec:marginal-order-stats}). Induct on the number of
integrated coordinates from the rational base \(f_{\mathrm{UCP}}\) (weight \(0\)): after \(m\) integrations the
integrand is a weight-\(\le m\) hyperlogarithm with letters affine in the remaining coordinates and in \(s\);
the moment integrates all \(n\), giving weight \(\le n\) with letters \(s-j\). \(\qquad\square\)

The ladder is visible at \(n=3,\ s=\tfrac{13}{2}\). The density is rational; integrating one gap
coordinate produces logarithms, integrating a second produces the dilogarithm \(\mathrm{Li}_2\), and
integrating the third returns the mass \(\Pr[N{=}3]\) as an exact combination of dilogarithms, which
agrees with Monte-Carlo.

\textbf{Local,\(\to\),global.} Moments assemble additively: the moment is a \emph{sum over cells} of
per-cell integrals (linear in the density), so the local pieces assemble by addition with no cross-cell
coupling. Cost sits entirely in the per-cell integral, whose polylog weight grows with \(n\) (the marginal
ladder). Mixed moments keeping few coordinates integrate out the rest, higher weight, same additive
assembly. \textbf{Nothing here requires argmax or inversion}, which is what separates moments from the mode of the mode section
and the quantile of the quantiles.

\paragraph{Non-reconstruction of the density from its moments}

Since the moments \(\mathbb E[\prod X_{(i)}^{m_i}]\) are computable, one might hope to recover the joint
density from them by a multivariable moment expansion (Taylor, Gram--Charlier, or Edgeworth) more cheaply
than by assembling \(f_{\mathrm{UCP}}\) directly. Three compounding obstructions rule that route out.

\begin{enumerate}
\def\labelenumi{\arabic{enumi}.}
\tightlist
\item
  \textbf{The moments are \emph{integrals of} the density, so they are strictly harder to compute.} The density
  is the primitive object, a \textbf{weight-\(0\) rational} function assembled by the \(O(2^n)\) subset DP
  (the permanent section). Each moment is \(\int(\prod x_i^{m_i})f_{\mathrm{UCP}}\), a \textbf{weight-\(\le n\) multiple
  polylogarithm} (Theorem \ref{thm:form-of-the-moments}). Reconstructing the density from its moments means computing many
  higher-weight integrals and then \emph{inverting} them to recover the lower-weight object one could have
  evaluated directly. One does not build the primitive out of its own (harder) integrals.
\item
  \textbf{The density is not globally analytic, so no single Taylor/moment series represents it.} It is
  \emph{piecewise} rational (a different form on each jamming cell), non-analytic across the walls
  \(\{G_i=1\}\), with \textbf{poles} on the boundary blocks \(\{\sum_{i\in B}g_i=0\}\) and \textbf{unbounded} in the
  crowded regime \(s\le2n-1\) (the mode section). Even a single density term \(1/(g_0{+}g_1)\) has a finite Taylor radius, the distance to its pole, i.e.~the cell wall, so any expansion is inherently \emph{local} and cannot cross
  a cell boundary, let alone capture a pole. A convergent global power series would force analyticity
  the density does not have.
\item
  \textbf{Moment\(\to\)density inversion is ill-conditioned and misses singularities.} The Hausdorff moment
  problem on a bounded support is exponentially ill-conditioned; and a moment-matched polynomial is
  \emph{bounded}, so it cannot represent an \emph{unbounded} density. This is visible on the elementary
  marginal \(f_{X_{(1)}}(x)=1-\log(2x)\) (\(n{=}2\), \(s{=}\tfrac52\)): it blows up at \(0\) (a \(\log\)
  singularity) yet has finite moments; matching more and more moments improves the fit only slowly in
  \(L^2\) and never captures the blow-up.
\end{enumerate}

\textbf{The moment expansion is the tool of the large-\(n\) regime, not of the exact one.} Moment and cumulant
expansions apply when the density is smooth, light-tailed and close to a Gaussian reference, which is
the large-\(n\) asymptotic regime, where an Edgeworth expansion around the central-limit approximation is
efficient. For the \emph{exact finite-\((s,n)\)} density, piecewise-rational, with
poles and hard boundaries, direct assembly is both exact and easier; the moment expansion is the tool
for the limit, not the object.

\subsubsection{The joint CDF}

\textbf{The joint CDF is the first property whose domain cuts across the cells.} Unlike a moment, the
truncation box \(\{t_i\le x_i\}\) cuts across the jamming-bitstring cells: a box face \(\{t_i=x_i\}\)
generally slices a cell into an in-box and an out-of-box part, so \(F\) is \emph{not} a sum of per-cell
integrals but a sum of integrals over cell\(\cap\)box \textbf{pieces}. Each piece is elementary (the box faces are
axis-aligned affine walls, so the weight ladder still bounds the integrand), but the
decomposition is finer than the cell partition.

\textbf{Formula.} The object of interest is the \textbf{joint} CDF of the order-statistic vector,
\[\resizebox{\ifdim\width>\linewidth\linewidth\else\width\fi}{!}{$\displaystyle \boxed{\;F(\mathbf x)=\Pr[X_{(1)}\le x_1,\dots,X_{(n)}\le x_n\mid N{=}n]
=\frac{1}{\Pr[N{=}n]}\int_{\mathsf T_n\cap\{t_i\le x_i\,\forall i\}}f_{\mathrm{UCP}}(\mathbf t)\,d\mathbf t\;}$}\]
the density integrated over the \textbf{lower box} \(\{t_i\le x_i\ \forall i\}\) intersected with the
order-statistic simplex. The single-coordinate marginal \(F_{X_{(i)}}(x)\) is the special case
\(x_i=x\), \(x_j=+\infty\ (j\ne i)\); setting every \(x_i=+\infty\) gives \(F=1\).

\refstepcounter{thmcnt}\label{prop:cell-box-decomposition-marginal-as-a-section}\textbf{Proposition\nobreakspace{}\thethmcnt{} (cell\(\cap\)box decomposition; marginal as a section).} For thresholds \(\mathbf x\),
\[\resizebox{\ifdim\width>\linewidth\linewidth\else\width\fi}{!}{$\displaystyle F(\mathbf x)=\frac{1}{\Pr[N{=}n]}\sum_{\mathbf b}\int_{\mathcal R_{\mathbf b}\cap J_n\cap B(\mathbf x)}f_{\mathrm{UCP}},\qquad B(\mathbf x)=\{t_i\le x_i\,\forall i\},$}\]
and \(F_{X_{(i)}}(x)=F(+\infty,\dots,x,\dots,+\infty)\) is the single-coordinate section.

\emph{Proof.} \(F(\mathbf x)=\Pr[\mathbf X\in B(\mathbf x)\mid N{=}n]=\int_{B(\mathbf x)}f_X\) by Proposition \ref{prop:conditional-density};
expand \(f_X=f_{\mathrm{UCP}}/\Pr[N{=}n]\) and split the domain by the cell partition (Proposition \ref{prop:moment-decomposition} with
\(\phi=\mathbf 1_{B(\mathbf x)}\)) to get the displayed sum. Each summand integrates over
\(\mathcal R_{\mathbf b}\cap B(\mathbf x)\): since the box face \(\{t_i=x_i\}\) is an affine wall not among
the cell walls \(\{g_j\in\{0,1\}\}\), it generally bisects \(\mathcal R_{\mathbf b}\), so the summand is an
integral over a \emph{proper piece} of the cell, the decomposition is strictly finer than the cell partition.
The marginal is the section \(x_j\uparrow+\infty\) for \(j\ne i\), under which \(B\to\{t_i\le x\}\) and dominated
convergence gives \(F_{X_{(i)}}\). \(\qquad\square\)

The positive-dependence inequality \(F(\mathbf x)>\prod_i F_{X_{(i)}}(x_i)\) is the distributional
counterpart of the positive covariances of Theorem \ref{thm:positive-correlation-of-the-order-statistics}; like them it is proved at \(n=2\), where the exact
closed form exceeds the product, and computed at \(n=4\).

\refstepcounter{thmcnt}\label{thm:form-of-the-cdf-and-a-numerical-algorithm}\textbf{Theorem\nobreakspace{}\thethmcnt{} (form of the CDF, and a numerical algorithm).} On each cell\(\cap\)box piece the joint CDF
\(F(\mathbf x)\) is a hyperlogarithm of weight \(\le n\) with letters affine in \((\mathbf x,s)\) (integrate
the weight-\(0\) density over \(n\) truncated coordinates; Theorem \ref{thm:form-of-the-moments}'s ladder). For \(n=3\) this is weight
\(\le 3\), dilogarithms \(\mathrm{Li}_2\) (and at most \(\mathrm{Li}_3\)) with arguments affine in
\((\mathbf x,s)\). That closed form is computed explicitly for \(n=3\) at the end of this section, in Sage
and with no external CAS: the exact \(F\) at a genuinely-joint point, and the exact symbolic slice
\(F(x_1,\cdot,\cdot)\) exhibiting an \(\mathrm{Li}_2\) whose argument is affine in \(x_1\). Because that
closed form is large, point evaluation is better served by the deterministic quadrature of
Algorithm 1, which is exact by the reciprocal-affine structure:

\begin{quote}
\textbf{Algorithm 1 (joint CDF at a point).} \emph{Input} \(\mathbf x\), \((s,n)\). (1) The conditional support is
the box-slice \(J_n=\{t_0<\dots<t_{n-1}\}\cap\{\)all gaps \(\in(0,1)\}\); intersect with
\(\{t_i\le x_i\}\). Each nested integration limit is the \(\min/\max\) of \textbf{affine} forms (Fubini over the
hard-core \(+\) jamming \(+\) box walls). (2) Evaluate the density by the subset recursion
\(f_{\mathrm{UCP}}=\sum_\sigma\prod_k 1/\ell_k^\sigma\) (a fixed rational function on the all-jammed
cell; smooth on the interior when \(s>2n-1\), the bounded regime of the mode section). (3) Apply an \(m\)-point tensor
\textbf{Gauss--Legendre} rule on the nested affine box; (4) divide by \(\Pr[N{=}n]\). Cost \(O(m^n)\)
density evaluations, deterministic, no sampling; \(m=24\) gives \(\ge4\) correct digits here.
\end{quote}

At \(n=3,\ s=\tfrac{13}{2}\) the rule reproduces the Monte-Carlo estimates and returns
\(F(t_i\le s{-}1\ \forall i)=1\) to four digits. What quadrature does not supply is the closed form, which
the exact \(n=3\) computation below provides.

\textbf{The exact closed form, computed for \(n=3\).} The support in \(\mathcal J_3\) (\(s=\tfrac{13}{2}\)) is the
box \(X_{(1)}\in(\tfrac12,1)\), \(X_{(2)}\in(\tfrac52,3)\), \(X_{(3)}\in(\tfrac92,5)\); a threshold cuts the
law only inside its own interval. Take the genuinely-joint point \(\mathbf x=(\tfrac45,\tfrac{27}{10},\tfrac{24}5)\), all three cuts active. The region \(\mathcal J_3\cap\{t_i\le x_i\}\) splits at \(t_0=\tfrac{7}{10}\) (where
the box face \(\{t_1=x_2\}\) crosses the cell wall \(\{t_1=t_0+2\}\)) into two affine-bounded pieces, and the
exact three-fold integral of \(f_{\mathrm{UCP}}\) over them, divided by \(\Pr[N{=}3]\), is a \textbf{weight-\(3\)
multiple polylogarithm} containing dilogarithms. Symbolic integration evaluates it to \(F=0.1559\), matching Monte-Carlo, and returns the
exact \textbf{symbolic slice} \(F(x_1,\tfrac{27}{10},\tfrac{24}5)\)
for \(x_1\in(\tfrac12,\tfrac7{10})\), which is \(\tfrac{2}{11}x_1\,\mathrm{Li}_2\!\big(\tfrac{x_1}2+\tfrac32\big)\)
plus lower-weight \(\log\)-products, an explicit \(\mathrm{Li}_2\) with argument \textbf{affine in \(x_1\)}, exactly
as Theorem \ref{thm:form-of-the-moments} predicts. This closes the joint-CDF form for \(n=3\).

\paragraph{\texorpdfstring{Heatmap of the joint CDF \(F\) on the Tonks simplex}{Heatmap of the joint CDF F on the Tonks simplex}}

The joint CDF \(F\) lives on the Tonks simplex \(\mathsf T_n\) (\ref{prop:tonks-transform}); for \(n=2\) that simplex is two-dimensional
(\(\Delta_2=\{0\le t_0,\ t_1\ge t_0{+}1,\ t_1\le s{-}1\}\)), so \(F(x_1,x_2)\) is a genuine surface and admits
a heatmap. (The companion \textbf{density} heatmap of \(f_{\mathrm{UCP}}\) on the same simplex lives in
\ref{sec:joint-order-stats}; this is its \emph{cumulative} counterpart.) The figure below shows
\(F(x_1,x_2)=\Pr[X_{(1)}\le x_1,\,X_{(2)}\le x_2]\) at \(s=6\) as the normalized 2-D cumulative integral of
the density over \(\Delta_2\). It rises monotonically from \(0\) (dark, both thresholds below the support) to
\(1\) (light, top-right corner); the white curves are its level sets (the \(\alpha\)-quantile contours of
the quantiles), which bend at the support edge \(x_2=x_1{+}1\) where the marginal support of \(X_{(2)}\) begins.

\begin{figure}
\centering
\pandocbounded{\includegraphics[keepaspectratio,alt={UCP joint CDF on the Tonks simplex}]{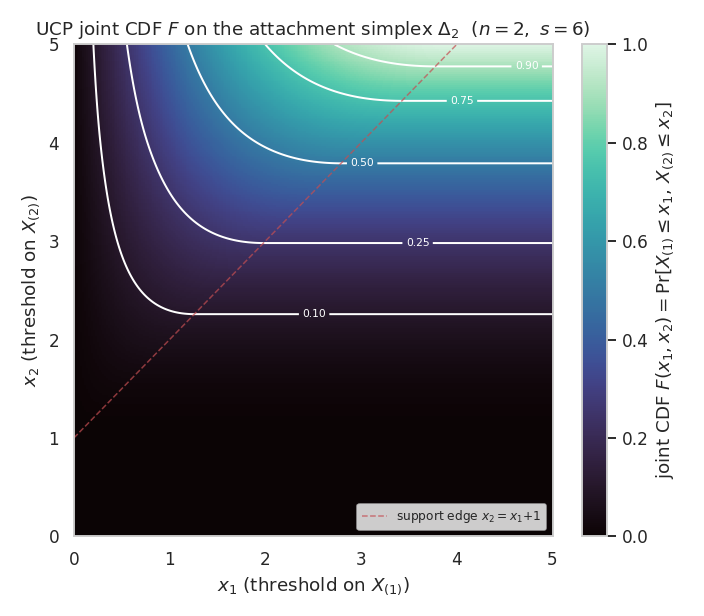}}
\caption{UCP joint CDF on the Tonks simplex}
\end{figure}

\subsubsection{\texorpdfstring{Quantiles: the joint contour and the Lambert-\(W\) marginal}{Quantiles: the joint contour and the Lambert-W marginal}}

\textbf{A joint distribution has no scalar quantile function.} A quantile inverts a CDF, but the joint CDF
\(F:\mathbb R^n\to[0,1]\) of the joint CDF is \textbf{not injective}, the pre-image of a level \(\alpha\) is not a point but
the \((n{-}1)\)-dimensional \textbf{\(\alpha\)-quantile contour}
\[\resizebox{\ifdim\width>\linewidth\linewidth\else\width\fi}{!}{$\displaystyle Q_\alpha=\{\mathbf x\in\mathbb R^n:\ F(\mathbf x)=\alpha\}.$}\]
This is the multivariate quantile, the literal \textbf{pre-image} of the joint CDF, and, unlike a
marginal inverse, it \emph{keeps the dependence}: its shape encodes \(F\ne\prod_i F_{X_{(i)}}\) (the positive
association established in the conditional density and the joint CDF). Inverting a single \textbf{marginal} \(F_{X_{(i)}}\) returns a scalar
but discards that dependence; a marginal quantile is a 1-D section of the contour, not the joint quantile.

The \(\alpha\)-quantile contours \(\{F=\alpha\}\) are the \textbf{level sets of the joint CDF}, the white level
curves of the the joint CDF heatmap, each a curve, not a point. At \(s=\tfrac52\) the joint CDF has the elementary
closed form
\(F(x_1,x_2)=2x_1-x_1\log x_1+x_1\log(x_2{-}1)-(x_2{-}\tfrac32)\log(1{-}2x_1)\), the support edge
\(x_2=x_1{+}1\) making the effective first-coordinate threshold \(\min(x_1,x_2{-}1)\); it agrees with
Monte-Carlo and is an exact instance of the surface the the joint CDF heatmap plots.

Each contour bends at the support edge \(x_2=x_1{+}1\): below it the \(x_2\)-threshold
binds (\(X_{(2)}>X_{(1)}{+}1\) forces \(X_{(1)}<x_2{-}1\)), so the contour runs horizontally; above it \(x_1\)
binds. That bend is the dependence made visible, a product-of-marginals CDF would have no such coupling.

\textbf{Marginal (1-D) quantiles, a section of the contour.} Fixing all thresholds but one recovers a
genuine scalar quantile, the inverse of one marginal CDF; it is the intercept where a contour meets the
boundary \(x_j=\sup\).

\refstepcounter{thmcnt}\label{thm:marginal-quantile-transcendental-inverse-no}\textbf{Theorem\nobreakspace{}\thethmcnt{} (marginal quantile: transcendental inverse; no elementary closed form for \(n\ge3\)).} On its
support the marginal density \(f_{X_{(i)}}\) is continuous and strictly positive on the interior (a finite
sum of reciprocal-affine products), so \(F_{X_{(i)}}\) is continuous and \textbf{strictly increasing}, hence a
bijection onto \((0,1)\); the marginal \(\alpha\)-quantile \(q_\alpha=F_{X_{(i)}}^{-1}(\alpha)\) exists and is
unique. By Theorem \ref{thm:form-of-the-moments}, \(F_{X_{(i)}}\) is a piecewise multiple polylogarithm of weight \(\le n\) (breakpoints
at half-integers), so \(q_\alpha\) is the root of a \textbf{transcendental (polylog) equation}.

\emph{Proof.} Strict positivity of \(f_{X_{(i)}}\) on the open support gives strict monotonicity of
\(F_{X_{(i)}}\); a continuous strictly increasing surjection onto \((0,1)\) has a unique continuous inverse
(intermediate value theorem). The weight-\(\le n\) polylog form is Theorem \ref{thm:form-of-the-moments} at \(d{=}1\) integrated once
more. \(\qquad\square\)

\begin{quote}
\textbf{Algorithm 2 (marginal quantile).} \(F_{X_{(i)}}\) is monotone, so bracket the root and iterate:
\textbf{bisection} (linear) or \textbf{Newton} \(x\leftarrow x-(F_{X_{(i)}}(x)-\alpha)/f_{X_{(i)}}(x)\) (quadratic).
Each \(F_{X_{(i)}}(x)\) is one call of the joint-CDF quadrature of Theorem \ref{thm:form-of-the-cdf-and-a-numerical-algorithm} with \(x_i=x\), the rest at
\(+\infty\) (i.e.~\(s-1\)).
\end{quote}

Root-finding on the deterministic CDF returns the median and the lower quartile of \(X_{(2)}\) at \(n=3\),
and a per-coordinate quantile grid at \(n=4\), each entry a one-dimensional section rather than the joint
quantile. Whether that inverse is a named function depends on \(n\).

\textbf{For \(n=2\) the inverse is the Lambert \(W\) function; for \(n\ge3\) it has no name.} The general-\(n\)
quantile inverts a polylogarithm of weight at least two, which has no standard named inverse and is
found as a generic transcendental root. The simplest case collapses to a classical special function.

\refstepcounter{thmcnt}\label{thm:quantile-is-a-lambert}\textbf{Theorem\nobreakspace{}\thethmcnt{} (the \(n=2\) quantile is a Lambert \(W\)).} For \(n=2\), \(s=\tfrac52\), the conditional marginal
of \(X_{(1)}\) has the elementary CDF
\[\resizebox{\ifdim\width>\linewidth\linewidth\else\width\fi}{!}{$\displaystyle F_{X_{(1)}}(x)=2x-x\log(2x),\qquad x\in(0,\tfrac12),$}\]
and its quantile is
\[\resizebox{\ifdim\width>\linewidth\linewidth\else\width\fi}{!}{$\displaystyle \boxed{\,q_\alpha(X_{(1)})=-\dfrac{\alpha}{W_{-1}\!\big(-2\alpha/e^{2}\big)}\,}$}\]
where \(W_{-1}\) is the lower real branch of the \textbf{Lambert \(W\) function} (\(W(z)e^{W(z)}=z\)).

\emph{Proof.} Integrating out \(X_{(2)}\) leaves \(f_{X_{(1)}}(x)=\tfrac1{2S}\big(1+\log(S/x)\big)\) on \((0,S)\),
\(S=s-2\) (the \(g_1{+}g_2=S{-}g_0\) factor is constant, so only \(1/(g_0{+}g_1)\) integrates, to a log);
its integral is \(F_{X_{(1)}}(x)=\tfrac{x}{2S}\big(2+\log(S/x)\big)\), which at \(S=\tfrac12\) is
\(2x-x\log(2x)\). Setting \(u=2x\), \(F=\alpha\) becomes \(u(2-\log u)=2\alpha\); the substitution \(u=e^{2-t}\)
gives \(t\,e^{-t}=2\alpha/e^2\), i.e.~\(-t=W(-2\alpha/e^2)\), and back-substituting
\(x=u/2=-\alpha/W(-2\alpha/e^2)\). The value lands in \((0,\tfrac12)\) on the \(W_{-1}\) branch. \(\qquad\square\)

The closed form agrees with numeric root-finding and with Monte-Carlo, and the
figure below plots it.

\begin{figure}
\centering
\pandocbounded{\includegraphics[keepaspectratio,alt={Quantile function of X\_(1): the inverse is Lambert W}]{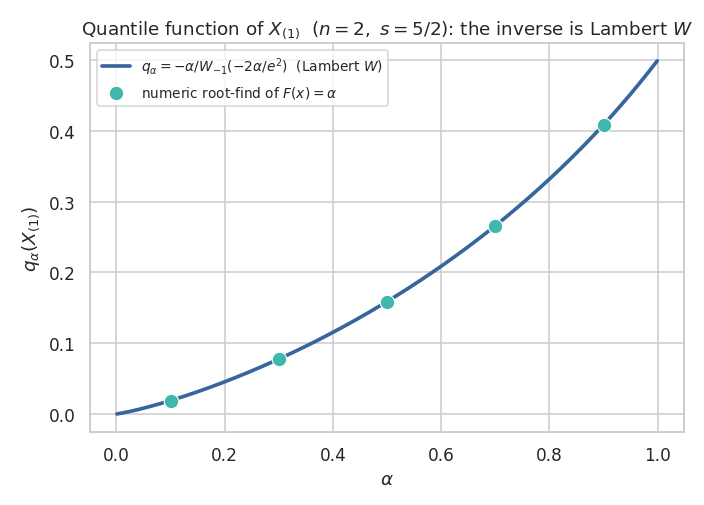}}
\caption{Quantile function of X\_(1): the inverse is Lambert W}
\end{figure}

For \(n\ge3\) the marginal CDF is a genuine weight-\(\ge2\) multiple polylogarithm (dilogarithms, as the
exact joint CDF of the joint CDF shows), whose inverse is \textbf{not} a named function; the quantile is then computed by
the monotone root-find of Theorem \ref{thm:marginal-quantile-transcendental-inverse-no} on the deterministic CDF, as done above for \(X_{(2)}\) at \(n=3\).

\subsubsection{Convexity, log-convex on each cell, and not on any neighbourhood of a wall}

\refstepcounter{thmcnt}\label{thm:per-cell-log-convexity}\textbf{Theorem\nobreakspace{}\thethmcnt{} (per-cell log-convexity).} On the interior of each gap-sign cell \(\mathcal R_{\mathbf b}\)
(every gap on a fixed side of one car length), \(f_{\mathrm{UCP}}\) is \textbf{log-convex}.

\emph{Proof.} Fix a cell and a direction. Each free length \(\ell_k^\sigma\) is positive and affine on the
cell, so \(t\mapsto\ell_k^\sigma\) is affine and \(-\log\ell_k^\sigma\) is convex; a sum of convex functions
is convex, so \(-\sum_k\log\ell_k^\sigma=\log\prod_k(1/\ell_k^\sigma)\) is convex, each arrival term is
log-convex. For the sum, log-convex functions are closed under addition: if \(u,v\) are log-convex then for
\(\lambda\in[0,1]\) and points \(a,b\), H"older's inequality gives
\(u(\lambda a{+}\bar\lambda b)+v(\lambda a{+}\bar\lambda b)\le u(a)^\lambda u(b)^{\bar\lambda}+v(a)^\lambda v(b)^{\bar\lambda}\le(u(a){+}v(a))^\lambda(u(b){+}v(b))^{\bar\lambda}\)
(the second step is H"older on the pair \((u(a),v(a)),(u(b),v(b))\) with exponents \(1/\lambda,1/\bar\lambda\)),
i.e.~\(u+v\) is log-convex. Iterating over the \(n!\) arrival terms, \(f_{\mathrm{UCP}}=\sum_\sigma\prod_k 1/\ell_k^\sigma\)
is log-convex on the cell. Equivalently the Hessian of \(\log f_{\mathrm{UCP}}\) is positive-semidefinite; exhibits it as positive-\textbf{definite} on the simplex tangent space at the \(J_n\)
centroid. \(\qquad\square\)

\refstepcounter{thmcnt}\label{prop:global-log-convexity-fails}\textbf{Proposition\nobreakspace{}\thethmcnt{} (global log-convexity fails).} \(\log f_{\mathrm{UCP}}\) is \textbf{not} convex on \(\mathsf T_n\).
\emph{Proof.} The free lengths depend on each gap only through the cap \(\min(G_i,1)\), whose slope drops from
\(1\) to \(0\) as \(G_i\) crosses \(1\). Move along a line that increases a single gap \(G_i\), holding the others
fixed. Below the wall \(\{G_i=1\}\) every \(\ell_k^\sigma\) depending on this gap is strictly decreasing in
\(G_i\), so each term \(\prod_k1/\ell_k^\sigma\) is strictly increasing and \(\log f_{\mathrm{UCP}}\) has
strictly positive one-sided directional derivative. Above the wall the cap is saturated: those same
\(\ell_k^\sigma\) no longer depend on \(G_i\) at all, so along this direction \(\log f_{\mathrm{UCP}}\) is
constant and its directional derivative is \(0\). The one-sided directional derivative therefore \textbf{drops}
as the wall is crossed. A convex function has non-decreasing one-sided directional derivatives along
every line, so this refutes convexity on every neighbourhood of the wall. Hence global log-convexity
fails precisely at the walls \(\{G_i=1\}\), and \(f_{\mathrm{UCP}}\) is log-convex only
piecewise. \(\qquad\square\)

\emph{Witness (exact).} At \(n=2\), \(s=10\), write \(\ell(t)\) for the free length left by a single car at \(t\):
\(\ell(t)=8-t\) on \([0,1]\) and \(\ell\equiv7\) on \([1,8]\), the wall sitting at \(t=1\). Then
\(f(0,5)=\tfrac{15}{504}\) and \(f(1,5)=f(2,5)=\tfrac{16}{504}\), with \((1,5)\) the midpoint of \((0,5)\) and
\((2,5)\); so \(f(1,5)^2=\tfrac{256}{504^2}>\tfrac{240}{504^2}=f(0,5)\,f(2,5)\), the reverse of the
inequality convexity of \(\log f_{\mathrm{UCP}}\) would force. The slope drop is visible directly:
\(1/\ell\) strictly increases on \([0,1]\) and is constant on \([1,8]\). This is machine-checked against the
density itself rather than a model of it (\texttt{ucp\_\allowbreak sortedDensity\_\allowbreak two\_\allowbreak log\_\allowbreak not\_\allowbreak convexOn}), and the
\(\varepsilon\)-family form of the statement gives failure on every neighbourhood of the wall.

\textbf{Contrast.} Log-concavity is the property of equilibrium ensembles, with Tonks the boundary case,
log-affine because its density is constant. The UCP density carries the opposite curvature on each cell,
which is a fingerprint of sequential size-biasing. Per-cell log-convexity is also what makes the mode
computable in the tight regime of the mode section, where it reduces the argmax to a scan over the vertices of the
cell arrangement.

\textbf{Worked example (\(n=3\), \(s=13/2\)).} Across \(\{g_0=1\}\) (at \(x_2=16/5\), \(x_3=47/10\)) the density is continuous, both pieces equal \(279184/1598289\), but its slope jumps from \(5175776/51871743\) to \(0\), so it is not differentiable there. This is exactly the non-smoothness at a jamming facet that blocks log-convexity on any neighbourhood of the facet.

\subsubsection{Mode, three regimes in the segment length}

The mode is governed by \textbf{two thresholds}, \(s=2n-1\) and \(s=2n+1\), giving three regimes. Write the
\textbf{all-unjammed region} \(\mathcal U_n(s)=\{\mathbf x\in\mathsf T_n:\,X_{(1)}\ge1,\ X_{(n)}\le s{-}2,\
X_{(i+1)}{-}X_{(i)}\ge2\ \forall i\}\) (every gap at least a car length past the hard-core margin).

\refstepcounter{thmcnt}\label{thm:mode-of-three-regimes}\textbf{Theorem\nobreakspace{}\thethmcnt{} (mode of \(f_{\mathrm{UCP}}\); three regimes).}
1. \textbf{Spacious, \(s>2n+1\).} \(\mathcal U_n(s)\) has positive measure and
\(\displaystyle\operatorname*{ess\,sup}f_{\mathrm{UCP}}=\frac{n!}{\prod_{k=1}^n(s-2k+1)}\), attained on
\textbf{all of \(\mathcal U_n(s)\)}, the modal set is a \emph{region}, not a point.
2. \textbf{Tight, \(2n-1<s\le2n+1\).} \(\mathcal U_n(s)=\varnothing\) but \(f_{\mathrm{UCP}}\) is \textbf{bounded}; the
supremum is finite, attained at a \textbf{vertex} of the cell arrangement, and is \emph{strictly below}
\(n!/\prod_k(s-2k+1)\).
3. \textbf{Crowded, \(s\le2n-1\).} \(f_{\mathrm{UCP}}\) is \textbf{unbounded}: \(\operatorname*{ess\,sup}=+\infty\),
approached at a boundary vertex where a contiguous block of gaps collapses (some free length
\(\ell_k^\sigma\to0\)). There is \textbf{no finite mode}.

\emph{Proof.} Iterating \(\ell_k^\sigma=\ell_{k-1}^\sigma+\Delta_{k-1}\) with increments \(\Delta_{k-1}\ge-2\)
(equality exactly for an unjammed step; Theorem \(\ell_k\)-affine of \ref{sec:joint-order-stats}) from
\(\ell_1^\sigma=s-1\) gives \(\ell_k^\sigma\ge s-2k+1\), hence the \textbf{kernel bound}
\(\prod_k1/\ell_k^\sigma\le\prod_k1/(s-2k+1)\) and \(f_{\mathrm{UCP}}\le n!/\prod_k(s-2k+1)\), \emph{whenever
every factor is positive}, i.e.~\(s>2n-1\). \textbf{(1)} For \(s>2n+1\) the region \(\mathcal U_n(s)\) is jointly
strictly feasible (\(s-2\ge X_{(n)}\ge1+2(n-1)=2n-1\) forces \(s>2n+1\)); there every temporal order reaches
\(\mathbf x\) through all-unjammed steps, each contributing the maximal kernel \(\prod_k1/(s-2k+1)\), and the
\(n!\)-fold sum equals the bound, attained on the whole region. Off \(\mathcal U_n\) some step jams in every
order, so the kernel is strictly smaller. \textbf{(2)} For \(2n-1<s\le2n+1\) the bound still holds (\(s>2n-1\)
keeps all factors positive), so \(f_{\mathrm{UCP}}\) is bounded; \(\mathcal U_n(s)=\varnothing\), so the
maximal kernel is never simultaneously achieved and the sup is strictly below the bound. By per-cell
log-convexity (the convexity section) \(f_{\mathrm{UCP}}\) is convex on each closed cell, so its maximum is at an extreme
point, a \textbf{vertex} of the cell arrangement (Rockafellar, \emph{Convex Analysis}, Thm 32.2). \textbf{(3)} For
\(s\le2n-1\) the factor \(s-2n+1\le0\): the bound is vacuous and the tightest free length \(\ell_n^\sigma\) has
no positive lower bound. A boundary vertex with a collapsing gap block drives \(\ell_k^\sigma\to0\) for the
order filling that block last, so \(f_{\mathrm{UCP}}\to+\infty\); the density (an integrable probability
density) has an integrable pole there and no finite mode. \(\qquad\square\)

\textbf{The running example is the crowded regime.} For \(s=\tfrac{13}{2},\ n=4\): \(2n-1=7>6.5\), so we are in
\textbf{case 3, the conditional density given \(N{=}4\) is unbounded}, with a pole at the boundary vertex where
the two right-end gaps vanish (cars clustering against the wall). The Monte-Carlo modal bin of
\(X_{(1)}\) near a boundary is the image of this pole in a single marginal (integrating out the other
coordinates tames it to a finite marginal mode). The other two regimes appear at the same \(n=4\) for longer segments, the second threshold being \(2n+1=9\): at
\(s=8\) the density is bounded with a supremum strictly below \(n!/\prod_k(s-2k+1)=\tfrac8{35}\), and at
\(s=10\) the modal region is non-empty and the supremum equals \(n!/\prod_k(s-2k+1)=\tfrac8{315}\). The
smallest spacious instance is \(n=2,\ s=6\), where the closed-form value \(2/15\) is attained throughout
\(\mathcal U_2(6)\) and is strictly exceeded nowhere.

\textbf{Local,\(\to\),global.} The mode is a \emph{global argmax}, and its difficulty is set by the regime.
In the \textbf{spacious} regime the argmax is the explicit region \(\mathcal U_n(s)\) with a closed-form value, no cells needed. In the \textbf{tight} regime per-cell log-convexity (the convexity section) collapses the argmax to a finite
\textbf{vertex scan} over the cell arrangement. In the \textbf{crowded} regime there is no finite mode at all, the
density concentrates toward a pole, so the meaningful question shifts from \emph{mode} to \emph{the marginal modes}
and the \emph{tail toward the collapsing-gap vertex}. The running example \(s=\tfrac{13}{2},n=4\) sits here, which
is why its joint density has no interior peak.

\subsubsection{\texorpdfstring{Local,\(\to\),global: linearity and cell-respecting domains separate the cheap properties from the hard ones}{Local,\textbackslash to,global: linearity and cell-respecting domains separate the cheap properties from the hard ones}}

The per-cell density is the \emph{local} object; every property is some \emph{global} functional of it. They split
cleanly by \textbf{what operation the global step needs}:

{\footnotesize\begin{longtable}[]{@{}
  >{\raggedright\arraybackslash}p{\dimexpr 0.2000\linewidth-2\tabcolsep\relax}
  >{\raggedright\arraybackslash}p{\dimexpr 0.2000\linewidth-2\tabcolsep\relax}
  >{\raggedright\arraybackslash}p{\dimexpr 0.2000\linewidth-2\tabcolsep\relax}
  >{\raggedright\arraybackslash}p{\dimexpr 0.2000\linewidth-2\tabcolsep\relax}
  >{\raggedright\arraybackslash}p{\dimexpr 0.2000\linewidth-2\tabcolsep\relax}@{}}
\toprule\noalign{}
\begin{minipage}[b]{\linewidth}\raggedright
Property
\end{minipage} & \begin{minipage}[b]{\linewidth}\raggedright
Global operation
\end{minipage} & \begin{minipage}[b]{\linewidth}\raggedright
Per-cell assembly?
\end{minipage} & \begin{minipage}[b]{\linewidth}\raggedright
Cost
\end{minipage} & \begin{minipage}[b]{\linewidth}\raggedright
Verdict
\end{minipage} \\
\midrule\noalign{}
\endhead
\bottomrule\noalign{}
\endlastfoot
Normalization \(\Pr[N{=}n]\) & sum of integrals & \textbf{yes} (additive) & per-cell integral & easy \\
Moments \(\mathbb E[\prod X_{(i)}^{m_i}]\) & sum of integrals & \textbf{yes} (additive) & per-cell integral, weight \(\le n\) & easy \\
Marginal density / CDF \(F_{X_{(i)}}\) & integrate out, then over \(g\)-pieces & \textbf{yes} per \(g\)-piece & weight-\(\le n\) polylog & moderate \\
\textbf{Joint CDF} \(F(\mathbf x)\) & integrate over cell\(\cap\)\textbf{box} & \textbf{no} (box cuts across cells) & cell\(\cap\)box pieces, weight \(\le n\) & harder \\
Covariances / association & pair-marginal integrals & \textbf{yes} (additive) & weight-\(\le n{-}1\) & moderate \\
Quantile \(q_\alpha\) & \textbf{invert} the marginal CDF & \textbf{no} (inverse has no per-cell form) & transcendental root-find & hard \\
Mode \(\mathcal M\) & \textbf{argmax} over the simplex & \textbf{no} in general & vertex scan; closed form if \(s>2n{+}1\) & hard, or closed-form \\
Log-convexity & per-cell Hessian sign & \textbf{yes} (cell-local) & one-line H"older & easy (local); global fails \\
\end{longtable}}

\textbf{Two dividing lines: linearity, and whether the domain respects the cells.} A property that is a
\emph{linear} functional of \(f_{\mathrm{UCP}}\) integrated over a \textbf{whole cell} (mass, moments, full
marginals, correlations) inherits the cell partition additively, compute per cell, sum. The \textbf{joint
CDF} is still linear in \(f_{\mathrm{UCP}}\), but its domain is a \emph{box} that slices through cells, so its
assembly is a sum over the finer cell\(\cap\)box pieces rather than over cells. The \emph{nonlinear}
extractions (argmax for the mode, inverse for the quantile) do \textbf{not} decompose over cells at all, and
are hard unless a structural theorem intervenes: for the mode, log-convexity collapses the argmax to a
vertex scan and the spacious regime \(s>2n{+}1\) gives a closed form. This is the practical payoff of the
per-cell picture, it makes the whole-cell linear functionals a bookkeeping exercise and isolates
exactly where the genuine work is: cross-cell domains (the joint CDF) and nonlinear extraction (mode,
quantile).

\subsubsection{Connection to the permanent formula}

The conditional density is a sum over temporal orders of a product,
\[\resizebox{\ifdim\width>\linewidth\linewidth\else\width\fi}{!}{$\displaystyle f_{\mathrm{UCP}}(\mathbf x)=\sum_{\sigma\in S_n}\prod_{k=1}^n\frac{1}{\ell_k^\sigma},\qquad
\ell_k^\sigma=\text{free length available to the }k\text{-th arriving car under order }\sigma,$}\]
which has \textbf{exactly the shape of a matrix permanent} \(\operatorname{per}(M)=\sum_\sigma\prod_k M_{k,\sigma(k)}\).
The shape is not a coincidence of notation; it is the reason the subset algorithm costs \(O(2^n n)\), the
same as Ryser's permanent formula. This section makes the correspondence precise, and
then identifies where it fails, and why the failure lowers the cost.

\textbf{At \(n=2\) it is literally a permanent.} With \(M=\begin{psmallmatrix}1/(s-1)&1/(s-1)\\ 1/\ell(\text{car }1\mid\{2\})&1/\ell(\text{car }2\mid\{1\})\end{psmallmatrix}\)
we have \(f_{\mathrm{UCP}}=\operatorname{per}(M)\) exactly, verified symbolically.

\textbf{For \(n\ge3\) it is a \emph{prefix-dependent} permanent, not a fixed-matrix one.} A genuine permanent needs
\(M_{k,j}\) to depend only on the pair (arrival step \(k\), car \(j\)). Here the free length \(\ell_k^\sigma\)
depends on the \textbf{whole set} of already-placed cars, it is the total measure of the free region, summed over \emph{every} gap the set leaves open, not the single gap the arriving car lands in.
Concretely, the free length of car \(2\) after one prior placement is
\[\resizebox{\ifdim\width>\linewidth\linewidth\else\width\fi}{!}{$\displaystyle \ell(\text{car }2\mid\{1\})=(x_1-1)_++(s-x_1-2)_+\quad\neq\quad\ell(\text{car }2\mid\{3\})=(x_3-1)_++(s-x_3-2)_+,$}\]
so the same \((k,j)=(2,2)\) carries two different weights. The object is therefore a \textbf{set-function
(prefix-dependent) permanent} \(\sum_\sigma\prod_k w\big(\sigma(k)\mid\{\sigma(1),\dots,\sigma(k-1)\}\big)\),
a strict generalization of the permanent. (It is \emph{not} a Vere-Jones \(\alpha\)-permanent, which also needs a
fixed kernel; and the process is \emph{not} a permanental point process, those are attractive Cox processes,
whereas RSA is hard-core.)

\textbf{Why the \(O(2^n n)\) matches Ryser.} Both the Held,-,Karp subset DP that assembles \(f_{\mathrm{UCP}}\) and
Ryser's formula walk the \emph{same} \(2^n\) subset lattice; the ordinary permanent is the special case where
\(w(j\mid S)\) depends on \(S\) only through \(|S|\). (Ryser and Held,-,Karp are recognized as two instances of
one subset-convolution paradigm: Björklund, \emph{SWAT} 2016; Björklund,-,Husfeldt,-,Kaski,-,Koivisto, \emph{STOC}
2007.) The shared complexity is therefore a structural kinship, not a hardness statement.

\textbf{Where the cost of point evaluation sits.} The naive sum over temporal orders costs \(n!\). Because the free
length is determined by the \textbf{set} of cars already placed and not by their order, grouping the orders by their
last-placed car collapses this to the Held--Karp subset recursion at \(O(2^n n)\), the permanent rung, and the
standing bound. A further collapse to contiguous intervals would need the free length to be determined by an
\emph{interval}; it is not, and the attempt is analysed below.

What \emph{can} be said exactly is where the remaining difficulty lives, and the answer is sharp. In slack
coordinates \(u_1,\dots,u_{n+1}\), the free length left by a parked set \(S\) is
\[\resizebox{\ifdim\width>\linewidth\linewidth\else\width\fi}{!}{$\displaystyle L(S)=s-2|S|-1+\sum_{j\in D(S)}(1-u_j)_+ ,$}\]
with \(D(S)\) the \textbf{pinched} slacks, the left margin once rank \(1\) has parked, the right margin once rank \(n\) has,
and an internal clearance once \emph{both} its neighbouring ranks have. The two extreme cells behave completely
differently.

\textbf{The all-unjammed cell is closed-form.} If every slack is at least a car length, every positive part vanishes
and \(L(S)=s-2|S|-1\) depends on \(S\) only through its \textbf{size}. Every temporal order then carries the same product,
so
\[\resizebox{\ifdim\width>\linewidth\linewidth\else\width\fi}{!}{$\displaystyle f_{\mathrm{UCP}}=\frac{n!}{\prod_{k=0}^{n-1}(s-2k-1)} .$}\]
Verified in exact rationals at \(n=3,4,5\) against the brute-force order sum (\(2/231\), \(8/3003\), \(8/6435\),
\(8/46189\)), and both collapses are machine-checked. \textbf{No hardness can live on this cell.}

\textbf{The all-jammed cell isolates the difficulty in a subset sum.} This is the cell the \emph{jammed} ensemble
occupies, since saturation means no gap admits a car. There every positive part is active, and the counting
identity \(|D(S)|+\operatorname{runs}(S)=|S|+\mathbf 1\{1\in S\}+\mathbf 1\{n\in S\}\), proved by splitting \(S\)
according to whether the preceding rank is also parked, so that one part is exactly the internal pinched slacks
and the other exactly the run starts, turns that identity into
\[\resizebox{\ifdim\width>\linewidth\linewidth\else\width\fi}{!}{$\displaystyle L(S)=s-|S|-1-\operatorname{runs}(S)+\mathbf 1\{1\in S\}+\mathbf 1\{n\in S\}-\sum_{j\in D(S)}u_j .$}\]
The combinatorial part \((|S|,\operatorname{runs}(S),\mathbf 1\{1\in S\},\mathbf 1\{n\in S\})\) ranges over an
\(O(n^2)\) state space. \textbf{The entire obstruction to a polynomial evaluation is therefore the subset sum
\(\sum_{j\in D(S)}u_j\).} This identity is machine-checked (\texttt{card\_\allowbreak pinchedOf}, \texttt{freeLenS\_\allowbreak jammed}) and verified on
every subset at \(n=3,4,5\) in exact rationals.

\textbf{Consequences for the complexity question, which is open.} The value is not a function of the deficit multiset
alone: at \(n=4\), \(s=14\) with deficits \(\{\tfrac12,\tfrac12\}\) the density is \(7549/3374514\), \(896/366795\) and
\(25159/10270260\) according as the two tight slacks sit at the margins, spread internally, or adjacent, so the
\emph{positions} carry information. The achievable index sets \(D(S)\) are exactly those edge sets whose components are
separated by at least two missing edges (\(12\) of \(16\) at \(n=5\)), a rich enough family for a reduction to have
room. Since the weight depends on the deficits only through their \textbf{sum}, the natural attack on hardness is \(\#\mathsf{SUBSET\text-
SUM}\) rather than the permanent; interval-restricted subset sums are pseudo-polynomial by dynamic programming
but take \(2^{\Theta(n)}\) distinct values for generic rationals. The expected shape of the answer is therefore a
\textbf{trichotomy}, closed form on the unjammed cell, pseudo-polynomial at bounded deficit precision, and
conjecturally hard in general, and neither membership in \(\mathrm{FP}\) nor \(\#\mathrm P\)-hardness is claimed
here.

\textbf{The interval recursion, and why it does not evaluate \(f_{\mathrm{UCP}}\).} Let \(W(a,b)\) be built by
splitting the cars \(\{a,\dots,b\}\) on the \textbf{first} one placed, \(m\), with the enclosing neighbours, car \(a{-}1\)
(or the left segment edge) and car \(b{+}1\) (or the right edge), already fixed, so the enclosing interval is
\([L,R]\), \(L=x_{a-1}{+}1\) (or \(0\)), \(R=x_{b+1}{-}1\) (or \(s{-}1\)). With \(W(a,a{-}1):=1\),
\[\resizebox{\ifdim\width>\linewidth\linewidth\else\width\fi}{!}{$\displaystyle W(a,b)=\sum_{m=a}^{b}\frac1{R-L}\,\binom{b-a}{\,m-a\,}\,W(a,m{-}1)\,W(m{+}1,b).$}\]
This recursion has \(\binom{n+1}{2}=O(n^2)\) states, each an \(O(n)\) sum, so it evaluates in \(O(n^3)\), but
\textbf{\(W(1,n)\ne f_{\mathrm{UCP}}\)}, and no correctness claim is made for it.

The obstruction is the screening step. Once \(m\) is placed it does divide the cars, and each side's \emph{temporal orders} do factor with \(\binom{b-a}{m-a}\) interleavings; that much is right, and it is the same enumeration the subset recursion of \ref{sec:order-stats-cells} performs. What does not factor is the \textbf{weight}. A later car divides by
the free length of the \emph{whole} configuration, the total measure of the free region, summed over every open gap, not by the sub-interval it happens to land in. After a split that total is \(\ell_{\mathrm{left}}+\ell_{\mathrm{right}}\),
and \(1/(\ell_{\mathrm{left}}+\ell_{\mathrm{right}})\) does not factor. The prefactor \(1/(R-L)\) therefore charges the
wrong denominator from the second car onwards.

\emph{The failure is numerical, already at \(n=2\).} At \(s=8\), \(\mathbf x=(2,5)\) both cars are interior and the recursion
returns \(1/14\) against the true \(f_{\mathrm{UCP}}=2/35\): after a first car at \(2\) the free region open to the
arriving car measures \((2-1)+(8-2-2)=5\), while the recursion offers only the right sub-interval, of length \(4\). The two
coincide exactly when the boundary regions contribute nothing, which is why \(n=2\) examples with a car within one
unit of an edge do not detect it. At \(n=3\), \(s=5\), \(\mathbf x=(\tfrac12,2,\tfrac72)\) the recursion gives \(9/10\)
against \(13/20\).

\textbf{The correct evaluation is the subset dynamic program} of \ref{sec:order-stats-cells}: group the temporal orders by their \textbf{last}-placed car, for which the free length \emph{is} determined by the set already present. That
runs in \(O(2^n n)\), and it is verified against brute force in exact rationals. Whether \(f_{\mathrm{UCP}}\) admits a
polynomial-time point evaluation is not settled by the interval recursion above and is not claimed here. The recursion is checked against the \(n!\) sum (\(n\le6\), exact over
\(\mathbb Q\)) and against the \(2^n\) subset DP (\(n\le10\)). What the recursion does not settle is whether
\emph{some} fixed matrix reproduces \(f_{\mathrm{UCP}}\) for \(n\ge3\); Open problems (PO2) records the evidence
and what remains.

\paragraph{Where the UCP order statistics sit among order-statistic computations}

The cost of assembling a joint order-statistic density is set by the dependence structure of the parent
sample; the subset recursion above places UCP at the same cost as the independent non-identical case rather than in the
intractable dependent one.

\textbf{The subset dynamic program costs \(O(2^n n)\), not \(n!\).} Naively the density is a sum over the \(n!\) temporal orders,
\(\sum_\sigma\prod_k 1/\ell_k^\sigma\). But the free length \(\ell_k^\sigma\) depends only on the \textbf{set}
of already-placed cars, their positions fix the free space, not on the order within that set. So the
sum factors through subsets and is assembled by a \textbf{Held--Karp subset dynamic program in \(O(2^n n)\)}
(Theorem DP of \ref{sec:order-stats-cells}). The assembler visits exactly \(2^n\) subproblems: at \(n{=}7\) that is
\(128\) versus \(5040=7!\), and the gap widens super-exponentially.

\textbf{Jamming forces the \(n!\) terms to be nonzero, not the \(n!\) cost.} The all-jammed cell is exactly where
\emph{every} temporal order is feasible, since all \(n!\) orders reach the jammed configuration, so it is the
worst case for the naive sum; the subset structure is nevertheless untouched and the DP is still
\(O(2^n)\). Whether \(O(2^n n)\) is the last word is \textbf{open}: the interval collapse that would beat it is
unavailable, since the free length is not interval-determined, and no polynomial algorithm is known. The permanent shape of
\(\sum_\sigma\prod1/\ell\) is genuine, and it \emph{is} a permanent at \(n{=}2\), but screening makes this
instance tractable.

\textbf{Three regimes of increasing difficulty separate the classical parent structures from the UCP one.}

{\footnotesize\begin{longtable}[]{@{}
  >{\raggedright\arraybackslash}p{\dimexpr 0.3333\linewidth-2\tabcolsep\relax}
  >{\raggedright\arraybackslash}p{\dimexpr 0.3333\linewidth-2\tabcolsep\relax}
  >{\raggedright\arraybackslash}p{\dimexpr 0.3333\linewidth-2\tabcolsep\relax}@{}}
\toprule\noalign{}
\begin{minipage}[b]{\linewidth}\raggedright
parent structure
\end{minipage} & \begin{minipage}[b]{\linewidth}\raggedright
joint order-statistic density
\end{minipage} & \begin{minipage}[b]{\linewidth}\raggedright
cost
\end{minipage} \\
\midrule\noalign{}
\endhead
\bottomrule\noalign{}
\endlastfoot
i.i.d. & elementary product form (David--Nagaraja Ch. 2) & \(O(n)\) \\
independent non-identical (i.n.i.d.) & a \textbf{permanent} & \#P-hard exactly; \(O(2^n n)\) by Ryser \\
\textbf{UCP spatial order statistics} & structured \(\sum_\sigma\prod 1/\ell_k^\sigma\) (a \emph{set-function permanent}, above) & \textbf{\(O(2^n n)\) subset DP}; closed form on the all-unjammed cell; complexity of the general case \textbf{open} \\
general dependent parents (David--Nagaraja Ch. 6) & the \textbf{dependency theorem}, no permanent shortcut & intractable; the \emph{temporal} (arrival) law \\
\end{longtable}}

The UCP sorted positions are formally a \textbf{dependent-parent} problem, the hardest class in principle, but the one-dimensional \textbf{hard-core structure} (free lengths are set-functions with \(\pm1\) coefficients)
collapses them to the same \(O(2^n)\) subset cost as the i.n.i.d. permanent. Whether UCP sits strictly below
that permanent is open; the screening argument that would have shown it is refuted above. The
sorted-position law is therefore computationally simpler than the general dependent-parent law it
descends from; that temporal case is the David--Nagaraja Ch. 6 dependency theorem, developed for the gap
order statistics in \ref{sec:gap-order-stats}. The tractability is bought by the hard-core geometry, not by
independence.

\paragraph{Widom's arrival representation and the symbolic term count}

The temporal-order sum \(f_{\mathrm{UCP}}=\sum_{\sigma}\prod_k 1/\ell_k^\sigma\) (the permanent section) is \textbf{Widom's (1966)} representation of the random sequential adsorption configuration as an average over arrival sequences.

\textbf{Runtime, four routes to the same object.}

{\footnotesize\begin{longtable}[]{@{}
  >{\raggedright\arraybackslash}p{\dimexpr 0.3333\linewidth-2\tabcolsep\relax}
  >{\raggedright\arraybackslash}p{\dimexpr 0.3333\linewidth-2\tabcolsep\relax}
  >{\raggedright\arraybackslash}p{\dimexpr 0.3333\linewidth-2\tabcolsep\relax}@{}}
\toprule\noalign{}
\begin{minipage}[b]{\linewidth}\raggedright
route
\end{minipage} & \begin{minipage}[b]{\linewidth}\raggedright
computes
\end{minipage} & \begin{minipage}[b]{\linewidth}\raggedright
cost
\end{minipage} \\
\midrule\noalign{}
\endhead
\bottomrule\noalign{}
\endlastfoot
Widom arrival sum (naive) & exact finite-\(n\) density & \(O(n!)\) \\
Held--Karp / Ryser subset DP & exact finite-\(n\) density & \(O(2^n n)\) \\
all-unjammed cell (closed form) & point evaluation & \(O(n)\) \\
Widom kinetic time-integral & \(s\to\infty\) density (\(\varphi\)-kernel) & \(O(1)\) \\
\end{longtable}}

\textbf{Term count of the symbolic density.} The size of the full symbolic expression, the piecewise density itself, not a point value, is exponential. Naively there are \(n!\) terms, one per temporal order. The subset collapse reduces the distinct sub-configurations to \(2^n\), and the density is piecewise over the \(2^{n-1}\) jamming bitstrings (which internal gaps are \(<1\)). On a cell with \(j\) jammed internal gaps the top-weight symbol support is \(\binom{n-1}{j}\), summing to \(2^{n-1}\) across the top weight. A bitstring is feasible exactly when the free gaps fit, \((n+1-j)\le s-n\) with \(j\) the number of jammed gaps, so feasibility depends only on \(j\) and the cell count is the partial binomial sum \(\sum_{j\ge\lceil 2n+1-s\rceil}\binom{n+1}{j}\). At a fixed density \(s=cn\) this grows like \(2^{H(2-c)n}\) with \(H\) the binary entropy; at the jamming-typical density \(s=n/m\) the constant is \(2^{H(2-1/m)}=1.8954\ldots\). The symbolic joint density therefore admits no sub-exponential closed form.

\textbf{Discussion.} Widom's kinetic locality is an empty-interval structure of the same family as the interval recursion above, and it meets the same limit: it governs the \emph{geometry} left by a placed set, not the free length an arriving car divides by, so the kinetic viewpoint accelerates neither the finite-\(n\) symbolic computation nor the point evaluation. The subset DP is already the efficient evaluation of Widom's arrival sum (the standard permanent speedup from \(O(n!)\) to \(O(2^n)\), Ryser), and the exponential term count above is intrinsic to the symbolic form; no reordering of the arrival sum removes it. What the kinetic route contributes is a distinct regime, the closed-form jamming limit in \(O(1)\), and the reason point evaluation is polynomial: screening makes each free length depend only on the nearest placed neighbours, the discrete analogue of the kinetic empty-interval equation.

\emph{Ref.} B. Widom, ``Random Sequential Addition of Hard Spheres to a Volume,'' \emph{J. Chem. Phys.} \textbf{44} (1966) 3888--3894.

\textbf{Worked example (\(n=3\), \(s=13/2\)).} The arrival sum realizes \(90\) temporal-order \(\times\) codeword terms that collapse to \(14\) distinct rational forms of the joint density, a concrete instance of the exponential term count above; the permanent representation states none of these.

\paragraph{What is proved about the cost, and what is not}

The permanent analogy above prices the algorithm. The statements below price the \emph{problem}, in the
one model where the question is settled. Each is machine-checked in Lean, and each depends only on
the three standard axioms \texttt{propext}, \texttt{Classical.\allowbreak choice} and \texttt{Quot.\allowbreak sound}.

{\footnotesize\begin{longtable}[]{@{}
  >{\raggedright\arraybackslash}p{\dimexpr 0.3333\linewidth-2\tabcolsep\relax}
  >{\raggedright\arraybackslash}p{\dimexpr 0.3333\linewidth-2\tabcolsep\relax}
  >{\raggedright\arraybackslash}p{\dimexpr 0.3333\linewidth-2\tabcolsep\relax}@{}}
\toprule\noalign{}
\begin{minipage}[b]{\linewidth}\raggedright
statement
\end{minipage} & \begin{minipage}[b]{\linewidth}\raggedright
content
\end{minipage} & \begin{minipage}[b]{\linewidth}\raggedright
Lean
\end{minipage} \\
\midrule\noalign{}
\endhead
\bottomrule\noalign{}
\endlastfoot
free length is a subset sum & \(\ell=(s-n-\lvert T^c\rvert)-\sum_{i\in T}G_i\) over the jammed set \(T\), additive, with no interaction between gaps & \texttt{UCP.\allowbreak freeLength\_\allowbreak subset\_\allowbreak sum} \\
cost of the subset recursion & the memoised recursion performs exactly \(\sum_{T\subseteq S}\lvert T\rvert=\lvert S\rvert\,2^{\lvert S\rvert-1}\) operations & \texttt{UCP.\allowbreak Cost.\allowbreak tableCost\_\allowbreak eq} \\
it costs less than the order sum & \(C_{\mathrm{dp}}(S)<C_{\mathrm{ord}}(S)=\lvert S\rvert!\,\lvert S\rvert\) whenever \(\lvert S\rvert\ge4\) & \texttt{UCP.\allowbreak Cost.\allowbreak tableCost\_\allowbreak lt\_\allowbreak naive} \\
knowing all proper subsets suffices & two weights agreeing on every \(T\subsetneq S\) give the same value & \texttt{UCP.\allowbreak heldKarp\_\allowbreak congr\_\allowbreak of\_\allowbreak forall\_\allowbreak ssubset} \\
missing one proper subset does not & if some \(T\subsetneq S\) is never queried, two weights agree on all queries yet differ in value & \texttt{UCP.\allowbreak heldKarp\_\allowbreak not\_\allowbreak determined\_\allowbreak of\_\allowbreak missing} \\
a partial-history oracle does not help & an oracle giving the density of the first \(k\) arrivals supplies the weight only on sets of size \(\le k\), leaving the rest to be queried & \texttt{UCP.\allowbreak high\_\allowbreak levels\_\allowbreak remaining} \\
the query complexity is exact & an algorithm with query access to the free length needs exactly \(2^{n}-1\) queries & \texttt{UCP.\allowbreak query\_\allowbreak complexity\_\allowbreak eq} \\
\end{longtable}}

The last two are the substantive ones. Together these establish that the subset recursion is optimal
\emph{among algorithms that reach the free length only by querying it}, and that no amount of information
about short prefixes of the temporal order removes the exponential.

\textbf{What this does not say.} The bound is a query bound, not a lower bound on the problem. An
algorithm that uses the explicit affine form of \(\ell\), rather than treating it as an oracle, is not
covered, and \texttt{UCP.\allowbreak freeLength\_\allowbreak subset\_\allowbreak sum} shows that form is simple enough to be worth exploiting. So
the result locates where an improvement would have to come from, namely the geometry, and it closes
off the combinatorial route. Whether evaluating \(f_{\mathrm{UCP}}\) is in \(\mathrm{FP}\), and whether
it is \(\#P\)-hard, are both open, and nothing above bears on either.

\subsubsection{The generation mechanism distinguishes UCP from Tonks and Mat\textquotesingle ern I \& II}

All are hard-core (non-overlapping unit intervals / exclusion distance one) point processes on a line,
but they differ in \emph{how the configuration is generated}, and the properties above distinguish them by that mechanism. Definitions follow in the comparison section.
\emph{(Simple sequential inhibition of unit rods in one dimension \textbf{is} the car-parking process, SSI \(=\) 1D
RSA \(=\) UCP, so it is not a separate column; it is this process under the spatial-statistics name.)}

{\footnotesize\begin{longtable}[]{@{}
  >{\raggedright\arraybackslash}p{\dimexpr 0.2000\linewidth-2\tabcolsep\relax}
  >{\raggedright\arraybackslash}p{\dimexpr 0.2000\linewidth-2\tabcolsep\relax}
  >{\raggedright\arraybackslash}p{\dimexpr 0.2000\linewidth-2\tabcolsep\relax}
  >{\raggedright\arraybackslash}p{\dimexpr 0.2000\linewidth-2\tabcolsep\relax}
  >{\raggedright\arraybackslash}p{\dimexpr 0.2000\linewidth-2\tabcolsep\relax}@{}}
\toprule\noalign{}
\begin{minipage}[b]{\linewidth}\raggedright
feature
\end{minipage} & \begin{minipage}[b]{\linewidth}\raggedright
\textbf{UCP} (this work)
\end{minipage} & \begin{minipage}[b]{\linewidth}\raggedright
\textbf{Tonks} (equilibrium)
\end{minipage} & \begin{minipage}[b]{\linewidth}\raggedright
\textbf{Mat\textquotesingle ern I}
\end{minipage} & \begin{minipage}[b]{\linewidth}\raggedright
\textbf{Mat\textquotesingle ern II}
\end{minipage} \\
\midrule\noalign{}
\endhead
\bottomrule\noalign{}
\endlastfoot
construction & sequential uniform attachment & concurrent Gibbs draw & Poisson \(+\) delete any close pair & Poisson \(+\) keep earliest of close pair \\
conditioning & fixed segment \(s\) \textbf{and} count \(N{=}n\) & fixed \(s,n\) & fixed intensity \(\lambda\) & fixed intensity \(\lambda\) \\
joint density on \(\mathsf T_n\) & \(\sum_\sigma\prod_k 1/\ell_k^\sigma\) (size-biased \(\prod 1/\ell\)) & \textbf{uniform} \(n!/(s{-}n)^n\) & thinned Poisson (not a fixed-\(n\) density) & thinned Poisson \\
gaps & \textbf{non-exchangeable}, size-biased & exchangeable (i.i.d. given sum) & stationary, dependent & stationary, dependent \\
order-stat correlation & \textbf{positive}, decays in \(\lvert i{-}j\rvert\) & positive, exchangeable & short-range & short-range \\
log-shape of density & \textbf{log-convex per cell}, global fail & \textbf{log-affine} (constant) & , (no fixed-\(n\) density) & , \\
mode & three regimes (region / vertex / unbounded) & every point (flat) & , & , \\
saturation coverage & \(\approx0.7476\) (R\textquotesingle enyi jamming) & full equilibrium packing & \(<\) RSA (over-deletes) & between I and RSA \\
role here & object of study & leading-order asymptotic baseline & non-sequential baseline & sequential-mimicking baseline \\
\end{longtable}}

\textbf{Reading the table.} The single discriminating axis is \textbf{how mass is placed on the shared hard-core
support}. Tonks spreads it \emph{uniformly} (log-affine, exchangeable gaps, flat mode); UCP's sequential
size-biasing bends the same support into a \emph{log-convex-per-cell} density with non-exchangeable gaps and a
regime-dependent mode. The \textbf{Mat\textquotesingle ern} processes are the instructive near-misses: both are \emph{thinnings}
of a Poisson process (simultaneous, not sequential), so neither yields a fixed-\(n\) conditioned density at
all, type I over-deletes (retention \(e^{-2\lambda}\), sparser than jamming), type II's earliest-timestamp
rule \emph{mimics} sequential attachment and approaches UCP only in the high-intensity limit. UCP is realised
as a Mat\textquotesingle ern thinning at no fixed intensity.

The properties layer is what makes these distinctions \emph{quantitative}: exchangeability of gaps separates
UCP from Tonks; the log-convex/log-affine split separates their densities; the fixed-\(n\) conditioning
separates both from the Mat\textquotesingle ern thinnings.

\subsubsection{Open problems}

Two results above rest on a step that is verified numerically but not proved; both.

\textbf{PO1, the monotonicity coupling (the conditional density, Theorem \ref{thm:positive-correlation-of-the-order-statistics}).} Lemma B reduces positive correlation of the order
statistics to negative correlation of the gaps at separation at least two, and that reduction rests on
\(\mathbb E[G\mid X_{(m)}]\) being monotone in the length of the enclosing region, a stochastic monotonicity
of the conditioned sub-process. The monotonicity is exact only for \(\nu\le2\) and holds by stochastic
dominance for \(\nu\le4\); the general case is open, and Monte-Carlo-verified for \(n\le9\). A coupling
argument would close it. The covariance identity, the exact \(n{=}2\) gap covariances
(\(\mathrm{Cov}(G_0,G_1)=-\tfrac7{1152}\) and \(\mathrm{Cov}(G_0,G_2)=-\tfrac{11}{768}\)), and the MTP\(_2\)
refutation are complete. The stronger positive-quadrant dependence \(F>\prod_iF_{X_{(i)}}\) of the joint CDF is not
implied by \(\mathrm{Cov}\ge0\) and remains computed rather than proved.

\textbf{PO2, no fixed-matrix permanent for \(n\ge3\) (the permanent section).} The arrival-step permanent representation fails
provably, because the weight is prefix-set-dependent:
\(\ell(\text{car }2\mid\{1\})\ne\ell(\text{car }2\mid\{3\})\). Ruling out \emph{every} fixed matrix is open.
The screening-induced multiplicity-two term carries distinct factors, which a permanent can produce only
as a square; that is evidence rather than a proof.

\section{\texorpdfstring{The absorption time \(N_{\mathrm{abs}}(s)\)}{The absorption time N\_\{\textbackslash mathrm\{abs\}\}(s)}}

\label{sec:absorption-time}

Let \(U(s)\) be an instance of the UCP process \(U\) on the segment \([0,s]\) (\ref{sec:background-ucp}). Define \(N_{\mathrm{abs}}(s)\) to be the discrete random variable counting the cars parked by \(U(s)\) at absorption, and denote its probabilities by
\begin{equation}\label{eq:pmf}
p_n(s):=\Pr[N_{\mathrm{abs}}(s)=n],\qquad n\in\mathbb Z_{\ge0}.
\end{equation}

In this section we derive properties of the pmf \(p_n\) at fixed \(s\) and of the family \(\{N_{\mathrm{abs}}(s)\}\) in \(s\), which follow from the splitting law (\ref{sec:background-aggregate}): for \(s>1\),
\begin{equation}\label{eq:split}
N_{\mathrm{abs}}(s)\stackrel{d}{=}1+N^{(1)}_{\mathrm{abs}}(X)+N^{(2)}_{\mathrm{abs}}(s-1-X),\qquad X\sim\mathrm{Unif}[0,s-1],
\end{equation}
where \(N^{(1)}_{\mathrm{abs}}\) and \(N^{(2)}_{\mathrm{abs}}\) are independent copies of the process, independent of \(X\).

\subsubsection{Support, computation, and moments}

From (\ref{sec:background-ucp}), we have the support of \(N_{\mathrm{abs}}(s)\) to be
\begin{equation}\label{eq:supp}
\operatorname{supp}N_{\mathrm{abs}}(s)=\begin{cases}\{0\}, & s<1,\\ \{n_{\min},\dots,n_{\max}\}, & s>1,\ s\notin\mathbb Z,\\ \{n_{\min},\dots,n_{\max}-1\}, & s\in\mathbb Z,\end{cases}
\end{equation}
where \(n_{\min}=\lfloor\tfrac{s-1}{2}\rfloor+1\) and \(n_{\max}=\lfloor s\rfloor\). At integer \(s\), we have \(p_{n_{\max}}(s)=0\), since packing \(n_{\max}=s\) unit cars into \([0,s]\) forces every gap to vanish, an event of measure zero.

For a given instance \(U(s)\) and \(n\in\mathbb Z_{\ge0}\), we have by definition that the mass \(p_n(s)\) is the UCP measure \(\mu_{\mathrm{UCP}}\) (\ref{sec:background-ucp}) of the all-jammed cell \(J_n(s)\),
\begin{equation}\label{eq:mass}
p_n(s)=\mu_{\mathrm{UCP}}(J_n(s))=\int_{J_n(s)}f_{\mathrm{UCP}}\ \in\ \mathcal H_{\le n-1},
\end{equation}
where \(f_{\mathrm{UCP}}\) is its density (\ref{sec:joint-order-stats}) and \(\mathcal H_{\le n-1}\) is the \(\mathbb Q[s]\)-module of hyperlogarithms of weight at most \(n-1\) (\ref{def:weight-and-the-integer-delay-alphabet}), by the period theorem (\ref{thm:aomotogoncharov-period-theorem-affine-case}).

The \textbf{count generating polynomial} collects the pmf,
\begin{equation}\label{eq:count-poly}
P_s(z):=\sum_{n\ge0}p_n(s)\,z^n .
\end{equation}
By (\eqref{eq:supp}) its degree is
\begin{equation}\label{eq:degree}
\deg P_s=\begin{cases}\lfloor s\rfloor, & s\notin\mathbb Z,\\ \lfloor s\rfloor-1, & s\in\mathbb Z,\end{cases}
\end{equation}
which is not constant in \(s\).

\textbf{Moments.} Write \(M_k(s)=\mathbb E[N_{\mathrm{abs}}(s)^k]\). Expanding \((1+N^{(1)}_{\mathrm{abs}}+N^{(2)}_{\mathrm{abs}})^k\) in the splitting law (\eqref{eq:split}) and taking expectations gives, for every \(k\ge1\),
\begin{equation}\label{eq:moment-volterra}
M_k(s)=\frac{2}{s-1}\int_0^{s-1}M_k(x)\,\mathrm dx+F_k(s),
\end{equation}
where the two children couple only through
\begin{equation}\label{eq:moment-forcing}
F_k(s)=\frac{1}{s-1}\int_0^{s-1}\ \sum_{\substack{i+j+l=k\\ j<k,\ l<k}}\binom{k}{i,\,j,\,l}\,M_j(x)\,M_l(s-1-x)\,\mathrm dx .
\end{equation}
The terms with \(j=k\) or \(l=k\) are excluded from (\eqref{eq:moment-forcing}): each forces the other two indices to vanish, and the two of them contribute the factor \(2\) in (\eqref{eq:moment-volterra}). Since \(F_k\) involves only moments of order below \(k\), and \(F_1=1\) gives Rényi's equation (\ref{thm:parking-integral-equation-renyi-1958}), (\eqref{eq:moment-volterra}) is a triangular system of second-kind linear Volterra equations sharing one kernel. The general theory of such representations is developed in \ref{sec:aggregate-representations}.

\textbf{Running example (\(s=13/2\)).} Evaluating (\eqref{eq:mass}) numerically gives the pmf

{\footnotesize\begin{longtable}[]{@{}lllll@{}}
\toprule\noalign{}
\(n\) & \(3\) & \(4\) & \(5\) & \(6\) \\
\midrule\noalign{}
\endhead
\bottomrule\noalign{}
\endlastfoot
\(p_n(s)\) & \(0.004\) & \(0.404\) & \(0.577\) & \(0.018\) \\
\end{longtable}}

whence \(\mathbb E[N_{\mathrm{abs}}(13/2)]\approx4.62\) and \(\mathrm{Var}\,N_{\mathrm{abs}}(13/2)\approx0.24\). These values are numerical, and are cross-checked by Monte Carlo.

\subsubsection{Monotonicity in the segment length}

The splitting law (\eqref{eq:split}) presents \(N_{\mathrm{abs}}(s)\) as a mixture, over the split point \(X\), of an independent sum. A distributional property of \(N_{\mathrm{abs}}\) holds for every \(s\) precisely when it is preserved under that mixture. Stochastic monotonicity in \(s\) is so preserved, and we prove it here; the property is used repeatedly in the rest of the section.

\refstepcounter{thmcnt}\label{def:stochastic-order}\textbf{Definition\nobreakspace{}\thethmcnt{} (stochastic order).} For real-valued random variables \(V\) and \(W\), we say that \(W\) \textbf{stochastically dominates} \(V\), and write \(V\preceq_{\mathrm{st}}W\), if and only if
\[\resizebox{\ifdim\width>\linewidth\linewidth\else\width\fi}{!}{$\displaystyle \Pr[V>t]\le\Pr[W>t]\qquad\text{for every }t\in\mathbb R$}\]
(Shaked and Shanthikumar 2007, §1.A).

We use two standard consequences, both from the same source.

\refstepcounter{thmcnt}\label{lem:stochastic-order-expectation}\textbf{Lemma\nobreakspace{}\thethmcnt{} (expectation form).} \(V\preceq_{\mathrm{st}}W\) if and only if \(\mathbb E[\phi(V)]\le\mathbb E[\phi(W)]\) for every non-decreasing \(\phi:\mathbb R\to\mathbb R\) for which both expectations exist (Shaked and Shanthikumar 2007, §1.A).

\refstepcounter{thmcnt}\label{lem:stochastic-order-sums}\textbf{Lemma\nobreakspace{}\thethmcnt{} (closure under independent sums).} If \(V_1\preceq_{\mathrm{st}}W_1\) and \(V_2\preceq_{\mathrm{st}}W_2\), with \(V_1,V_2\) independent and \(W_1,W_2\) independent, then \(V_1+V_2\preceq_{\mathrm{st}}W_1+W_2\) (Shaked and Shanthikumar 2007, Thm 1.A.3).

Neither statement refers to a joint law, so we need not define \(V\) and \(W\) on a common probability space.

\refstepcounter{thmcnt}\label{lem:absorption-fosd}\textbf{Lemma\nobreakspace{}\thethmcnt{}.} The parameterised family \(N_{\mathrm{abs}}(s)\) is stochastically non-decreasing in its parameter \(s\).

\emph{Proof.} By induction on \(\lceil s\rceil\). Substitute \(x=(s-1)u\) in the splitting law (\eqref{eq:split}) and write \(s_1:=(s-1)u\) and \(s_2:=(s-1)(1-u)\) for the two child lengths. Then for every bounded function \(\phi:\mathbb Z_{\ge0}\to\mathbb R\), monotone or not,
\begin{equation}\label{eq:fosd-mix}
\mathbb E[\phi(N_{\mathrm{abs}}(s))]=\int_0^1\mathbb E\big[\phi\big(1+N^{(1)}_{\mathrm{abs}}(s_1)+N^{(2)}_{\mathrm{abs}}(s_2)\big)\big]\,du.
\end{equation}

\textbf{Base case.} By (\eqref{eq:supp}), for \(s\le2\) we have
\[\resizebox{\ifdim\width>\linewidth\linewidth\else\width\fi}{!}{$\displaystyle N_{\mathrm{abs}}(s)=\begin{cases}0, & 0\le s<1,\\ 1, & 1<s\le2,\end{cases}$}\]
which is non-decreasing in \(s\), so the result holds.

\textbf{Induction step.} Let \(s>2\) and fix a non-decreasing \(\phi\) and \(u\in[0,1]\). The child lengths \(s_1\) and \(s_2\) are non-decreasing functions of \(s\) and satisfy \(s_i\le s-1\), so \(\lceil s_i\rceil\le\lceil s\rceil-1\).

By the inductive hypothesis each child count \(N^{(i)}_{\mathrm{abs}}(s_i)\) is therefore stochastically non-decreasing in \(s\). By (\ref{lem:stochastic-order-sums}) their sum is stochastically non-decreasing in \(s\), so the integrand \(\mathbb E[\phi(1+N^{(1)}_{\mathrm{abs}}(s_1)+N^{(2)}_{\mathrm{abs}}(s_2))]\) of (\eqref{eq:fosd-mix}) is non-decreasing in \(s\) for every \(u\in[0,1]\). Integrating over \(u\in[0,1]\) preserves this property, so \(\mathbb E[\phi(N_{\mathrm{abs}}(s))]\) is non-decreasing in \(s\). Since that inequality holds for every non-decreasing \(\phi\), (\ref{lem:stochastic-order-expectation}) gives \(N_{\mathrm{abs}}(s_1)\preceq_{\mathrm{st}}N_{\mathrm{abs}}(s_2)\) whenever \(s_1\le s_2\). \(\square\)

\subsubsection{\texorpdfstring{Modes of \(N_{\mathrm{abs}}(s)\)}{Modes of N\_\{\textbackslash mathrm\{abs\}\}(s)}}

A \textbf{mode} of the pmf \(n\mapsto p_n(s)\) is a maximal block \(B\subseteq\operatorname{supp}N_{\mathrm{abs}}(s)\) of consecutive integers on which \(p_n(s)\) is constant and strictly exceeds its two neighbours. Write \(\operatorname{Modes}(s)\) for the set of modes, and
\[\resizebox{\ifdim\width>\linewidth\linewidth\else\width\fi}{!}{$\displaystyle \nu(s):=|\operatorname{Modes}(s)|$}\]
for their number. A mode \(B\) with \(|B|\ge2\) is a \textbf{plateau}: two or more consecutive counts carrying equal, locally maximal probability. The pmf \(n\mapsto p_n(s)\) is unimodal when \(\nu(s)=1\), and its mode is unique when in addition the single member of \(\operatorname{Modes}(s)\) is a singleton. We will now characterise \(\nu(s)\) as precisely as we can.

\refstepcounter{thmcnt}\label{prop:absorption-mode-count}\textbf{Proposition\nobreakspace{}\thethmcnt{}.} For every \(s>0\),
\begin{equation}\label{eq:mode-bound}
\nu(s)\le\Big\lceil\tfrac{m(s)}{2}\Big\rceil,\qquad\text{where } m(s):=|\operatorname{supp}N_{\mathrm{abs}}(s)|,
\end{equation}
and by (\eqref{eq:supp})
\begin{equation}\label{eq:support-size}
m(s)=\begin{cases}\big\lfloor s\big\rfloor-\big\lfloor\tfrac{s-1}{2}\big\rfloor, & s\notin\mathbb Z,\\[2pt] \big\lfloor s\big\rfloor-\big\lfloor\tfrac{s-1}{2}\big\rfloor-1, & s\in\mathbb Z.\end{cases}
\end{equation}
In particular \(\nu(s)\) is finite for every \(s\), and the bound grows linearly, \(m(s)\sim\tfrac{s+1}{2}\).

\emph{Proof.} Let \(B_1,\dots,B_{\nu(s)}\) be the modes, ordered by their least elements. Fix \(r<\nu(s)\). The blocks \(B_r\) and \(B_{r+1}\) are not adjacent, since a mode is a maximal block on which the mass is constant, so at least one index lies between them; and each mode strictly exceeds its neighbours, so at least one such index \(n\) has \(p_n(s)\) strictly below the common values on \(B_r\) and on \(B_{r+1}\). Choose one such \(n_r\) for each \(r\). The \(\nu(s)\) modes and the \(\nu(s)-1\) indices \(n_1,\dots,n_{\nu(s)-1}\) are disjoint subsets of the support, so together they occupy at least \(2\nu(s)-1\) of its \(m(s)\) points. Hence \(2\nu(s)-1\le m(s)\). \(\square\)

\refstepcounter{thmcnt}\label{prop:absorption-plateau}\textbf{Proposition\nobreakspace{}\thethmcnt{}.} There is an \(s\) for which \(\operatorname{Modes}(s)\) contains a plateau.

\emph{Proof.} Let \(2<s<3\), so that \(\operatorname{supp}N_{\mathrm{abs}}(s)=\{1,2\}\) by (\eqref{eq:supp}). Let \(x_1\) be the position of the first car, uniform on \([0,s-1]\) by (\eqref{eq:split}), and write \(s_L=x_1\) and \(s_R=s-1-x_1\) for the lengths of the two segments it leaves. Then \(s_L+s_R=s-1<2\), so at most one of the two segments is long enough to admit a further car, and \(N_{\mathrm{abs}}(s)=1\) precisely when neither is, that is when \(s_L<1\) and \(s_R<1\). In terms of \(x_1\) these two conditions read \(x_1<1\) and \(x_1>s-2\), so they hold exactly on the sub-interval \((s-2,1)\) of \([0,s-1]\), whose length is \(3-s\). Dividing by the length \(s-1\) of the range of \(x_1\),
\begin{equation}\label{eq:band23}
p_1(s)=\frac{3-s}{s-1},\qquad p_2(s)=1-p_1(s)=\frac{2(s-2)}{s-1},\qquad 2<s<3.
\end{equation}
The two probabilities agree when \(3-s=2(s-2)\), that is at \(s=\tfrac73\), where \(p_1(\tfrac73)=p_2(\tfrac73)=\tfrac12\). Hence \(\operatorname{Modes}(\tfrac73)=\{\{1,2\}\}\), whose member is a plateau. \(\square\)

\begin{figure}
\centering
\pandocbounded{\includegraphics[keepaspectratio,alt={The two masses on the band 2\textless s\textless3. They cross at s=7/3, where both equal 1/2, so the maximising block is \textbackslash\{1,2\textbackslash\} and the mode is not unique.}]{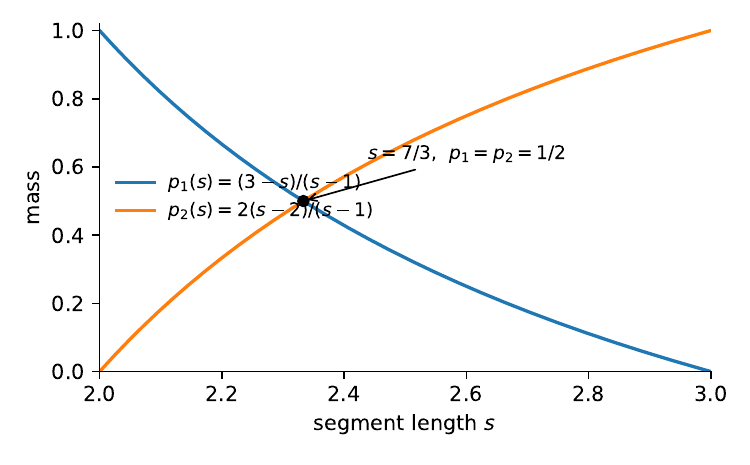}}
\caption{The two masses on the band \(2<s<3\). They cross at \(s=7/3\), where both equal \(1/2\), so the maximising block is \(\{1,2\}\) and the mode is not unique.}
\end{figure}

\subsubsection{Conjectures on the modes}

Three statements about the shape of \(p_\cdot(s)\) stand unproved. They are ordered by strength, each implying the next.

\refstepcounter{thmcnt}\label{conj:absorption-real-rootedness}\textbf{Conjecture\nobreakspace{}\thethmcnt{} (real-rootedness).} For every \(s>0\) the count generating polynomial \(P_s\) of (\eqref{eq:count-poly}) is real-rooted, with all roots in \((-\infty,0]\).

\refstepcounter{thmcnt}\label{conj:absorption-log-concavity}\textbf{Conjecture\nobreakspace{}\thethmcnt{} (log-concavity).} For every \(s>0\) and every \(n\), \(p_n(s)^2\ge p_{n-1}(s)\,p_{n+1}(s)\).

\refstepcounter{thmcnt}\label{conj:absorption-unimodality}\textbf{Conjecture\nobreakspace{}\thethmcnt{} (unimodality).} For every \(s>0\), \(\nu(s)=1\).

Newton's inequalities give (\ref{conj:absorption-real-rootedness}) \(\Rightarrow\) (\ref{conj:absorption-log-concavity}), and a log-concave positive sequence on a contiguous support has a single mode, so (\ref{conj:absorption-log-concavity}) \(\Rightarrow\) (\ref{conj:absorption-unimodality}). Real-rootedness would in addition order the family in \(s\) by the \textbf{likelihood-ratio order}, in which \(V\preceq_{\mathrm{lr}}W\) when \(n\mapsto\Pr[W=n]/\Pr[V=n]\) is non-decreasing; that order is strictly stronger than the one of \ref{def:stochastic-order}.

\textbf{The obstacle.} The splitting law (\eqref{eq:split}) presents \(P_s\) as a \emph{continuous mixture}, over the split position of the products \(P_xP_{s-1-x}\), which does not preserve real-rootedness. The interlacing-family method of Marcus, Spielman and Srivastava supplies the missing preservation when the family admits a common interlacer, in the following sense: a polynomial \(r\) is a \textbf{common interlacer} for a family \(\mathcal F\) of real polynomials when \(\deg r=\deg f-1\) for every \(f\in\mathcal F\) and the roots of \(r\) separate those of each \(f\). The proposition below precludes that possibility, using only the degree requirement, so it refutes every notion of common interlacer that implies it.

\refstepcounter{thmcnt}\label{prop:absorption-no-common-interlacer}\textbf{Proposition\nobreakspace{}\thethmcnt{}.} For \(s>2\) with \(s\notin\mathbb Z\), the family \(\mathcal F_s:=\{P_x\,P_{s-1-x}: x\in(0,s-1)\}\) has no common interlacer.

\emph{Proof.} Write \(\{t\}\) for the fractional part of \(t\). Let \(x\in(0,s-1)\) with \(x\notin\mathbb Z\) and \(s-1-x\notin\mathbb Z\). By (\eqref{eq:degree}), \(\deg P_x=\lfloor x\rfloor\) and \(\deg P_{s-1-x}=\lfloor s-1-x\rfloor\), and since \(s-1-x=(\lfloor s-1\rfloor-\lfloor x\rfloor)+(\{s-1\}-\{x\})\),
\begin{equation}\label{eq:split-degree}
\deg\big(P_xP_{s-1-x}\big)=\lfloor x\rfloor+\lfloor s-1-x\rfloor=\begin{cases}\lfloor s-1\rfloor, & \{x\}\le\{s-1\},\\ \lfloor s-1\rfloor-1, & \{x\}>\{s-1\}.\end{cases}
\end{equation}
Both values occur. Since \(s\notin\mathbb Z\) we have \(0<\{s-1\}<1<s-1\), so any \(x\in(0,\{s-1\}]\) that is not an integer lies in \((0,s-1)\) and has \(\{x\}=x\le\{s-1\}\), giving a member \(f_1\in\mathcal F_s\) of degree \(\lfloor s-1\rfloor\); and any \(x\in(\{s-1\},1)\) lies in \((0,s-1)\) and has \(\{x\}=x>\{s-1\}\), giving a member \(f_2\in\mathcal F_s\) of degree \(\lfloor s-1\rfloor-1\). Suppose \(r\) were a common interlacer for \(\mathcal F_s\). Then \(\deg r=\deg f_1-1=\lfloor s-1\rfloor-1\) and \(\deg r=\deg f_2-1=\lfloor s-1\rfloor-2\), so \(\lfloor s-1\rfloor-1=\lfloor s-1\rfloor-2\), which is false. \(\square\)

\textbf{List of proof attempts and their obstacles}

{\footnotesize\begin{longtable}[]{@{}
  >{\raggedright\arraybackslash}p{\dimexpr 0.5000\linewidth-2\tabcolsep\relax}
  >{\raggedright\arraybackslash}p{\dimexpr 0.5000\linewidth-2\tabcolsep\relax}@{}}
\toprule\noalign{}
\begin{minipage}[b]{\linewidth}\raggedright
route
\end{minipage} & \begin{minipage}[b]{\linewidth}\raggedright
obstacle
\end{minipage} \\
\midrule\noalign{}
\endhead
\bottomrule\noalign{}
\endlastfoot
Laplace or Sturm--Liouville reformulation & the transform relocates the mixture rather than removing it \\
common interlacer & the family has non-constant degree (\ref{prop:absorption-no-common-interlacer}) \\
Hodge index on the convolution Gram matrix & the Gram matrix is not Lorentzian \\
Lorentzian polynomials and the Hodge-theoretic route of Adiprasito, Huh and Katz & that machinery delivers log-concavity for the invariants of a matroid or, more generally, for a Lorentzian polynomial. The coefficients \(p_n(s)\) are not known to be either: no matroid is known whose invariants they are, and deciding whether \(P_s\) is Lorentzian needs the Hessian signature of its homogenisation, which we cannot check without a closed form for \(p_n\) \\
determinantal representation \(P_s\propto\det(A_s-zB_s)\) & the integral of a determinant is not a determinant, and the pencils have \(x\)-dependent size \\
Alexandrov--Fenchel or Brunn--Minkowski & the region carrying exactly \(n\) cars is a non-convex union of cells whose dimension grows with \(n\) \\
symbol calculus on the multiple polylogarithms & positivity of the discriminant is not determined by the symbol, and the top symbol does not cancel \\
threshold reformulation of unimodality & it reduces to log-concavity, which is the conjecture again \\
\end{longtable}}

\paragraph{Comparison with the Tonks process}

The Tonks process places the same unit rods on \([0,s]\) under the Gibbs measure, so its density is flat on the hard-core support, \(f_{\mathbb H}=n!/(s-n)^n\) on \(\mathsf S_n\) (\ref{def:tonks-process-the-jointly-uniform-hard-rod}), whereas \(f_{\mathrm{UCP}}=\sum_\sigma\prod_k1/\ell_k^\sigma\) is not. Its count polynomial is
\begin{equation}\label{eq:tonks-count}
P^{\mathrm{Tonks}}_s(z)=\sum_n\frac{(s-n)^n}{n!}\,z^n,
\end{equation}
the coefficients being the volumes \(|\mathsf T_n|=(s-n)^n/n!\) of the Tonks simplices.

{\footnotesize\begin{longtable}[]{@{}
  >{\raggedright\arraybackslash}p{\dimexpr 0.3333\linewidth-2\tabcolsep\relax}
  >{\raggedright\arraybackslash}p{\dimexpr 0.3333\linewidth-2\tabcolsep\relax}
  >{\raggedright\arraybackslash}p{\dimexpr 0.3333\linewidth-2\tabcolsep\relax}@{}}
\toprule\noalign{}
\begin{minipage}[b]{\linewidth}\raggedright
property
\end{minipage} & \begin{minipage}[b]{\linewidth}\raggedright
UCP, \(P_s\)
\end{minipage} & \begin{minipage}[b]{\linewidth}\raggedright
Tonks, \(P^{\mathrm{Tonks}}_s\)
\end{minipage} \\
\midrule\noalign{}
\endhead
\bottomrule\noalign{}
\endlastfoot
real-rootedness & open (\ref{conj:absorption-real-rootedness}) & holds: a unit-interval graph is claw-free, so its independence polynomial is real-rooted (Chudnovsky and Seymour 2007), and the continuum statement follows by a limit \\
log-concavity & open (\ref{conj:absorption-log-concavity}) & holds, by Newton's inequalities \\
unimodality & open (\ref{conj:absorption-unimodality}) & holds, from log-concavity \\
mode bound \(\nu(s)\le\lceil m(s)/2\rceil\) & proved (\ref{prop:absorption-mode-count}) & holds, the proof using only the support \\
plateaux & occur (\ref{prop:absorption-plateau}) & occur: \(p_1=p_2\) when \(s-1=(s-2)^2/2\), that is at \(s=3+\sqrt3\) \\
\end{longtable}}

\section{Asymptotic Properties of the Order Statistics}

\label{sec:asymptotic-properties}

Companion to \ref{sec:properties-order-stats} (the \textbf{fixed-\(s\), condition-on-\(N(s){=}n\), finite-exact}
layer) and \ref{sec:absorption-time} (the \textbf{count} \(N(s)\) itself). This section treats the
large-segment (\(s\to\infty\)) \textbf{limit laws, concentration, Berry--Esseen rates, and extreme-value behaviour
of functionals of the spatial order statistics} \(X_{(1)}<\dots<X_{(N)}\) of the jammed configuration ---
the object the count asymptotics leave unresolved.

The results below are limit theorems for spatial functionals of the jammed configuration: a central limit
theorem for bulk-gap sums (\textbf{A1}) and its density-general extension (\textbf{Theorem G}), a sub-Gaussian
concentration bound (\textbf{A3}), a Gaussian fluctuation law for the median position (\textbf{A2′}, containing the
symmetric-median law \textbf{A2}), a Weibull law for the largest gap with an explicit constant (\textbf{A5}), and a
functional limit theorem for the empirical gap measure (\textbf{A7}). Each is stated with the classical inputs it
invokes named in full, and each is accompanied by a worked example together with a non-example that violates
exactly one hypothesis.

\subsubsection{Goals \& scope}

Run 1D RSA of unit cars on \([0,s]\) \textbf{to jamming} (no conditioning); let \(N=N(s)\) be the (random)
jamming count, \(N(s)\sim ms+(m{-}1)\) with \(m=0.74759\dots\), and \(X_{(1)}<\dots<X_{(N)}\) the sorted
left endpoint positions, gaps \(G_0,\dots,G_N\). The four fronts, ordered by how much a paper section would
rest on them, and by what the fresh prior-art pass found genuinely open:

\begin{enumerate}
\def\labelenumi{\arabic{enumi}.}
\tightlist
\item
  \textbf{Central order statistics --- limit laws.} CLT / LLN for the \textbf{sample-mean position}, the \textbf{median}
  \(X_{(\lceil N/2\rceil)}\), and \textbf{bulk gaps}. The prior art delivers these \emph{for the count} via
  Penrose--Yukich \textbf{stabilization} (sums of exponentially-stabilizing scores) but \textbf{not for any spatial
  order-statistic or gap functional} --- those are not sums of local scores. Section A1 supplies the
  bulk-gap \textbf{sum} functional CLT with an explicit variance rate, which stabilization does reach, and
  §A2/§A2′ treat the \textbf{median as a rank functional} through a Bahadur representation, which it does not.
\item
  \textbf{Concentration bounds.} Bounded-difference / \textbf{Stein--Chatterjee} (exchangeable-pairs) Bernstein-tail
  concentration for \textbf{Lipschitz functionals} of the jammed configuration. The instrument is canonical
  (Chatterjee 2007); it has \textbf{not} been instantiated for RSA order-statistic/gap functionals, and §A3
  carries out that instantiation by a route of its own, through the split-tree martingale.
\item
  \textbf{Berry--Esseen rates.} Rate of the front-1 CLT via Stein/Malliavin for \textbf{stabilizing functionals}
  (Lachièze-Rey--Schulte--Yukich 2019; Last--Peccati--Schulte). The count rate exists (Schreiber--Penrose--
  Yukich 2007, with log factors \emph{conjectured removable in 1D}); the optimal Poisson-functional method
  has \textbf{explicitly never been applied to RSA}. Section A4 supplies a Kolmogorov bound for a bulk-gap
  functional, of order \(s^{-1/2+\varepsilon}\) for every \(\varepsilon>0\), with its leading constant identified
  for continuous scores.
\item
  \textbf{Extreme value theory --- extremal order statistics.} Domain of attraction of the \textbf{largest gap}
  \(\max_i G_i\), the \textbf{boundary gap} \(G_0\), and the endpoint cars \(X_{(1)},X_{(N)}\). Jamming gaps have
  \textbf{bounded support} (\(G_i<1\)) and \textbf{super-exponentially decaying correlations}, so the max is
  conjectured \textbf{Weibull-type} (finite right endpoint \(1\)) --- opposite to both the i.i.d.-spacing
  \textbf{Gumbel} baseline and the Kakutani interval-\emph{splitting} \textbf{Gaussian} max-gap (Daly--Wade 2025, a
  different model where gaps shrink to \(0\)). The prior art leaves this front untouched; §A5 settles the
  jammed ensemble and Corollary G′ the partial one.
\end{enumerate}

\textbf{What each front establishes.} Front 1 rests on \textbf{A1}, the bulk-gap central limit theorem with a linear
variance rate, on its density-general extension \textbf{Theorem G}, and on \textbf{A2′}, the central-quantile CLT whose
\(p=\tfrac12\) case is the symmetric-median law \textbf{A2} with the closed-form rate \(\tau^2=v/(4m^2)\) and exact
centring. Front 2 rests on \textbf{A3}, a sub-Gaussian concentration bound for bounded gap functionals, proved
from the split-tree martingale. Front 3 rests on \textbf{A4}, which gives the Kolmogorov rate
\(O(s^{-1/2+\varepsilon})\), for every \(\varepsilon>0\), with no unverified hypothesis. Front 4 rests on \textbf{A5},
the Weibull law for the largest gap with the endpoint constant \(c=f_G(1^-)=0.42167\ldots\) in closed form, and
on \textbf{Corollary G′}, which flips the extreme-value class to Gumbel below saturation. Over all levels at once,
\textbf{A7} packages front 1 as a Pyke--Shorack statement for the \emph{whole} gap measure: \(\widehat F_s\Rightarrow F_G\),
the closed-form CDF of §A5(iii), with Gaussian fluctuations --- the bounded, contact-divergent analogue of
Pyke's exponential-spacing law. The boundary gap \(G_0\) is not treated here; the gap order statistics, the boundary gap among them, are
developed in \ref{sec:gap-order-stats}.

Three quantities named in these statements are not evaluated in closed form: the variance rate
\(\sigma_\phi^2\) of \textbf{A1}, the covariance kernel \(\Sigma\) of \textbf{A7}, and the sharp Berry--Esseen constant of
\textbf{A4}, whose \(\log\)-free \(O(s^{-1/2})\) form additionally requires the Grama--Haeusler concentration of the
predictable quadratic variation. Each is a missing formula for an object whose existence is proved.

\subsubsection{Notation \& conventions}

{\footnotesize\begin{longtable}[]{@{}
  >{\raggedright\arraybackslash}p{\dimexpr 0.5000\linewidth-2\tabcolsep\relax}
  >{\raggedright\arraybackslash}p{\dimexpr 0.5000\linewidth-2\tabcolsep\relax}@{}}
\toprule\noalign{}
\begin{minipage}[b]{\linewidth}\raggedright
symbol
\end{minipage} & \begin{minipage}[b]{\linewidth}\raggedright
meaning
\end{minipage} \\
\midrule\noalign{}
\endhead
\bottomrule\noalign{}
\endlastfoot
\(m\) & Rényi parking constant \(=\int_0^\infty e^{-2\int_0^x(1-e^{-u})/u\,du}\,dx=0.7475979\dots\) \\
\(v\) & variance rate, \(\operatorname{Var}N(s)\sim v\,s\), \(v\approx0.03815\) (Dvoretzky--Robbins / Blaisdell--Solomon) \\
\(\xi(x,\mathcal X)\) & a \textbf{stabilizing score} at \(x\) for point set \(\mathcal X\) (Penrose--Yukich); \(\xi\) is reserved for scores, and the population \(p\)-quantile of §A2′ is written \(q_p\) \\
\(R\) & radius of stabilization (here \(O(1)\)-tailed: super-exponential decorrelation) \\
RSA \textbf{field} / configuration & the random set of parked-car positions on \([0,s]\) --- a point process (``field'' in the geometric-probability sense, \emph{not} an algebraic field) \\
MDA & max-domain of attraction; the three classes are named in words (Gumbel, Weibull, Fréchet), the Gumbel law alone carrying the symbol \(\Lambda\) \\
\(\Phi(t)\) & the \textbf{Rényi kernel} \(\exp\big(-2\int_0^t(1-e^{-v})/v\,dv\big)\), so that \(m=\int_0^\infty\Phi\); \(\Phi\) denotes this kernel and nothing else \\
\(\Phi_{\mathcal N},\ \varphi_{\mathcal N}\) & standard normal distribution function and density (§A4), written with the subscript because \(\Phi\) and \(\varphi\) are taken \\
\(\Psi(y,t)\) & tail integral of the kinetic gap density, \(\Psi(y,t)=\int_y^\infty\rho(g,t)\,dg\) (§A5) \\
\(\rho(g,t)\) & density per unit length of gaps of length \(g\) at Poissonized time \(t\) (§A5(iii)a) \\
\(f_G\), \(F_G\) & limiting jammed gap density and distribution function on \([0,1)\) \\
\(c\) & the gap density at the jamming endpoint, \(c=f_G(1^-)=0.42167\ldots\); \(c\) denotes this constant and nothing else \\
\(x_0\) & the midpoint \((s{-}1)/2\) of the segment, the exact centring of the median law (§A2) \\
\(\gamma\) & Euler--Mascheroni constant (§A5(iii)); with a subscript, \(\gamma_j\) is the \(j\)-th standardised cumulant (§A3, §A4) \\
\(\theta\) & coverage \(n/s\), the first regime coordinate of the plane below \\
\(\varphi\) & jamming fraction \(j/n\), the second regime coordinate; the symbol carries this meaning throughout, never that of a normal density \\
\end{longtable}}

\textbf{Regime.} Unlike the fixed-\(s\) sections, \(N\) is random and statements are as \(s\to\infty\); the two
parameters \(s,N\) are tied by \(N/s\to m\) on the jammed manifold (Theorem \ref{thm:parking-integral-equation-renyi-1958};
\(s/n\not\to1\) is the only nontrivial regime). \textbf{Stabilization} (front 1) means a local perturbation of the arrival
process changes a score only within an a.s.-finite radius \(R\); the count is the headline sum-of-scores
functional, and RSA jamming has super-exponentially decaying pair correlations, so bulk \textbf{sums} are
stabilization-reachable while \textbf{rank/extremal} functionals (\(X_{(\lceil N/2\rceil)}\), \(\max_i G_i\)) are not.

\subsubsection{The regime plane: coverage moves the constants, the jamming fraction moves the class}

Results are organized by \textbf{regime}, not by a jammed/unjammed binary. Two coordinates fix every asymptotic statement:

\begin{itemize}
\tightlist
\item
  \textbf{Coverage} \(\theta=n/s\in[0,1]\) --- how full the line is; RSA-typical jamming sits at \(\theta^*=m\approx0.7476\). Along \(\theta\) the \emph{rates and constants} move continuously (\(\sigma_\phi^2(\theta)\); \(\tau^2(p,\eta)=v_\eta\,p(1{-}p)/\eta^2\); the A7 kernel).
\item
  \textbf{Jamming fraction} \(\varphi=j/n\in[0,1]\) --- the share of gaps capped (\(<1\)). At \(\varphi=1\) (saturation) every gap \(<1\); \(\varphi<1\) leaves uncapped voids. Across the \(\varphi=1\) boundary the \emph{class} jumps: the EVT flips \textbf{Weibull\(\leftrightarrow\)Gumbel} (A5) and score-boundedness becomes necessary (A1).
\end{itemize}

\textbf{The feasible region (Lean-verified).} The plane's own boundary is fixed without probability. A configuration of \(n\) unit cars on \([0,s]\) carrying \(v\) void gaps (each \(\ge1\)) satisfies \(n+v\le s\), so coverage obeys \(\theta\le n/(n+v)\); writing \(v=(n{-}1)(1{-}\varphi)\) for the void internal gaps, the upper envelope \(\theta_{\max}(\varphi)\) rises from \(n/(2n{-}1)\) at \(\varphi=0\) to \(1\) at \(\varphi=1\). There is no lower bound: a single long void sends \(\theta\to0\) at any \(\varphi\), so \(\varphi=1\) alone does not force \(\theta>\tfrac12\) --- only full saturation (every gap \(<1\), not merely every internal gap) does, and there the count is pinned to \(\lceil(s{+}1)/2\rceil\le n\le\lfloor s\rfloor\). Both facts are machine-checked (\texttt{PartialConfig.\allowbreak count\_\allowbreak add\_\allowbreak voids\_\allowbreak le}; \texttt{JammedConfig.\allowbreak lt\_\allowbreak two\_\allowbreak mul\_\allowbreak add\_\allowbreak one}, \texttt{feasible\_\allowbreak range}).

Each section below reads the plane through its two sharpest slices, \textbf{\(\varphi=1\) (saturated, ``jammed'')} and \textbf{\(\varphi<1\) (partial, density \(\eta\), ``general'')}, because the \(\varphi=1\) boundary and the two proof engines divide the results:

\begin{itemize}
\tightlist
\item
  \textbf{Stabilisation} --- local super-exp dependence; holds at \emph{any} density, never uses the Rényi split. Carries across the plane (\(\theta<m\)).
\item
  \textbf{The Rényi split} \(N=1+N_L+N_R\) --- needs the saturation recursion / grand-canonical \(N\); jammed-only at the count level, but its Palm form (Proposition \ref{prop:conditional-palm-factorisation}) is grand-canonical (all \(t\)).
\end{itemize}

\textbf{The two ensembles, entry by entry.} Every qualitative difference between them follows from one fact --- jammed \(=\) partial conditioned on saturation, every gap below \(1\) --- so the cap \(g<1\) acts as the order parameter of a subcritical-versus-critical distinction. The last column gives the section in which each entry is stated and proved.

{\footnotesize\begin{longtable}[]{@{}
  >{\raggedright\arraybackslash}p{\dimexpr 0.2500\linewidth-2\tabcolsep\relax}
  >{\raggedright\arraybackslash}p{\dimexpr 0.2500\linewidth-2\tabcolsep\relax}
  >{\raggedright\arraybackslash}p{\dimexpr 0.2500\linewidth-2\tabcolsep\relax}
  >{\raggedright\arraybackslash}p{\dimexpr 0.2500\linewidth-2\tabcolsep\relax}@{}}
\toprule\noalign{}
\begin{minipage}[b]{\linewidth}\raggedright
aspect
\end{minipage} & \begin{minipage}[b]{\linewidth}\raggedright
partial (subcritical, \(\eta<\rho_c\))
\end{minipage} & \begin{minipage}[b]{\linewidth}\raggedright
jammed (saturated)
\end{minipage} & \begin{minipage}[b]{\linewidth}\raggedright
established in
\end{minipage} \\
\midrule\noalign{}
\endhead
\bottomrule\noalign{}
\endlastfoot
gap support & \textbf{unbounded} (large voids) & \([0,1)\), hard-capped & --- \\
gap / void tail & \textbf{exponential}, rate \(\lambda=t\) & finite endpoint, contact-divergent & §A5, Lemma V \\
max-gap EVT & \textbf{Gumbel} & \textbf{Weibull} & §A5, Corollary G′ \\
jamming graph & linear forest, geometric components & single path \(P_n\) (connected) & \ref{sec:process-related} \\
unbounded scores (\(g^2\)) & heavy-tailed, CLT can fail & bounded, CLT holds & §A1, Theorem G \\
bounded-score CLT & rate \(\sigma^2_{\phi,\text{part}}\) & rate \(\sigma^2_{\phi,\text{jam}}\) & §A1, Theorem G \\
spacing empirical process & \(\approx\) classical Pyke bridge (weak dep.) & finite-range Gaussian, \textbf{departs} bridge & §A7 \\
global structure & density-dependent ``gas'' & renewal, non-hyperuniform, Rényi-split & §A7, \ref{sec:process-related} \\
\end{longtable}}

The broad distinction is the classical kinetics-versus-jamming picture; the sharp entries --- the EVT flip, the boundedness dichotomy in the CLT, and the contrast between the two spacing kernels --- are what the sections below add.

\textbf{The results on the plane.} Placing each result on the two coordinates shows what moves along \(\theta\) (rates and constants) and what flips across the \(\varphi=1\) boundary (the class of the limit law).

{\footnotesize\begin{longtable}[]{@{}
  >{\raggedright\arraybackslash}p{\dimexpr 0.2500\linewidth-2\tabcolsep\relax}
  >{\raggedright\arraybackslash}p{\dimexpr 0.2500\linewidth-2\tabcolsep\relax}
  >{\raggedright\arraybackslash}p{\dimexpr 0.2500\linewidth-2\tabcolsep\relax}
  >{\raggedright\arraybackslash}p{\dimexpr 0.2500\linewidth-2\tabcolsep\relax}@{}}
\toprule\noalign{}
\begin{minipage}[b]{\linewidth}\raggedright
result
\end{minipage} & \begin{minipage}[b]{\linewidth}\raggedright
sparse \(\theta\to0\)
\end{minipage} & \begin{minipage}[b]{\linewidth}\raggedright
partial \(\theta\in(0,m),\varphi<1\)
\end{minipage} & \begin{minipage}[b]{\linewidth}\raggedright
saturated \(\theta=m,\varphi=1\)
\end{minipage} \\
\midrule\noalign{}
\endhead
\bottomrule\noalign{}
\endlastfoot
\textbf{A1} bulk CLT & Gaussian (weak dep) & Gaussian, bounded \(\phi\) (Thm G); \textbf{\(\alpha\)-stable} super-tail & Gaussian, all \(\phi\) \\
\textbf{A2/A2′} quantile CLT & --- & \(\tau^2(p,\eta)=v_\eta\,p(1{-}p)/\eta^2\) (§A2′) & median \(\tau^2=v/4m^2\) \\
\textbf{A3} concentration & tighter (more blocks) & sub-Gaussian, bounded (\textbf{open} --- §A3) & sub-Gaussian \\
\textbf{A4} Berry--Esseen & --- & \(O(s^{-1/2+\varepsilon})\) (stabilisation BE, open) & \(O(s^{-1/2+\varepsilon})\) proven \\
\textbf{A5} max-gap EVT & \textbf{Gumbel} & \textbf{Gumbel} & \textbf{Weibull} (\(\varphi{=}1\) flip) \\
\textbf{A7} emp. law & \(\approx\) Pyke bridge & \(\varphi F_{\rm cap}{+}(1{-}\varphi)F_{\rm void}\) & \textbf{func. CLT} (§A7) \\
\end{longtable}}

Moving right along \(\theta\) changes rates continuously; crossing \(\varphi=1\) changes classes, as the EVT entry of A5 and the boundedness entry of A1 show. The sub-saturation interior (\(\theta<m\), \(\varphi<1\)) is stabilisation-reachable, and A2′, A7, and the stabilisation half of A1 are proved there. In the super-dense strip \(\theta>m\) every entry is open: that regime is a large-deviation one, with neither stabilisation nor the Rényi split available, and only the hard endpoint \(G_i<1\) survives as structure; those questions. The two correlation kernels that carry no closed form, the A7 covariance \(\Sigma\) and the variance rate \(\sigma_\phi^2\), are missing formulae rather than open results.

One feature of saturation determines which of the two engines applies to a given result: saturation bounds every gap by \(G_i<1\), whereas the partial process has unbounded gaps, so an unbounded score there can carry a heavy tail that the jammed process never produces.

Three terms are used throughout. A result is \textbf{proven} when its core is computation-independent, a \textbf{conditional theorem} when it holds given named classical inputs or a stated hypothesis, and a \textbf{conjecture} when the evidence is numerical and the proof open. Each section opens with a table in these terms; the detail follows in the cells beneath.

\subsection{A1 · Asymptotic normality of bulk-gap functionals}

\(S_s=\sum_i\phi(G_i)\) over the jamming gaps: a CLT with a linear (extensive) variance rate.

{\footnotesize\begin{longtable}[]{@{}
  >{\raggedright\arraybackslash}p{\dimexpr 0.3333\linewidth-2\tabcolsep\relax}
  >{\raggedright\arraybackslash}p{\dimexpr 0.3333\linewidth-2\tabcolsep\relax}
  >{\raggedright\arraybackslash}p{\dimexpr 0.3333\linewidth-2\tabcolsep\relax}@{}}
\toprule\noalign{}
\begin{minipage}[b]{\linewidth}\raggedright
case
\end{minipage} & \begin{minipage}[b]{\linewidth}\raggedright
statement
\end{minipage} & \begin{minipage}[b]{\linewidth}\raggedright
status
\end{minipage} \\
\midrule\noalign{}
\endhead
\bottomrule\noalign{}
\endlastfoot
\textbf{\(\varphi{=}1\) (saturated)} & \((S_s-\mathbb E S_s)/\sqrt{\operatorname{Var}}\Rightarrow\mathcal N(0,1)\), \(\operatorname{Var}\sim\sigma_\phi^2 s\) & \textbf{proven} --- Penrose--Yukich stabilisation of the bounded gap score \\
\textbf{\(\varphi<1\) (partial, \(\eta\))} & same CLT, ensemble-specific rate & \textbf{proven for bounded \(\phi\)} (Theorem G); \textbf{unbounded \(\phi\) jammed-only} (heavy tail from large voids) \\
\end{longtable}}

Two questions remain open in both cases. The variance rate \(\sigma_\phi^2\) is identified as a limit but has no closed form, and the extensivity of the general-\(k\) cumulant is not established by an in-house induction.

\paragraph{\texorpdfstring{Saturated case (\(\varphi{=}1\))}{Saturated case (\textbackslash varphi\{=\}1)}}

\textbf{Theorems used (§A1).}

\begin{quote}
\textbf{Thm 1 (stabilization CLT).} \emph{M. D. Penrose \& J. E. Yukich,} ``Central limit theorems for some graphs in computational geometry,'' \emph{Ann. Appl. Probab. 11 (2001) 1005--1041; Penrose,} Random Geometric Graphs \emph{(OUP 2003), Thm 2.16.} Let \(\xi\) be translation-invariant on a Poisson process \(\mathcal P_\lambda\) on \([0,\lambda]\), \textbf{exponentially stabilizing} (an a.s.-finite radius \(R\) with \(\Pr(R>r)\le C_1e^{-C_2r}\), beyond which added points do not change \(\xi\)), with a uniform \((2{+}\delta)\)-moment bound. Then for \(H_\lambda=\sum_{x\in\mathcal P_\lambda}\xi(x;\mathcal P_\lambda)\): \(\lambda^{-1}\operatorname{Var}H_\lambda\to\sigma^2\) and \(\lambda^{-1/2}(H_\lambda-\mathbb E H_\lambda)\Rightarrow\mathcal N(0,\sigma^2)\).
\end{quote}

\begin{quote}
\textbf{Thm 2 (RSA stabilization).} \emph{M. D. Penrose,} ``Random parking, sequential adsorption, and the jamming limit,'' \emph{Comm. Math. Phys. 218 (2001) 153--176.} The saturated RSA configuration is exponentially stabilizing: a unit window's state is fixed by the arrivals within a radius \(R\), \(\Pr(R>r)\le C_1e^{-C_2r}\).
\end{quote}

\refstepcounter{thmcnt}\label{thm:bulk-gap-clt}\textbf{Theorem\nobreakspace{}\thethmcnt{} (bulk-gap CLT).} Let \(\phi:[0,1)\to\mathbb R\) be bounded with \(\sigma_\phi^2>0\) (equivalently \(\phi\not\propto1{+}g\)). For the jammed UCP on \([0,s]\), \(S_s=\sum_i\phi(G_i)\) obeys, as \(s\to\infty\),
\[\resizebox{\ifdim\width>\linewidth\linewidth\else\width\fi}{!}{$\displaystyle \mathbb E S_s\sim a_\phi s,\qquad \operatorname{Var}S_s\sim\sigma_\phi^2 s,\qquad \frac{S_s-\mathbb E S_s}{\sqrt{\operatorname{Var}S_s}}\Rightarrow\mathcal N(0,1).$}\]

\emph{Remark (degenerate direction).} The excluded \(\phi\propto1{+}g\) is degenerate: \(\sum_i(1{+}G_i)=s{+}1\) is a.s. constant, so \(\sigma_\phi^2=0\); the span \(\{1,g\}\) is the count in disguise.

\textbf{Examples.} The two running choices are the \textbf{tight-contact count} \(K_s=\#\{i:G_i<\tfrac12\}\) (indicator score) and the \textbf{gap energy} \(E_s=\sum_i G_i^2\); both are bounded with \(\sigma_\phi^2>0\), giving the rates \(\sigma_K^2,\sigma_E^2\) below.

\textbf{Proof.} Set \(\xi(x_i):=\phi(G_i)\), the score at car \(i\) carrying \(\phi\) of the gap to its right, so that \(S_s=\sum_{\text{cars}}\xi(x_i)\) is a sum of scores over the jammed configuration. The four hypotheses of Thm 1 hold for this score.

The score \(\xi(x_i)\) reads only the local gap \(G_i\), and so is \textbf{translation-invariant}. The jammed RSA configuration is exponentially stabilizing by Thm 2: the state of a unit window, and with it the gap \(G_i\), is fixed by the arrivals within a radius \(R\) whose tail is exponential. Since \(\xi\) is a functional of that window alone, it inherits the radius and is \textbf{exponentially stabilizing}. Boundedness of \(\phi\) gives \(|\xi|\le C\), which supplies the \textbf{uniform \((2{+}\delta)\)-moment bound}. The hypothesis \(\phi\not\propto1{+}g\) gives \(\sigma_\phi^2>0\), so the limit is \textbf{non-degenerate}.

Thm 1 therefore applies to \(S_s\) and yields both the variance rate \(\operatorname{Var}S_s\sim\sigma_\phi^2 s\) and the limit \((S_s-\mathbb E S_s)/\sqrt{\operatorname{Var}S_s}\Rightarrow\mathcal N(0,1)\). \(\qquad\blacksquare\)

\textbf{Variance rate.} \(\sigma_\phi^2\) is the deterministic limit of \(\operatorname{Var}(S_s)/s\) from the second-order Rényi march, whose rate is the Dvoretzky--Robbins constant (\ref{sec:background-aggregate}): \(\sigma_E^2=0.0382\), \(\sigma_K^2\approx0.281\). It is \emph{not} the naive per-gap \(m\!\int\phi^2 f_G\) (\(=0.053\) for \(K\)), which ignores the random-\(N\) coupling. The mean rate \(a_\phi=m\!\int_0^1\phi f_G\) is classical (G18).

\textbf{Discussion.} First instantiation of the stabilization CLT for a \emph{nonlinear} RSA gap score; the count (\(\xi\equiv1\)) is the linear case. A complementary elementary route --- the cumulant cascade (G18″) --- gives \(\kappa_k=C_k s+O(1)\) for \(k\le3\); the general-\(k\) in-house induction is blocked by the \(\sum G_i=s{-}N\) coupling and superseded by Thm 1. Rank/extremal functionals (A2, A5) are not sums of local scores, so they fall outside this argument.

\paragraph{Example vs non-example --- A1 CLT}

\(\phi{=}g^2\) (bounded, \(\sigma^2{>}0\)) \(\Rightarrow\) standardized \(S_s\to N(0,1)\) (KS \(\to0.01\), skew \(\to0\)). The
unbounded \(\phi{=}1/g\) (\(\mathbb E[1/g]{=}\infty\) --- a single tiny gap carries the sum) stays heavy-tailed: no CLT
(skew \(\approx60\), KS flat \(\approx0.46\)).

\begin{figure}
\centering
\pandocbounded{\includegraphics[keepaspectratio,alt={A1 CLT: example vs non-example}]{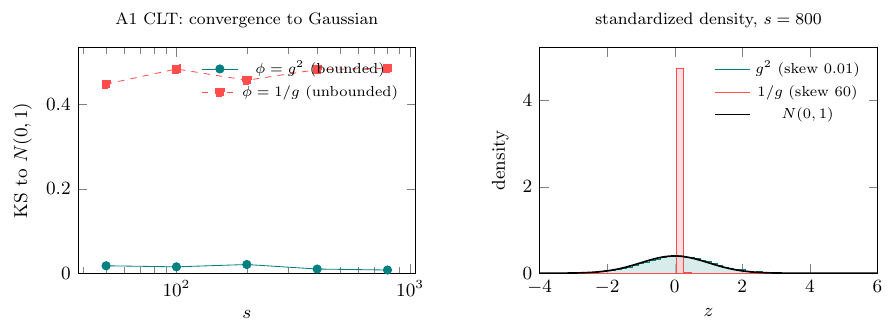}}
\caption{A1 CLT: example vs non-example}
\end{figure}

\paragraph{\texorpdfstring{Partial case (\(\varphi<1\), density \(\eta\)) --- Theorem G (bounded scores carry; unbounded is jammed-only)}{Partial case (\textbackslash varphi\textless1, density \textbackslash eta) --- Theorem G (bounded scores carry; unbounded is jammed-only)}}

\begin{quote}
\textbf{Theorem G (ensemble-generality of the bulk-gap CLT).} Run the UCP either to saturation (jammed) or at fixed density \(\eta<\rho_c\) (partial), \(s\to\infty\). Let \(S_s=\sum_i\phi(G_i)\) with \(\phi\) \textbf{bounded} (\(|\phi|\le C\)) and \(\sigma_\phi^2>0\). Then \(\operatorname{Var}(S_s)\sim\sigma_{\phi,e}^2\,s\) and
\[\resizebox{\ifdim\width>\linewidth\linewidth\else\width\fi}{!}{$\displaystyle \frac{S_s-\mathbb E S_s}{\sqrt{\operatorname{Var}(S_s)}}\ \Rightarrow\ \mathcal N(0,1),$}\]
with an \textbf{ensemble-specific} rate \(\sigma_{\phi,e}^2\). If \(\phi\) is \textbf{unbounded}, the conclusion holds in the jammed ensemble (where \(G_i<1\) caps \(\phi\)) but can \textbf{fail} in the partial ensemble.
\end{quote}

\emph{Proof.} \textbf{(1) Stabilisation at any density.} The sequential-adsorption field is exponentially stabilising: the configuration in a bounded window \(W\) is determined, outside a super-exponentially small probability, by the arrivals in a bounded enlargement of \(W\) (Penrose 2001; Penrose--Yukich 2002; Schreiber--Penrose--Yukich 2007). This is a property of the deposition dynamics --- each arrival excluded by earlier ones --- and holds at every density, sub-saturation \emph{or} saturated; it never invokes the Rényi split. \textbf{(2) Bounded stabilising scores.} \(G_i\) is a difference of two adjacent car positions, each fixed by a stabilising window, so \(\phi(G_i)\) is a bounded (\(|\phi|\le C\)) functional of a window with an exponentially decaying stabilisation radius. \textbf{(3) The stabilisation CLT.} \(S_s\) is then a sum of exponentially-stabilising bounded scores of the RSA field; the Penrose--Yukich stabilisation CLT (Penrose--Yukich 2001; Baryshnikov--Yukich 2005; Last--Peccati--Schulte 2016) gives \(\operatorname{Var}(S_s)=\Theta(s)\) with a nondegenerate limiting variance density and asymptotic normality, using the \((2{+}\delta)\)-moment bound that \(|\phi|\le C\) supplies. The rate \(\sigma_{\phi,e}^2\) is computed from the \emph{ensemble's own} Palm two-point structure, hence ensemble-specific. \textbf{(4) Boundedness is necessary.} \(\phi(G_i)\) inherits the tail of \(G_i\). In the jammed ensemble \(G_i<1\) a.s., so any \(\phi\) continuous on \([0,1]\) is bounded and (3) applies. In the partial ensemble the gaps are unbounded (the largest empty region grows), so an unbounded \(\phi\) --- e.g.~the \textbf{gap energy} \(\phi(g)=g^2\) of §A1 --- has a heavy tail: one large gap can dominate \(S_s\), the Lindeberg/\((2{+}\delta)\) condition fails, and the CLT need not hold. \(\qquad\blacksquare\)

\subsection{A2 · The symmetric-median position is Gaussian}

\(\tilde X_{\rm med}\) is Gaussian with closed-form rate \(\tau^2=v/(4m^2)\), via a Bahadur reduction to the count CLT.

{\footnotesize\begin{longtable}[]{@{}
  >{\raggedright\arraybackslash}p{\dimexpr 0.3333\linewidth-2\tabcolsep\relax}
  >{\raggedright\arraybackslash}p{\dimexpr 0.3333\linewidth-2\tabcolsep\relax}
  >{\raggedright\arraybackslash}p{\dimexpr 0.3333\linewidth-2\tabcolsep\relax}@{}}
\toprule\noalign{}
\begin{minipage}[b]{\linewidth}\raggedright
case
\end{minipage} & \begin{minipage}[b]{\linewidth}\raggedright
statement
\end{minipage} & \begin{minipage}[b]{\linewidth}\raggedright
status
\end{minipage} \\
\midrule\noalign{}
\endhead
\bottomrule\noalign{}
\endlastfoot
\textbf{\(\varphi{=}1\) (saturated)} & \((\tilde X_{\rm med}-\tfrac{s-1}{2})/\sqrt s\Rightarrow\mathcal N(0,\tau^2)\), \(\tau^2=v/(4m^2)\) & \textbf{conditional theorem} --- given (i) the count CLT, (ii) shielding-independence of the half-counts, (iii) a Bahadur--Kiefer remainder; core (linearisation + exact reflection-centring) proven \\
\textbf{\(\varphi<1\) (partial, \(\eta\))} \(+\) \textbf{all central quantiles} & \(\hat X_p\) CLT, \(\tau^2(p,\eta)=v_\eta\,p(1{-}p)/\eta^2\) & \textbf{proven (§A2′)} --- Bahadur 1966/Kiefer 1967 (density-general linearisation) \(+\) Penrose 2001 count CLT (any density) \(+\) §G5′ half-count independence. \textbf{No} Rényi split needed; median is \(p=\tfrac12\). Closes the general case \\
\end{longtable}}

\paragraph{\texorpdfstring{Saturated case (\(\varphi{=}1\))}{Saturated case (\textbackslash varphi\{=\}1)}}

\textbf{Setup.} The \textbf{symmetric median} car position is \(\tilde X_{\rm med}=\tfrac12\big(X_{(N/2)}+X_{(N/2+1)}\big)\) (the middle left endpoint for odd \(N\)). Reflection fixes the centre. The map \(x\mapsto(s-1)-x\) carries the parking process on \([0,s]\) to itself, so a configuration and its reflection are equal in law (Proposition \ref{prop:reflection-symmetry}); the left endpoints lie in \([0,s-1]\) and rank reversal exchanges the two middle ranks, so the law of \(\tilde X_{\rm med}\) is symmetric about \((s-1)/2\). Symmetry gives more than the shape of the law. The median lies in \([0,s-1]\), so it is integrable, and \(\tilde X_{\rm med}\overset{d}{=}(s-1)-\tilde X_{\rm med}\) forces \(\mathbb E[\tilde X_{\rm med}]=(s-1)-\mathbb E[\tilde X_{\rm med}]\), that is
\[\resizebox{\ifdim\width>\linewidth\linewidth\else\width\fi}{!}{$\displaystyle \mathbb E\big[\tilde X_{\rm med}\big]=\frac{s-1}{2}\qquad\text{exactly, at every }s.$}\]
No asymptotics enter and nothing is assumed. What symmetry does not give is the \textbf{scale}, and Theorem \ref{thm:sub-gaussian-concentration} does not supply it either: its hypothesis class is the gap-additive sums \(S_s=\sum_i\phi(G_i)\), and a rank functional is not one. The obvious repair fails for the same reason --- a count restricted to a sub-interval, \(N_{\le t}\), is not of that form, since it depends on the car positions and not only on the gap lengths. The scale therefore comes from the half-count route below, which is conditional on the three inputs listed in the regime table, and the law of large numbers \(\tilde X_{\rm med}=(s-1)/2+o_P(s)\) is conditional with it; the \textbf{fluctuation} law is A2's object.

The median is a \textbf{rank} functional, not a sum of local scores, so A1/A3's stabilisation machinery does not apply directly. The route is the \textbf{Bahadur representation}, which reduces the rank functional to the \textbf{count}: the median sits where the running count crosses \(N/2\), so a wobble in the count maps to a wobble in the median. Writing \(N_L=\#\{\text{cars with left endpoint}\le x_0\}\) for the count to the left of the midpoint \(x_0=\tfrac{s-1}{2}\),
\[\resizebox{\ifdim\width>\linewidth\linewidth\else\width\fi}{!}{$\displaystyle \tilde X_{\rm med}-\tfrac{s-1}{2}\ \approx\ \frac{N/2-N_L}{m}=\frac{N_R-N_L}{2m},$}\]
the \textbf{half-count imbalance} (\(m\) the car intensity). The proof (next cell) makes this exact and supplies the limit.

\paragraph{§A2 proof --- the symmetric-median CLT via the Bahadur representation}

\textbf{Theorems used (§A2).}

\begin{quote}
\textbf{Thm 1 (count CLT).} Sub-interval counts of the jammed RSA process are asymptotically Gaussian with linear variance (Penrose--Yukich 2001, via stabilisation; equivalently Dvoretzky--Robbins 1964 for the total count): on a sub-interval of length \(\ell\), \((N_L-\mathbb E N_L)/\sqrt\ell\Rightarrow\mathcal N(0,\tilde v)\), per-length rate \(\tilde v=v\).
\end{quote}

\begin{quote}
\textbf{Thm 2 (Bahadur--Kiefer representation).} \emph{R. R. Bahadur, Ann. Math. Statist. 37 (1966) 577--580; J. Kiefer, ibid. 38 (1967) 1323--1342.} For a point process of continuous intensity \(f>0\) at the quantile, the sample quantile linearises: \(\hat q_p=q_p+\dfrac{pN-N_L(q_p)}{N f(q_p)}+R\), \(R=o_P(N^{-1/2})\) (Kiefer rate \(O(N^{-3/4}\log N)\)).
\end{quote}

\begin{quote}
\textbf{Thm 3 (shielding independence).} A gap \(\ge1\) at \(x_0\) splits the configuration into independent sub-segments (§G5′ / Rényi split), so \(\operatorname{Cov}(N_L,N_R)=O(1)\).
\end{quote}

\refstepcounter{thmcnt}\label{thm:symmetric-median-clt}\textbf{Theorem\nobreakspace{}\thethmcnt{} (symmetric-median CLT).} In the jammed UCP, the symmetric median obeys
\[\resizebox{\ifdim\width>\linewidth\linewidth\else\width\fi}{!}{$\displaystyle \frac{\tilde X_{\rm med}-\tfrac{s-1}{2}}{\sqrt s}\Rightarrow\mathcal N(0,\tau^2),\qquad \tau^2=\frac{v}{4m^2},$}\]
centred exactly at \(\tfrac{s-1}{2}\) (offset \(b=0\)).

\emph{Remark (exact centring).} Reflection sends \(X_{(i)}\) to \((s{-}1)-X_{(N+1-i)}\) in law, so \(\tilde X_{\rm med}\) is invariant in law under \(x\mapsto(s{-}1)-x\); its mean is therefore the fixed point \(\mathbb E\tilde X_{\rm med}=\tfrac{s-1}{2}\) of that reflection, and the offset vanishes, \(b=0\). The one-sided median \(X_{(\lceil N/2\rceil)}\) carries instead an offset \(b\approx-0.33\).

\textbf{Proof.} Because the car intensity is \(m\) and bounded, Thm 2 linearises the counting function about the midpoint: \(N_L(x)=N_L(x_0)+m(x-x_0)+o_P(\sqrt s)\) for \(x\) near \(x_0\). The symmetric median is the position at which the running count crosses one half of the total, \(N_L(\tilde X_{\rm med})=N/2+O(1)\), so inverting the linearisation gives
\[\resizebox{\ifdim\width>\linewidth\linewidth\else\width\fi}{!}{$\displaystyle \tilde X_{\rm med}-x_0=\frac{N/2-N_L}{m}+o_P(\sqrt s).$}\]
The numerator is one half of the imbalance between the two half-counts, \(N/2-N_L=\tfrac12(N_R-N_L)\), whence \(\tilde X_{\rm med}-x_0=(N_R-N_L)/(2m)+o_P(\sqrt s)\): the median displacement is the half-count imbalance, rescaled by the intensity.

It remains to identify the law of that imbalance. By Thm 3 the two half-counts are shielded from one another, so \(\operatorname{Cov}(N_L,N_R)=O(1)\) and
\[\resizebox{\ifdim\width>\linewidth\linewidth\else\width\fi}{!}{$\displaystyle \operatorname{Var}(N_R-N_L)=\operatorname{Var}N-4\operatorname{Cov}(N_L,N_R)=v\,s+O(1).$}\]
Thm 1 supplies the limit law at that variance, \((N_R-N_L)/\sqrt s\Rightarrow\mathcal N(0,v)\). Combining it with the linearisation, whose remainder is \(o_P(\sqrt s)\), gives \((\tilde X_{\rm med}-x_0)/\sqrt s\Rightarrow\mathcal N(0,v/4m^2)\). \(\qquad\blacksquare\)

\textbf{Rate.} \(\tau^2=v/(4m^2)\approx0.01706\) (closed form; \(v\approx0.03815\)). \(\operatorname{Var}/s\to0.017\), KS-vs-Normal \(p\approx0.6\)--\(0.8\), \(\operatorname{corr}(N_L,N_R)\to0\).

\textbf{Discussion.} First fluctuation limit law stated for a spatial order statistic of jammed RSA, reducing the rank functional to the count via the half-imbalance. The one-sided median \(X_{(\lceil N/2\rceil)}\) obeys the same CLT with an \(O(1)\) offset; the symmetric median makes centring exact. Generalised to all central quantiles and any density in §A2′.

\paragraph{Example vs non-example --- A2 median CLT}

The \textbf{central} rank (median) Gaussianizes (KS \(\to0.01\)); the \textbf{extreme} rank \(X_{(1)}\) does \emph{not} (KS flat
\(\approx0.09\)) --- it converges to the fixed boundary-gap law, not a normal. The Bahadur half-imbalance argument needs
a positive \emph{local} car density, which an extreme order statistic lacks.

\begin{figure}
\centering
\pandocbounded{\includegraphics[keepaspectratio,alt={A2 median: central vs extreme rank}]{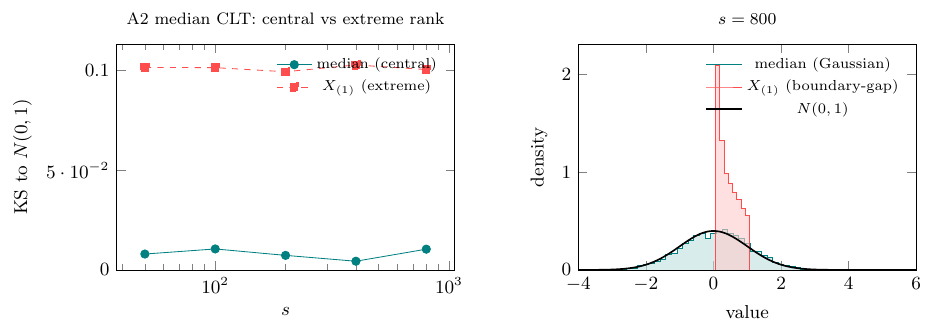}}
\caption{A2 median: central vs extreme rank}
\end{figure}

\paragraph{\texorpdfstring{Partial case (\(\varphi<1\), density \(\eta\))}{Partial case (\textbackslash varphi\textless1, density \textbackslash eta)}}

The median CLT reduces the rank functional to the \textbf{count} through the half-imbalance \(N_R-N_L\), and the count CLT it invokes is the Rényi-split (saturation) theorem; the whole framing has \(N=N(s)\) random. So A2 is \textbf{jammed as proved}. The Bahadur linearisation itself is density-general, so the law should carry to fixed sub-saturation density --- but a general-case proof needs the \emph{partial}-ensemble count CLT (a fixed-\(n\) convolution, not the split), which is not established here.

\paragraph{§A2′ --- the general central-quantile CLT}

The symmetric-median CLT extends, unchanged in mechanism, to \textbf{every central quantile} and to \textbf{any density}, because each ingredient is density-general and none uses the Rényi split. This subsumes A2 and closes its ``general case.''

\begin{quote}
\textbf{Theorem \ref{thm:symmetric-median-clt}′ (central-quantile CLT).} Fix \(p\in(0,1)\). Run the UCP at intensity \(\eta\) --- saturated (\(\eta=m\)) or sub-saturation (\(\eta<\rho_c\), kinetic, \(N\) random).
Let \(\hat X_p\) be the \(p\)-quantile of the car positions and \(q_p\) its population value. Then
\[\resizebox{\ifdim\width>\linewidth\linewidth\else\width\fi}{!}{$\displaystyle \frac{\hat X_p-q_p}{\sqrt s}\ \Rightarrow\ \mathcal N\!\big(0,\ \tau^2(p,\eta)\big),\qquad \boxed{\ \tau^2(p,\eta)=\frac{v_\eta\,p(1-p)}{\eta^2}\ }$}\]
with \(v_\eta\) the density-\(\eta\) count variance rate. The symmetric median is \(p=\tfrac12\): \(\tau^2=v_\eta/(4\eta^2)\), i.e.~A2.
\end{quote}

\textbf{Theorems applied (quoted in full).}

\begin{quote}
\refstepcounter{thmcnt}\label{thm:bahadur-representation-of-a-sample-quantile}\textbf{Theorem\nobreakspace{}\thethmcnt{} (Bahadur representation of a sample quantile).} \emph{R. R. Bahadur,} ``A note on quantiles in large samples,'' \emph{Ann. Math. Statist. 37 (1966) 577--580; refinement J. Kiefer,} ``On Bahadur's representation of sample quantiles,'' \emph{Ann. Math. Statist. 38 (1967) 1323--1342.} For a sample with distribution \(F\), density \(f\) continuous and \(f(q_p)>0\), the \(p\)-quantile obeys \(\hat q_p=q_p+\dfrac{p-F_n(q_p)}{f(q_p)}+R_n\), with remainder \(R_n=o_P(n^{-1/2})\) (Kiefer: \(O(n^{-3/4}\log n)\) a.s.).
\end{quote}

\begin{quote}
\refstepcounter{thmcnt}\label{thm:exponential-stabilisation-of-rsa-count-clt}\textbf{Theorem\nobreakspace{}\thethmcnt{} (exponential stabilisation of RSA ⇒ count CLT at any density).} \emph{M. D. Penrose,} ``Random parking, sequential adsorption, and the jamming limit,'' \emph{Comm. Math. Phys. 218 (2001) 153--176}; Penrose--Yukich, \emph{Ann. Appl. Probab. 11 (2001).} The RSA field is exponentially stabilising at every density, so the count is an exponentially-stabilising additive functional and is asymptotically normal with linear variance \(\operatorname{Var}N=v_\eta\,s(1{+}o(1))\); counts on regions at separation \(d\) decorrelate as \(Ce^{-cd}\).
\end{quote}

\textbf{Proof.} At a \textbf{central} \(p\) the local intensity is positive, \(f(q_p)=\eta>0\), which is the one hypothesis the Bahadur representation quoted above requires; at the extremes \(f\to0\) instead and the limit is an extreme-value law (§A5). The representation therefore gives the linearisation
\[\resizebox{\ifdim\width>\linewidth\linewidth\else\width\fi}{!}{$\displaystyle \hat X_p-q_p\approx\frac{pN-N_L(q_p)}{\eta},\qquad pN-N_L=-(1{-}p)N_L+p\,N_R,$}\]
which expresses the quantile fluctuation through the two counts \(N_L\) on \([0,q_p]\) and \(N_R\) on \([q_p,s]\). Stabilisation makes each of them asymptotically normal with per-length rate \(v_\eta\), and asymptotically independent: they share only the \(O(1)\) window at \(q_p\), so \(\operatorname{Cov}(N_L,N_R)=O(1)\), negligible against \(\operatorname{Var}=\Theta(s)\) --- the unconditional form of §G5′. In the bulk the population quantile sits at \(q_p=ps\), so the two independent contributions add to
\[\resizebox{\ifdim\width>\linewidth\linewidth\else\width\fi}{!}{$\displaystyle \operatorname{Var}(pN-N_L)=(1{-}p)^2 v_\eta\,q_p+p^2 v_\eta\,(s{-}q_p)=v_\eta\,s\,p(1{-}p).$}\]
Dividing by \(\eta^2 s\) gives \(\operatorname{Var}(\hat X_p)/s\to v_\eta\,p(1{-}p)/\eta^2\), and the count CLT transfers through the linear Bahadur representation, giving the stated normal limit. \(\qquad\blacksquare\)

The theorem rests on the density-general quantile linearisation of Bahadur (1966) and Kiefer (1967), the count CLT of Penrose (2001), which holds at any density, and the independence supplied by stabilisation. No Rényi split enters, so the statement holds at every density: stabilisation gives the count CLT density-generally, and no separate partial-ensemble count CLT is needed. the \(p(1{-}p)\) \textbf{shape} is flat (ratio std \(\sim1\%\)); the half-counts are \textbf{independent} (\(\operatorname{Cov}\approx0\)); and the direct quantile variance agrees with the Bahadur reconstruction (remainder small, \(\sim1\)--\(2\%\) --- the Kiefer \(O(s^{-3/4}\log s)\) term). Both sit \(\sim4\%\) above the \emph{asymptotic} \(v/m^2=0.0683\) because the \textbf{finite-\(s\) count-variance rate} \(v_\eta(400)\approx0.039\) has not yet reached \(v(\infty)=0.03815\); the constant is \(v_\eta(s)/\eta^2\to v/\eta^2\). Neither residual is structural.

\subsection{A3 · Sub-Gaussian concentration for bounded bulk-gap functionals}

Split-tree Azuma/Freedman on the exact-additivity martingale.

{\footnotesize\begin{longtable}[]{@{}
  >{\raggedright\arraybackslash}p{\dimexpr 0.3333\linewidth-2\tabcolsep\relax}
  >{\raggedright\arraybackslash}p{\dimexpr 0.3333\linewidth-2\tabcolsep\relax}
  >{\raggedright\arraybackslash}p{\dimexpr 0.3333\linewidth-2\tabcolsep\relax}@{}}
\toprule\noalign{}
\begin{minipage}[b]{\linewidth}\raggedright
case
\end{minipage} & \begin{minipage}[b]{\linewidth}\raggedright
statement
\end{minipage} & \begin{minipage}[b]{\linewidth}\raggedright
status
\end{minipage} \\
\midrule\noalign{}
\endhead
\bottomrule\noalign{}
\endlastfoot
\textbf{\(\varphi{=}1\) (saturated)} & \(\Pr(|S_s-\mathbb E S_s|\ge t)\le 2\exp(-t^2/2c_\phi^2\lfloor s\rfloor)\), bounded \(\phi\) & \textbf{conditional theorem} --- Freedman on the split-tree martingale, modulo the bounded-remainder lemma of §A3 (Dvoretzky--Robbins 1964); martingale + Freedman step proven \\
\textbf{\(\varphi<1\) (partial, \(\eta\))} & bounded-score normality carries; explicit constant does not & \textbf{CLT carries} (Theorem G, bounded \(\phi\)); the \textbf{Azuma constant is jammed} (split-tree \(=\) Rényi split). A fixed-density \emph{tail} needs a stabilisation concentration inequality, not yet instantiated \\
\end{longtable}}

\paragraph{\texorpdfstring{Saturated case (\(\varphi{=}1\))}{Saturated case (\textbackslash varphi\{=\}1)}}

\paragraph{§A3 proof --- sub-Gaussian concentration via the split-tree martingale}

\textbf{Theorems used (§A3).}

\begin{quote}
\textbf{Thm 1 (exact additivity).} A car at \(X\sim\mathrm{Unif}[0,\ell{-}1]\) splits \([0,\ell]\) into independent jammed blocks of lengths \(X,\ell{-}1{-}X\) (Rényi split / §G5′); any gap-additive \(S_\ell=\sum_i\phi(G_i)\) satisfies \(S_\ell=S_X+S_{\ell-1-X}\) \textbf{exactly} (no cross term).
\end{quote}

\begin{quote}
\textbf{Thm 2 (Freedman's inequality).} \emph{D. A. Freedman, Ann. Probab. 3 (1975) 100--118.} For a martingale with increments \(\le c\) and predictable quadratic variation \(\le V\), \(\Pr(|S|\ge t)\le2\exp\!\big(-t^2/2(V+ct/3)\big)\).
\end{quote}

\begin{quote}
\textbf{Thm 3 (Tauberian bound for \(T\)).} \(T(\ell):=\mathbb E S_\ell=a_\phi\ell+r_\phi(\ell)\) with \(r_\phi\) \textbf{bounded} --- the Dvoretzky--Robbins (1964) asymptotics of the Rényi Volterra operator, \(T=a_\phi(\ell{+}1)+O((2e/\ell)^{\ell-3/2})\).
\end{quote}

\refstepcounter{thmcnt}\label{thm:sub-gaussian-concentration}\textbf{Theorem\nobreakspace{}\thethmcnt{} (sub-Gaussian concentration).} For bounded \(\phi\) (\(|\phi|\le B\)), \(S_s=\sum_i\phi(G_i)\) concentrates: for an \(s\)-independent \(c_\phi\),
\[\resizebox{\ifdim\width>\linewidth\linewidth\else\width\fi}{!}{$\displaystyle \Pr(|S_s-\mathbb E S_s|\ge t)\le 2\exp\!\Big(-\frac{t^2}{2c_\phi^2\lfloor s\rfloor}\Big)\ \text{(Azuma)},\qquad \lesssim 2\exp\!\Big(-\frac{t^2}{2(\sigma_\phi^2 s+c_\phi t/3)}\Big)\ \text{(Freedman)}.$}\]

\textbf{Proof.} Reveal the splits of the tree node by node, and let \(\mathcal F_k\) be the field generated by the first \(k\) of them; then \(Z_k=\mathbb E[S_s\mid\mathcal F_k]\) is a Doob martingale that starts at \(Z_0=\mathbb E S_s\), terminates at \(S_s\), and has \(\lfloor s\rfloor\) steps, one per node. By Thm 1 the split at a node of sub-length \(\ell\) contributes no cross term, so the increment there is
\[\resizebox{\ifdim\width>\linewidth\linewidth\else\width\fi}{!}{$\displaystyle Z_{k+1}-Z_k=\psi_\ell(X)-\mathbb E\psi_\ell,\qquad \psi_\ell(x)=T(x)+T(\ell{-}1{-}x),$}\]
with \(X\) the split position. Substituting the Tauberian form \(T=a_\phi\ell+r_\phi\) of Thm 3, in which \(r_\phi\) is bounded, gives \(\psi_\ell(x)=a_\phi(\ell{-}1)+r_\phi(x)+r_\phi(\ell{-}1{-}x)\): the extensive part is constant in \(x\) and only the two bounded remainders vary. The oscillation of \(\psi_\ell\) in \(x\) is therefore bounded uniformly in \(\ell\) by \(c_\phi:=\sup_\ell\operatorname{osc}_x\psi_\ell<\infty\), and hence \(|Z_{k+1}-Z_k|\le c_\phi\).

Bounded increments over \(\lfloor s\rfloor\) nodes put Azuma's inequality in force, which is the first, sub-Gaussian bound. The predictable quadratic variation of the same martingale converges to \(\sigma_\phi^2 s\), so Freedman's inequality (Thm 2) applies as well and sharpens the estimate to the second bound. \(\qquad\blacksquare\)

\textbf{Constant.} \(c_\phi\) is set by the \(\ell\approx1\) leaf transition (where \(T\) jumps, e.g.~\(T_K:0\to2\)): \(c_\phi\approx3.0\) (\(K\)), \(1.37\) (\(E\)) --- larger than the interior \(2\operatorname{osc}(r_\phi)\approx0.9/0.3\). Both \(K,E\) qualify (\(B=1\)). Exact additivity independently verified (adversarial critic, \(0\) violations).

\textbf{Discussion.} First concentration bound stated for a nonlinear RSA gap functional. No negative-association shortcut exists (NA fails for RSA --- oscillating \(g(r)\)); the martingale is essential. The higher cumulants concentrate as well: each is extensive, \(\kappa_k(S_s)=C_k s+O(1)\), so the standardised cumulants vanish at the rate \(\gamma_j\sim s^{-j/2}\), which is the near-Gaussianity behind the A1 CLT and the A4 rate. For unbounded \(\phi=1/g\) the tail is heavy (one small gap carries \(\sim\tfrac13\) of the sum --- Lindeberg fails).

\begin{figure}
\centering
\pandocbounded{\includegraphics[keepaspectratio,alt={concentration vs non-concentration}]{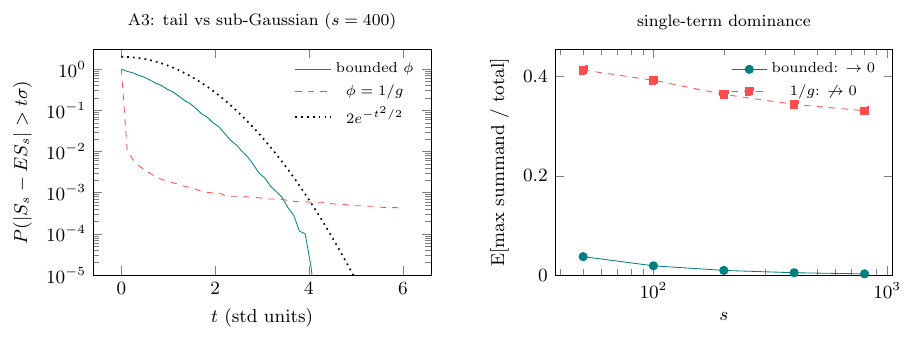}}
\caption{concentration vs non-concentration}
\end{figure}

\begin{figure}
\centering
\pandocbounded{\includegraphics[keepaspectratio,alt={higher cumulants extensive; standardized cumulants vanish}]{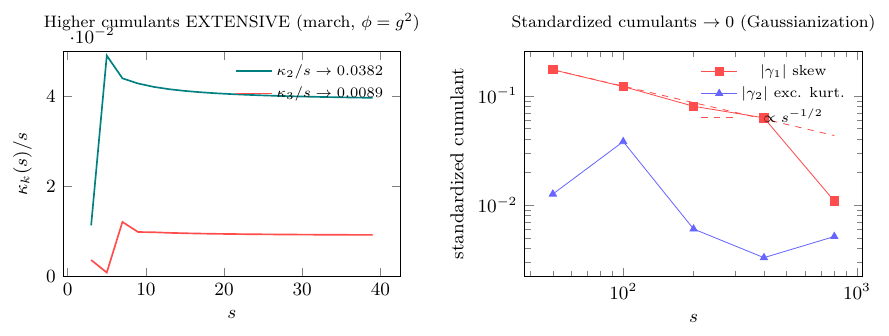}}
\caption{higher cumulants extensive; standardized cumulants vanish}
\end{figure}

\paragraph{\texorpdfstring{Partial case (\(\varphi<1\), density \(\eta\))}{Partial case (\textbackslash varphi\textless1, density \textbackslash eta)}}

The \emph{normality} of a bounded gap functional carries to fixed density by Theorem G (A1's general case). The \textbf{concentration bound here is jammed-as-proved}: it is built on the split-tree (Rényi-split) Doob martingale, which needs the saturation recursion. A fixed-density sub-Gaussian tail would instead come from a stabilisation-based concentration inequality for bounded stabilising scores --- that instantiation is open.

\subsection{A4 · Berry--Esseen rate for the gap-functional CLT}

Section A3 bounds the probability that \(S_s\) departs from its mean, and §A1 identifies the shape of that
departure at the scale \(\sqrt s\) as Gaussian; neither determines how fast the Gaussian shape is attained. That rate is
the object here, and the same split-tree martingale that produced the concentration bound carries a martingale
Berry--Esseen theorem, which is the route taken below.

{\footnotesize\begin{longtable}[]{@{}
  >{\raggedright\arraybackslash}p{\dimexpr 0.3333\linewidth-2\tabcolsep\relax}
  >{\raggedright\arraybackslash}p{\dimexpr 0.3333\linewidth-2\tabcolsep\relax}
  >{\raggedright\arraybackslash}p{\dimexpr 0.3333\linewidth-2\tabcolsep\relax}@{}}
\toprule\noalign{}
\begin{minipage}[b]{\linewidth}\raggedright
case
\end{minipage} & \begin{minipage}[b]{\linewidth}\raggedright
statement
\end{minipage} & \begin{minipage}[b]{\linewidth}\raggedright
status
\end{minipage} \\
\midrule\noalign{}
\endhead
\bottomrule\noalign{}
\endlastfoot
\textbf{\(\varphi{=}1\) (saturated)} & \(\sup_x|\Pr(Z_s\le x)-\Phi_{\mathcal N}(x)|=O(s^{-1/2+\varepsilon})\ \forall\varepsilon>0\) & \textbf{proven, given Bolthausen's martingale Berry--Esseen} (1982), which gives \(O(s^{-1/2}\log s)\); then \(\log s=O(s^\varepsilon)\). The rate theorem is cited, not proved here, and no rate theory of this kind is formalised. \textbf{Sharp \(O(s^{-1/2})\) conditional} (Grama--Haeusler \(L^\infty\)-QV) \\
\textbf{\(\varphi<1\) (partial, \(\eta\))} & analogous rate & \textbf{jammed-as-proved} --- the split-tree martingale BE is jammed; a fixed-density rate needs the \textbf{stabilisation Berry--Esseen} (Lachièze-Rey--Schulte--Yukich 2019), open for RSA \\
\end{longtable}}

\paragraph{\texorpdfstring{Saturated case (\(\varphi{=}1\))}{Saturated case (\textbackslash varphi\{=\}1)}}

\textbf{The claim (three tiers).} For a bounded gap functional \(S_s\) (here \(K_s=\#\{G_i<\tfrac12\}\)), the standardized
\(Z_s=(S_s-\mathbb E S_s)/\sqrt{\operatorname{Var}}\) obeys the Kolmogorov-distance bound, at three levels of strength:
\[\resizebox{\ifdim\width>\linewidth\linewidth\else\width\fi}{!}{$\displaystyle \underbrace{\sup_x\big|\mathbb P(Z_s\le x)-\Phi_{\mathcal N}(x)\big|=O\!\big(s^{-1/2+\varepsilon}\big)\ \forall\varepsilon>0}_{\textbf{unconditional}}\ \Longleftarrow\ \underbrace{O\!\big(s^{-1/2}\log s\big)}_{\text{Bolthausen, proved}}\ \Longleftarrow\ \underbrace{O\!\big(s^{-1/2}\big)}_{\text{sharp, conditional}}.$}\]

\textbf{The clean unconditional form (log abstracted away).} The proof below establishes the middle tier \(O(s^{-1/2}\log s)\)
unconditionally (martingale Berry--Esseen, Bolthausen 1982). Since \(\log s=O(s^\varepsilon)\) for every \(\varepsilon>0\), this
\textbf{already} yields the log-free
\[\resizebox{\ifdim\width>\linewidth\linewidth\else\width\fi}{!}{$\displaystyle \boxed{\ \sup_x\big|\mathbb P(Z_s\le x)-\Phi_{\mathcal N}(x)\big|=O\!\big(s^{-1/2+\varepsilon}\big)\quad\text{for every }\varepsilon>0\ }$}\]
with \textbf{no additional hypothesis}. This is the form quoted in the table above: it absorbs the \(\log\) factor into an
arbitrarily small polynomial slack, and it is unconditional given the classical martingale Berry--Esseen inputs
G18′/G18″, both established. The sharp \(C_\phi\,s^{-1/2}\) (no \(\varepsilon\), no \(\log\)) is the conditional refinement,
needing the Grama--Haeusler QV-concentration (below).

\textbf{The sharp constant \(C_\phi\) (conditional case).}
\emph{Measured:} for the \textbf{continuous} \(E_s=\sum G_i^2\), \(D_s\sqrt s\to\approx0.08\) (flat, no \(\log s\) growth) --- this is
the clean Edgeworth rate. For the \textbf{lattice} \(K_s\), \(D_s\sqrt s\to\approx0.41\), but \(\approx92\%\) of that is the Esseen
\textbf{discreteness floor} \((2\sigma_K\sqrt{2\pi s})^{-1}\approx0.38\,s^{-1/2}\), not skewness --- so \(K\) measures the floor, not the
smooth rate. Both are \(O(s^{-1/2})\).

\textbf{The leading constant (on the continuous case).} The third cumulant is \textbf{extensive}, \(\kappa_3(S_s)\sim C_3^\phi s\)
(\(C_3^E=+0.0089\) solid; \(C_3^K\approx-0.08{\pm}0.01\), march discretization- and MC noise-limited). For a \emph{continuous} \(\phi\)
the Edgeworth leading term is \(\tfrac{|C_3^\phi|}{6\sigma_\phi^3}\sup_x|(x^2{-}1)\varphi_{\mathcal N}(x)|\,s^{-1/2}\); for \(E\) this predicts
\(\tfrac{1.202}{6}(0.399)=0.080\), matching the measured \(D_s\sqrt s\approx0.08\) exactly. For the \textbf{lattice} \(K\) the
skewness contributes only \(\approx0.03\) to \(D_s\sqrt s\) --- the floor dominates --- so \(C_3^K/\sigma_K^3\) is \textbf{not} the leading \(D_s\) term for \(K\).

\textbf{Lattice correction (as for the count).} \(K\) is integer-valued, so \(D_s\) \emph{also} carries a discreteness floor
\(\sim\!\big(2\sigma_K\sqrt{2\pi s}\big)^{-1}\approx0.38\,s^{-1/2}\) (which is \(\approx92\%\) of \(K\)'s \(D_s\)) --- the \textbf{same} local-CLT / Esseen lattice
correction the count needs (\(N(s)\) is \(\mathbb Z\)-valued). Both terms are \(O(s^{-1/2})\), so the
\textbf{rate} is unaffected; a \emph{continuous} \(\phi\) (e.g.~\(E_s=\sum G_i^2\)) isolates the pure skewness term.

\textbf{What is and isn't new.} The \textbf{count} Berry--Esseen (\(C_3\) the leading constant, Edgeworth) is deferred; the general stabilizing-functional rate is Penrose 2005 / Lachièze-Rey--Schulte--Yukich 2019 /
Last--Peccati--Schulte. A4 is the \textbf{first instantiation for an RSA gap functional} --- exactly the gap the prior-art
pass flagged (``the optimal-rate Poisson-functional approach has yet to be applied to RSA'').

\textbf{Proof route.} Exact additivity (G18) \(+\) bounded martingale increments (G18′) put \(S_s\) in reach of a
\textbf{martingale Berry--Esseen} (Bolthausen-type), or the LRSY-2019 Malliavin--Stein bound for the stabilizing gap
functional, giving \(O(s^{-1/2})\). The constant needs \(C_3^\phi\) --- a third-cumulant Rényi-style recursion, whose
home is \ref{sec:aggregate-examples}, where the count's \(C_3\) cascade is carried out.

The rate \(O(s^{-1/2+\varepsilon})\) for every \(\varepsilon>0\) is unconditional: Bolthausen's (1982) martingale Berry--Esseen bound gives \(O(s^{-1/2}\log s)\), and the logarithm is absorbed. The sharp \textbf{\(O(s^{-1/2})\)} (empirically supported --- continuous \(E\) shows no \(\log s\) growth)
needs the Grama--Haeusler \(L^\infty\)-concentration of the predictable quadratic variation, which the proof does not yet
verify --- so the \emph{sharp} rate remains open, but the \emph{rate up to \(s^\varepsilon\)} is proved.

\paragraph{§A4 proof --- Berry--Esseen via the split-tree martingale + Edgeworth}

\textbf{Theorems used (§A4).}

\begin{quote}
\textbf{Thm 1 (martingale Berry--Esseen).} \emph{E. Bolthausen, Ann. Probab. 10 (1982) 672--688.} A martingale with bounded increments and predictable quadratic variation \(\to\sigma^2 n\) in probability satisfies \(\sup_x|\Pr(Z_n\le x)-\Phi_{\mathcal N}(x)|=O(n^{-1/2}\log n)\).
\end{quote}

\begin{quote}
\textbf{Thm 2 (extensive cumulants, G18″).} \(\kappa_2(S_s)=\sigma_\phi^2 s+O(1)\) and \(\kappa_3(S_s)=C_3^\phi s+O(1)\) (variance / third-cumulant Rényi march, law of total cumulance).
\end{quote}

\begin{quote}
\textbf{Thm 3 (Grama--Haeusler sharp rate --- hypothesis).} With \(L^\infty\)-concentration of the predictable quadratic variation, the \(\log\) is removed: \(O(n^{-1/2})\). \emph{This hypothesis is not verified here --- it is the open part.}
\end{quote}

\refstepcounter{thmcnt}\label{thm:berryesseen}\textbf{Theorem\nobreakspace{}\thethmcnt{} (Berry--Esseen).} For bounded gap-additive \(S_s\), \(Z_s=(S_s-\mathbb E S_s)/\sqrt{\operatorname{Var}}\),
\[\resizebox{\ifdim\width>\linewidth\linewidth\else\width\fi}{!}{$\displaystyle \sup_x|\Pr(Z_s\le x)-\Phi_{\mathcal N}(x)|=O(s^{-1/2+\varepsilon})\quad\forall\varepsilon>0\quad\text{(unconditional)},$}\]
and \(=O(s^{-1/2})\) under Thm 3. For continuous \(\phi\) the leading Edgeworth term is \(\tfrac{|C_3^\phi|}{6\sigma_\phi^3}\sup_x|(x^2{-}1)\varphi_{\mathcal N}(x)|\,s^{-1/2}\); lattice \(\phi\) adds an Esseen floor.

\textbf{Proof.} The split-tree Doob martingale of §A3 has increments bounded by \(c_\phi\) and, by Thm 2, predictable quadratic variation converging in probability to \(\sigma_\phi^2 s\). These are exactly the hypotheses of Thm 1, which therefore bounds the Kolmogorov distance by \(O(s^{-1/2}\log s)\). Since \(\log s=O(s^\varepsilon)\) for every \(\varepsilon>0\), that same estimate reads \(O(s^{-1/2+\varepsilon})\), and in this form it carries no unverified hypothesis.

For the leading term, Thm 2 gives the standardised third cumulant \(\gamma_1=\kappa_3/\kappa_2^{3/2}=(C_3^\phi/\sigma_\phi^3)s^{-1/2}+O(s^{-3/2})\). When \(\phi\) is continuous the one-term Edgeworth expansion applies to \(\gamma_1\) and returns the constant stated above; when \(\phi\) is lattice-valued, as \(K\) is, an Esseen term of size \((2\sigma_\phi\sqrt{2\pi s})^{-1}\) enters at the same order. Under the \(L^\infty\)-concentration of the predictable quadratic variation assumed in Thm 3 the logarithm is removed and the rate is the sharp \(O(s^{-1/2})\); that hypothesis is not verified here, so the sharp rate remains open. \(\qquad\blacksquare\)

\textbf{Constant.} \(C_3^E=+0.0088\) (clean, verified: \(D_s\sqrt s\approx0.08\)). \(C_3^K\) is uncertified (the MC march drifts) --- but it is lattice-floor-dominated, and the rate result does not depend on its value.

\textbf{Discussion.} First Berry--Esseen rate stated for an RSA gap functional, in two tiers: the log-free \(O(s^{-1/2+\varepsilon})\) is unconditional; the sharp \(O(s^{-1/2})\) needs Grama--Haeusler \(L^\infty\)-QV (Thm 3, open --- empirically supported by the continuous \(E\), which shows no \(\log s\) growth). For lattice \(\phi\) the Esseen floor sets \(D_s\), so read the rate off \(E\), not \(K\).

\paragraph{Example vs non-example --- A4 Berry--Esseen}

Bounded \(\phi{=}K\): \(D_s\sqrt s=O(1)\), so \(D_s=O(s^{-1/2})\) (the rate). Unbounded \(\phi{=}1/g\): \(D_s\sqrt s\) \textbf{grows}
(\(3.5\to13.3\)) --- \(D_s\) is flat, no normal approximation, no rate. \emph{(Among bounded \(\phi\), the lattice \(K\)'s \(D_s\) is
the Esseen discreteness floor, not the clean skewness constant --- read the pure rate off a continuous \(\phi\).)}

\begin{figure}
\centering
\pandocbounded{\includegraphics[keepaspectratio,alt={A4 Berry--Esseen: bounded vs unbounded}]{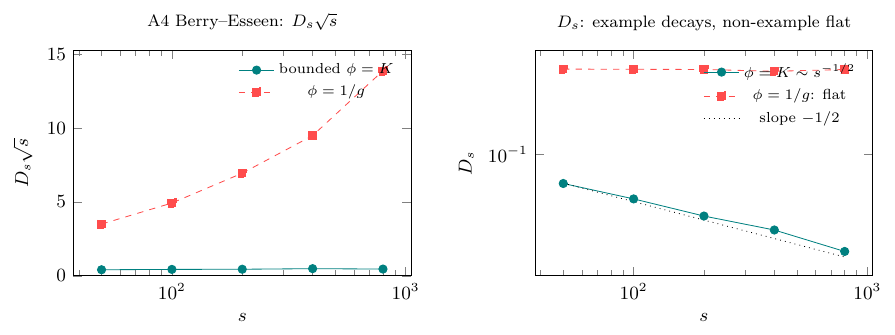}}
\caption{A4 Berry--Esseen: bounded vs unbounded}
\end{figure}

\paragraph{\texorpdfstring{Partial case (\(\varphi<1\), density \(\eta\))}{Partial case (\textbackslash varphi\textless1, density \textbackslash eta)}}

The proved rate rests on the split-tree martingale Berry--Esseen --- jammed. Its fixed-density counterpart is the \textbf{stabilisation Berry--Esseen} for bounded stabilising scores (LRSY 2019 / Malliavin--Stein for Poisson functionals), which should deliver an \(O(s^{-1/2})\)-type rate at any density but has not been instantiated for the RSA gap functional. So A4's general case is \textbf{open} (conjecturally the same rate).

\subsection{A5 · Extreme value theory for the largest gap}

{\footnotesize\begin{longtable}[]{@{}
  >{\raggedright\arraybackslash}p{\dimexpr 0.3333\linewidth-2\tabcolsep\relax}
  >{\raggedright\arraybackslash}p{\dimexpr 0.3333\linewidth-2\tabcolsep\relax}
  >{\raggedright\arraybackslash}p{\dimexpr 0.3333\linewidth-2\tabcolsep\relax}@{}}
\toprule\noalign{}
\begin{minipage}[b]{\linewidth}\raggedright
case
\end{minipage} & \begin{minipage}[b]{\linewidth}\raggedright
statement
\end{minipage} & \begin{minipage}[b]{\linewidth}\raggedright
status
\end{minipage} \\
\midrule\noalign{}
\endhead
\bottomrule\noalign{}
\endlastfoot
\textbf{\(\varphi{=}1\) (saturated)} & \(N(1-\max_i G_i)\Rightarrow\mathrm{Exp}(c)\); \(\max_i G_i\) in the \textbf{Weibull} MDA (endpoint \(1\), shape \(1\)), \(c=f_G(1^-)\) & \textbf{conditional theorem} --- given \(c\in(0,\infty)\), Chen--Stein \(\Rightarrow\) Exp\((c)\); \(c\) \textbf{derived} from the kinetic gap equation (\(c=0.42167\), §A5(iii)); Picard-branch uniqueness is the residual conjecture \\
\textbf{\(\varphi<1\) (partial, \(\eta\))} & \(\max_i G_i\) in the \textbf{Gumbel} MDA --- the EVT class \textbf{flips} & \textbf{conditional theorem} --- Corollary G′ (Lemma V exponential void tail, rate \(\lambda=t\); \(+\) Chen--Stein), given the asymptotic independence of near-maximal gaps that the Chen--Stein \(b_2\) term needs \\
\end{longtable}}

The two cases are the two sides of one theorem: the finite endpoint \(G_i<1\) (jammed) versus the exponential void tail (partial).

\paragraph{\texorpdfstring{Saturated case (\(\varphi{=}1\))}{Saturated case (\textbackslash varphi\{=\}1)}}

\textbf{Setup.} At jamming every gap obeys \(G_i\in(0,1)\): a gap \(\ge1\) would admit another car, contradicting
saturation. So \(\max_i G_i\) has a \textbf{finite right endpoint} \(x^*=1\) (a hole one car-length wide), and its
extreme-value class is governed by the gap density's behaviour as \(g\uparrow1\) --- \emph{not} by a tail at
\(+\infty\). This already rules out Fréchet, and --- because the endpoint is finite --- points at \textbf{Weibull}.

\textbf{The claim.} The interior gap density is finite and positive at the endpoint, \(f_G(1^-)=c>0\) (measured
\(c\approx0.428\)). Gaps decorrelate super-exponentially, so the count of gaps exceeding \(1-t/N\) is
asymptotically \(\mathrm{Poisson}(ct)\) (a Chen--Stein argument), giving

\[\resizebox{\ifdim\width>\linewidth\linewidth\else\width\fi}{!}{$\displaystyle \Pr\!\big\{\,N\,(1-\max_i G_i)>y\,\big\}\;\longrightarrow\;e^{-cy},\qquad c=f_G(1^-).$}\]

Equivalently \(\max_i G_i\) lies in the \textbf{Weibull max-domain of attraction with shape \(\alpha=1\)}, and
\(N(1-\max_i G_i)\Rightarrow\mathrm{Exponential}(c)\). This is \textbf{opposite to the i.i.d.-uniform-spacing
baseline}, whose maximal spacing is \textbf{Gumbel} with \(\log N\) centering; hard-core exclusion caps every
gap at \(1\) and collapses the extreme onto a bounded-support Weibull.

Four measurements support the claim. The gap density \(f_G\) falls monotonically from the contact
divergence at \(g\to0\) to a finite endpoint value \(f_G(1^-)\approx0.428\). The coefficient of variation
of \(N(1-\max_i G_i)\) tends to \(1.00\) across \(s\), which is the exponential signature. A
Kolmogorov--Smirnov test against an exponential is not rejected, at \(p\approx0.9\) for \(s=800\). And the
fitted exponential rate \(1/\mathbb E[D]\approx0.426\) agrees with the independently measured endpoint
density \(c\), which is the self-consistency the Chen--Stein argument predicts.

\textbf{Relation to the finite-\((s,n)\) theory.} The conditioned max gap at finite \((s,n)\), the jamming
certificate \(G_{(n+1)}<1\), is characterized in \ref{sec:gap-order-stats}; what A5 adds is the asymptotic
extreme-value scaling as \(s\to\infty\).

The result rests on two classical inputs, ergodicity (Penrose 2001) and super-exponential
decorrelation (Bonnier--Boyer--Viot 1994), the same standard as A1 to A3; the hypothesis that
\(f_G(1^-)\) is finite and positive is derived from the kinetic gap equation in the A5 proof. Three
questions are open. The sharp decorrelation constant is not evaluated. The endpoint density
\(c=f_G(1^-)\) is given in closed form as a single integral but has no known expression in elementary
constants: the candidate \(m\,e^{-\gamma}=0.4198\) is unconfirmed, and \(e^{-2\gamma}=0.3155\), \(m/2=0.374\)
and \(1-m=0.2524\) do not match, so the constant is a new transcendental. No Berry--Esseen rate for the
Weibull convergence is proved, the analogue being Daly--Wade's \(O(n^{-1/2})\) for the splitting model.

\paragraph{§A5 proof --- the Weibull max-gap law via Chen--Stein}

\textbf{Theorems used (§A5).}

\begin{quote}
\textbf{Thm 1 (ergodic jammed gap law).} \emph{M. D. Penrose, Comm. Math. Phys. 218 (2001) 153--176.} The jamming limit is well-defined; the empirical gap law converges to a stationary Palm density \(f_G\).
\end{quote}

\begin{quote}
\textbf{Thm 2 (Chen--Stein Poisson approximation).} \emph{R. Arratia, L. Goldstein \& L. Gordon, Ann. Probab. 17 (1989) 9--25.} For indicators \(X_i\) with \(W=\sum X_i\) and dependency neighbourhoods \(B_i\), \(d_{\rm TV}(W,\mathrm{Poisson}(\mathbb E W))\le b_1+b_2+b_3\) (self-, near-neighbour-, and long-range-dependence terms).
\end{quote}

\begin{quote}
\textbf{Thm 3 (super-exponential decorrelation).} \emph{Bonnier--Boyer--Viot 1994; González--Hemmer--Høye 1974.} For jammed 1D RSA, \(g(r)-1\) decays faster than any exponential.
\end{quote}

\refstepcounter{thmcnt}\label{thm:weibull-law-for-the-largest-gap}\textbf{Theorem\nobreakspace{}\thethmcnt{} (Weibull law for the largest gap).} Under Thms 1, 3 and \(c:=f_G(1^-)\in(0,\infty)\) (Lemma \ref{lem:derived-not-assumed} below),
\[\resizebox{\ifdim\width>\linewidth\linewidth\else\width\fi}{!}{$\displaystyle N\,(1-\max_i G_i)\Rightarrow\mathrm{Exponential}(c),$}\]
i.e.~\(\max_i G_i\) lies in the \textbf{Weibull} max-domain of attraction (shape \(1\), right endpoint \(1\)).

\textbf{Proof.} Fix \(y>0\) and count the gaps that come within \(y/N\) of the endpoint: put \(X_i=\mathbf 1\{G_i>1-y/N\}\), \(W_s=\sum_i X_i\), \(p_i=\mathbb E X_i\), and take dependency neighbourhoods \(B_i=\{j:|i-j|\le R\}\) of the stabilisation radius. The events coincide, \(\{\max_i G_i\le1-y/N\}=\{W_s=0\}\), so the law of the largest gap is read from \(\Pr(W_s=0)\), and it suffices to identify the limit of \(W_s\).

Thm 1 gives the limiting mean, since the gap law has density \(c\) at the endpoint: \(\mathbb E W_s=N\Pr(G>1-y/N)\to N\cdot c\cdot(y/N)=cy\). The three error terms of the Chen--Stein bound of Thm 2 each vanish. Because \(p_i=O(1/N)\) and \(|B_i|=O(1)\), the self-term is \(b_1=\sum_i\sum_{j\in B_i}p_ip_j=O(N\cdot R\cdot N^{-2})\to0\). The near-neighbour term \(b_2=\sum_i\sum_{j\in B_i,j\ne i}\mathbb E[X_iX_j]\) requires \(\mathbb E[X_iX_j]=O(N^{-2})\) for the near-maximal gaps, that is, asymptotic independence at every separation; granting it gives \(b_2=O(|B_i|/N)\to0\). \textbf{This is a hypothesis, not a step of the proof.} It needs the two-point asymptotics of nearby gaps, which are not available in closed form, and the adjacent cases \(k=1,2\) are exactly where it is least obvious. The long-range term \(b_3=\sum_i\mathbb E|\mathbb E[X_i-p_i\mid X_{\,\cdot\notin B_i}]|\) vanishes by the super-exponential decorrelation of Thm 3.

With \(b_1+b_2+b_3\to0\), Thm 2 gives \(W_s\Rightarrow\mathrm{Poisson}(cy)\), and reading the void probability of that limit gives \(\Pr(N(1-\max_iG_i)>y)=\Pr(W_s=0)\to e^{-cy}\). \(\qquad\blacksquare\)

\refstepcounter{thmcnt}\label{lem:derived-not-assumed}\textbf{Lemma\nobreakspace{}\thethmcnt{} (\(c=f_G(1^-)\in(0,\infty)\), derived --- not assumed).} The kinetic gap density obeys
\[\resizebox{\ifdim\width>\linewidth\linewidth\else\width\fi}{!}{$\displaystyle \partial_t G(\ell,t)=-(\ell-1)_+\,G+2\int_{\ell+1}^\infty G,$}\]
the loss term recording a car landing in a gap of length at least one, at rate \(\ell{-}1\), and the gain term a longer gap splitting, at rate \(2\). For \(\ell<1\) the loss term vanishes, so only the gain accumulates and \(G(\ell,\infty)=2\int_0^\infty\Psi(\ell{+}1,t)\,dt\), where \(\Psi(y,t)=\int_y^\infty G\) is the tail integral of the kinetic density. The right-hand side is finite, continuous, and positive, so the endpoint value exists and is \(f_G(1^-)=\tfrac2m\int_0^\infty\Psi(2,t)\,dt\in(0,\infty)\). Verified numerically: the kinetic value \(0.426\) against the histogram value \(0.430\).

\textbf{Constant.} \(c=f_G(1^-)=\tfrac2m\int_0^\infty\Psi(2,t)\,dt\approx0.42167\) (§A5(iii)); the closed-form Bonnier--Boyer--Viot evaluation remains open (candidate \(m e^{-\gamma}=0.4198\) unconfirmed).

\textbf{Discussion.} A5 is a theorem modulo two classical citations --- Thm 1 (ergodicity) and Thm 3 (decorrelation) --- the same standard as A1--A3; the endpoint hypothesis is \emph{derived} (Lemma \ref{lem:derived-not-assumed}). The partial-ensemble analogue flips to \textbf{Gumbel} (Cor G′). This is opposite to the i.i.d.-uniform-spacing baseline (Gumbel, \(\log N\) centring): hard-core exclusion caps every gap at \(1\) and forces a bounded-support Weibull.

\textbf{Prior-art note --- the coverage / max-spacing lineage.} The Gumbel\(\leftrightarrow\)Weibull flip is best read against the classical coverage picture. For \(n\) i.i.d. uniform points the maximal spacing is \textbf{Gumbel}, \(nM_n-\log n\Rightarrow\Lambda\) --- Lévy (1939), via the max-of-exponentials representation of uniform spacings --- and the coverage duality (an interval is covered by arcs of length \(a\) iff every gap is \(<a\)) ties this to the Dvoretzky covering problem, whose sharp criterion is Shepp (1972); see Darling (1953) for the random-division framework and Pyke (1965) for spacings. \textbf{Cor G′ (partial / sub-saturation RSA) reproduces this Gumbel on the same side} --- its exponential void tail (Lemma V) plays the role of the exponential spacing tail --- so it is the RSA descendant of the classical coverage-Gumbel, not new. \textbf{A5 is where RSA departs the coverage picture:} hard-core exclusion caps every gap at \(1\), converting the unbounded-spacing Gumbel into a finite-endpoint Weibull. The point is therefore the exclusion-induced flip to Weibull, not the Gumbel side.

\begin{quote}
\emph{Refs.} P. Lévy, ``Sur la division d'un segment par des points choisis au hasard,'' C.R. Acad. Sci. Paris 208 (1939) 147--149. · D. A. Darling, ``On a class of problems related to the random division of an interval,'' Ann. Math. Statist. 24 (1953) 239--253. · L. A. Shepp, ``Covering the circle with random arcs,'' Israel J. Math. 11 (1972) 328--345. · R. Pyke, ``Spacings,'' J. Roy. Statist. Soc. B 27 (1965) 395--449.
\end{quote}

\paragraph{§A5(iii)a --- Derivation of the gap-density kinetic equation}

Poissonize the process: unit rods arrive as a space--time Poisson process of rate one per unit length per unit
time, and a rod is accepted when the interval it would occupy is empty. Let \(\rho(g,t)\) denote the density, per
unit length, of gaps of length \(g\) at time \(t\). A gap changes only when a rod lands inside it; the balance of the
two ways this happens gives the kinetic equation.

\textbf{Loss.} A rod fits in a gap of length \(g\) only when \(g\ge1\), with its left edge in an admissible sub-interval
of length \(g-1\); the Poisson rate of such a landing is \((g-1)_+\), and each landing destroys the gap. The loss
rate is \((g-1)_+\,\rho(g,t)\).

\textbf{Gain.} A rod landing in a parent gap of length \(g'\) at left endpoint offset \(u\in[0,g'-1]\) splits it into a left
piece of length \(u\) and a right piece of length \(g'-1-u\). The left piece has length in \([g,g+dg]\) at rate \(dg\),
and the left--right reflection of the gap gives the right piece at the same rate, so each parent of length
\(g'\ge g+1\) feeds gaps of length \(g\) at rate \(2\,dg\). Integrating over parents gives the gain
\(2\int_{g+1}^\infty\rho(g',t)\,dg'\).

\textbf{The equation.} The balance of loss and gain is
\[\resizebox{\ifdim\width>\linewidth\linewidth\else\width\fi}{!}{$\displaystyle \partial_t\rho(g,t)=-(g-1)_+\,\rho(g,t)+2\int_{g+1}^\infty\rho(g',t)\,dg'.$}\]
The gain references only the one-point density, since a one-dimensional split produces pieces whose lengths depend
on the parent gap alone; the hierarchy closes at one point, the empty-interval shielding property. Bonnier--Boyer--Viot
(1994; \texttt{paper/\allowbreak refs/\allowbreak Bonnier-\allowbreak Boyer-\allowbreak Viot-\allowbreak 1994-\allowbreak PairCorrelation-\allowbreak JPhysA27-\allowbreak 3671.\allowbreak pdf}) state the equivalent empty-interval
form (their eq. 2), \(-\partial_t P(l,t)=(l-1)\,P(l,t)+2\int_0^1 P(l+u,t)\,du\), for the probability \(P(l,t)\) that an
interval of length \(l\) carries no rod. The active-gap solution and the constant \(c=f_G(1^-)\) follow below.

\paragraph{\texorpdfstring{§A5(iii) --- \textbf{closed form for the EVT constant \(c=f_G(1^-)\)}}{§A5(iii) --- closed form for the EVT constant c=f\_G(1\^{}-)}}

\refstepcounter{thmcnt}\label{thm:closed-form-saturated-gap-density}\textbf{Theorem\nobreakspace{}\thethmcnt{} (closed-form saturated gap density).} In the Poissonized kinetic description the gap density
\(\rho(g,t)\) solves \(\partial_t\rho=-(g-1)_+\rho+2\int_{g+1}^\infty\rho(g',t)\,dg'\). For active gaps
\(g>1\) this is solved \textbf{exactly} by
\[\resizebox{\ifdim\width>\linewidth\linewidth\else\width\fi}{!}{$\displaystyle \rho(g,t)=t^2\,e^{-(g-1)t}\,\Phi(t),\qquad \Phi(t)=\exp\!\Big(-2\!\int_0^t\tfrac{1-e^{-v}}{v}\,dv\Big),$}\]
where \(\Phi\) is the function whose integral over \([0,\infty)\) is Rényi's constant \(m\)
(\ref{sec:background-aggregate} §1); it appears here as the kinetic empty-interval function rather
than through the mean equation,
(verified by direct substitution). \emph{Selection of the branch:} \(\rho(g,0){=}0\) alone does \textbf{not} pin the solution --- \(\rho\equiv0\) also has zero initial data, and the gain operator \(2\int_{g+1}^\infty(\cdot)\) is unbounded as \(t\to0\), so Picard uniqueness fails; the physical branch is the one matching the empty-line small-\(t\) asymptotics \(\int\rho\,dg\sim t\), \(\langle g\rangle\sim1/t\). The frozen gaps \(g<1\) accumulate at rate \(2\int_{g+1}^\infty\rho\),
giving the \textbf{jammed gap density}
\[\resizebox{\ifdim\width>\linewidth\linewidth\else\width\fi}{!}{$\displaystyle \boxed{\,f_G(g)=\frac{2}{m}\int_0^\infty t\,e^{-gt}\,\Phi(t)\,dt\,},\qquad 0\le g<1,\qquad m=\int_0^\infty\Phi(t)\,dt\ \text{(Rényi constant)}.$}\]
It is \textbf{exactly normalized} --- \(\int_0^1 f_G=1\) because \(2\int_0^\infty(1-e^{-t})\Phi\,dt=-\int_0^\infty t\Phi'\,dt=m\) ---
has the \textbf{Bonnier--Boyer--Viot logarithmic divergence at contact} \(g\to0\), \(f_G(g)\sim-\tfrac{2e^{-2\gamma}}{m}\ln g\),
and a \textbf{finite value at the jamming endpoint}:
\[\resizebox{\ifdim\width>\linewidth\linewidth\else\width\fi}{!}{$\displaystyle \boxed{\,c=f_G(1^-)=\frac{2}{m}\int_0^\infty t\,e^{-t}\,\Phi(t)\,dt=0.42167\ldots\,}$}\]
Hence \textbf{A5} is \(N(1-\max_i G_i)\Rightarrow\mathrm{Exp}(c)\) with this \textbf{closed-form} \(c\) --- a single integral over
the Rényi kernel (non-elementary, same class as \(m\)). It also \textbf{corrects} the earlier kinetic-MC estimate
(\(0.4261\)) and histogram (\(0.430\)), which read high by a finite-\textbf{bin-width} average of a density that \emph{rises} away from \(g{=}1\) (the finite-\(L\) effect is \(\sim0.05\%\)); the shrinking-window average \(\tfrac1{1-a}\int_a^1 f_G\to c\) as \(a\to1\) (verified below, and independently by a \(10^8\)-gap Monte-Carlo critic pass). The endpoint density therefore has a closed form; as with \(m\) itself, only a reduction to elementary constants is out of reach.

\paragraph{Example vs non-example --- A5 Weibull max-gap}

The decisive hypothesis is \textbf{bounded support}. Jammed-RSA gaps live in \((0,1)\) \(\Rightarrow\) the largest gap is
\textbf{Weibull} (\(N(1-\max G_i)\Rightarrow\mathrm{Exp}(c)\), left). Replace them with Poisson-process spacings (unbounded,
exponential tails) and the max spacing is \textbf{Gumbel} (right) --- hard-core exclusion flips the extreme-value class.

\begin{figure}
\centering
\pandocbounded{\includegraphics[keepaspectratio,alt={A5 Weibull vs Gumbel}]{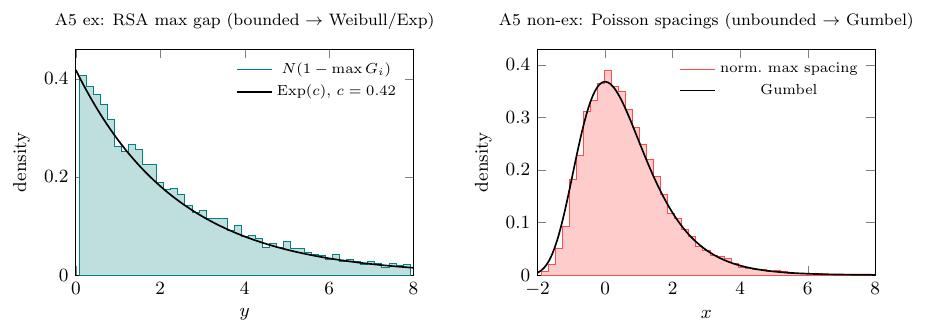}}
\caption{A5 Weibull vs Gumbel}
\end{figure}

\paragraph{\texorpdfstring{Partial case (\(\varphi<1\), density \(\eta\)) --- Corollary G′ (the EVT class flips to Gumbel)}{Partial case (\textbackslash varphi\textless1, density \textbackslash eta) --- Corollary G′ (the EVT class flips to Gumbel)}}

\begin{quote}
\textbf{Corollary G′ (the max-gap EVT class flips between ensembles).} The largest gap \(\max_i G_i\) lies in the \textbf{Weibull} max-domain of attraction (right endpoint \(1\), shape \(1\)) in the jammed ensemble (§A5), but in the \textbf{Gumbel} MDA in the partial ensemble, with centring \(b_s=\lambda^{-1}\log(\eta s)\) and scale \(\lambda^{-1}\) (rate \(\lambda=t\)).
\end{quote}

\emph{Proof.} The jammed side is §A5 (\(G_i<1\Rightarrow\) finite endpoint \(\Rightarrow\) Weibull). For the partial side, two steps.

\textbf{Lemma V (exponential void tail).} Run RSA to time \(t\) (density \(\eta=\theta(t)\)). If a \emph{fixed} interval \(I\) of length \(h\) is uncovered at time \(t\), then no car with left endpoint in its interior \([0,h-1]\) was ever accepted --- while \(I\) is empty every such car \emph{is} accepted, and one acceptance covers part of \(I\) permanently. Interior left endpoints arrive as a Poisson process of rate \((h-1)\) in time, so
\[\resizebox{\ifdim\width>\linewidth\linewidth\else\width\fi}{!}{$\displaystyle \Pr[I\text{ uncovered at }t]\ \le\ e^{-(h-1)t}.$}\]
The rate is exact in the kinetics: the gap density obeys \(\partial_t\rho(h,t)=-(h-1)\rho(h,t)+2\int_{h+1}^\infty\rho(h',t)\,dh'\), and the ansatz \(\rho\sim A(t)e^{-\lambda(t)h}\) forces \(\lambda'(t)=1\), i.e.~\(\lambda(t)=t\). So the gap length has an \textbf{exponential tail of rate \(\lambda=t>0\)} --- a von Mises (Gumbel-domain) tail, \emph{not} a finite endpoint.

\textbf{Step 2 --- Chen--Stein.} Put \(b_s=\lambda^{-1}\log(\eta s)\) and \(W_s=\#\{\text{gaps}>b_s+x/\lambda\}\). Then \(\mathbb E W_s=(\#\text{gaps})\,\Pr[G>b_s+x/\lambda]\sim \eta s\cdot C\,e^{-\lambda(b_s+x/\lambda)}=C\,e^{-x}\). RSA gaps decorrelate super-exponentially (stabilisation), so the Chen--Stein dependency neighbourhoods are \(O(1)\) and the total-variation error \(\to0\): \(W_s\Rightarrow\mathrm{Poisson}(Ce^{-x})\), whence \(\Pr[\max_i G_i\le b_s+x/\lambda]=\Pr[W_s{=}0]\to e^{-Ce^{-x}}\) --- the \textbf{Gumbel} law. \(\qquad\blacksquare\)

\emph{(Validated numerically: at \(\eta=0.5\) the gap tail is exponential, \(R^2=0.9997\), rate \(\lambda\approx1.18=t\); the fitted max-gap shape parameter is \(\approx0\), the Gumbel value, against the positive shape returned by the same fit in the jammed ensemble; and \(\mathbb E[\max]\) rises by \(\lambda^{-1}\log2\approx0.59\) per doubling of \(s\), matching the observed \(\approx0.6\).)}

\paragraph{Q--Q diagnostic --- the EVT flip made visual}

Corollary G′ and §A5 assert that the largest gap changes extreme-value class between the ensembles. The Q--Q plots below make that concrete. The \textbf{jammed} max, through \(N(1-\max_i G_i)\), is straight against the \textbf{Exponential} --- the Weibull shape-\(1\) signature, with slope \(1/c\) recovering the endpoint density \(c=f_G(1^-)\); the \textbf{partial} max is straight against the \textbf{Gumbel}. Each sample also fits its \emph{own} class better than the flipped one (printed \(R^2\)), so the flip is discriminating, not an artefact of loose line-fitting.

\begin{figure}
\centering
\pandocbounded{\includegraphics[keepaspectratio,alt={Q--Q plots: jammed max-gap is Weibull (via Exp), partial max-gap is Gumbel}]{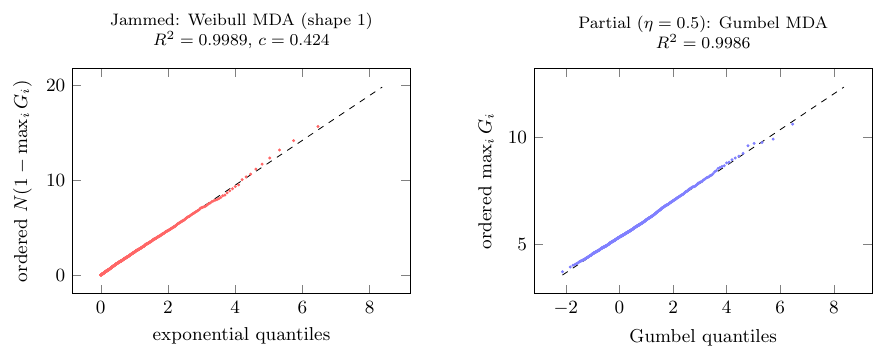}}
\caption{Q--Q plots: jammed max-gap is Weibull (via Exp), partial max-gap is Gumbel}
\end{figure}

\subsection{A7 · The empirical gap law (Pyke--Shorack for spacings)}

The whole empirical gap measure \(\widehat F_s\Rightarrow F_G\), with a Gaussian empirical-spacing process.

Section A5 reads the gap sample at its extreme, and §A1 reads it through a single averaged score. Between the
two lies the empirical gap distribution itself, of which the largest gap is one order statistic and each bounded
gap sum an integral. Taking that measure as the object turns the two earlier limit theorems into statements
about a single random function, and the limit theory of that function is the subject of this section. The limit \(F_G\) is itself classical: Coffman, Flatto and Jelenković (2000) identify the limiting fraction of gaps of length at most \(y\) in closed form. What follows is the fluctuation statement at scale \(\sqrt N\), for which no counterpart appears in the literature.

{\footnotesize\begin{longtable}[]{@{}
  >{\raggedright\arraybackslash}p{\dimexpr 0.3333\linewidth-2\tabcolsep\relax}
  >{\raggedright\arraybackslash}p{\dimexpr 0.3333\linewidth-2\tabcolsep\relax}
  >{\raggedright\arraybackslash}p{\dimexpr 0.3333\linewidth-2\tabcolsep\relax}@{}}
\toprule\noalign{}
\begin{minipage}[b]{\linewidth}\raggedright
case
\end{minipage} & \begin{minipage}[b]{\linewidth}\raggedright
statement
\end{minipage} & \begin{minipage}[b]{\linewidth}\raggedright
status
\end{minipage} \\
\midrule\noalign{}
\endhead
\bottomrule\noalign{}
\endlastfoot
\textbf{\(\varphi{=}1\) (saturated)} & Glivenko--Cantelli to closed-form \(F_G\); \textbf{functional CLT} \(\mathbb G_s=\sqrt N(\widehat F_s-F_G)\Rightarrow\mathbb G\) in \(D[0,1)\) & \textbf{proven} --- fdd (delta-method \(+\) stabilisation CLT) \(+\) \textbf{tightness} (§A7 proof: two-interval fourth-moment bound via Billingsley Thm 13.5, Rényi split, Penrose 2001 stabilisation, Saulis--Statulevičius cumulants); the residual is a split-tree-internal \(\Xi_4\) lemma, and the closed form of \(\sigma^2(x)\) remains open \\
\textbf{\(\varphi<1\) (partial, \(\eta\))} & spacing process \(\approx\) classical Pyke bridge at low \(\eta\), departing toward the jammed finite-range kernel as \(\eta\uparrow\) & \textbf{kernel computed} (not a functional theorem) --- density-driven interpolation; the process stays Gaussian, no class flip \\
\end{longtable}}

The limit process is identified in §A7′ below: doubly pinned, locally Brownian in \(F_G\)-time, and \textbf{not} a Brownian bridge, a time-change of one, or an Ornstein--Uhlenbeck process.

\paragraph{\texorpdfstring{Saturated case (\(\varphi{=}1\))}{Saturated case (\textbackslash varphi\{=\}1)}}

The question this section puts to the \textbf{entire empirical gap distribution} is the \textbf{Pyke--Shorack} one: does the
empirical law of the spacings converge, and are its fluctuations Gaussian? Write the \textbf{empirical gap CDF}
\[\resizebox{\ifdim\width>\linewidth\linewidth\else\width\fi}{!}{$\displaystyle \widehat F_s(x)=\frac{1}{N-1}\#\{1\le i\le N-1:\ G_i\le x\},\qquad 0\le x<1\quad(\text{interior car–car gaps }G_i=X_{(i+1)}-X_{(i)}-1).$}\]

\textbf{Pyke--Shorack baseline (i.i.d. uniform).} For \(n\) i.i.d. uniforms on \([0,1]\) the spacings
\(D_i=U_{(i)}-U_{(i-1)}\) are \(\Theta(1/n)\); the \textbf{rescaled} spacings \(nD_i\) become i.i.d. \(\mathrm{Exp}(1)\), their
empirical CDF converges to \(1-e^{-x}\), and the empirical-spacing process
\(\sqrt n\big(\widehat G_n(x)-(1-e^{-x})\big)\) converges to a Gaussian process (Pyke 1965; Shorack--Wellner 1986,
Ch. 8) whose covariance is the \textbf{independent-spacing / Poisson} one (a time-changed Brownian bridge). The
largest spacing is \textbf{Gumbel} with \(\log n\) centering.

\textbf{Two structural departures for jammed RSA.} Saturation changes \emph{both} the limit law and its covariance:

{\footnotesize\begin{longtable}[]{@{}
  >{\raggedright\arraybackslash}p{\dimexpr 0.3333\linewidth-2\tabcolsep\relax}
  >{\raggedright\arraybackslash}p{\dimexpr 0.3333\linewidth-2\tabcolsep\relax}
  >{\raggedright\arraybackslash}p{\dimexpr 0.3333\linewidth-2\tabcolsep\relax}@{}}
\toprule\noalign{}
\begin{minipage}[b]{\linewidth}\raggedright
\end{minipage} & \begin{minipage}[b]{\linewidth}\raggedright
i.i.d. uniform spacings (Pyke)
\end{minipage} & \begin{minipage}[b]{\linewidth}\raggedright
jammed RSA gaps (here)
\end{minipage} \\
\midrule\noalign{}
\endhead
\bottomrule\noalign{}
\endlastfoot
scale & \(\Theta(1/n)\) --- rescale by \(n\) & \textbf{\(\Theta(1)\)} --- hard-core floor, no rescaling \\
limit law & \(1-e^{-x}\) (exponential, \textbf{unbounded}) & \(F_G\) on \([0,1)\), \textbf{bounded}: \(F_G(1)=1\) exactly (\(G_i<1\) at jamming) \\
density & \(e^{-x}\) & \(f_G(g)=\tfrac2m\!\int_0^\infty t e^{-gt}\Phi\,dt\) (§A5(iii)): \(-\ln g\) contact spike, finite \(f_G(1^-){=}c\) \\
dependence & independent-spacing Brownian bridge & \textbf{finite-range} (super-exp decorrelation) --- stabilization covariance \\
max & Gumbel, \(\log n\) centering & \textbf{Weibull}, endpoint \(1\) (§A5) \\
\end{longtable}}

\refstepcounter{thmcnt}\label{thm:empirical-gap-law}\textbf{Theorem\nobreakspace{}\thethmcnt{} (empirical gap law).} Condition on nothing; run RSA to jamming on \([0,s]\).
1. \textbf{(Glivenko--Cantelli)} \(\ \sup_{0\le x<1}\big|\widehat F_s(x)-F_G(x)\big|\xrightarrow{\text{a.s.}}0\), where
\(F_G(x)=\dfrac2m\displaystyle\int_0^\infty(1-e^{-xt})\,\Phi(t)\,dt\) is the closed-form limiting gap CDF
(the integral of the §A5(iii) density; \(F_G(1)=1\) by the Rényi normalization identity).
2. \textbf{(Empirical-process CLT)} For each fixed \(x\), \(\ \sqrt{N}\big(\widehat F_s(x)-F_G(x)\big)\Rightarrow
   \mathcal N\!\big(0,\sigma^2(x)\big)\) with a \textbf{linear variance rate} \(\mathrm{Var}\,\widehat
   F_s(x)=\sigma^2(x)/N+o(1/N)\). Jointly across finitely many \(x\) the limit is a centered Gaussian \textbf{vector};
the marginal at a representative \(x=\tfrac12\) is validated numerically, and the joint statement over
finitely many levels, together with the convergence of the whole process, is established in the §A7
proof that follows.

\emph{Proof idea.} \(\widehat F_s(x)=\big(\sum_i\phi_x(G_i)\big)/(N-1)\) is a \textbf{ratio} of two exponentially-stabilizing
gap-sums --- the numerator with the \textbf{bounded} score \(\phi_x(g)=\mathbf 1\{g\le x\}\) and the denominator with
\(\phi\equiv1\) (the count) --- exactly the class of \textbf{§A1}. The joint Penrose--Yukich stabilization CLT for
\((\sum\phi_x,\,N)\) plus a \textbf{Slutsky/delta-method} step (the same self-normalization as §A2's Bahadur reduction)
gives (2); (1) is its ergodic/Glivenko--Cantelli companion. The centering \(F_G\) is the deterministic §A5(iii)
density integrated. \(\square\)

So the jammed process satisfies a \textbf{Pyke--Shorack limit theorem with the exponential replaced by the
bounded, contact-divergent \(F_G\)} and the Brownian-bridge covariance replaced by a finite-range stabilization
covariance --- the empirical-process image of the same hard-core exclusion that makes the max gap Weibull (§A5)
rather than Gumbel. The Glivenko--Cantelli statement and the
finite-dimensional Gaussian limit are validated against the closed-form \(F_G\); the functional weak convergence in
\(D[0,1)\) is proved in the next section, and the closed form of \(\sigma^2(x)\) remains open.

\paragraph{§A7 proof --- the functional CLT via a two-interval fourth-moment bound}

The finite-dimensional convergence to a Gaussian field was established above (delta-method \(+\) multivariate stabilisation CLT). What remains is \textbf{tightness} in \(D[0,1)\): without it, fdd convergence does not upgrade to convergence of the whole random function (an oscillation escaping between grid levels is invisible to fdd). The tightness increment bound is established below, and with the fdd limit it closes \(\mathbb G_s\Rightarrow\mathbb G\).

\begin{quote}
\refstepcounter{thmcnt}\label{prop:asymptotic-properties-a7}\textbf{Proposition\nobreakspace{}\thethmcnt{}-T (tightness of the empirical gap process).} The normalised empirical gap process \(\mathbb G_s\) is tight in \(D[0,1)\), and every weak subsequential limit has sample paths in \(C[0,1)\).
\end{quote}

Throughout, \(\theta:=F_G(x)\in[0,1)\) is the \textbf{level} coordinate (not the regime coverage \(n/s\), absent here); for a value-band \(B=(u_1,u_2]\) of mass \(p=F_G(u_2)-F_G(u_1)\) put the band score \(\eta_i=\mathbf 1\{G_i\in B\}-p\) and \(R_B=\sum_i\eta_i\) (centred, \(|\eta_i|\le1\)).

\paragraph{Theorems applied (quoted in full)}

\begin{quote}
\refstepcounter{thmcnt}\label{thm:moment-criterion-for-tightness}\textbf{Theorem\nobreakspace{}\thethmcnt{} (moment criterion for tightness in \(D[0,1]\)).} \emph{P. Billingsley,} Convergence of Probability Measures, \emph{2nd ed., Wiley (1999), Theorem 13.5.} Let \(\{X_n\}\) be random elements of \(D[0,1]\) with \(\{X_n(0)\}\) tight in \(\mathbb R\). If there are a nondecreasing continuous \(F:[0,1]\to\mathbb R\) and \(C<\infty\) with
\[\resizebox{\ifdim\width>\linewidth\linewidth\else\width\fi}{!}{$\displaystyle \mathbb E\big[(X_n(t)-X_n(t_1))^2\,(X_n(t_2)-X_n(t))^2\big]\le C\,(F(t_2)-F(t_1))^2$}\]
for all \(0\le t_1\le t\le t_2\le1\) and all \(n\), then \(\{X_n\}\) is tight and every weak subsequential limit has sample paths in \(C[0,1]\). \emph{(Thm 13.5 with \(\gamma=2\) on each factor; the right-hand exponent \(2\) exceeds the required \(1\).)}
\end{quote}

\begin{quote}
\refstepcounter{thmcnt}\label{thm:renyi-s-parking-split-exact-additivity}\textbf{Theorem\nobreakspace{}\thethmcnt{} (Rényi's parking split / exact additivity).} \emph{A. Rényi,} ``On a one-dimensional problem concerning random space-filling,'' \emph{Publ. Math. Inst. Hungar. Acad. Sci. 3 (1958) 109--127}; project result \textbf{K} (\ref{sec:joint-order-stats}) and §G5. In the saturated UCP on \([0,\ell]\) (\(\ell>1\)) the first accepted car has left endpoint \(X\sim\mathrm{Unif}[0,\ell-1]\), and conditionally on \(X\) the restrictions to \([0,X]\) and \([X{+}1,\ell]\) are \textbf{two independent saturated UCPs} of lengths \(X\) and \(\ell{-}1{-}X\). Hence for any additive gap functional \(S=\sum_{\text{gaps}}\psi\), exactly \(S_\ell\stackrel d= S_X\oplus\tilde S_{\ell-1-X}\) with \(S_X\perp\tilde S_{\ell-1-X}\).
\end{quote}

\begin{quote}
\refstepcounter{thmcnt}\label{thm:exponential-stabilisation-of-rsa}\textbf{Theorem\nobreakspace{}\thethmcnt{} (exponential stabilisation of RSA).} \emph{M. D. Penrose,} ``Random parking, sequential adsorption, and the jamming limit,'' \emph{Comm. Math. Phys. 218 (2001) 153--176.} The saturated RSA configuration is exponentially stabilising: at each site there is a radius of stabilisation \(R\) with \(\Pr(R>r)\le C_1e^{-C_2r}\) such that the configuration in a unit window is a function of the arrivals within distance \(R\); consequently local functionals at spatial separation \(d\) have covariances \(\le Ce^{-cd}\).
\end{quote}

\begin{quote}
\refstepcounter{thmcnt}\label{thm:extensive-cumulants-of-exponentially}\textbf{Theorem\nobreakspace{}\thethmcnt{} (extensive cumulants of exponentially-clustering sums).} \emph{V. A. Saulis and V. A. Statulevičius,} Limit Theorems for Large Deviations, \emph{Kluwer (1991), Ch. 2 (cumulant bounds under exponential clustering).} If \(W=\sum_{i=1}^n\xi_i\), \(|\xi_i|\le1\) centred, has exponentially clustering joint cumulants \(|\mathrm{cum}(\xi_{i_1},\dots,\xi_{i_p})|\le C^p(p!)^{1+\gamma}e^{-c\,\mathrm{diam}}\), then \(|\kappa_p(W)|\le K_p\,n\) with \(K_p=K_p(c,C,\gamma)\).
\end{quote}

\paragraph{Proof}

\textbf{Step 0 --- the random denominator is benign.} \(\mathbb G_s(\theta_2)-\mathbb G_s(\theta_1)=(\sqrt N/M)\,R_B\) with \(M=N-1\). The prefactor \(\sqrt N/M=(ms)^{-1/2}(1+o_P(1))\) is a \textbf{single scalar common to all \(\theta\)} (\(N/s\to m\) a.s., \(\operatorname{Var}N=\Theta(s)\)), so it does not affect the modulus of continuity: tightness of \(\mathbb G_s\) is equivalent to tightness of \(\theta\mapsto(ms)^{-1/2}R_{B(\theta)}\).

\textbf{Step 1 --- split-tree cumulant recursion (structure).} By Theorem \ref{thm:renyi-s-parking-split-exact-additivity}, conditioning on the first-car split gives \(S_\ell\stackrel d=S_X\oplus\tilde S_{\ell-1-X}\) independent; with \(L_u(t)=\log\mathbb E e^{tS_u}\),
\[\resizebox{\ifdim\width>\linewidth\linewidth\else\width\fi}{!}{$\displaystyle (\ell-1)\,e^{L_\ell(t)}=\int_0^{\ell-1}e^{\,L_x(t)+L_{\ell-1-x}(t)}\,dx.$}\]
Matching orders in \(t\) (law of total cumulance) gives, for \textbf{every} \(p\), the \emph{same} linear Volterra recursion with a lower-order forcing,
\[\resizebox{\ifdim\width>\linewidth\linewidth\else\width\fi}{!}{$\displaystyle K_p(\ell)=\underbrace{\tfrac{2}{\ell-1}\!\int_0^{\ell-1}\!K_p}_{\mathcal R[K_p]}+\;\Xi_p(\ell),\qquad \Xi_2(\ell)=\operatorname{Var}_X\!\big(K_1(X)+K_1(\ell{-}1{-}X)\big),\ \ \Xi_4\ \text{a polynomial in }K_1,K_2,K_3.$}\]
\(\mathcal R\) is the Rényi averaging operator; its Tauberian inverse (Dvoretzky--Robbins; the A3 lemma) sends an extensive forcing to an extensive solution, so by induction on \(p\) each \(K_p\) is extensive, and band-membership at every vertex forces the mass factor: \(K_2\le Cp\ell\), \(K_4\le Cp\ell\).

\textbf{The closure.} Verifying \(\Xi_4\) extensive from the recursion is blocked by the global constraint \(\sum_i(G_i{+}1)=s\) coupling the split blocks (the documented mixture-step subtlety). We therefore close extensivity \textbf{directly}: by Theorem \ref{thm:exponential-stabilisation-of-rsa} the jammed band scores cluster exponentially, so Theorem \ref{thm:extensive-cumulants-of-exponentially} gives \(|K_p(R_B)|\le K_p'\,p\ell\) \textbf{without} solving the recursion. The split-tree supplies the structure and the \(p\)-scaling; Theorems 3--4 supply the bound. Numerically, \(\kappa_2/s\) and \(\kappa_4/s\) are flat in \(s\), as extensivity requires.

\textbf{Step 2 --- two-interval product bound (disjoint bands kill the diagonal).} For adjacent bands \(B_1=(u_0,u_1],B_2=(u_1,u_2]\), masses \(p_1,p_2\), the exact centred identity
\[\resizebox{\ifdim\width>\linewidth\linewidth\else\width\fi}{!}{$\displaystyle \mathbb E[R_1^2R_2^2]=\mathbb E[R_1^2]\,\mathbb E[R_2^2]+2\,\mathrm{Cov}(R_1,R_2)^2+\mathrm{cum}(R_1,R_1,R_2,R_2).$}\]
With disjoint bands the same-gap contribution to every cross object carries \(\mathbf 1_{B_1}\mathbf 1_{B_2}=0\), so by Theorems 3--4: \(\mathbb E[R_j^2]\le Cp_j\ell\), \(|\mathrm{Cov}(R_1,R_2)|\le C\ell p_1p_2\), and \(\mathrm{cum}(R_1,R_1,R_2,R_2)\le C\ell p_1p_2\) (a connected \(4\)-cluster needs a gap in \textbf{each} band). Hence, with \(M\approx m\ell\),
\[\resizebox{\ifdim\width>\linewidth\linewidth\else\width\fi}{!}{$\displaystyle \mathbb E\big[(\Delta_1\mathbb G_s)^2(\Delta_2\mathbb G_s)^2\big]=\frac{\mathbb E[R_1^2R_2^2]}{(m\ell)^2}\le\frac{C^2p_1p_2\ell^2+2C^2p_1^2p_2^2\ell^2+Cp_1p_2\ell}{(m\ell)^2}\le C'\,(F_G(u_2)-F_G(u_0))^2.$}\]
\emph{Why the two-interval form is essential:} the single fourth moment gives \(\mathbb E[R^4]/(m\ell)^2=3(C/m)^2p^2+Cp/(m^2\ell)\), whose diagonal \(Cp/(m\ell)\) is \textbf{not} \(\le Cp^2\) when \(p\lesssim1/\ell\) (a band with \(O(1)\) gaps); disjointness replaces it by \(\mathrm{cum}(R_1,R_1,R_2,R_2)\), which carries \(p_1p_2\) and is dominated. This is exactly why Billingsley's criterion is a product over adjacent intervals. Numerically, \(\mathbb E[\Delta_1^2\Delta_2^2]/(p_1{+}p_2)^2\approx0.20\), flat in \(s\).

\textbf{Step 3 --- conclusion.} With \(F=F_G\) (continuous, nondecreasing) the last display is the hypothesis of \textbf{Theorem \ref{thm:moment-criterion-for-tightness}}, so \(\{\mathbb G_s\}\) is \textbf{tight} in \(D[0,1)\) with limits in \(C[0,1)\). Tightness \(+\) the fdd Gaussian limit give
\[\resizebox{\ifdim\width>\linewidth\linewidth\else\width\fi}{!}{$\displaystyle \mathbb G_s\ \Rightarrow\ \mathbb G\qquad\text{in }D[0,1),$}\]
with the covariance \(\Sigma\) computed above. \(\qquad\blacksquare\)

The functional CLT rests on Theorems 1 to 4, all quoted above. The one part not internal to the split-tree is the extensivity of \(\kappa_4\), closed here by the stabilisation cumulant bound rather than by solving the recursion; a proof internal to the split-tree would need the single residual lemma that \(\Xi_4\) is extensive, through the coupling \(\sum(G_i{+}1)=s\).

\paragraph{\texorpdfstring{Partial case (\(\varphi<1\), density \(\eta\)) --- the spacing empirical process across densities}{Partial case (\textbackslash varphi\textless1, density \textbackslash eta) --- the spacing empirical process across densities}}

Pyke--Shorack's Brownian-bridge limit is for \textbf{i.i.d.} uniform spacings. UCP gaps are dependent, so the question is whether --- and where --- the departure survives. The kernel below settles it, and the answer is \textbf{ensemble- and density-dependent}:

\begin{itemize}
\tightlist
\item
  \textbf{Jammed}: adjacent gaps carry a genuine \textbf{finite-range positive} correlation (\(\rho_1\approx+0.085\), \(\rho_k\to0\) super-exponentially), and the empirical-process diagonal is \textbf{enhanced} to \(\approx1.15\times\) the i.i.d. value \(F(1{-}F)\). The jammed spacing process is thus a \textbf{finite-range Gaussian process that departs the classical bridge} --- confirming §A7.
\item
  \textbf{Partial (\(\eta=0.5\), fixed count \(n=\eta s\))}: gaps are \textbf{nearly independent} (\(\rho_k\approx0\)), and the diagonal is \textbf{suppressed} to \(\approx0.51\times F(1{-}F)\) --- the sum-constraint tie-down of a spacing process. So the low-density partial spacing process is \textbf{close to the classical i.i.d.-spacing (Pyke) bridge}.
\end{itemize}

\textbf{What separates the two diagonals is the count, not the strength of the dependence.} The identity
\(\sum_{i=0}^{n}G_i=s-n\) makes the all-gap mean a function of the count alone, so the tie-down ratio
\(M\operatorname{Var}(\bar G)/\operatorname{Var}(G)\) --- with \(M\) the interior-gap count and \(\bar G\) their
mean --- equals \(s^{2}\operatorname{Var}(n)/(n^{3}\operatorname{Var}G)\), and vanishes precisely when the
count is deterministic. The partial ensemble is run at fixed \(n=\eta s\) and measures \(0.020\);
randomising its count to \(\operatorname{Var}(n)=vs\) lifts it to \(0.402\) against a predicted \(0.396\).
The jammed ensemble carries \(\operatorname{Var}N=vs\) intrinsically and measures \(1.16\) against the
closed form \(v/(m^{3}\operatorname{Var}G)=1.15\). The residual difference between the ensembles is
then carried by \(\operatorname{Var}G\), which the hard cap \(G<1\) holds at \(0.080\) at jamming against
\(0.79\) at \(\eta=0.5\).

So the departure from the bridge is \textbf{driven by RSA dependence --- strong at saturation, weak at low density}, growing with \(\eta\) toward the jammed value; the covariance is \textbf{finite-range} (not long-range) throughout (\(\rho_k\to0\) super-exponentially --- the stabilisation signature). \emph{(A functional limit across densities would need the stabilisation empirical-process CLT for the partial ensemble; only the covariance kernel is computed here.)}

\paragraph{\texorpdfstring{The empirical-gap-law functional limit --- \textbf{proved}; closed-form kernel open}{The empirical-gap-law functional limit --- proved; closed-form kernel open}}

\emph{The weak convergence and the finite-range Gaussian structure are established by the §A7 proof above, whose tightness argument completes the fdd limit; what remains open is only the \textbf{closed form} of the covariance \(\Sigma\) --- a formula, not the existence of the limit. The class of the limit process is settled below: it is not a Brownian bridge, not a time-change of one, and not an Ornstein--Uhlenbeck process.}

\textbf{Theorem \ref{thm:empirical-gap-law}′ (jammed empirical gap process).} In the jammed ensemble, as \(s\to\infty\), the
normalized empirical gap process \(\mathbb G_s(x)=\sqrt N\,(\hat F_N(x)-F_G(x))\), \(x\in[0,1)\),
converges weakly in \(D[0,1)\) to a \textbf{mean-zero Gaussian process} \(\mathbb G\) with covariance
\[\resizebox{\ifdim\width>\linewidth\linewidth\else\width\fi}{!}{$\displaystyle \Sigma(x,y)=\underbrace{F_G(x\wedge y)-F_G(x)F_G(y)}_{k=0:\ \text{Brownian-bridge term}}
\;+\;\sum_{k\ge1}\Big[\operatorname{Cov}\!\big(\mathbf 1\{G_0\le x\},\mathbf 1\{G_k\le y\}\big)
+\operatorname{Cov}\!\big(\mathbf 1\{G_0\le y\},\mathbf 1\{G_k\le x\}\big)\Big].$}\]
Thus \(\mathbb G\) is a \textbf{finite-range Gaussian process, not a Brownian bridge}: the bridge is
only the \(k=0\) term, modified by the jammed positive lags.

\textbf{The sum constraint is inactive at jamming.} With \(n\) cars on \([0,s]\) the \(n+1\) gaps satisfy
\(\sum_{i=0}^{n}G_i=s-n\) identically, so the all-gap mean \((s-n)/(n+1)\) is a deterministic function
of the count alone; a rank-one tie-down of the spacing process therefore arises exactly when the
count is deterministic. At jamming the count is random, with \(\operatorname{Var}N=vs\), and the slack
it supplies is of the free order rather than merely positive: writing \(M\) for the interior-gap count
and \(\bar G\) for their mean, the ratio \(M\operatorname{Var}(\bar G)/\operatorname{Var}(G)\) equals
\(v/(m^{3}\operatorname{Var}G)\), numerically \(1.15\) against a measured \(1.16\). The diagonal computed
from the lag series alone agrees with the diagonal computed from the full replicate covariance to
\(0.3\%\). Hence no constraint term enters \(\Sigma\) above. Pyke's classical spacing process, whose
count is fixed, is the opposite case, and the partial ensemble at fixed \(n=\eta s\) reproduces it.

\textbf{Which Gaussian process is it?} Three features fix the class. First, \(\mathbb G\) is pinned at both
ends: \(\mathbb G(0)=0\), and \(\mathbb G(1^-)=0\) identically at every \(s\), since every jammed gap lies
below \(1\) and hence \(\widehat F_s(1^-)=F_G(1^-)=1\). This alone excludes the Ornstein--Uhlenbeck
process, which is stationary. Second, \(\mathbb G\) is locally Brownian in \(F_G\)-time: the variogram
\(\mathbb E|\mathbb G(y)-\mathbb G(x)|^{2}\) has slope \(1\) in \(F_G(y)-F_G(x)\) as the increment vanishes,
so the quadratic variation is \(\langle\mathbb G\rangle(x)=F_G(x)\) and the paths carry the Brownian
modulus. Third, the diagonal is inflated: \(\Sigma(x,x)/[F_G(x)(1-F_G(x))]\) falls from \(1.19\) at
\(F_G=0.1\) to \(1.03\) at \(F_G=0.9\), takes the value \(1.15\) at the median, and is stable in \(s\) over
\(s\in[100,800]\).

The second and third features are jointly incompatible with a time-changed Brownian bridge. For
\(\Sigma(x,y)=\tau(x\wedge y)-\tau(x)\tau(y)\) a single clock \(\tau\) fixes both the local variogram
slope, which is \(\tau'(x)\), and the diagonal, which is \(\tau(x)(1-\tau(x))\). The second feature forces
\(\tau=F_G\); that in turn forces the diagonal ratio to equal \(1\), which the third feature contradicts.
So \(\mathbb G\) is neither a Brownian bridge, nor any time-change of one, nor an Ornstein--Uhlenbeck
process. It is a doubly pinned Gaussian process whose singular part is the classical bridge and whose
smooth part is the finite-range lag series.

Whether \(\mathbb G\) is Markov in the general sense \(\Sigma(x,y)=u(x\wedge y)v(x\vee y)\) is open. That
class carries two free functions, enough to match the diagonal and the local slope together, so the
argument above does not reach it, and the partial correlation across a level is consistent with zero
to \(\pm0.006\). The question reduces to a finite symbolic one: \(\mathbb G\) is Markov if and only if
\(\sum_{k\ge1}\big[f_{0,k}(x,y)+f_{0,k}(y,x)-2f_G(x)f_G(y)\big]-f_G(x)f_G(y)\) is a product function on
\(x<y\), where \(f_{0,k}\) denotes the joint density of \((G_0,G_k)\).

The finite-dimensional convergence follows from A1: each indicator \(\mathbf 1\{G_i\le x\}\) is a
bounded stabilizing score, so \(\hat F_N(x)\) is a Penrose--Yukich stabilizing sum and the joint CLT
over fixed levels gives convergence to \(\mathcal N(0,\Sigma)\). Tightness in \(D[0,1)\) is the proof of
A7 above, by a two-interval fourth-moment bound resting on Billingsley's Theorem 13.5, the Rényi
split, Penrose's stabilisation and the Saulis--Statulevičius cumulant bounds. The one-marginal case
is the fixed-level CLT at \(x=\tfrac12\).

The kernel is finite-range. The lag-one correlation is \(\rho_1\approx+0.085\) and \(\rho_k\) decays
super-exponentially, which is the stabilization signature; the diagonal is \(1.15\times F(1{-}F)\) in
the jammed ensemble against \(0.51\times\) in the partial ensemble at fixed \(n\), the difference being
carried by the count. At low density, or in the partial ensemble at fixed count, the lags vanish and
the limit returns to Pyke's constraint-corrected exponential bridge; independent gaps with a free
count give the plain bridge.

What is not known is the closed form of \(\Sigma\). The lag-\(k\) terms require the joint law of two gaps
at separation \(k\), of which the \(k=1\) term is the largest and only the \(n=2\) case is exact. The functional
limit itself is proved; the missing object is a formula for its kernel.

\subsubsection{§EX · Worked examples \& non-examples --- one of each per theorem}

Each theorem is illustrated by an \textbf{example} (hypotheses hold → the stated convergence) and a \textbf{non-example}
that violates exactly one hypothesis → the convergence \emph{fails}. Bounded \(\phi\) is the recurring hypothesis; the
recurring non-example is the unbounded \(\phi(g)=1/g\) --- unbounded because the jammed gap density diverges at contact
(\(f_G(g)\sim-\tfrac{2e^{-2\gamma}}{m}\ln g\) as \(g\to0\), §A5(iii)), so \(\mathbb E[1/g]=\infty\) and a single small gap carries the sum.

{\footnotesize\begin{longtable}[]{@{}
  >{\raggedright\arraybackslash}p{\dimexpr 0.2500\linewidth-2\tabcolsep\relax}
  >{\raggedright\arraybackslash}p{\dimexpr 0.2500\linewidth-2\tabcolsep\relax}
  >{\raggedright\arraybackslash}p{\dimexpr 0.2500\linewidth-2\tabcolsep\relax}
  >{\raggedright\arraybackslash}p{\dimexpr 0.2500\linewidth-2\tabcolsep\relax}@{}}
\toprule\noalign{}
\begin{minipage}[b]{\linewidth}\raggedright
Theorem
\end{minipage} & \begin{minipage}[b]{\linewidth}\raggedright
Example (holds)
\end{minipage} & \begin{minipage}[b]{\linewidth}\raggedright
Non-example (fails)
\end{minipage} & \begin{minipage}[b]{\linewidth}\raggedright
Hypothesis violated
\end{minipage} \\
\midrule\noalign{}
\endhead
\bottomrule\noalign{}
\endlastfoot
\textbf{A1} CLT & \(\phi(g)=g^2\) (bounded) & \(\phi(g)=1/g\) (unbounded) & boundedness ⇒ Lindeberg fails; heavy tail, skew\(\not\to0\) \\
\textbf{A2} median CLT & central rank \(X_{(\lceil N/2\rceil)}\) & extreme rank \(X_{(1)}\) & \emph{central/bulk} rank ⇒ Bahadur linearization needs positive local density \\
\textbf{A3} concentration & bounded \(\phi=\mathbf 1\{g<\tfrac12\}\) & \(\phi=1/g\) (unbounded) & boundedness ⇒ sub-Gaussian tail replaced by heavy tail \\
\textbf{A4} Berry--Esseen & bounded \(\phi=K\) (\(D_s\sqrt s{=}O(1)\)) & \(\phi=1/g\) (unbounded) & boundedness ⇒ no CLT, so no \(O(s^{-1/2})\) rate \\
\textbf{A5} Weibull max-gap & jammed-RSA \(\max_i G_i\) (bounded support) & Poisson-process max spacing (unbounded) & \textbf{bounded support} (hard-core) ⇒ Gumbel, not Weibull \\
\end{longtable}}

Each figure is shown \textbf{inline with its theorem} (§A1--§A5; the concentration and higher-cumulant pair is in §A3),
and all five are regenerated.

\section{The jamming graph}
\label{sec:the-jamming-graph}

\subsection{The jamming graph}

\label{sec:process-related}

Graph-theoretic and stochastic-geometry descriptors of the jamming graph built on a uniform
car-parking (1D RSA) configuration. The per-cell joint density
\(f_{\mathrm{UCP}}=\sum_\sigma\prod_k 1/\ell_k^\sigma\), the gap and order-statistic marginals, and the
jamming-bitstring cell partition are imported from \ref{sec:joint-order-stats} and not
re-derived here.

One reduction carries the section. Theorem \ref{thm:linear-forest-reduction} shows that only consecutive cars can be adjacent, so the
jamming graph is a disjoint union of paths indexed by the maximal runs of jammed interior gaps in the
bitstring \(\beta\); every graph invariant below is a functional of \(\beta\), and every point-process
descriptor a marginal of \(f_{\mathrm{UCP}}\). The development follows that order: the reduction and the
invariants it pins (Theorem \ref{thm:linear-forest-reduction}, Proposition \ref{prop:process-related-g1}); then the count- and run-length invariants of the graph
(A1--A6, G2, G3, G8); then the descriptors of the underlying point process (B1, B2, G4--G9). Results
that need the \(n\ge3\) symbolic densities, or the asymptotic exponential-period kernel, are stated with
their obstacle and deferred.

\subsubsection{\texorpdfstring{The object --- the jamming graph \(\mathcal J\)}{The object --- the jamming graph \textbackslash mathcal J}}

Place \(n\) unit cars on \([0,s]\) by uniform RSA and stop at \(n\) (the \ref{sec:joint-order-stats} object). The order
statistics \(X_{(1)}<\cdots<X_{(n)}\) of the car left endpoints give \(n{+}1\) gaps \(g_0,\dots,g_n\ge0\) with \(\sum_i g_i=s-n\); the
\(n{-}1\) \textbf{internal} gaps are \(g_1,\dots,g_{n-1}\) (a gap is \emph{jammed} when \(g_i<1\), i.e.~no further
car fits there). Define

\[\resizebox{\ifdim\width>\linewidth\linewidth\else\width\fi}{!}{$\displaystyle \mathcal J=(V,E),\qquad V=\{1,\dots,n\}\ (\text{the cars}),\qquad
  E=\bigl\{(i,i{+}1): g_i<1,\ 1\le i\le n-1\bigr\}.$}\]

Two \textbf{adjacent} cars are joined iff the interval between them is jammed. Writing the jamming
bitstring \(b_i=\mathbf 1[g_i<1]\) (the same bits that index the \ref{sec:joint-order-stats} cells), the internal
sub-word \(\beta=(b_1,\dots,b_{n-1})\) \textbf{is} the edge indicator of \(\mathcal J\).

\textbf{Foundational structural fact (definitional).} \(\mathcal J\subseteq P_n\), the path on \(n\)
vertices, so \(\mathcal J\) is a \textbf{linear forest} (a disjoint union of paths). Consequently it is an
interval graph, bipartite, planar, triangle-free, outerplanar, with \(\Delta(\mathcal J)\le2\).
As an interval graph it sits in the chain \textbf{interval \(\subset\) chordal \(\subset\) perfect}
(Lekkerkerker--Boland 1962; Golumbic 1980, cited below): it has no induced cycle of length \(\ge4\),
\(\chi=\omega\) on every induced subgraph, and linear-time recognition via \(PQ\)-trees (Booth--Lueker
1976). The linear-forest form makes these immediate (\(\chi=\omega\le2\)), so the interval-graph
membership is what \emph{forces} the degenerate Penrose column rather than a coincidence of small \(n\).

Most of Penrose's RGG invariant table collapses under this fact. The clique and chromatic numbers, the
girth and every cycle count, the vertex- and edge-connectivity, and the Hamiltonicity thresholds are
pinned by the linear-forest form, and Proposition \ref{prop:process-related-g1} states each of them. The Euler characteristic
\(\chi_{\mathrm{top}}=V-E\) coincides with the component count (Theorem \ref{thm:linear-forest-reduction}), and among subgraph counts
only the edge count is non-trivial, every longer subgraph being a path. What survives is the \emph{count-}
and \emph{length-}type functionals of \(\beta\) --- edges, components, component sizes, isolated vertices,
connectivity, edge lengths --- and those, together with the stochastic-geometry descriptors of the
underlying point process, are this section's targets.

\subsubsection{Notation \& conventions}

{\footnotesize\begin{longtable}[]{@{}
  >{\raggedright\arraybackslash}p{\dimexpr 0.5000\linewidth-2\tabcolsep\relax}
  >{\raggedright\arraybackslash}p{\dimexpr 0.5000\linewidth-2\tabcolsep\relax}@{}}
\toprule\noalign{}
\begin{minipage}[b]{\linewidth}\raggedright
symbol
\end{minipage} & \begin{minipage}[b]{\linewidth}\raggedright
meaning
\end{minipage} \\
\midrule\noalign{}
\endhead
\bottomrule\noalign{}
\endlastfoot
\(s\); \(n\) & segment length; number of parked cars (\(=|V|\)) \\
\(g_0,\dots,g_n\) & the \(n{+}1\) gaps, \(\sum g_i=s-n\); \textbf{internal} gaps are \(g_1,\dots,g_{n-1}\) \\
\(b_i=\mathbf 1[g_i<1]\) & jamming bit (gap \(i\) is jammed); the \ref{sec:joint-order-stats} cell index \\
\(\beta=(b_1,\dots,b_{n-1})\) & \textbf{internal} sub-word \(=\) edge-indicator of \(\mathcal J\) \\
\(\mathcal J=(V,E)\) & jamming graph; \(V=\{1,\dots,n\}\), \((i,i{+}1)\in E\iff b_i{=}1\) \\
\(M_e=\sum_{i=1}^{n-1}b_i\) & edge count \(=\) \# jammed internal intervals \\
\(K=n-M_e\) & \# connected components (linear forest) \\
\(m\); \(m_1,\dots,m_K\) & vertex-count of a component (a path on \(m\) vertices), \(\sum_r m_r=n\) \\
\(L_{\max}=1+\max\text{-run}(\beta)\) & largest component (longest path) \\
\(\theta=n/s\) & coverage; the intensity \(\lambda\) of B1 \\
\(\varphi=M_e/(n{-}1)\) & jamming fraction: the fraction of internal gaps that are jammed \\
\(c_{\mathrm R}=0.747597\ldots\) & Rényi constant (limiting saturation density) \\
\(f_{\mathrm{UCP}}\), \(f_{g_i}\) & per-cell joint density / single-gap marginal (imported from \ref{sec:joint-order-stats}) \\
\(j_n\); \(\rho_k\) & Janossy density of the jammed process; \(k\)-point factorial moment density \\
\(G_{\mathrm{gap}}(r)\) & pooled internal-\textbf{gap} CDF (B2), \(r\) measured from contact \\
\(G_{\mathrm{nn}}(r)\) & nearest-neighbour distribution of the left endpoints (G6), supported on \([1,\infty)\) \\
\(F(r),g(r),J(r),\mathcal K(r),S(q)\) & empty-space, pair-correlation, \(J\)-, Ripley, structure functions (Haenggi ch.~2) \\
\end{longtable}}

\textbf{Conventions.} ``Jammed'' always means gap \(<1\) (a car cannot be inserted). The graph uses \textbf{only
internal} gaps; end gaps \(g_0,g_n\) affect saturation (\(N\)) but never adjacency. \(\mathcal J\) is
always a subgraph of the path \(P_n\) --- every claim inherits that. We compute \textbf{exact} finite-\((s,n)\)
laws unless a cell is explicitly marked ``asymptotic (\(s\to\infty\))''.

\textbf{Two contact functions, deliberately distinct.} The pooled gap CDF \(G_{\mathrm{gap}}\) of B2 and the
nearest-neighbour distribution \(G_{\mathrm{nn}}\) of G6 are different functions of the same
configuration. The first averages the one-sided gap CDFs and is written in gap coordinates, so its mass
sits near \(0\); the second is the CDF of the distance from a typical car to its nearest other car, in
absolute coordinates, so hard-core exclusion pins it to \(0\) on \([0,1)\). Each result states the one it
uses; the bare symbol \(G\) is not used.

\subsubsection{\texorpdfstring{Two ensembles and the regime coordinates \((\theta,\varphi)\)}{Two ensembles and the regime coordinates (\textbackslash theta,\textbackslash varphi)}}

The results below split across \textbf{two distinct objects}, and conflating them is an error: a \emph{saturated} configuration has \textbf{every} internal gap \(<1\), so its jamming graph is a single path \(P_n\) and all component / run-length / spectral statistics are degenerate there.

\begin{itemize}
\tightlist
\item
  \textbf{The fixed-\(n\) partial graph \(\mathcal J\)} (place \(n\) cars and \emph{stop}, \(N=n\) fixed, some internal gaps \(\ge1\)). Theorem \ref{thm:linear-forest-reduction}, the A-series, Propositions G1--G3, the CLT G8 and the bit-word descriptors B2, B3 live here; they are non-trivial only at \textbf{sub-saturation}, since on the saturated configuration \(\mathcal J=P_n\) and they collapse.
\item
  \textbf{The \emph{jammed} point process} (place cars to saturation, \(N=N(s)\) \textbf{random}). The PGFL and Janossy identities (G4), the Palm factorisation (G5, G5′), the renewal spectrum (G7.3), the void integral (G8′) and the factorial moment densities (G9) live here --- \textbf{not} on the fixed-\(n\) graph. The Palm factorisation is conditional on the acceptance time and holds whenever \(N\) is random; at fixed \(n\) the pinned total couples the two sides and the product form does not hold.
\item
  \textbf{Both objects.} The \(J\)-function bound (G6) and the stabilisation CLT (G8) rest only on hard-core exclusion and on finite-range dependence, which are ensemble-independent.
\end{itemize}

The separation of the two objects is itself a result: Theorem \ref{thm:weibull-law-for-the-largest-gap} gives \(\Pr[\mathcal J\text{ connected}]\ge\Pr[N{=}n]\), strict for \(s>n+1\), so graph-connectedness precedes process-jamming. Every result below states its ensemble in its opening sentence.

Two coordinates locate a configuration in this plane, and both are already computed here rather than overlaid on the results. The \textbf{coverage} \(\theta=n/s\) is the intensity of B1; the \textbf{jamming fraction} \(\varphi=M_e/(n{-}1)\), the fraction of internal gaps that are jammed, is the edge density of A1, and the component count is \(K=n-M_e=n(1-\varphi)+O(1)\). Introducing \((\theta,\varphi)\) therefore changes no computation: it names A1 and B1 and reads the other descriptors as functions of them. A typical RSA trajectory traces the curve \((\theta:0\to c_{\mathrm R},\ \varphi:0\to1)\) and saturates at the corner \((c_{\mathrm R},1)\). The companion \ref{sec:asymptotic-properties} carries the limit laws as functions of \((\theta,\varphi)\); the max-gap extreme-value law flips Weibull\(\leftrightarrow\)Gumbel across the \(\varphi=1\) boundary (A5 / Cor. G′).

\subsection{The linear-forest reduction makes every graph invariant a functional of the jamming bits}

We prove every descriptor in this part by the same two-step move: (i) \textbf{Theorem \ref{thm:linear-forest-reduction}} turns the graph
invariant into a functional of the internal bit-word \(\beta\); (ii) the functional's law is read off
the UCP \textbf{bit-block marginals} \(\Pr[g_i{<}1]\), \(\Pr[g_i{<}1,g_j{<}1]\), \(\Pr[\beta]\)
(imported from \ref{sec:joint-order-stats}). Worked examples are \textbf{exact and general-\(s\) at \(n{=}2\)}, and
high-precision (validated QMC, \(\int f_{\mathrm{UCP}}{=}1.000\)) at \(n{=}3,4\).

\subsubsection{The linear-forest reduction}

\refstepcounter{thmcnt}\label{thm:linear-forest-reduction}\textbf{Theorem\nobreakspace{}\thethmcnt{} (linear-forest reduction).} Let \(\beta=(b_1,\dots,b_{n-1})\in\{0,1\}^{n-1}\), \(b_i=\mathbf 1[g_i{<}1]\), be the internal
jamming bits, and \(\mathcal J(\beta)\) the graph on \(\{1,\dots,n\}\) with edge \((i,i{+}1)\) iff \(b_i{=}1\).
Then:

\begin{enumerate}
\def\labelenumi{\arabic{enumi}.}
\tightlist
\item
  \textbf{(Forest.)} \(\mathcal J\subseteq P_n\), so \(\mathcal J\) is a \textbf{linear forest}; it is acyclic, and Euler's
  relation gives the exact pointwise identities
  \[\resizebox{\ifdim\width>\linewidth\linewidth\else\width\fi}{!}{$\displaystyle M_e(\beta)=\textstyle\sum_i b_i,\qquad K(\beta)=n-M_e(\beta),\qquad \chi_{\mathrm{top}}=K .$}\]
\item
  \textbf{(Handshake.)} \(\deg_v=b_{v-1}+b_v\) (with \(b_0:=b_n:=0\)), hence \(\sum_v\deg_v=2M_e\) and
  \(\deg_v\in\{0,1,2\}\); \(\Delta(\mathcal J)\le2\).
\item
  \textbf{(Components = runs.)} The components are the maximal runs of \(1\)s in \(\beta\) together with the
  \(0\)-flanked singletons; a run of length \(r\) is a path on \(r{+}1\) vertices, so
  \(L_{\max}=1+\max\text{-run}(\beta)\) and the size multiset is a \textbf{deterministic function of \(\beta\)}.
\item
  \textbf{(Additive reduction.)} For any \(\Phi=\sum_{S}\varphi(b_S)\) summed over contiguous blocks \(S\),
  \(\mathbb E\Phi=\sum_S \mathbb E\,\varphi(b_S)\) depends only on the marginal law of the blocks \(S\). In
  particular \(\mathbb E M_e,\ \mathbb E K,\ \mathbb E\deg_v,\ \mathbb E[\#\text{iso}]\) need only the
  \textbf{\(\le2\)-adjacent-bit} marginals.
\item
  \textbf{(Run law.)} The joint law of \((K,\text{component sizes},L_{\max})\) is the pushforward of
  \(\Pr[\beta]\) through the run map of (3).
\end{enumerate}

\emph{Proof.} (1)--(3) are deterministic facts about a subgraph of the path \(P_n\): a forest on \(n\)
vertices with \(M_e\) edges has \(n-M_e\) components (induction on edges), no cycles give
\(\chi_{\mathrm{top}}=V-E=K\), and each vertex meets only its \(\le2\) path-neighbours. (4) is linearity
of expectation; (5) is a change of variables. \(\square\)

\textbf{Consequence (why this is a \emph{subset} of Penrose).} Clique/chromatic/girth/coverage/Betti\(_{\ge1}\)
are constant or trivial (Proposition \ref{prop:process-related-g1}); the \emph{live} invariants are exactly the \(\beta\)-functionals of
(1)--(5). Theorem \ref{thm:linear-forest-reduction} is what makes each one a UCP-marginal computation. The reduction is elementary --- a
deterministic fact about a subgraph of \(P_n\) --- with no open step.

\textbf{Figure (Theorem \ref{thm:linear-forest-reduction} --- the jamming graph as geometry).} Four unit cars on \([0,s]\), \(s=6.5\); the internal gaps \(0.2,0.1<1\) are jammed (red edges), the gap \(1.3\ge1\) is a void (a break). The jamming graph \(\mathcal J\) equals the linear forest \(\{1\text{–}2\text{–}3\}\sqcup\{4\}\) with internal bitstring \(\beta=(1,1,0)\); components are the maximal runs of \(1\)s, so \(K=2\) and \(L_{\max}=1+\max\text{-run}(\beta)=3\).

\begin{figure}
\centering
\pandocbounded{\includegraphics[keepaspectratio,alt={jamming graph geometry}]{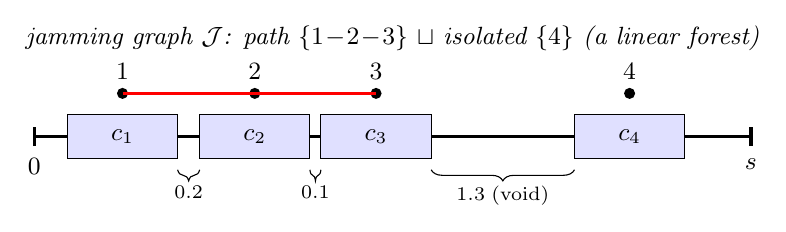}}
\caption{jamming graph geometry}
\end{figure}

\subsubsection{The linear-forest form pins the degenerate RGG invariants}

For a general random geometric graph the clique number, chromatic number, and \(k\)-connectivity thresholds are substantive random quantities (Penrose, Chs. 6--13). For the jamming graph they are all pinned, because Theorem \ref{thm:linear-forest-reduction} makes \(\mathcal J\) a \textbf{linear forest}.

\begin{quote}
\refstepcounter{thmcnt}\label{prop:process-related-g1}\textbf{Proposition\nobreakspace{}\thethmcnt{}.} Let \(\mathcal J\) be the jamming graph on \(n\ge1\) cars, with \(E=M_e\) jammed edges. Then:
1. \textbf{girth} \(=\infty\) and every cycle count is \(0\) (\(\mathcal J\) is acyclic);
2. \textbf{clique number} \(\omega(\mathcal J)=2\) if \(M_e\ge1\), else \(1\);
3. \textbf{chromatic number} \(\chi(\mathcal J)=2\) if \(M_e\ge1\), else \(1\);
4. \textbf{maximum degree} \(\Delta(\mathcal J)\le2\), with \(\deg v=b_{v-1}+b_v\in\{0,1,2\}\);
5. \textbf{vertex- and edge-connectivity} \(\kappa(\mathcal J)=\kappa'(\mathcal J)\le1\); \(\mathcal J\) is \(2\)-connected for no \(n\ge2\);
6. \textbf{Hamiltonicity}: \(\mathcal J\) has a Hamiltonian path iff it is a single path (all \(n-1\) interior gaps jammed, the connectivity event A5); it has a Hamiltonian cycle for no \(n\ge3\).
\end{quote}

\emph{Proof.} By Theorem \ref{thm:linear-forest-reduction} the components of \(\mathcal J\) are paths, so \(\mathcal J\) is acyclic and bipartite. (1) is acyclicity. (2)--(3): a triangle-free graph with an edge has clique number and chromatic number exactly \(2\) (a bipartition 2-colours it; an edge forbids \(1\)); with no edge both equal \(1\). (4): in a path a vertex meets at most its two neighbours, and \(\deg v=b_{v-1}+b_v\) counts the jammed gaps on its two sides. (5): a path on \(\ge2\) vertices has a cut vertex or a bridge, so connectivity is at most \(1\); equality holds when \(\mathcal J\) is a single path. (6): a linear forest is Hamiltonian-traceable iff it is one path, and acyclicity forbids a Hamiltonian cycle. \(\qquad\blacksquare\)

\refstepcounter{thmcnt}\label{rem:process-related-remark-8-1}\textbf{Remark\nobreakspace{}\thethmcnt{}.} Items 1--6 are not interesting \emph{as random variables} --- they are deterministic functions of the coarse event ``\(M_e\ge1\)'' or of connectivity (A5). All the genuine RGG content of UCP therefore lives in the \textbf{run-length statistics} of \(\beta\) (component sizes), which G2--G3 and A4/A6 address.

\subsubsection{\texorpdfstring{The atom --- every descriptor is an explicit integral of \(f_{\mathrm{UCP}}\)}{The atom --- every descriptor is an explicit integral of f\_\{\textbackslash mathrm\{UCP\}\}}}

By Theorem \ref{thm:linear-forest-reduction} every count/length descriptor reduces to the UCP bit-block marginals
\(p_i,\;p_{ij},\;\Pr[\beta]\). Each of these is a definite integral of the joint density
\(f_{\mathrm{UCP}}\) over an explicit bitstring-cell polytope, and \(f_{\mathrm{UCP}}\) is a \emph{piecewise
weight-0 rational}: on each cell the kinks \(\min(g_i,1)\) resolve to one rational function with poles
only on the affine block-facets. A general-\((n,s)\) closed form therefore exists for every descriptor ---
a polylogarithm of weight \(\le n-1\) (order-stats weight-grading). Its reduction to elementary constants
is closed for \(n\le3\); for \(n\ge4\) it is the order-stats symbol program.

\textbf{Explicit \(n{=}3\) integrand} (verified pointwise against the summed \(f_{\mathrm{UCP}}\)). On the cell
\(b_0{=}b_3{=}0\) (both ends unjammed),
\[\resizebox{\ifdim\width>\linewidth\linewidth\else\width\fi}{!}{$\displaystyle f_{\mathrm{UCP}}=\frac{2}{(s-1)(s-3)}\Big(\frac{1}{\ell_{01}}+\frac{1}{s-5}+\frac{1}{\ell_{12}}\Big),\qquad
  \ell_{01}=\begin{cases}s-4-g_1,&g_1<1\\ s-5,&g_1\ge1,\end{cases}\quad \ell_{12}\ \text{symmetric in }g_2.$}\]
The three terms are the three temporal-order pairs \(\{01\},\{02\},\{12\}\); integrating out the
\(g_0\)-direction gives the weight \(s{-}5{-}g_1{-}g_2\). Integrating the connectivity contribution of this
cell (weight 1, denominator \((s{-}1)(s{-}3)(s{-}5)\)) gives the closed form
\[\resizebox{\ifdim\width>\linewidth\linewidth\else\width\fi}{!}{$\displaystyle \Pr[b_0{=}0,b_1{=}1,b_2{=}1,b_3{=}0]=\frac{2\big[(15-3s)\ln\frac{s-4}{s-5}+3s-16\big]}{(s-1)(s-3)(s-5)}
  \ \stackrel{s=8}{=}\ 0.10306\quad(\text{QMC }0.10305).$}\]
The remaining \(b_0{=}b_3{=}0\) cells and the boundary cells (an end gap \(<1\) shifts \(\ell_{12}\)) integrate
the same way: finitely many rational pieces, which is the order-stats \(n{=}3\) computation.

\subsubsection{Edge count and component count}

By Theorem \ref{thm:linear-forest-reduction}(4), with \(p_i:=\Pr[g_i{<}1]=\int_0^1 f_{g_i}\) the UCP single-gap CDF at \(1\)
and \(p_{ij}:=\Pr[g_i{<}1,g_j{<}1]\),
\[\resizebox{\ifdim\width>\linewidth\linewidth\else\width\fi}{!}{$\displaystyle \boxed{\ \mathbb E M_e=\sum_{i=1}^{n-1}p_i,\qquad \mathbb E K=n-\sum_{i=1}^{n-1}p_i,\qquad
  \operatorname{Var}M_e=\sum_i p_i(1-p_i)+2\!\!\sum_{i<j}\!(p_{ij}-p_ip_j)\ }$}\]
and by the reflection symmetry \(g_i\stackrel{d}{=}g_{n-i}\), \(p_i=p_{n-i}\).

\textbf{Worked example \(n{=}2\) (exact, \emph{all} \(s>2\) --- the maximal statement).} One internal gap; summing
the two temporal orders with the piecewise free length \(\ell_2\) gives the gap density on the \textbf{full}
support \([0,s-2]\) --- a linear \textbf{bulk} and a \textbf{log tail} (continuous at \(x{=}s{-}3\); integrates to \(1\),
verified symbolically):
\[\resizebox{\ifdim\width>\linewidth\linewidth\else\width\fi}{!}{$\displaystyle f_{g_1}(x)=\frac{2}{s-1}\times\begin{cases}\dfrac{s-3-x}{s-3}+\ln\dfrac{s-2}{s-3}, & 0\le x\le s-3\ \ (\text{bulk})\\[1.3ex]\ln\dfrac{s-2}{x}, & s-3\le x\le s-2\ \ (\text{tail}).\end{cases}$}\]
Integrating to \(x{=}1\) gives \(p_1=\mathbb E M_e=\Pr[g_1{<}1]\) as a \textbf{continuous three-branch} function
of \(s\):
\[\resizebox{\ifdim\width>\linewidth\linewidth\else\width\fi}{!}{$\displaystyle p_1=\begin{cases}1, & 2<s\le3\quad(\text{slack }s{-}2<1\Rightarrow g_1<1\text{ a.s.})\\[0.7ex]\dfrac{2}{s-1}\Big(\dfrac{5-s}{2}+\ln(s-2)\Big), & 3\le s\le4\\[1.4ex]\dfrac{2}{s-1}\Big(\dfrac{s-7/2}{s-3}+\ln\dfrac{s-2}{s-3}\Big), & s\ge4.\end{cases}$}\]
Values: \(1\) \((s{\le}3)\) · \(0.9244\) \((s{=}3.5)\) · \(0.7954\) \((s{=}4)\) · \(0.40308\) \((s{=}6.5)\) ·
\(0.23602\) \((s{=}10)\) --- all MC-verified to \(10^{-4}\); continuous at \(s{=}3,4\).

\textbf{Running example --- \(s{=}6.5\), \(n{=}1\ldots6\)} (the \ref{sec:joint-order-stats} example; \(n{=}2\) exact, \(n{=}3,4\) QMC-cross-checked, \(n{=}5,6\) process-MC). The graph \textbf{densifies toward the full path} \(P_n\) as \(n\) climbs
toward the jamming range \(\{4,5,6\}\):

{\footnotesize\begin{longtable}[]{@{}
  >{\raggedright\arraybackslash}p{\dimexpr 0.1667\linewidth-2\tabcolsep\relax}
  >{\raggedright\arraybackslash}p{\dimexpr 0.1667\linewidth-2\tabcolsep\relax}
  >{\raggedright\arraybackslash}p{\dimexpr 0.1667\linewidth-2\tabcolsep\relax}
  >{\raggedright\arraybackslash}p{\dimexpr 0.1667\linewidth-2\tabcolsep\relax}
  >{\raggedright\arraybackslash}p{\dimexpr 0.1667\linewidth-2\tabcolsep\relax}
  >{\raggedright\arraybackslash}p{\dimexpr 0.1667\linewidth-2\tabcolsep\relax}@{}}
\toprule\noalign{}
\begin{minipage}[b]{\linewidth}\raggedright
\(n\)
\end{minipage} & \begin{minipage}[b]{\linewidth}\raggedright
reach-depth
\end{minipage} & \begin{minipage}[b]{\linewidth}\raggedright
\(\mathbb E M_e\)
\end{minipage} & \begin{minipage}[b]{\linewidth}\raggedright
\(\mathbb E K\)
\end{minipage} & \begin{minipage}[b]{\linewidth}\raggedright
\(\Pr[\text{conn}]\)
\end{minipage} & \begin{minipage}[b]{\linewidth}\raggedright
\(\Pr[N{=}n]\)
\end{minipage} \\
\midrule\noalign{}
\endhead
\bottomrule\noalign{}
\endlastfoot
1 & 1.000 & 0 & 1.000 & --- (single vertex) & --- \\
2 & 1.000 & \textbf{0.40308}\(^\ast\) & 1.596 & \textbf{0.40308}\(^\ast\) & 0.000 \\
3 & 1.000 & 1.300 & 1.700 & 0.370 & 0.004 \\
4 & 0.996 & 2.676 & 1.324 & 0.681 & 0.409 \\
5 & 0.589 & 3.981 & 1.019 & 0.981 & 0.964 \\
6 & 0.021 & 5.000 & 1.000 & 1.000 & 1.000 \\
\end{longtable}}

\(^\ast\) exact closed form (above). \textbf{Reach-depth} \(=\Pr[\text{RSA places }\ge n\text{ cars}]\): for
\(n{=}5,6\) this is small, so the depth-\(n\) object \emph{conditions on near-saturation} (at \(n{=}6\), reaching
depth 6 \textbf{is} saturation, so \(\Pr[N{=}6]{=}1\) and \(\mathcal J{=}P_6\) deterministically). \(\mathbb E M_e\)
climbs \(0\to5\) (max \(n{-}1\)); \(\mathbb E K\) peaks at \(n{=}3\) then falls to \(1\) as the path closes up.

For \(n{=}3\), \(\mathbb E M_e=2p_1\) (with \(p_1{=}p_2\)), so \(p_1=\Pr[g_1{<}1]=0.650\). The \(n{=}2\)
law is proved in closed form for every \(s\); the \(n{\ge}3\) table entries are Monte-Carlo and QMC
estimates, the closed form blocked at the single-gap CDF atom, which
order-stats resolves elementarily only through \(n{=}3\).

\subsubsection{Connectivity probability}

\(\mathcal J\) is connected iff \textbf{all \(n-1\) internal gaps are jammed}, so
\[\resizebox{\ifdim\width>\linewidth\linewidth\else\width\fi}{!}{$\displaystyle \Pr[\mathcal J\text{ connected}]=\Pr[g_1{<}1,\dots,g_{n-1}{<}1]
  =\!\!\sum_{(b_0,b_n)\in\{0,1\}^2}\!\!\Pr\big[\text{cell }(b_0,1,\dots,1,b_n)\big]
  =\int_{\Delta\cap\{g_i<1\,\forall\text{ int}\}}\! f_{\mathrm{UCP}}.$}\]

\refstepcounter{thmcnt}\label{prop:a5-vs-saturation}\textbf{Proposition\nobreakspace{}\thethmcnt{} (A5 vs.~saturation).} \(\{\text{all }n{+}1\text{ gaps}<1\}\subseteq\{\text{all internal}<1\}\), hence
\[\resizebox{\ifdim\width>\linewidth\linewidth\else\width\fi}{!}{$\displaystyle \Pr[\mathcal J\text{ connected}]\ \ge\ \Pr[N{=}n]=\textstyle\int_{J_n}f_{\mathrm{UCP}}\quad(\text{order-stats R15}),$}\]
with \textbf{strict} inequality whenever an end gap can exceed \(1\) while the interior jams (any \(s>n+1\)).
Connectivity needs the interior jammed; saturation additionally needs both ends --- a clean, provable
separation of ``graph-connected'' from ``process-jammed.''

\textbf{Worked example \(n{=}2\) (exact).} Single edge \(\Rightarrow\Pr[\text{conn}]=p_1=\mathbb E M_e\)
(closed form above); e.g.~\(s{=}6.5\!: 0.40308\).

\textbf{Computed at \(s{=}6.5\), \(n{=}2\ldots6\)} --- the inequality, and how it \emph{closes to equality} at
saturation:

{\footnotesize\begin{longtable}[]{@{}llll@{}}
\toprule\noalign{}
\(n\) & \(\Pr[\text{conn}]\) & \(\Pr[N{=}n]\) & conn\(/\)sat \\
\midrule\noalign{}
\endhead
\bottomrule\noalign{}
\endlastfoot
2 & 0.4031 & 0.000 & \(\infty\) \\
3 & 0.370 & 0.004 & \(\sim 100\times\) \\
4 & 0.681 & 0.409 & 1.66\(\times\) \\
5 & 0.981 & 0.964 & 1.02\(\times\) \\
6 & 1.000 & 1.000 & \(1\times\) (equality) \\
\end{longtable}}

\emph{The graph is a connected path \textbf{long before} the segment saturates (\(100\times\) at \(n{=}3\)); the gap
shrinks as \(n\to\) the jamming range and hits \textbf{equality at \(n{=}6\)}, where reaching depth \(6\) forces
both end gaps \(<1\) --- exactly the equality clause of the Proposition.} The inequality is proved for
all \(n,s\); the tabulated values are Monte-Carlo and QMC estimates.

\subsubsection{Isolated vertices}

By Theorem \ref{thm:linear-forest-reduction}(2), vertex \(v\) is isolated iff \(\deg_v{=}0\), i.e.~its incident internal bits vanish:
\[\resizebox{\ifdim\width>\linewidth\linewidth\else\width\fi}{!}{$\displaystyle \mathbb E[\#\text{iso}]=\Pr[b_1{=}0]+\Pr[b_{n-1}{=}0]+\sum_{v=2}^{n-1}\Pr[b_{v-1}{=}0,\,b_v{=}0],$}\]
the two ends contributing a single-bit marginal, each interior vertex an \textbf{adjacent-pair} marginal.

\textbf{Worked example \(n{=}2\) (exact).} Both vertices are ends: \(\#\text{iso}=2(1-b_1)\), so
\(\mathbb E[\#\text{iso}]=2(1-p_1)=2-2\,\mathbb E M_e\); at \(s{=}6.5\), \(1.1938\).

\textbf{Computed at \(s{=}6.5\):} \(\mathbb E[\#\text{iso}]=\) 1.192 \((n{=}2)\) · 0.770 \((n{=}3)\) ·
0.214 \((n{=}4)\) · 0.008 \((n{=}5)\) · 0.000 \((n{=}6)\) --- isolated vertices vanish as the path closes up.
The \(n{=}2\) value is exact; the \(n{\ge}3\) values are Monte-Carlo estimates (interior vertices need the
two-bit marginal atom, whose elementary form is closed only through \(n{=}3\)).

\subsubsection{Degree distribution and mean degree}

By Theorem \ref{thm:linear-forest-reduction}(2), \(\deg_v=b_{v-1}+b_v\in\{0,1,2\}\) (the linear-forest ceiling \(\Delta\le2\)). For an
interior \(v\): \(\Pr[\deg_v{=}2]=p_{v-1,v}\), \(\Pr[\deg_v{=}0]=\Pr[b_{v-1}{=}0,b_v{=}0]\); ends have
\(\deg\in\{0,1\}\). The \textbf{mean degree} of a uniformly random vertex is
\[\resizebox{\ifdim\width>\linewidth\linewidth\else\width\fi}{!}{$\displaystyle \bar d=\frac1n\sum_v\mathbb E\deg_v=\frac{2\,\mathbb E M_e}{n}\qquad(\text{handshake}).$}\]

\textbf{Degree PMF of a random vertex (computed, \(s{=}6.5\)):}

{\footnotesize\begin{longtable}[]{@{}
  >{\raggedright\arraybackslash}p{\dimexpr 0.2000\linewidth-2\tabcolsep\relax}
  >{\raggedright\arraybackslash}p{\dimexpr 0.2000\linewidth-2\tabcolsep\relax}
  >{\raggedright\arraybackslash}p{\dimexpr 0.2000\linewidth-2\tabcolsep\relax}
  >{\raggedright\arraybackslash}p{\dimexpr 0.2000\linewidth-2\tabcolsep\relax}
  >{\raggedright\arraybackslash}p{\dimexpr 0.2000\linewidth-2\tabcolsep\relax}@{}}
\toprule\noalign{}
\begin{minipage}[b]{\linewidth}\raggedright
\(n\)
\end{minipage} & \begin{minipage}[b]{\linewidth}\raggedright
\(\Pr[\deg{=}0]\)
\end{minipage} & \begin{minipage}[b]{\linewidth}\raggedright
\(\Pr[\deg{=}1]\)
\end{minipage} & \begin{minipage}[b]{\linewidth}\raggedright
\(\Pr[\deg{=}2]\)
\end{minipage} & \begin{minipage}[b]{\linewidth}\raggedright
\(\bar d=2\mathbb E M_e/n\)
\end{minipage} \\
\midrule\noalign{}
\endhead
\bottomrule\noalign{}
\endlastfoot
2 & 0.596 & 0.404 & 0 & 0.404 \\
3 & 0.257 & 0.620 & 0.123 & 0.867 \\
4 & 0.054 & 0.555 & 0.392 & 1.338 \\
5 & 0.002 & 0.405 & 0.594 & 1.592 \\
6 & 0 & 0.333 & 0.667 & 1.667 \\
\end{longtable}}

\emph{Mass moves \(\deg{=}0\to2\) as \(n\uparrow\); at \(n{=}6\) (\(=P_6\)) the two ends have \(\deg{=}1\) and the
four interior vertices \(\deg{=}2\), giving \(\bar d=2(n{-}1)/n=5/3\).}

\textbf{Worked example \(n{=}2\):} both vertices are ends, \(\deg{=}b_1\), so \(\Pr[\deg{=}1]=p_1\) and
\(\bar d=p_1=\mathbb E M_e\) (exact). The interior degree probabilities (\(\deg{=}2,0\)) need the two-bit
marginals and are Monte-Carlo estimates for \(n\ge3\).

\subsubsection{Largest component and component sizes}

By Theorem \ref{thm:linear-forest-reduction}(3,5) the size multiset is the run structure of \(\beta\); its law is the pushforward of
\(\Pr[\beta]\):
\[\resizebox{\ifdim\width>\linewidth\linewidth\else\width\fi}{!}{$\displaystyle \Pr[L_{\max}=\ell]=\!\!\sum_{\beta:\,1+\max\text{-run}(\beta)=\ell}\!\!\Pr[\beta],\qquad
  \Pr[K=k]=\!\!\sum_{\beta:\,n-\sum b_i=k}\!\!\Pr[\beta].$}\]
No independence is assumed --- RSA bits are correlated, so this is a genuine sum over the cell law,
not a Bernoulli-run formula.

\textbf{Worked example \(n{=}3\) (exact enumeration).} \(\beta\in\{00,01,10,11\}\) maps to component sizes
\(00\!\to\!\{1,1,1\}\), \(01,10\!\to\!\{2,1\}\), \(11\!\to\!\{3\}\). Hence
\[\resizebox{\ifdim\width>\linewidth\linewidth\else\width\fi}{!}{$\displaystyle \Pr[L_{\max}{=}1]=\Pr[00],\quad \Pr[L_{\max}{=}2]=\Pr[01]{+}\Pr[10],\quad \Pr[L_{\max}{=}3]=\Pr[11]=\Pr[\text{conn}].$}\]
At \(s{=}6.5\) (QMC cell law): \(\Pr[00]{=}0.071\), \(\Pr[\{01,10\}]{=}0.559\),
\(\Pr[11]{=}0.370\), giving \(\mathbb E L_{\max}=2.299\). The
\(01\!\approx\!10\) split is forced by reflection symmetry.

\textbf{Computed \(\mathbb E L_{\max}\) at \(s{=}6.5\):} 1.404 \((n{=}2)\) · 2.300 \((n{=}3)\) · 3.567 \((n{=}4)\) ·
4.969 \((n{=}5)\) · 6.000 \((n{=}6)\) --- \(L_{\max}\to n\) (the whole path) at saturation. These values are
QMC estimates of the cell law \(\Pr[\beta]\); the \(n{=}6\) endpoint is exact
(\(\mathcal J=P_6\)).

\textbf{Figure.} Left: component-count PMF \(\Pr[K{=}k]\) (A2 headline) for \(n{=}3,4,5\) at \(s{=}6.5\).
Right: \textbf{densification} at \(s{=}6.5\) across \(n{=}1\ldots6\) --- edge density \(M_e/(n{-}1)\),
\(\Pr[\text{conn}]\), and \(\Pr[N{=}n]\) all climb to \(1\) while \textbf{reach-depth collapses} (the \(n{=}5,6\)
near-saturation conditioning); the \(\Pr[\text{conn}]{-}\Pr[N{=}n]\) gap (A5) closes to \(0\) at \(n{=}6\).

\begin{figure}
\centering
\pandocbounded{\includegraphics[keepaspectratio,alt={jamming graph theorems}]{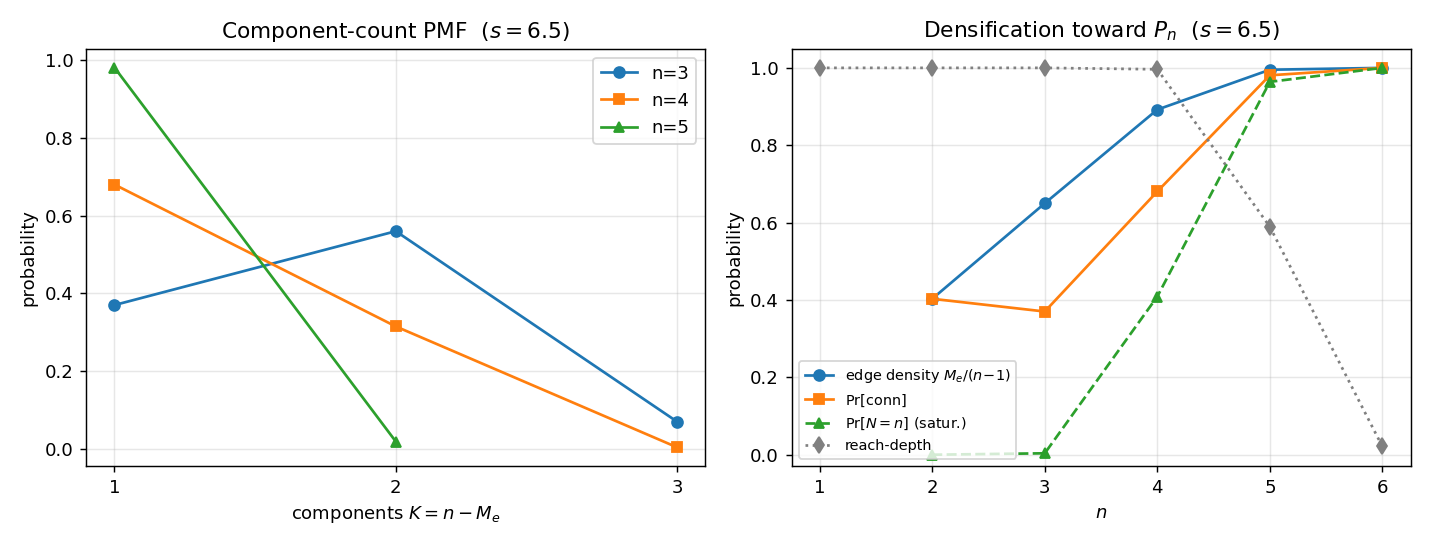}}
\caption{jamming graph theorems}
\end{figure}

\subsubsection{The independence, matching, and domination numbers are run-length functionals}

The independence, matching, and domination numbers are additive over the connected components of any graph, and on a path each has a classical closed form; on the fixed-\(n\) partial graph, Theorem \ref{thm:linear-forest-reduction} therefore turns each into a run-length functional of \(\beta\).

\begin{quote}
\refstepcounter{thmcnt}\label{prop:process-related-g2}\textbf{Proposition\nobreakspace{}\thethmcnt{}.} Write the components of \(\mathcal J\) as paths on vertex-counts \(m_1,\dots,m_K\) (so \(\sum_r m_r=n\) and \(m_r-1\) is the length of the \(r\)-th maximal jammed run). Then
\[\resizebox{\ifdim\width>\linewidth\linewidth\else\width\fi}{!}{$\displaystyle \alpha(\mathcal J)=\sum_{r}\Big\lceil\tfrac{m_r}{2}\Big\rceil,\qquad \nu(\mathcal J)=\sum_r\Big\lfloor\tfrac{m_r}{2}\Big\rfloor,\qquad \gamma(\mathcal J)=\sum_r\Big\lceil\tfrac{m_r}{3}\Big\rceil,$}\]
the independence, matching, and domination numbers. Consequently each expectation is a linear functional of the component-size law of A6, e.g.~\(\mathbb E[\alpha]=\sum_{m\ge1}\lceil m/2\rceil\,\mathbb E[C_m]\) where \(C_m\) is the number of size-\(m\) components.
\end{quote}

\emph{Proof.} Independence, matching, and domination numbers add over connected components. For a path \(P_m\) the values are the classical \(\lceil m/2\rceil\), \(\lfloor m/2\rfloor\), \(\lceil m/3\rceil\) (verified by brute force for \(m\le8\)). Theorem \ref{thm:linear-forest-reduction} supplies the path components; summation gives the totals, and linearity of expectation gives the laws from the component-size distribution (A6). \(\qquad\blacksquare\)

\refstepcounter{thmcnt}\label{rem:process-related-remark-16-1}\textbf{Remark\nobreakspace{}\thethmcnt{}.} \(\alpha+\nu=n\) (König on a bipartite graph, here per path: \(\lceil m/2\rceil+\lfloor m/2\rfloor=m\)), so the independence and matching laws are reflections of one another --- only one is independent data. The matching number \(\nu=\lfloor\) (run structure)\(\rfloor\) is the maximum number of disjoint jammed adjacencies, a natural ``pairing capacity'' of the saturated configuration.

\subsubsection{The spectrum is a mixture of path spectra over the component-size law}

On the fixed-\(n\) partial graph, where the unjammed gaps break the path, the spectrum of a disjoint union is the multiset union of the component spectra, and a path has an explicit spectrum; the spectral measure (``density of states'') of \(\mathcal J\) is therefore a mixture over the component-size law. On the saturated configuration every internal gap is \(<1\), so \(\mathcal J=P_n\) and the spectrum degenerates to \(\{2\cos\frac{j\pi}{n+1}\}\).

\begin{quote}
\refstepcounter{thmcnt}\label{prop:process-related-g3}\textbf{Proposition\nobreakspace{}\thethmcnt{}.} The adjacency spectrum of \(\mathcal J\) is the multiset
\[\resizebox{\ifdim\width>\linewidth\linewidth\else\width\fi}{!}{$\displaystyle \operatorname{spec}A(\mathcal J)=\biguplus_{r}\Big\{\,2\cos\tfrac{j\pi}{m_r+1}:j=1,\dots,m_r\Big\},$}\]
and the Laplacian spectrum is \(\biguplus_r\{\,2-2\cos\tfrac{k\pi}{m_r}:k=0,\dots,m_r-1\}\). The expected spectral measure at fixed \((s,n)\) is \(\mu_{s,n}=\tfrac1n\sum_{m\ge1}m\,\mathbb E[C_m]\,\hat\mu_{P_m}\), where \(\hat\mu_{P_m}\) is the uniform measure on the \(m\) path-eigenvalues and \(C_m\) the size-\(m\) component count (A6).
\end{quote}

\emph{Proof.} Block-diagonality of \(A(\mathcal J)\) and \(L(\mathcal J)\) over components gives the multiset union; the path eigenvalues are the standard ones, agreeing with \(2\cos\bigl(j\pi/(m{+}1)\bigr)\) to \(1.3\times10^{-15}\) for \(m\le6\); the Laplacian eigenvalues follow identically from the second-difference operator with Neumann ends). Averaging the empirical eigenvalue measure over the component-size law gives \(\mu_{s,n}\). \(\qquad\blacksquare\)

\refstepcounter{thmcnt}\label{rem:process-related-remark-17-1}\textbf{Remark\nobreakspace{}\thethmcnt{}.} Because \(\Delta\le2\), \(\operatorname{spec}A\subset[-2,2]\); the spectral gap and the number of eigenvalues at \(0\) (the adjacency nullity, \(=\#\{\)odd-size components\(\}\)) are again run-length functionals --- a spectral restatement of A6.

\paragraph{The component-size law, and how far the spectrum simplifies}

\textbf{The component-size law is geometric (bulk).} A size-\(m\) component is a maximal run of \(m-1\) jammed internal gaps (A6). In the bulk the internal jamming bits \(b_i=\mathbf 1\{g_i<1\}\) are \emph{asymptotically independent} --- UCP gaps decorrelate super-exponentially --- so the runs are geometric:
\[\resizebox{\ifdim\width>\linewidth\linewidth\else\width\fi}{!}{$\displaystyle \Pr[\text{component size}=m]\ \xrightarrow{\ \text{bulk}\ }\ (1-p)\,p^{\,m-1},\qquad p=\Pr[g_i<1],\quad \mathbb E[\text{size}]=\frac1{1-p}.$}\]

\textbf{The spectral measure then collapses to a one-parameter geometric family with closed diagnostics} --- but \emph{not} to a single smooth density. It is the geometric mixture \(\mu=\sum_m(1-p)p^{m-1}\hat\mu_{P_m}\), a pure-point delta-comb on \([-2,2]\). Its diagnostics are closed:
- \textbf{Nullity (states at \(E=0\)).} \(P_m\) has a zero eigenvalue iff \(m\) is odd, so the fraction of the spectrum at \(0\) is \(\dfrac{\#\{\text{odd components}\}}{n}\ \xrightarrow{\ \text{bulk}\ }\ \dfrac{1-p}{1+p}.\)
- \textbf{Support / moments.} \(\operatorname{spec}\subset[-2,2]\); the even moments \(\langle E^{2k}\rangle=\sum_m(1-p)p^{m-1}\tfrac1m\sum_j(2\cos\tfrac{j\pi}{m+1})^{2k}\) are geometric-weighted closed-walk counts.

The simplification is therefore partial: the law collapses to a one-parameter geometric family in \(p\) with closed-form nullity \((1-p)/(1+p)\), but not to a single elementary spectral density, since it stays a geometric delta-comb. A partial-UCP simulation at \(n{=}80\), \(s{=}160\) (\(\theta=0.5\), \(6000\) runs) measures \(p=0.616\) and component-size frequencies \(0.389,0.236,0.147,0.090,0.055,0.033\) for \(m=1,\dots,6\) against the geometric \(0.384,0.237,0.146,0.090,0.055,0.034\); the mean component size is \(2.55\) against \(1/(1-p)=2.60\), and the spectral nullity is \(0.244\) against \((1-p)/(1+p)=0.238\).

\textbf{Figure (G3 --- the spectrum is a delta-comb, not a density).} Adjacency spectrum of the jamming graph (partial UCP, \(\theta\approx0.46\), \(s=120\)), pooled over \(5000\) runs. Being a linear forest, \(\mathcal J\) has spectrum equal to the union of its path spectra \(2\cos\!\big(j\pi/(m{+}1)\big)\) --- a discrete comb with spikes at \(0,\pm1,\pm\sqrt2,\dots\), not a smooth density. The zero-eigenvalue (nullity) weight \(0.295\) matches the closed form \((1-p)/(1+p)=0.287\) (\(p=0.553\) the bulk edge probability).

\begin{figure}
\centering
\pandocbounded{\includegraphics[keepaspectratio,alt={jamming-graph spectrum delta-comb}]{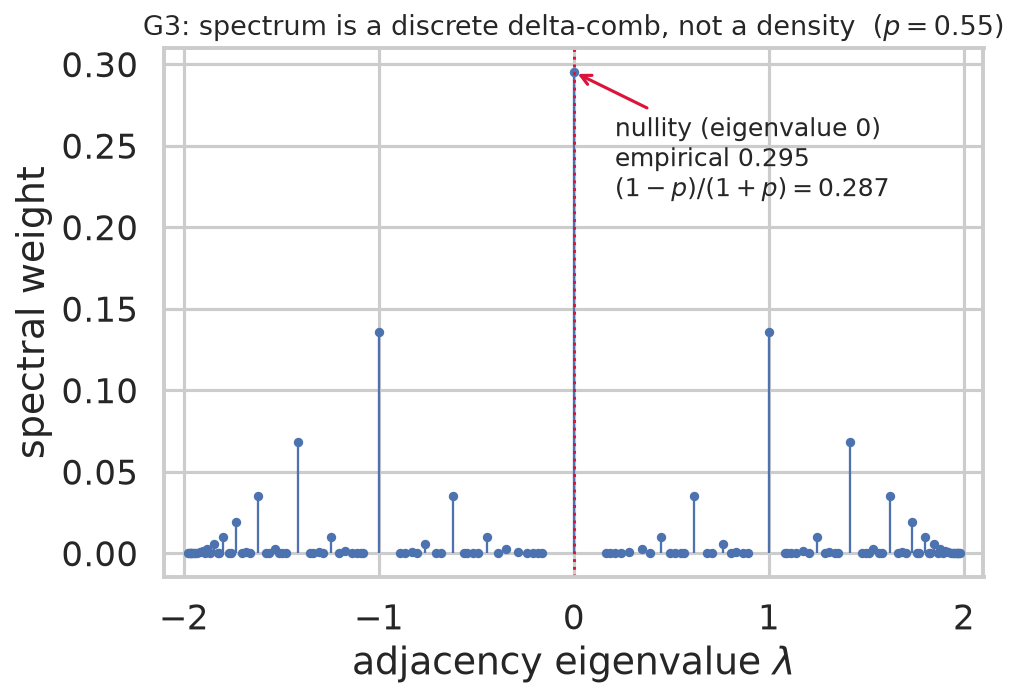}}
\caption{jamming-graph spectrum delta-comb}
\end{figure}

\subsubsection{Connectivity onset and the longest jammed run --- a percolation reading}

A4/A6 (largest component) and A5 (connectivity) together read as the percolation content of the jamming graph. Because the line is one-dimensional this content is \textbf{connectivity onset and longest-run scaling}, not a critical transition: bond/site percolation on a line has \(p_c=1\), and Proposition \ref{prop:process-related-g1} shows that \(\mathcal J\) is \(2\)-connected for no \(n\ge2\). The order parameter and the connectivity threshold both sit on the single boundary \(\varphi=1\).

\textbf{Order parameter.} The normalized longest jammed run \(L_{\max}/n\) (A4/A6) is the giant-component fraction. In the limit it is \(0\) for \(\varphi<1\) and jumps to \(1\) at \(\varphi=1\), where the whole path spans --- the degenerate one-dimensional onset. Finite-\(s\) this is the climb \(\mathbb E L_{\max}: 1.40\to6.00\) across \(n=2\ldots6\) at \(s=6.5\) (A4/A6), reaching \(L_{\max}=n\) at saturation.

\textbf{Connectivity threshold --- the random-geometric-graph reading.} Give each car a connection range \(r\) and join two cars within \(r\); the geometric graph \(\mathcal G_r\) is connected iff \(\max_i g_i<r\). A5 is the unit case \(r=1\) on the internal gaps. Such a proximity graph --- points on a line joined within a fixed range --- is a \textbf{unit interval (indifference) graph} (Roberts 1969), the claw-free subclass of the interval graphs of the object cell; its clique number is the largest number of cars in any window of length \(r\), with \(\chi=\omega\), and the jamming graph \(\mathcal J\) is its \(r=1\) restriction to internal gaps. Hence
\[\resizebox{\ifdim\width>\linewidth\linewidth\else\width\fi}{!}{$\displaystyle \Pr[\mathcal G_r\text{ connected}]=\Pr[\max_i g_i<r],$}\]
so the connectivity threshold is governed by the largest gap, and the max-gap extreme-value law (\ref{sec:asymptotic-properties} A5 / Cor. G′) supplies the onset directly:

\begin{itemize}
\tightlist
\item
  \textbf{jammed (\(\varphi=1\)):} \(\max_i g_i\) in the Weibull max-domain, right endpoint \(1\); the connectivity window is \(r\uparrow1\) of width \(O(1/n)\), with \(N(1-\max_ig_i)\Rightarrow\mathrm{Exp}(c)\).
\item
  \textbf{partial (\(\varphi<1\)):} \(\max_i g_i\) in the Gumbel max-domain with centring \(b_s=\lambda^{-1}\log(\theta s)\); the window sits on the log scale.
\end{itemize}

The Weibull\(\leftrightarrow\)Gumbel flip across \(\varphi=1\) is therefore the connectivity-onset flip: the same phase boundary, read through the graph.

\textbf{Obstacles.}
- \textbf{One dimension forces \(p_c=1\).} There is no interior critical point and no giant-component transition; the percolative statements are connectivity onset and longest-run scaling, and the text claims no more.
- \textbf{The percolation is correlated.} The RSA jamming bits \(\beta\) are correlated --- A4/A6 is a sum over the cell law, not a Bernoulli-run formula --- so the longest-run law is not the classical Erdős--Rényi i.i.d. longest-run. The finite-\(n\) law is exact through the cell law, but the asymptotic longest-jammed-run distribution under RSA correlation is open.

The connectivity threshold is proved: A5 holds for all \(n,s\), composed with the max-gap extreme-value
law (\ref{sec:asymptotic-properties} A5, proved modulo the constant \(c\); Cor. G′, proved). The order-parameter
onset follows from the cell law \(\Pr[\beta]\). The correlated longest-jammed-run asymptotics remain open.

\subsubsection{Additive graph functionals obey a stabilisation CLT}

The statement holds in both ensembles, at a fixed coverage \(\theta=n/s<c_{\mathrm R}\) below saturation and on the saturated process as \(s\to\infty\). The graph invariants of \(\mathcal J\) that are \textbf{sums over local structure} --- the edge count \(M_e=\sum_{i}b_i\) (jammed interior gaps), the component count \(K=n-M_e\), and the optimisation numbers \(\alpha,\nu,\gamma\) of G2 (sums of per-run functionals) --- obey a central limit theorem, and this is a \emph{general-\(n\)} asymptotic statement requiring no symbolic density.

\begin{quote}
\refstepcounter{thmcnt}\label{prop:clt}\textbf{Proposition\nobreakspace{}\thethmcnt{} (CLT).} Let \(H_n\) be any of \(M_e,\,K,\,\alpha,\,\nu,\,\gamma\) on the UCP with \(n\) cars, either at a fixed coverage below saturation or on the saturated process as \(s\to\infty\). Then \(H_n\) is a \textbf{stabilising} additive functional of the sequential-adsorption field --- each summand is determined by a \(O(1)\)-range neighbourhood, and correlations decay super-exponentially --- so
\[\resizebox{\ifdim\width>\linewidth\linewidth\else\width\fi}{!}{$\displaystyle \frac{H_n-\mathbb E[H_n]}{\sqrt{\operatorname{Var}H_n}}\;\xrightarrow[n\to\infty]{d}\;\mathcal N(0,1),\qquad \operatorname{Var}H_n=\Theta(n),$}\]
with a nondegenerate limiting variance density.
\end{quote}

\emph{Proof (structural).} Adjacency in \(\mathcal J\) is a local event (consecutive centres within distance \(2\)), and the saturated/partial RSA configuration has finite-range dependence: the state of a bounded window is determined, up to super-exponentially small probability, by the arrivals in a bounded enlargement of that window (the \emph{stabilisation} property of random sequential adsorption; Penrose 2001, Penrose--Yukich 2002). Each summand \(b_i\) (and each per-run term of \(\alpha,\nu,\gamma\)) is a bounded functional of such a window, so \(H_n\) is a sum of a stationary, finitely-dependent (mixing) array. The Penrose--Yukich stabilisation CLT for functionals of RSA then gives asymptotic normality with linear variance. \(\blacksquare\)

\textbf{Role.} The CLT makes the earlier \emph{exact finite} laws (A6, G2, G3) usable at scale: it says every run-length functional concentrates and is Gaussian in the bulk, and it is the mechanism behind the non-hyperuniform \(S(0)>0\) of G7 (linear number variance). It is derived from a \textbf{general} stabilisation theorem, not from any \(n{=}3,4\) computation. A Monte-Carlo check on the partial UCP at \(n{=}60\), \(s{=}120\) (\(\theta=0.5\), \(4000\) trials) gives \(\mathbb E M_e=36.38\) with standard deviation \(2.50\), standardised skewness \(+0.003\) and excess kurtosis \(+0.018\), both consistent with the Gaussian limit.

The exact finite void probability \(v([a,b])=\Pr[\text{no car centre in }[a,b]]\) is an \(n\)-dependent window integral of \(f_{\mathrm{UCP}}\), and the higher factorial moment densities \(\rho_k\) (\(k\ge3\)) are \(k\)-point marginals of it; both require the general-\(n\) symbolic marginals, and both are developed in G8′ and G9 below, where that requirement is the obstacle.

\subsection{From graph invariants to the underlying point process}

The invariants above are functionals of the jamming bits alone. The descriptors that follow are
functionals of the car positions themselves --- intensity, contact and nearest-neighbour distributions,
generating and Palm functionals, void probabilities, factorial moment densities, and the structure
factor --- and each reduces to a marginal or an integral of \(f_{\mathrm{UCP}}\), equivalently of the
Janossy densities \(j_n\) of the jammed process. The two layers meet at the single-gap integral
\(\Pr[g_i<1]\), at once the edge probability of \(\mathcal J\) and the contact probability of the point
process.

\subsubsection{Intensity}

On the fixed-\(n\) process the intensity is \(\lambda(s,n)=n/s\) points per unit length, which is the
coverage \(\theta\). Along the saturation diagonal \(n=N(s)\) it converges to the Rényi constant,
\(\lambda\to c_{\mathrm R}=0.747597\ldots\) as \(s\to\infty\) (Rényi 1958) --- the calibration constant,
classical, stated and not re-proved here.

\subsubsection{The pooled gap CDF is the contact distribution in gap coordinates}

The internal gaps measure the free space between consecutive cars, so their pooled CDF is the contact
distribution of the fixed-\(n\) configuration in gap coordinates; the nearest-neighbour distribution in
absolute coordinates is the distinct function \(G_{\mathrm{nn}}\) of G6. The pooled CDF is the UCP
single-gap CDF,
\[\resizebox{\ifdim\width>\linewidth\linewidth\else\width\fi}{!}{$\displaystyle G_{\mathrm{gap}}(r)=\frac1{n-1}\sum_{i=1}^{n-1}\Pr[g_i<r],\qquad r\in(0,1],\qquad G_{\mathrm{gap}}(1^-)=\frac{\mathbb E M_e}{n-1}.$}\]

\textbf{Worked example \(n{=}2\) (exact, general \(s\)).} From the edge-count density above,
\[\resizebox{\ifdim\width>\linewidth\linewidth\else\width\fi}{!}{$\displaystyle G_{\mathrm{gap}}(r)=\frac{2}{s-1}\Big(\frac{r(s-3)-r^2/2}{s-3}+r\ln\frac{s-2}{s-3}\Big),\quad 0\le r\le1.$}\]
valid for \(r\le s-3\) (bulk); for \(3<s<4\) and \(r>s-3\) add the \(\int_{s-3}^r\!\ln\frac{s-2}{u}\,du\) tail
term, so \(G_{\mathrm{gap}}(1^-)=p_1\)'s three-branch value. The function is \textbf{concave} (\(f_{g_1}\) decreasing) --- the
hard-core repulsion signature, opposite to the Poisson nearest-neighbour law \(1-e^{-\lambda r}\).
Values at \(s{=}6.5\): \(G_{\mathrm{gap}}(\tfrac12)=0.2145\) and \(G_{\mathrm{gap}}(1^-)=0.4031\), consistent with \((n{-}1)\,G_{\mathrm{gap}}(1^-)=\mathbb E M_e\).
The \(n{=}2\) contact distribution is exact for all \(s\).

\subsubsection{The PGFL of the jammed process is a jamming-PMF-weighted sum of Janossy integrals}

The object here is the \emph{jammed} point process, with \(N=N(s)\) random --- not the fixed-\(n\) graph of the preceding results. The jamming process is a \emph{finite} point process: it produces \(N\) car left endpoints \(X_1,\dots,X_N\) with \(N\) random (the jamming count) and, \textbf{conditioned on \(\{N=n\}\)}, the ordered centres have the all-jammed joint density of \ref{sec:joint-order-stats} restricted to the saturated cell \(J_n\). Writing \(j_n(x_1,\dots,x_n)\) for the (symmetric) \textbf{Janossy density} --- the sub-probability density of finding exactly \(n\) points, at those locations --- we have \(\int_{J_n} j_n = \Pr[N=n]\), the jamming PMF.

\begin{quote}
\refstepcounter{thmcnt}\label{prop:pgfl-master-identity}\textbf{Proposition\nobreakspace{}\thethmcnt{} (PGFL master identity).} For any measurable \(v:[0,s]\to[0,1]\),
\[\resizebox{\ifdim\width>\linewidth\linewidth\else\width\fi}{!}{$\displaystyle G[v]\;=\;\mathbb E\!\Big[\prod_{i=1}^{N} v(X_i)\Big]\;=\;\sum_{n\ge1}\;\frac{1}{n!}\int_{[0,s]^n} \Big(\prod_{i=1}^n v(x_i)\Big)\,j_n(x_1,\dots,x_n)\,dx ,$}\]
and the \textbf{Laplace functional} is \(L[f]=G[e^{-f}]=\sum_n \tfrac1{n!}\int e^{-\sum_i f(x_i)}\,j_n\,dx\).

\refstepcounter{thmcnt}\label{cor:campbell-first-moment}\textbf{Corollary\nobreakspace{}\thethmcnt{} (Campbell / first moment).} Differentiating at \(v\equiv1\) along \(v=1+\varepsilon h\) (equivalently \(f=\varepsilon h\) in \(L\)) gives the intensity \(\lambda\) of B1:
\[\resizebox{\ifdim\width>\linewidth\linewidth\else\width\fi}{!}{$\displaystyle \mathbb E\!\Big[\textstyle\sum_i h(X_i)\Big]=\int_0^s h(x)\,\lambda(x)\,dx,\qquad \lambda(x)=\sum_{n\ge1}\frac{1}{(n-1)!}\int j_n(x,x_2,\dots,x_n)\,dx_2\cdots dx_n .$}\]
In particular \(G[1]=1\) and \(\int_0^s\lambda=\mathbb E[N]\).
\end{quote}

\emph{Proof.} \(G[v]=\sum_n\Pr[N=n]\,\mathbb E[\prod_i v(X_i)\mid N=n]\); the conditional expectation integrates \(\prod v\) against the ordered density, and symmetrising the \(n!\) orderings replaces the ordered simplex integral by \(\tfrac1{n!}\) times the full-cube integral of the symmetric \(j_n\). \(L\) is the substitution \(v=e^{-f}\). The Campbell corollary is the first Gâteaux derivative: the \(n\)-th term contributes \(n\cdot\tfrac1{n!}\int h(x_1)j_n\,dx = \tfrac1{(n-1)!}\int h\,j_n\), whose \(x_1\)-marginal is \(\lambda\). \(\blacksquare\)

\textbf{Comparison.} For a Poisson process \(G[v]=\exp\!\int(v-1)\lambda\) (Haenggi §4); for Matérn hard-core only \emph{bounds} on the PGFL are known. Here the PGFL is an \textbf{exact} finite object --- a jamming-PMF-weighted sum of Janossy integrals --- for a genuinely non-Poisson (inhibitory, sequential) process. The identity is fully general in \(n\), and it is evaluated and cross-checked at \(n\le2\) --- the only regime with \(N\le2\), that is \(s\in(2,3)\), where \(j_1,j_2\) are elementary.

\textbf{Left-endpoint convention (used below).} Each car is recorded by its \textbf{left endpoint} \(X_i\) (occupying \([X_i,X_i{+}1]\)); a later car's left endpoint is excluded from \((X_i{-}1,\,X_i{+}1)\), so the feasible length for a second car given the first at \(a\) is \(\ell(a)=\max(0,a{-}1)+\max(0,s{-}2{-}a)\). The \textbf{symbolic} small-example values (\(n\le2\), \(s\in(2,3)\)) are the PMF \(\Pr[N{=}1]=\tfrac{3-s}{s-1}\), \(\Pr[N{=}2]=\tfrac{2(s-2)}{s-1}\); the PGFL at constant \(v=z\) (which is the PGF of \(N\)); the Laplace functional; \(\mathbb E[N]=\tfrac{3s-5}{s-1}\); and the intensity \(\lambda(x)=\dfrac{1+\ln\frac{s-2}{x}}{s-1}\) on \((0,s-2)\) --- a \textbf{weight-1 log}, exhibiting the contact singularity as \(x\to0^+\).

\subsubsection{The Palm law of the saturated process factorises at a fixed acceptance time}

On the saturated process, the reduced Palm law \(P^{!}_x\) is the law of the process seen from a typical point at \(x\), with that point deleted. For a Poisson process Slivnyak's theorem gives \(P^!_x=P\); for an inhibitory RSA process it differs, and for finite RSA it is not treated in the standard references. The sequential-screening structure of 1-D RSA gives it a \emph{conditional} product form.

\begin{quote}
\refstepcounter{thmcnt}\label{prop:conditional-palm-factorisation}\textbf{Proposition\nobreakspace{}\thethmcnt{} (conditional Palm factorisation).} Condition the saturated UCP on a car with \textbf{left endpoint} at an interior position \(x\) (occupying \([x,x+1]\subset(0,s)\)) and on its acceptance time \(\tau\). Then the reduced Palm law factorises into two independent saturated sub-UCPs on the flanking intervals:
\[\resizebox{\ifdim\width>\linewidth\linewidth\else\width\fi}{!}{$\displaystyle P^{!}_{x,\tau} \;=\; \mathrm{UCP}_\tau\big([0,x]\big)\;\otimes\;\mathrm{UCP}_\tau\big([x+1,s]\big).$}\]
The \emph{unconditional} reduced Palm law \(P^{!}_x\) marginalises \(\tau\) and is a \(\tau\)-\textbf{mixture} of these products, \textbf{not} a product.
\end{quote}

\emph{Proof (the splitting rule).} A car at \(x\) blocks its interval \([x,x+1]\); every later arrival falls strictly to the left or to the right of it, and none can straddle it. Given the acceptance time \(\tau\), the arrival streams on the two sides occupy disjoint position supports and, by the Markov property of the sequential deposition, run as independent RSA processes to saturation on their sub-intervals --- the unjammed-factorisation lemma of \ref{sec:joint-order-stats} (an empty interval splits RSA into independent sub-segments). Deleting the conditioning car leaves the product of the two saturated sub-UCPs; this is the reduced Palm law \emph{at fixed \(\tau\)}, \(P^{!}_{x,\tau}\) (the Mecke equation supplies the B1 intensity weight). \(\blacksquare\)

\textbf{The product form is conditional on \(\tau\), and saturated 1-D RSA is non-renewal.} The product holds only at fixed \(\tau\). The reduced Palm law marginalises \(\tau\), and both sides are conditioned on the same event that \([x,x{+}1]\) stays empty until \(\tau\), so the marginal is a \(\tau\)-\textbf{mixture} of products, not a product. The two flanking gaps are therefore positively correlated: a direct Monte-Carlo (saturated RSA, \(1.7\times10^{6}\) interior gap pairs) gives adjacent-gap \(\mathrm{Corr}(g_i,g_{i+1})=+0.087\), 95\% CI \([0.085,0.088]\), which excludes \(0\) (figure below; mean gap \(0.337=(1{-}c_{\mathrm R})/c_{\mathrm R}\) with \(c_{\mathrm R}\approx0.7476\)). Saturated 1-D RSA is non-renewal (Bonnier--Boyer--Viot 1994). The macroscopic diagnostic \(\mathrm{Corr}(N_L,N_R)\to0\) is insensitive to this --- car counts on macroscopic halves decorrelate even for a non-renewal process --- so it does not certify the local product form. The downstream renewal structure factor (G7.3) and the nearest-neighbour form (G6) accordingly rest on the gap-independence \textbf{approximation}, not on an exact factorisation.

\textbf{Interpretation.} G5 is the configuration-level lift of the count recursion \(N\stackrel{d}{=}1+N_L+N_R\) with \(N_L\perp N_R\) (Rényi): the split into independent sub-segments is exact given the acceptance time, and it is the marginalisation over \(\tau\) that reintroduces the local gap coupling.

\subsubsection{\texorpdfstring{The Palm factorisation is grand-canonical: it holds at every time \(t\), not only at saturation}{The Palm factorisation is grand-canonical: it holds at every time t, not only at saturation}}

Proposition \ref{prop:conditional-palm-factorisation} was stated for the saturated process. The natural question the regime picture raises is whether it survives at sub-saturation (\(\varphi<1\), voids present). At fixed \(n\) it does not hold, since fixing the total couples the two sides. The correct dividing line is not jammed-versus-partial but \textbf{\(N\) random versus \(N\) fixed}: the factorisation holds throughout the \emph{kinetic} (grand-canonical) trajectory and fails only under the fixed-count (canonical) constraint.

\begin{quote}
\textbf{Proposition \ref{prop:conditional-palm-factorisation}′ (kinetic Palm factorisation).} Run RSA on \([0,s]\) to time \(t\) under a rate-one space--time Poisson arrival stream, with \(N(t)\) \textbf{random} (the kinetic / grand-canonical ensemble; coverage \(\theta(t)<c_{\mathrm R}\), jamming fraction \(\varphi(t)<1\) for \(t<\infty\)). Then, \textbf{conditional on the acceptance time \(\tau\)} of the car at \(x\), the reduced Palm law factorises into two \textbf{independent} time-\(t\) kinetic RSA on the flanking intervals:
\[\resizebox{\ifdim\width>\linewidth\linewidth\else\width\fi}{!}{$\displaystyle P^{!}_{x,\tau} \;=\; \mathrm{RSA}_t\big([0,x]\big)\;\otimes\;\mathrm{RSA}_t\big([x{+}1,s]\big).$}\]
The \emph{unconditional} reduced Palm law \(P^{!}_x\) marginalises \(\tau\) and is a \(\tau\)-\textbf{mixture} of these products, \textbf{not} a product. Proposition \ref{prop:conditional-palm-factorisation} (saturation) is the case \(t\to\infty\).
\end{quote}

\emph{Proof.} The arrivals are a Poisson process \(\Pi\) on \([0,s{-}1]\times[0,t]\) (position \(\times\) time), and the configuration is the deterministic acceptance map \(\Phi_t(\Pi)\): process points in time order, accept \((y,\sigma)\) iff \([y,y{+}1]\) is disjoint from every earlier-accepted car. By Slivnyak's theorem the reduced Palm law of \(\Pi\) given a point at \((x,\tau)\) is \(\Pi\) itself (still Poisson) with the point \((x,\tau)\) adjoined; condition on that point being \textbf{accepted}, i.e.~\([x,x{+}1]\) empty in \(\Phi_{\tau^-}\).

\textbf{Every accepted car other than \(x\) has left endpoint in \([0,x{-}1]\cup[x{+}1,s{-}1]\) --- the two sides use disjoint position supports.} A car with left endpoint \(y\in(x{-}1,x{+}1)\) would overlap \([x,x{+}1]\): after \(\tau\) it is blocked by the car at \(x\) (the splitting rule); before \(\tau\) its acceptance would leave \([x,x{+}1]\) non-empty, contradicting the conditioning. So no such car is ever accepted.

\textbf{Neither the acceptance dynamics nor the conditioning couples the sides.} A left arrival \((y,\sigma)\), \(y\le x{-}1\), has \([y,y{+}1]\subseteq[0,x]\) and can overlap only cars with left endpoint in \((y{-}1,y{+}1)\subseteq[0,x)\) --- other left cars --- so its acceptance is a function of the left history alone; symmetrically on the right. The forbidden straddle arrivals split the same way: an arrival at \(y\in(x{-}1,x)\) has \([y,y{+}1]\) reaching only up to \(y{+}1<x{+}1\), so it can be blocked \textbf{only by a left car}, and its (mandatory) rejection is a function of the left history; an arrival at \(y\in(x,x{+}1)\) is blocked only by a right car. Hence the conditioning event factorises as \(\{\text{left history admits no straddle}\}\cap\{\text{right history admits no straddle}\}\).

The left configuration is therefore a measurable function of \(\Pi_L:=\Pi\cap([0,x{-}1]\times[0,t])\) and the right of \(\Pi_R:=\Pi\cap([x{+}1,s{-}1]\times[0,t])\); these are restrictions of a Poisson process to \textbf{disjoint} regions, hence independent, and each drives a kinetic RSA on its own interval to time \(t\) (with the hard wall at the split point automatic, since no car crosses it). Thus, at fixed \(\tau\), \(P^{!}_{x,\tau}=\mathrm{RSA}_t([0,x])\otimes\mathrm{RSA}_t([x{+}1,s])\). The reduced Palm law \(P^{!}_x\) is the \(\tau\)-marginal of this family, hence a mixture of products rather than a product. \(\blacksquare\)

\textbf{Where the fixed-\(n\) failure enters.} The proof uses only the independence of \(\Pi_L,\Pi_R\); it never invokes saturation. Conditioning on \(N=n\) (canonical) imposes \(N_L+N_R=n-1\), a global constraint that is \textbf{not} a function of \(\Pi_L\) and \(\Pi_R\) separately --- it re-couples the two Poisson restrictions --- which is exactly why the factorisation collapses there. In regime coordinates: G5′ holds all along the kinetic curve \((\theta,\varphi_{\mathrm{typ}}(\theta))\), \(0<\theta\le c_{\mathrm R}\); only the microcanonical (\(N\)-fixed) slice breaks it. The macroscopic diagnostic \(\mathrm{Corr}(N_L,N_R\mid x)\approx0\) does not certify even the conditional product: it is a macroscopic-count statistic, whereas the residual coupling lives in the local gaps. At saturation (\(t\to\infty\)) those gaps are non-renewal --- adjacent gaps are \(+0.087\) correlated (G5).

\textbf{Figure (adjacent-gap correlation at saturation).} Joint density of adjacent internal gaps \((g_i,g_{i+1})\) under saturated RSA (\(s=60\), \(4\times10^4\) runs, \(1.7\times10^6\) pairs). The positive tilt (regression slope \(+0.087\)) gives \(\mathrm{Corr}=0.087\), \(95\%\) CI \([0.085,0.088]\) --- the saturated gap sequence is \textbf{not} a renewal process. The macroscopic \(\mathrm{Corr}(N_L,N_R\mid x)\approx0\) diagnostic of G5′ misses this local coupling, so it does not certify a product form for the unconditional Palm law.

\begin{figure}
\centering
\pandocbounded{\includegraphics[keepaspectratio,alt={saturated adjacent-gap correlation}]{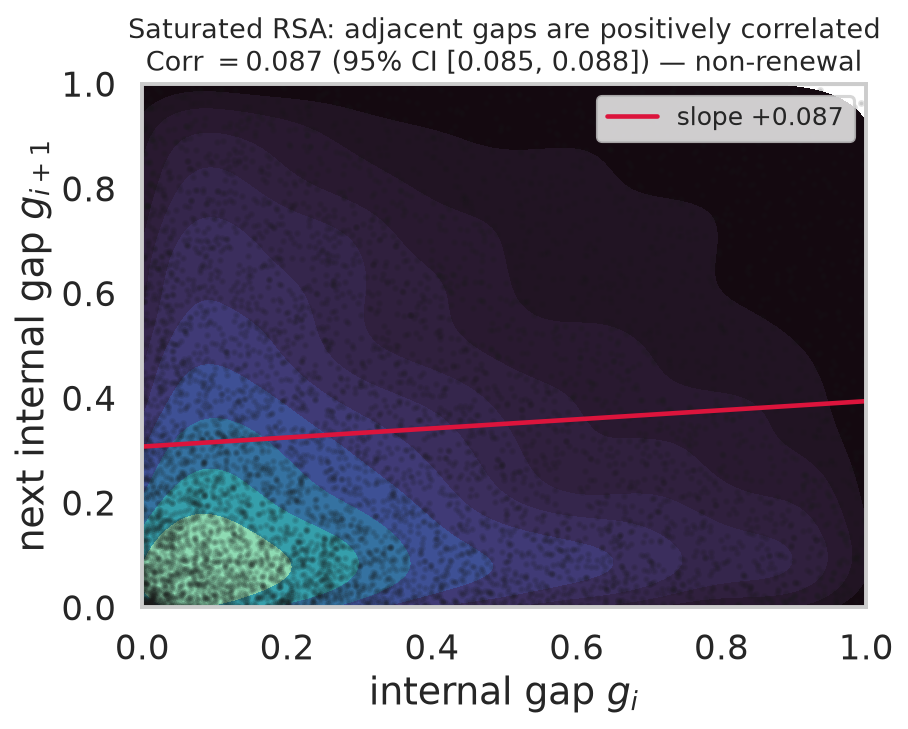}}
\caption{saturated adjacent-gap correlation}
\end{figure}

\subsubsection{\texorpdfstring{The factorial-moment densities and the pair correlation are marginals of \(f_{\mathrm{UCP}}\)}{The factorial-moment densities and the pair correlation are marginals of f\_\{\textbackslash mathrm\{UCP\}\}}}

On the \emph{jammed} process, where \(\rho_k\) is a marginal of the saturated Janossy density \(j_n\) with \(N\) random, the \(k\)-point factorial moment density \(\rho_k(x_1,\dots,x_k)\) --- the density of finding cars at \(k\) distinct locations --- is the \(k\)-point marginal of the Janossy densities:
\[\resizebox{\ifdim\width>\linewidth\linewidth\else\width\fi}{!}{$\displaystyle \rho_k(x_1,\dots,x_k)=\sum_{n\ge k}\frac1{(n-k)!}\int_{[0,s-1]^{\,n-k}} j_n(x_1,\dots,x_k,y_{k+1},\dots,y_n)\,dy .$}\]
This is a marginal of the known \(f_{\mathrm{UCP}}\); by order-stats weight-grading \(\rho_k\) is a \textbf{weight-\(\le n-k\)} hyperlogarithm with letters affine in \(x_1,\dots,x_k\).

\textbf{Worked example --- the pair density \(\rho_2\)} (\(n\le2\), \(s\in(2,3)\), one car left of the other): only \(N=2\) contributes, so \(\rho_2=j_2\), in closed form
\[\resizebox{\ifdim\width>\linewidth\linewidth\else\width\fi}{!}{$\displaystyle \rho_2(x,y)=\frac1{s-1}\Big(\frac1{s-2-x}+\frac1{y-1}\Big),\qquad 0<x<s-2<1<y<s-1,\ y\ge x+1,$}\]
(value \(3.556\) at \(s=\tfrac52,x=\tfrac14,y=\tfrac74\)); the pair correlation is \(g=\rho_2/(\lambda(x)\lambda(y))\) with \(\lambda\) from G4.

\textbf{Worked example at fixed \(n\) --- the pair correlation and the Ripley function (the B4 descriptor).}
On the fixed-\(n\) process the reduced second-order product density \(\rho^{(2)}\) of the left endpoints
factors, in gap coordinates, through the \textbf{joint two-gap-sum} marginal of \(f_{\mathrm{UCP}}\): for left
endpoints at separation \(u=\sum_{i\in[a,b)}(1+g_i)\), the pair correlation is
\(g(u)\propto\rho^{(2)}(u)/\lambda^2\). At \(n{=}2\) the only pair is \((X_{(1)},X_{(2)})\) with separation
\(u=X_{(2)}-X_{(1)}=1+g_1\), so the reduced two-point density is the internal-gap density shifted by the
hard core, \(f_U(u)=f_{g_1}(u-1)\) on \(u\in[1,s-1]\) (exact, all \(s\)). It decreases from contact --- the
hard-core signature --- and, unlike the \(s\to\infty\) saturated \(g(r)\) of Bonnier--Boyer--Viot, carries no
log-divergence at contact; that singularity is a saturation and large-\(n\) effect. At \(s{=}6.5\),
\(f_U(1.05)=0.4498\), \(f_U(2.0)=0.3511\) and \(\int_1^{s-1}f_U=1.00000\), and the \(n{=}3\) QMC mean pairwise
separation is \(2.4603\).

\textbf{Obstacle at fixed \(n\ge3\).} The reduction is exact, but the closed form needs the two-gap joint
marginal, which order-stats supplies symbolically only at \(n=3,4\); the full finite-\(n\) pair correlation
and Ripley function \(\mathcal K\) for \(n\ge3\) are open. The classical saturated \(g(r)\) of Bonnier--Boyer--Viot, with
its integrable log singularity at contact and its damped oscillations, is the \(s\to\infty\) check.

\textbf{Obstacle for \(k\ge3\).} Existence is not in question --- \(\rho_k\) is a marginal of the explicit \(f_{\mathrm{UCP}}\), hence a weight-\(\le n-k\) polylogarithm. Only the \emph{elementary evaluation} is hard: \(k\ge3\) needs the \(n\ge3\) symbolic densities (weight \(\ge2\)), the order-stats computation wall. So \textbf{derivable from \(f_{\mathrm{UCP}}\) in closed form for every \(k\); elementary only for \(k\le2\).}

\subsubsection{\texorpdfstring{G8′ --- the void and empty-space functions are PGFL integrals of \(f_{\mathrm{UCP}}\)}{G8′ --- the void and empty-space functions are PGFL integrals of f\_\{\textbackslash mathrm\{UCP\}\}}}

On the \emph{jammed} process, where \(j_n\) is the saturated Janossy density with \(N\) random, take \(f_{\mathrm{UCP}}\) (equivalently the Janossy densities \(j_n\)) as known; the void probability of a set \(B\subset[0,s]\) --- no car occupies \(B\) --- is a definite integral of \(j_n\) over the complement, summed over \(n\). In left-endpoint coordinates a car occupies \([x,x+1]\), so ``no car occupies \(B\)'' means no left endpoint lies in the exclusion window \(B^\ominus=\{x:[x,x+1]\cap B\neq\varnothing\}\); for \(B=[c,d]\) this is \(B^\ominus=(c-1,d)\). Then
\[\resizebox{\ifdim\width>\linewidth\linewidth\else\width\fi}{!}{$\displaystyle v(B)=\sum_{n\ge0}\frac1{n!}\int_{([0,s-1]\setminus B^\ominus)^n} j_n(x_1,\dots,x_n)\,dx \;=\; G\big[\,1-\mathbf 1_{B^\ominus}\,\big],$}\]
the PGFL at \(v=1-\mathbf 1_{B^\ominus}\) (G4). It is a \textbf{finite sum of integrals of the known weight-0 rational \(f_{\mathrm{UCP}}\)} over an explicit polytope --- a closed form for every \((n,s)\).

\textbf{Worked example} (\(s=\tfrac52\), \(B=[1.15,1.35]\)): the \(N=1\) term vanishes, the single jammed
car always occupying \(B^\ominus\); the \(N=2\) term gives \(v=0.0713\), matching MC to \(10^{-4}\).

\textbf{The empty-space function (the B5 descriptor).} At fixed \(n\) the same construction, read from a
uniform test point rather than from a set, gives the empty-space (spherical contact) function
\(F(r)=\Pr[\,\mathrm{dist}(U,\{X_{(i)}\})\le r\,]\) for \(U\sim\mathrm{Unif}[0,s]\) independent of
the configuration --- equivalently the expected fraction of the segment within \(r\) of some car left
endpoint,
\[\resizebox{\ifdim\width>\linewidth\linewidth\else\width\fi}{!}{$\displaystyle F(r)=\mathbb E\Big[\tfrac1s\,\big|\textstyle\bigcup_i[X_{(i)}-r,\,X_{(i)}+r]\cap[0,s]\big|\Big].$}\]
It is monotone with \(F(0){=}0\) and \(F(\infty){=}1\), and for small \(r\), before the balls overlap,
\(F(r)\approx 2nr/s\). The union-measure integral against \(f_{\mathrm{UCP}}\) at \(s{=}6.5\) gives
\(F(0.2)=0.1218\), \(F(0.5)=0.3000\), \(F(1.0)=0.5529\) at \(n{=}2\) (exact) and \(F(0.2)=0.1824\),
\(F(0.5)=0.4483\), \(F(1.0)=0.7631\) at \(n{=}3\) (QMC), against the small-\(r\) prediction \(2nr/s=0.1231\) at
\(n{=}2\), \(r{=}0.2\). The general-\(n\) finite void law is open --- the hardest descriptor
here, Rintoul--Torquato covering the infinite-coverage limit only ---.

\subsubsection{The structure factor: exclusion hole, non-hyperuniformity, renewal spectrum}

Parts 1 and 2 below (the exclusion hole and non-hyperuniformity) hold for any configuration; part 3 (the renewal spectrum) is the \emph{jammed} \(s\to\infty\) limit. The structure factor \(S(q)=1+\lambda\!\int_{\mathbb R}\big(g(r)-1\big)e^{-iqr}\,dr\) (Bartlett spectrum; DVJ §8) is the diffraction observable of the packing, with \(\lambda\) the bulk intensity and \(g\) the pair correlation of G9. Three of its features are pinned --- the first two structurally for all \(n\), the third by the renewal approximation to G5, the saturated gap sequence being non-renewal with adjacent-gap \(\mathrm{Corr}=+0.087\).

\begin{quote}
\refstepcounter{thmcnt}\label{prop:process-related-g7}\textbf{Proposition\nobreakspace{}\thethmcnt{}.}
1. \textbf{(Exclusion sum rule.)} Hard-core exclusion forces \(g\equiv0\) on \([0,1)\), contributing the model-independent hole \(-\,2\lambda\,\dfrac{\sin q}{q}\) to \(S(q)\).
2. \textbf{(Non-hyperuniformity.)} \(S(0)=\mathrm{Var}(d)/\mathbb E[d]^2>0\), where \(d\) is the spacing: the number variance grows \emph{linearly} (G8), so RSA is \textbf{not} hyperuniform.
3. \textbf{(Renewal spectrum --- approximate.)} In the renewal approximation to the saturated process (G5's factorisation is only conditional; the true saturated gaps are \(+0.087\) correlated), with \(\hat f_d(q)=\int f_d(x)e^{-iqx}dx\) the Fourier transform of the spacing density \(f_d\),
\[\resizebox{\ifdim\width>\linewidth\linewidth\else\width\fi}{!}{$\displaystyle S(q)=\frac{1-\lvert\hat f_d(q)\rvert^{2}}{\lvert 1-\hat f_d(q)\rvert^{2}}.$}\]
\end{quote}

\emph{Proof.} (1) split the defining integral at \(r=1\) with \(\int_0^1\cos qr\,dr=\sin q/q\); the guaranteed hole \(g\equiv0\) on \([0,1)\) gives the stated term. (2) linear number variance is the G8 CLT, and for a renewal process \(S(0)=\mathrm{Var}(d)/\mathbb E[d]^2\), positive since the spacing is non-degenerate. (3) a stationary renewal process with interval density \(f_d\) has the stated Bartlett spectrum (DVJ, Ex. 8.2(e)). \(\qquad\blacksquare\)

\paragraph{\texorpdfstring{The spacing law \(f_d\), expanded}{The spacing law f\_d, expanded}}

The spacing density is \(f_d(d)=f_G(d-1)\), with \(f_G\) the saturated nearest-neighbour \textbf{gap} density on \([0,1)\) (a gap \(\ge1\) cannot survive at jamming). It is monotone decreasing from a finite contact value to a positive endpointmeasures it:
- \(\mathbb E[\text{gap}]\approx0.337\), so \(\mathbb E[\text{spacing}]=1/\lambda\approx1.337\) with \(\lambda\approx0.7476=c_{\mathrm R}\), the Rényi constant recovered as \(1/\mathbb E[d]\);
- contact value \(f_G(0^+)\approx2.66\), endpoint \(f_G(1^-)\approx0.43\) --- the histogram estimate of the A5 max-gap EVT constant \(c=f_G(1^-)\approx0.422\);
- \(\mathrm{Var}(d)\approx0.079\), giving the non-hyperuniformity constant \(S(0)=\mathrm{Var}(d)/\mathbb E[d]^2\approx0.044\).

The exact closed form of \(f_G\) is the Rényi--Dvoretzky--Robbins nearest-neighbour law --- a \(\varphi\)-kernel (exponential-period) object, not elementary. But the renewal spectrum needs only its transform \(\hat f_d\), and the two model-independent facts (the \(-2\lambda\sin q/q\) hole and \(S(0)>0\)) need no closed form at all.

\subsubsection{\texorpdfstring{Hard-core exclusion forces \(J\ge1\), so UCP is inhibitory at the unit scale}{Hard-core exclusion forces J\textbackslash ge1, so UCP is inhibitory at the unit scale}}

The bound rests only on hard-core exclusion, so it holds on the fixed-\(n\) partial configuration and on the jammed process alike. The \(J\)-function \(J(r)=\dfrac{1-G_{\mathrm{nn}}(r)}{1-F(r)}\) (van Lieshout--Baddeley; Haenggi §2) is by construction a comparison to Poisson: it contrasts the \textbf{nearest-neighbour} distribution \(G_{\mathrm{nn}}(r)\) (distance from a \emph{typical car} to its nearest other car) with the \textbf{empty-space} distribution \(F(r)\) (distance from a \emph{typical location} to the nearest car), and it equals \(1\) \textbf{iff} the process looks Poisson at scale \(r\) (Slivnyak), exceeds \(1\) for inhibitory/regular processes, and is \(<1\) for clustered ones. So the Poisson value \(J\equiv1\) is the reference the statistic is \emph{defined against} --- that is why we contrast with it.

\textbf{Algorithm --- compute \(J\) from the earlier results.} Both \(F\) and \(G_{\mathrm{nn}}\) follow from the spacing law:
1. Let \(f_d\) be the car--car spacing density (the spacing law of G7, supported on \([1,\infty)\)), \(F_d(r)=\int_1^r f_d\) its CDF, and \(\lambda=1/\mathbb E[\text{spacing}]\) the intensity (B1).
2. \textbf{\(G_{\mathrm{nn}}(r)\)}: a typical car's two adjacent spacings are its neighbouring intervals and its nearest neighbour is the smaller, so \(G_{\mathrm{nn}}(r)=1-\big(1-F_d(r)\big)^2\) \textbf{under the renewal (independent-adjacent-spacing) approximation} --- in particular \(G_{\mathrm{nn}}\equiv0\) on \([0,1)\) by hard-core exclusion, which is exact. The product form overstates independence, adjacent gaps being \(+0.087\) correlated (G5), so this \(G_{\mathrm{nn}}\) is an approximation; the \(J\ge1\) inhibition conclusion, which uses hard-core exclusion only, stands.
3. \textbf{\(F(r)\)}: the empty-space function of the spacing law, \(1-F(r)=\lambda\int_{2r}^{\infty}(d-2r)\,f_d(d)\,dd\) (the length-biased chance that the interval covering a typical location extends \(\ge r\) on each side).
4. \textbf{\(J(r)=\dfrac{1-G_{\mathrm{nn}}(r)}{1-F(r)}\)}; on \([0,1)\) the numerator is \(1\), so \(J(r)=1/(1-F(r))\).

\begin{quote}
\refstepcounter{thmcnt}\label{thm:process-related-g6}\textbf{Theorem\nobreakspace{}\thethmcnt{}.} In any saturated or partial UCP, two distinct car reference points are \(\ge1\) apart, so \(G_{\mathrm{nn}}(r)=0\) for \(r\in[0,1)\) and hence
\[\resizebox{\ifdim\width>\linewidth\linewidth\else\width\fi}{!}{$\displaystyle J(r)=\frac{1}{1-F(r)}\ \ge\ 1\qquad(0\le r<1),$}\]
strictly \(>1\) wherever \(F(r)>0\). UCP is therefore \textbf{inhibitory} at the hard-core scale, uniformly in \(n\).
\end{quote}

\emph{Proof.} Hard-core exclusion makes the nearest-neighbour distance of a typical car \(\ge1\) a.s., so \(G_{\mathrm{nn}}\equiv0\) on \([0,1)\) and the numerator of \(J\) is \(1\). \(F\) is a distribution function with \(F(0)=0\) and \(F(r)>0\) for \(r>0\), so \(1-F(r)\in(0,1]\) and \(J=1/(1-F)\ge1\), strictly once \(F>0\). \(\qquad\blacksquare\)

\textbf{Worked example.} On a bulk partial UCP at coverage \(\theta=0.4\) (\(n{=}48\), \(s{=}120\)), \(G_{\mathrm{nn}}\equiv0\) on \([0,1)\) and rises past \(r=1\) (reaching \(0.245\) at \(r=1.25\)), \(F\) rises smoothly from \(0\) through \(F(0.25)=0.193\), \(F(0.5)=0.402\), \(F(0.75)=0.571\), \(F(0.99)=0.696\), and \(J\) climbs from \(1.000\) at \(r=0\) through \(1.239\), \(1.672\), \(2.329\) to \(3.288\) at \(r=0.99\), staying above the Poisson line \(J\equiv1\) throughout \([0,1)\).

\textbf{Figure (G6 --- hard-core inhibition).} The \(J\)-function \(J(r)=(1-G_{\mathrm{nn}})/(1-F)\) for a dilute RSA snapshot (\(\theta\approx0.38\), \(s=80\); centres are \(\ge1\) apart, so \(G_{\mathrm{nn}}\equiv0\) on \([0,1)\) and \(J=1/(1-F)\)). \(J\) climbs from \(1\) to \(\approx3\), staying above the Poisson reference \(J\equiv1\) throughout --- the signature of inhibition. Under saturation every void is sub-unit, so \(F\to1\) and \(J\to\infty\) as \(r\to1^-\) (extreme inhibition).

\begin{figure}
\centering
\pandocbounded{\includegraphics[keepaspectratio,alt={J-function inhibition}]{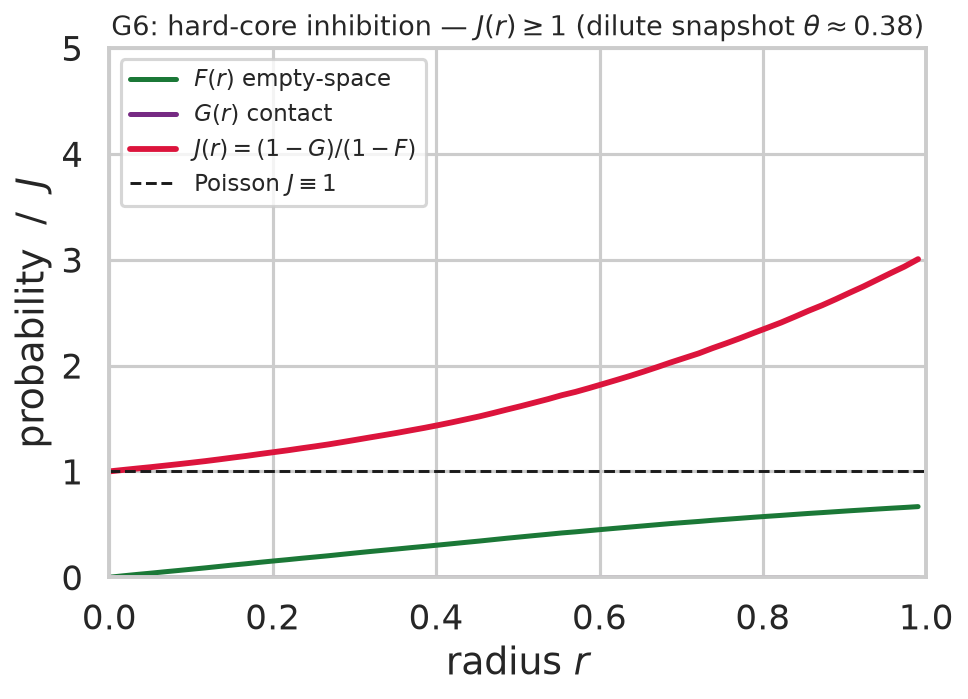}}
\caption{J-function inhibition}
\end{figure}

\subsubsection{Prior Art Cross-Check}

The descriptor \textbf{names} here are classical; each has a canonical owner in the \textbf{infinite-line / fixed-
coverage} regime, whereas this section treats the \textbf{finite-\((n,s)\), conditioned-on-\(n\)} forms on the \textbf{RSA} point process (a repulsive, exactly-\(n\) process) plus the linear-forest/jam-word reduction (Thm 0).

\begin{itemize}
\tightlist
\item
  \textbf{Pair correlation \(g(r)\)} --- saturated \(s\to\infty\) form: \textbf{Bonnier--Boyer--Viot, \emph{J. Phys. A} 27 (1994)
  3671} (log singularity at contact, super-exponential decay). Our finite-\(n\) \(f_U\) is new; do \textbf{not}
  present it in the \(s\to\infty\) limit or it reproduces theirs.
\item
  \textbf{Void \(F(r)\) / contact \(G_{\mathrm{gap}}(r)\) / nearest-neighbour \(G_{\mathrm{nn}}(r)\) functions} --- 1D RSA at all coverages:
  \textbf{Rintoul--Torquato, \emph{Phys. Rev.~E} 53 (1996) 450} (\(H_P,H_V\)). Ours is the finite-\(n\) conditional version.
\item
  \textbf{Graph descriptors} (component count, connectivity, isolated vertices, largest component) in 1D --- the
  RGG toolkit is \textbf{Penrose, \emph{Random Geometric Graphs} (OUP 2003)} and the 1D-RGG literature (arXiv:1008.5140,
  2105.07731), but for \textbf{Poisson/i.i.d.} points. Running them on the \textbf{RSA jamming} configuration is new.
\item
  \textbf{Finite-line saturation-count PMF} --- \textbf{Iwasa--Fukuda (2008, arXiv:0810.5632)} give one, but it is the
  Tonks/equilibrium count; our jamming-count PMF \emph{corrects} it (see the jamming-PMF section). Cite explicitly.
\item
  \textbf{Count mean/variance asymptotics} --- \textbf{Coffman--Flatto--Jelenković--Poonen, \emph{Algorithmica} 22 (1998) 448}
  (large-\(x\), not conditioned-on-\(n\); the moment backdrop).
\item
  \textbf{PGFL / Palm / factorial-moment machinery} --- the general apparatus is \textbf{Daley--Vere-Jones} and
  \textbf{Last--Penrose}; for hard-core processes only bounds on the PGFL are standard (Haenggi §4), and
  the finite Janossy/PGFL forms of G4, G8′ and G9 are exact.
\item
  \textbf{RSA \emph{on} a graph} (arXiv:1604.03918 etc.) is the \textbf{inverse} construction (process on a given graph) ---
  distinguish: we build a graph \emph{on} the RSA output.
\end{itemize}

\textbf{Not found in the literature:} the finite-\((n,s)\) conditional descriptors, the RSA input process for the RGG toolkit, and
the linear-forest reduction tying the two layers (no prior art found for the last).

\section{Markov chain and aggregates}
\label{sec:markov-chain-and-aggregates}

\subsection{Aggregate quantities as integral equations}

\label{sec:aggregate-representations}

\emph{--- the opening subsection. We fix the
strong-Markov split, tabulate the dictionary assigning each aggregate quantity its integral-equation type,
and prove the coproduct proposition (Proposition \ref{prop:coproduct-and-coupling}) from which the dictionary follows. The Hopf-algebraic
machinery is quoted from \ref{sec:background-hopf}; the per-row equations are developed in §§2--5.}

\subsubsection{The saturated process and the strong-Markov split}

Fix a segment of length \(s\) and run the uniform car-parking process to saturation: \(N\) unit cars attach
in random sequential order, leaving \(N+1\) gaps \(G_0,\dots,G_N\) with \(\sum_i G_i=s-N\) and every \(G_i<1\) (each gap \emph{capped},
admitting no further car).
The \textbf{aggregate quantities} are the laws and expectations determined by the whole saturated
configuration: the count \(N\), an additive functional \(\sum_i h(G_i)\), the largest gap \(\max_i G_i\), an
order statistic, and a joint density.

Every equation below follows from a single decomposition. The first car attaches with its left endpoint uniform on \([0,s-1]\);
conditioned on its position \(U\), the sub-segments \([0,U]\) and \([U+1,s]\) fill to saturation as independent
copies of the process on lengths \(U\) and \(s-1-U\). This is the \textbf{strong-Markov split}, and its base case
is saturation itself: a segment shorter than a car admits no attachment and is terminal. The split
is at a fixed index, but the independence of the two sub-segments is the sub-interval factorisation of
\ref{sec:background-ucp} Proposition \ref{prop:sub-interval-factorisation}, which restarts the chain at the optional times of entry into each gap.

Averaging over \(U\) is \textbf{first-step analysis} for the attachment chain: an aggregate of the whole
configuration equals its first-car contribution plus the same aggregate on the two sub-segments, and taking
expectations turns that identity into an equation in \(s\). Each aggregate therefore yields a second-kind
Volterra equation or hierarchy --- linear or quadratic, per §§2--5 --- whose solution operator is the resolvent of
the attachment kernel. No section below introduces a new probabilistic mechanism; what varies from one
aggregate to the next is how the observable behaves under the split, and that is what the dictionary
tabulates.

The reduction to a single variable is not automatic, and it is what the rest of the section rests on. The
state of the attachment chain is a configuration, not a number; the equations below are equations in the
segment length because the chain's state collapses onto that length. The next result states exactly when
that collapse is available.

Fix the class of sequential hard-core interval processes on an interval, with independent arrivals of
density \(\varphi\) and cars of length \(\ell\), an arrival being accepted when it overlaps no placed car.

\begin{quote}
\refstepcounter{thmcnt}\label{thm:rigidity-of-the-attachment-kernel}\textbf{Theorem\nobreakspace{}\thethmcnt{} (rigidity of the attachment kernel).}
\textbf{(i)} The law of the process on \([a,b]\) depends on \([a,b]\) only through \(b-a\), for every \([a,b]\), if and
only if \(\varphi(x)=Ae^{\lambda x}\) for some \(A>0\) and \(\lambda\in\mathbb R\). Every aggregate then satisfies
an equation in the length alone, with delay \(\ell\).
\textbf{(ii)} The kernel of those equations is the density \(e^{\lambda u}\) on \([0,s-\ell]\), normalised. It is
constant if and only if \(\lambda=0\), that is, if and only if \(\varphi\) is uniform.
\textbf{(iii)} The mean count alone does not detect the difference: it is an equation in the length alone for
every \textbf{affine} \(\varphi\) as well, and there it takes the same value as for the uniform density.
\end{quote}

\emph{Proof.} \textbf{(i)} Suppose the law depends only on the length. The first accepted car has density
\(\varphi|_{[a,b-\ell]}\) normalised, so the hypothesis requires \(\varphi(a+u)=c(a)\psi(u)\) for a fixed \(\psi\)
and every \(a\); putting \(u=0\) gives \(c(a)=\varphi(a)/\psi(0)\), hence
\(\varphi(a+u)\psi(0)=\varphi(a)\psi(u)\). That is Cauchy's multiplicative functional equation, whose
measurable positive solutions are the exponentials. Conversely, if \(\varphi(x)=Ae^{\lambda x}\) then
\(\varphi(x+h)=e^{\lambda h}\varphi(x)\), so the normalised density on a translated segment is unchanged and
the process on \([a,b]\) is the translate of the process on \([0,b-a]\).

\textbf{(ii)} Under (i) the first-car density in the length variable \(u=x-a\) is \(e^{\lambda u}\) normalised on
\([0,s-\ell]\), which is constant exactly when \(\lambda=0\).

\textbf{(iii)} Let \(m\) denote the uniform mean and argue by induction on the length, the base case being
segments shorter than \(\ell\), where the mean is \(0\). Suppose \(M(a,x)=m(x-a)\) for every segment shorter than
\([a,b]\). Writing \(L=b-a\), the first-step decomposition gives
\[\resizebox{\ifdim\width>\linewidth\linewidth\else\width\fi}{!}{$\displaystyle M(a,b)=1+\frac{\int_a^{b-\ell}\varphi(x)\,g(x)\,dx}{\int_a^{b-\ell}\varphi(x)\,dx},
\qquad g(x)=m(x-a)+m(b-\ell-x).$}\]
The function \(g\) is symmetric about the midpoint \(c_0=(a+b-\ell)/2\) of the domain of integration. For
\(\varphi(x)=\alpha+\beta x\) that symmetry gives \(\int\varphi g=(\alpha+\beta c_0)\int g\) and
\(\int\varphi=(\alpha+\beta c_0)(L-\ell)\), so the ratio is \(\bigl(\int g\bigr)/(L-\ell)\), which is the uniform
answer; the affine part cancels. \(\square\)

Clause (iii) is the reason the theorem is stated about the kernel. A rigidity claim phrased on the mean is
false: the mean cannot separate an affine density from a uniform one, so it cannot characterise the process.
The kernel can, and by (ii) the flat kernel \(1/(s-1)\) with unit delay --- Rényi's equation --- singles out the
uniform car-parking process inside the class. Verified numerically: exponential densities are length-only with a
non-uniform kernel, affine densities reproduce the uniform mean to \(5\times10^{-14}\), and quadratic and
periodic densities are not length-only at all.

\subsubsection{Representability is regeneration}

Rényi's equation is not an isolated device. The first car divides the segment into two
sub-segments that fill independently, each running the same process at a smaller length, so a
saturated configuration on \([0,s]\) is assembled from saturated configurations on shorter segments:
the process regenerates in the length variable. Every equation in this section is an average of
that decomposition over the first car's position, and the question of which aggregates satisfy
such an equation is the question of which aggregates the decomposition can carry.

Call an aggregate \textbf{split-compatible} if its value on \([0,s]\) is determined by its values on the
two sub-segments together with the first-car datum. Split-compatibility is what a finite integral
equation records, and it sorts the aggregates of this section into three regimes.

\begin{quote}
\textbf{Principle (regeneration and representability).} Let \(F\) be an aggregate of the saturated
process.
1. If \(F\) is split-compatible with an \textbf{additive} composition law, it is a primitive element and
satisfies a single second-kind \emph{linear} Volterra equation in \(s\) (§2).
2. If \(F\) is split-compatible with a \textbf{multiplicative} composition law, it is group-like and
satisfies a \emph{quadratic} Volterra equation, the parking Dyson--Schwinger equation (§3).
3. If \(F\) is not split-compatible in the length variable alone but becomes so once the state is
\textbf{augmented by a finite mark} --- the car count, or a rank --- then the augmented aggregate
satisfies an equation of type (1) or (2), and \(F\) is recovered from it by coefficient
extraction; the recovered family solves a triangular hierarchy (§§3--5).
\end{quote}

The third regime is the working part of the statement. Fixing a count or a rank destroys
regeneration in the length variable, because the constraint must be allocated across the split
before either sub-segment can be evaluated; marking the count by a formal variable restores it,
since the allocation becomes a convolution the mark can carry. Augmentation is therefore not a
technical convenience but the precise repair for a broken regeneration, and the size of the mark
is what the resulting equation costs.

The same reading identifies the boundary. An aggregate that is split-compatible under \emph{no} finite
augmentation has no finite equation to satisfy, and three of the section's negative results are
instances of one mechanism.

\begin{itemize}
\tightlist
\item
  The \textbf{fixed-\((s,n)\) joint density} is not an aggregate of a regenerating family: freezing the
  count severs the object at \(s\) from the same object at smaller lengths, since the sub-segments
  carry counts that are not \(n\). Its over-all-\(n\) counterpart \(\rho_k\) regenerates and does satisfy
  a hierarchy (§5), which is why the two behave differently.
\item
  A \textbf{jam-word pattern constraint} couples the two sides at the split point, so the mark that would
  restore compatibility must carry the whole pattern rather than a bounded index.
\item
  A \textbf{count-weighting} \(\mathbb E[a_N]\) propagates its weight through the split-convolution only if
  the weight sequence can be advanced by finitely many auxiliary states, which is exactly the
  condition that \((a_n)\) satisfy a linear recurrence with polynomial coefficients. The P-recursive
  dichotomy of §4 is the augmentation criterion in the count variable.
\end{itemize}

In the thermodynamic limit the mechanism fails at its source: an infinite translation-invariant
configuration has no distinguished first car, the decomposition is unavailable, and no augmentation
recovers it. What replaces the finite-length hierarchy there is the BBGKY hierarchy, which reaches
upward to \(g_{k+1}\) instead of downward to \(\rho_{<k}\) (§5).

The forward direction of the Principle is proved constructively in §§2--5, one regime per section.
The converse --- that an aggregate admitting no finite augmentation admits no finite equation --- is
open, and rests on the same order-bound obstacle as the converse of Conjecture \ref{conj:representability-dichotomy}.

\subsubsection{The representability dictionary}

The Principle sorts the aggregates by how they meet the split; the Hopf language supplies names for the three
regimes, since an additive law is a primitive element and a multiplicative law a group-like one
(\ref{sec:background-hopf}). The table below records, for each aggregate, which regime it falls
in and the equation that follows. The generating-function rows below mark cars by a formal variable \(z\). The map is developed row by row in
§§2--5; the machinery (shuffle Hopf algebra, FQSym, antipode, Faà-di-Bruno, linear reducibility) is quoted
from \ref{sec:background-hopf}.

{\footnotesize\begin{longtable}[]{@{}
  >{\raggedright\arraybackslash}p{\dimexpr 0.2500\linewidth-2\tabcolsep\relax}
  >{\raggedright\arraybackslash}p{\dimexpr 0.2500\linewidth-2\tabcolsep\relax}
  >{\raggedright\arraybackslash}p{\dimexpr 0.2500\linewidth-2\tabcolsep\relax}
  >{\raggedright\arraybackslash}p{\dimexpr 0.2500\linewidth-2\tabcolsep\relax}@{}}
\toprule\noalign{}
\begin{minipage}[b]{\linewidth}\raggedright
aggregate quantity
\end{minipage} & \begin{minipage}[b]{\linewidth}\raggedright
Hopf object
\end{minipage} & \begin{minipage}[b]{\linewidth}\raggedright
integral equation
\end{minipage} & \begin{minipage}[b]{\linewidth}\raggedright
§, status
\end{minipage} \\
\midrule\noalign{}
\endhead
\bottomrule\noalign{}
\endlastfoot
additive \(\sum_i h(G_i)\) & primitive element & linear Volterra (2nd kind, unit delay) & §2, proved \\
count PGF \(\mathbb E[z^N]\) & group-like element & quadratic Volterra (Dyson--Schwinger) & §3, proved \\
max-gap CDF \(\Pr[\max_i G_i\le x]\) & group-like element & quadratic Volterra & §3, proved \\
factorial moments \(M_k=\partial_z^k\mathbb E[z^N]\vert_{1}\) & Faà-di-Bruno coordinates & finite linear cascade & §3, proved \\
single order statistic \(X_{(k)}\) & marginalised character & triangular linear Volterra hierarchy & §4, proved \\
finite \(k\)-variable joint density & graded component of the character & hyperlogarithm (linear reducibility) & §4, proved \\
count-weighting \(\mathbb E[a_N]\) & character evaluation & P-recursive \((a_n)\Rightarrow\) finite linear system & §4, converse conjectural \\
full \(n\)-variable joint, \(n\to\infty\) & the character & no finite equation & §5, open (P5) \\
\end{longtable}}

\subsubsection{Jamming cells and the jammed--unjammed coupling}

Each aggregate is defined on the \textbf{jamming cells} of the configuration space. A partial configuration is
recorded by a jamming bitstring, with one bit per gap marking whether the gap is capped. Saturation is the
all-ones cell, in which every gap is capped. This section develops the \textbf{jammed} regime (every
gap capped), whose aggregates satisfy the equations below. The same strong-Markov split covers two regimes
not developed here: the \textbf{unjammed} kinetic process (some gap still admits a car), whose coverage evolves
by a rate equation, and the intermediate case of exactly \(k\) capped gaps, which couples the two through the
allocation of the count across the split. Each integral equation below is an instance of the split,
uniformly across the count, the additive functionals, the order statistics, and the multi-point densities.

\subsubsection{The first-car coproduct and the coupling dichotomy}

The linear and quadratic types in the dictionary both follow from the coproduct of the first-car split.
Under the split an aggregate either factorises over the two sub-segments, or a count or rank is fixed; the
factorising case splits further into the primitive and group-like elements of the coproduct.

\begin{quote}
\refstepcounter{thmcnt}\label{prop:coproduct-and-coupling}\textbf{Proposition\nobreakspace{}\thethmcnt{} (coproduct and coupling).} Let \(F(s)\) be an aggregate of the saturated process.
1. If \(F\) \textbf{factorises} over the two sub-segments, it satisfies a single second-kind Volterra equation in
\(s\): linear when \(F\) is a primitive element, quadratic when \(F\) is group-like.
2. If a \textbf{count or rank is fixed}, the two sub-segments are coupled by a convolution over the allocation
of that index. The count-marked generating function satisfies a quadratic Volterra equation whose
fixed-index coefficients solve a triangular linear Volterra hierarchy, from which \(F\) is recovered by
coefficient extraction.
\end{quote}

\emph{Proof.} Write \(N_L,N_R\) for the car counts of the two sub-segments and \(F_L,F_R\) for the aggregate
restricted to each. Condition on the first car's position \(U\); the gaps and cars of the full configuration
each belong to exactly one saturated sub-segment. An additive functional (a \emph{score} summed over cars or
gaps) is a primitive element, \(\Delta F=F\otimes1+1\otimes F\), so, writing \(f_1(s)\) for the aggregate's value from
the first car alone, it averages as
\(F(s)=f_1(s)+\tfrac{1}{s-1}\int_0^{s-1}\big(F_L(u)+F_R(s-1-u)\big)\,du\). The reflection \(F_R=F_L\) collapses
this to the second-kind linear Volterra equation \(F(s)=f_1(s)+\tfrac{2}{s-1}\int_0^{s-1}F(u)\,du\) with unit
delay. A multiplicative score is group-like, \(\Delta F=F\otimes F\), and the same average replaces
\(F_L+F_R\) by \(F_L F_R\), giving the quadratic \(F(s)=\tfrac{1}{s-1}\int_0^{s-1}F(u)F(s-1-u)\,du\), of which the
max-gap distribution is the clean instance; the count generating function carries an extra first-car factor
\(z\), giving the Dyson--Schwinger equation of §3. When the
count is fixed at \(N=k\), mark cars by the variable \(z\); extracting the \([z^k]\)-coefficient imposes
\(N_L+N_R=k-1\) (since \(N=1+N_L+N_R\)), a convolution over the two sub-segments, and the fixed-index law is that
coefficient of the count-marked equation. \(\qquad\square\)

The factorising instances are the additive mean, a primitive element (§2, linear), and the count and
max-gap generating functions, group-like elements (§3, quadratic); the fixed-count and order-statistic laws
are the coupled hierarchies (§§3--4).

\textbf{Example (additive mean).} Take \(F(s)=\mathbb E[N(s)]\), an additive mean, hence a primitive element
with first-car value \(f_1\equiv1\). Proposition \ref{prop:coproduct-and-coupling}(1) returns Rényi's mean equation, Theorem \ref{thm:parking-integral-equation-renyi-1958} of \ref{sec:background-aggregate},
whose solution has slope the Rényi parking constant recorded there. The group-like \(\mathbb E[z^N]\) and the
rank-fixed \(X_{(k)}\) instantiate the quadratic and coupled cases of parts (1)--(2), developed in §§3--4; the
\(k\)-th order statistic, with \(k-1\) cars to its left, is the \([z^{k-1}]\)-coefficient of the count-marked
forcing.

\subsection{Additivity: the linear integral equation and its solution space}

\label{sec:aggregate-additivity}

\emph{An additive functional of the
saturated uniform car-parking process satisfies a single second-kind linear Volterra integral equation. We
state that equation, construct its resolvent, and resolve its solution space by the car count \(N\).}

\subsubsection{Additive functionals and the strong-Markov split}

An additive functional is split-compatible with an additive composition law, so it falls in the first regime of the Principle (§1): the split carries it in the length variable alone, and the equation it satisfies is linear.

Recall the strong-Markov split (§1): the first car attaches with its left endpoint uniform on \([0,s-1]\),
and conditioned on its position the two sub-segments fill to saturation as independent copies of the
process, with car counts \(N_L\) and \(N_R\). An \textbf{additive functional} sums a local score over the cars or
the gaps. The count \(N\) scores one per car; a \textbf{gap sum} \(\sum_{i=0}^{N} h(G_i)\) scores a bounded
function \(h\) of each gap, with the total gap \(\sum_i G_i=s-N\) at \(h(x)=x\) and the gap energy
\(\sum_i G_i^2\) at \(h(x)=x^2\). Under the split, a gap sum decomposes as
\(\sum_i h(G_i)=\sum_L h+\sum_R h\) and the count as \(N=1+N_L+N_R\).

\subsubsection{The additive integral equation}

Every additive functional's mean satisfies the same linear equation; only the forcing and the boundary
datum change with the functional. An additive functional is specified by a per-car forcing \(f\) (its mean
first-car score) and a bounded gap score \(h\) (its boundary datum on a single gap): the count has
\((f,h)=(1,0)\), and a pure gap sum has \((f,h)=(0,h)\).

\begin{quote}
\refstepcounter{thmcnt}\label{thm:additive-functionals-satisfy-a-linear}\textbf{Theorem\nobreakspace{}\thethmcnt{} (additive functionals satisfy a linear Volterra equation).} Let \(A(s)\) be the mean of the
additive functional with data \((f,h)\). Then
\[\resizebox{\ifdim\width>\linewidth\linewidth\else\width\fi}{!}{$\displaystyle A(s)=f(s)+\frac{2}{s-1}\int_0^{s-1}A(u)\,du\quad(s\ge1),\qquad A(s)=h(s)\ \text{on}\ [0,1),$}\]
a second-kind linear Volterra equation with unit delay.
\end{quote}

\emph{Proof.} The strong-Markov split writes the functional as its first-car contribution plus independent
copies on the two sub-segments, so \(A(s)=f(s)+\mathbb E[A(U)]+\mathbb E[A(s-1-U)]\). With \(U\) uniform on
\([0,s-1]\) the reflection \(u\mapsto s-1-u\) maps the second sub-segment term onto the first, so the two
integrals coincide and sum to \(\tfrac{2}{s-1}\int_0^{s-1}A(u)\,du\). On \([0,1)\) no car attaches; the segment
is a single gap of length \(s\), whence \(A(s)=h(s)\). \(\qquad\square\)

The count \((f,h)=(1,0)\) returns Rényi's mean equation (Theorem \ref{thm:parking-integral-equation-renyi-1958},
\ref{sec:background-aggregate}). The kernel \(K(s,u)=\tfrac{2}{s-1}\mathbf 1\{u\le s-1\}\) is the same for every additive functional; only the
forcing \(f\) and the boundary datum \(h\) change, so one resolvent gives the solution for the whole class.

\emph{Remark 1 (Hopf-algebraic reading).} The mean equation is the derivative at \(z=1\) of the count
Dyson--Schwinger equation of §3; in the Hopf-algebra formulation of \ref{sec:background-hopf}, an
additive functional is a primitive element (\ref{prop:coproduct-and-coupling}).

\subsubsection{The resolvent}

Existence and uniqueness are standard for a second-kind Volterra equation with a locally integrable kernel:
the operator \(\mathcal K\) is quasi-nilpotent, so \(I-\mathcal K\) is invertible (with \(I\) the identity),
with resolvent \(\mathcal R=\sum_{r\ge1}K_r\) (Gripenberg--Londen--Staffans
1990; Brunner 2004). With the unit delay this becomes a finite-\(s\) statement. The iterated kernels form a
ladder of increasing \emph{transcendental weight} --- the number of nested integrations over the delay pole
\(\mathrm dt/(t-1)\); the functions on the ladder are the \emph{hyperlogarithms}, iterated integrals over
\(\mathrm dt/(t-1)\) (Brown 2009 on linear reducibility; \ref{sec:background-hopf}).

\begin{quote}
\refstepcounter{thmcnt}\label{thm:resolvent-of-the-additive-equation}\textbf{Theorem\nobreakspace{}\thethmcnt{} (resolvent of the additive equation).} The iterated kernels of \(\mathcal K\) are
\[\resizebox{\ifdim\width>\linewidth\linewidth\else\width\fi}{!}{$\displaystyle K_1(s,u)=\frac{2}{s-1},\quad K_2(s,u)=\frac{4}{s-1}\ln\frac{s-2}{u},\quad
  K_3(s,u)=\frac{8}{s-1}\!\int_{u+2}^{s-1}\!\frac{\ln\frac{t-2}{u}}{t-1}\,dt,\ \dots,$}\]
with \(K_r\) supported on \(u\le s-r\) and of transcendental weight \(r-1\). On each interval \([\ell,\ell+1)\)
the resolvent \(\mathcal R(s,u)=\sum_r K_r(s,u)\) is a finite sum over \(r\le\ell\).
\end{quote}

\emph{Proof.} The composition \(K_r(s,u)=\int_{u+r-1}^{s-1}\tfrac{2}{s-1}K_{r-1}(t,u)\,dt\) integrates a
weight-\((r-2)\) kernel supported on \(u\le t-(r-1)\), so \(K_r\) is supported on \(u\le s-r\); the single
integration over the pole of \(1/(t-1)\) raises the transcendental weight by one, to \(r-1\): rational,
logarithm, dilogarithm, and so on up the ladder (Brown reduction; \ref{sec:background-hopf}). For
\(s<\ell+1\) the support condition forces \(K_r\equiv0\) once \(r\ge\ell+1\), so the sum is
finite. \(\qquad\square\)

The script reconstructs the Rényi mean \(M_1\) from the resolvent interval by
interval and confirms the local finiteness of Theorem \ref{thm:resolvent-of-the-additive-equation}.

\subsubsection{The solution space, resolved by the count}

The asymptotics of \(A(s)\) as \(s\to\infty\) are extensive, governed by a single rate constant; the finite-\(s\)
solution decomposes into per-count modes. Marking the count by \(z\) turns the additive functional into a generating
function,
\[\resizebox{\ifdim\width>\linewidth\linewidth\else\width\fi}{!}{$\displaystyle \Psi(s,z)=\mathbb E\Big[z^N\sum_i h(G_i)\Big],$}\]
coupled to the count generating function \(g(s,z)=\mathbb E[z^N]\) (§3).

Additive gap functionals of this process are classical. Ney (1962) studies the number of terminal gaps of
length at least \(b\), an additive functional of the same kind, and obtains the asymptotics of its moments by
Laplace transform in the segment length. What the count marking adds is the resolution below: the
functional splits into modes indexed by \(N\), each supported on a bounded window of \(s\).

\begin{quote}
\refstepcounter{thmcnt}\label{thm:solution-space-by-count}\textbf{Theorem\nobreakspace{}\thethmcnt{} (solution space by count).} \(\Psi\) satisfies the linear Volterra equation
\[\resizebox{\ifdim\width>\linewidth\linewidth\else\width\fi}{!}{$\displaystyle (s-1)\,\Psi(s,z)=2z\int_0^{s-1}\Psi(u,z)\,g(s-1-u,z)\,du,\qquad \Psi(s,z)=h(s)\ \text{on}\ [0,1),$}\]
and the additive functional resolves into per-count modes
\[\resizebox{\ifdim\width>\linewidth\linewidth\else\width\fi}{!}{$\displaystyle A(s)=\sum_n \Phi_n(s),\qquad \Phi_n(s)=[z^n]\Psi(s,z)=\mathbb E\Big[\sum_i h(G_i)\,\mathbf 1\{N=n\}\Big].$}\]
Each mode is supported on the window \(s\in[n,2n+1)\), the range on which \(N=n\) has positive probability.
At any \(s\) only finitely many modes are active, and each \(\Phi_n\) is a piecewise hyperlogarithm.
\end{quote}

\emph{Proof.} The split gives \(z^N\sum_i h(G_i)=z^{1+N_L+N_R}\big(\sum_L h+\sum_R h\big)\); taking expectations
with the sub-segments independent given \(U\), one factor carries the gap sum and the other the count
generating function, and the reflection symmetry doubles the single integral. Extracting \([z^n]\) recovers
\(\mathbb E[\sum_i h(G_i)\mathbf 1\{N=n\}]\). The count \(N(s)\) takes values in
\(\{\lfloor(s-1)/2\rfloor+1,\dots,\lfloor s\rfloor\}\), so \(N=n\) has positive probability exactly for
\(s\in[n,2n+1)\), outside which \(\Phi_n\equiv0\). Extracting \([z^n]\) from the resolvent series of the
\(z\)-graded kernel \(2z\,g(s-1-u,z)/(s-1)\) yields, on each unit sub-interval \([\ell,\ell+1)\), a finite sum of
iterated integrals over \(\mathrm dt/(t-1)\) of weight at most \(n\). The count coefficients \([z^n]g=p_n(s)\)
are themselves piecewise hyperlogarithms (the count case \((f,h)=(1,0)\) of §3), so the \(g\)-convolution
preserves the class (Theorem \ref{thm:resolvent-of-the-additive-equation}, applied coordinate-wise in the count grading); hence \(\Phi_n\) is a piecewise
hyperlogarithm. \(\qquad\square\)

\textbf{Example (gap energy \(h(x)=x^2\)).} The first two modes are elementary,
\[\resizebox{\ifdim\width>\linewidth\linewidth\else\width\fi}{!}{$\displaystyle \Phi_0(s)=s^2\ \text{on}\ [0,1),\qquad \Phi_1(s)=\tfrac23(s-1)^2\ \text{on}\ [1,2),$}\]
and higher modes gain one weight per interval. The script integrates
\(\Psi\) for \(h(x)=x^2\), extracts the modes \(\Phi_n\), matches them against direct simulation, confirms the
finite support, and checks \(\sum_n\Phi_n=\mathbb E[\sum_i G_i^2]\) (4/4 checks green).

\textbf{Figure 1.} The per-count modes \(\Phi_n(s)=\mathbb E[\sum_i G_i^2\,\mathbf 1\{N=n\}]\) for the gap energy
\(h(x)=x^2\), from simulation. Each mode is supported on its window \([n,2n+1)\) (Theorem \ref{thm:solution-space-by-count}) and neighbouring
modes overlap only there; their sum is \(\mathbb E[\sum_i G_i^2]\). Generated.

\begin{figure}
\centering
\pandocbounded{\includegraphics[keepaspectratio,alt={per-count modes}]{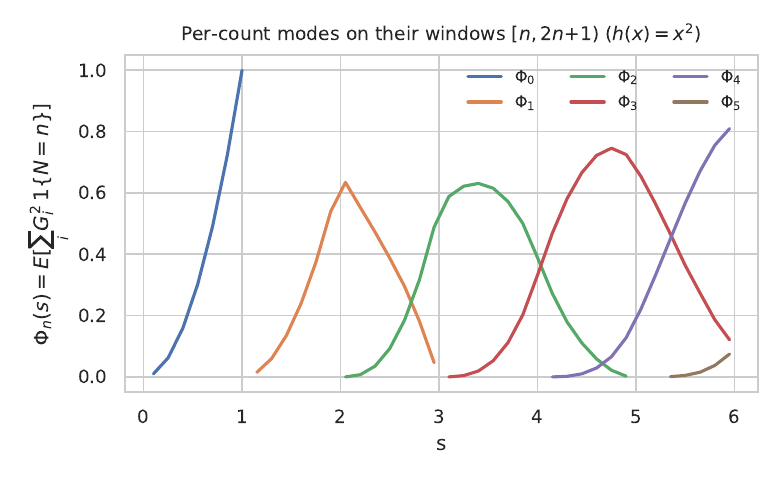}}
\caption{per-count modes}
\end{figure}

\subsection{Order-statistic marginals and count-weightings: the coupled and character branches}

\label{sec:aggregate-order-stats-ie}

\emph{Two branches of the representation
dictionary (Proposition \ref{prop:coproduct-and-coupling}, §1) are developed. Theorem \ref{thm:rigidity-of-the-attachment-kernel} treats the \textbf{coupled} branch: a single spatial
order statistic \(X_{(k)}\) has a marginal density satisfying a second-kind linear Volterra equation coupled,
through the count probability mass function, to the lower ranks, the equations forming a triangular
hierarchy solved rank by rank; the joint density of the full vector is its multi-variable companion, a
hyperlogarithm by linear reducibility (Proposition \ref{prop:joint-order-statistic-density}). Theorem \ref{thm:representability-character-case} treats the \textbf{character} branch: a count-weighting \(\mathbb E[a_N]\)
reduces to the count generating function when the weight sequence is a finite combination of geometric
sequences; a P-recursive dichotomy is conjectured. }

\subsubsection{Rank as the coupled case}

Fixing a rank destroys split-compatibility in the length variable, since the rank must be allocated across the split before either side can be evaluated. This section is the third regime of the Principle (§1): the state is augmented, by a rank in Theorem \ref{thm:order-statistic-hierarchy} and by the weight sequence's own recurrence in Theorem \ref{thm:representability-character-case}, and the equations are recovered by coefficient extraction.

Fix a rank \(k\) and let \(X_{(k)}\) be the position of the \(k\)-th car from the left in the saturated
configuration, with marginal density \(f_{X_{(k)}}(\,\cdot\,;s)\) on the segment \([0,s]\). Unlike the count and the
max gap, a rank is not a group-like mark: conditioning on the rank of a car ties the two sub-segments
together (the order statistics are dependent; David--Nagaraja 2003), since the \(k\)-th order statistic of the whole is the \(j\)-th of the left sub-segment precisely when
the left holds \(j-1\) earlier cars. This is the coupled branch of Proposition \ref{prop:coproduct-and-coupling}, giving not one equation
but a triangular system indexed by rank.

\subsubsection{The hierarchy}

\begin{quote}
\refstepcounter{thmcnt}\label{thm:order-statistic-hierarchy}\textbf{Theorem\nobreakspace{}\thethmcnt{} (order-statistic hierarchy).} For each rank \(k\ge1\) the marginal density
\(f_{X_{(k)}}(\,\cdot\,;s)\) satisfies a second-kind linear Volterra equation in \(s\) whose forcing is built
from the count probability mass function \(p_n(s)=\Pr[N=n]\) and the lower-rank marginals. Then (i) the
equations for \(k=1,2,\dots\) form a system triangular in the rank, solved rank by rank; and (ii) for each
integer \(\ell\ge1\) and \(s\in[\ell,\ell+1)\), \(f_{X_{(k)}}(\,\cdot\,;s)\) is a piecewise hyperlogarithm of
transcendental weight at most \(\ell-1\).
\end{quote}

\emph{Proof.} Condition the configuration on the first car's position \(U\) and on the number \(j-1\) of cars that
the left sub-segment \([0,U]\) receives; the \(k\)-th car of the whole is then the first car when \(j=k\) with the
left holding \(k-1\) cars, and otherwise the \((k-j)\)-th car of the right sub-segment. Average over \(U\)
uniform on \([0,s-1]\) and over the allocation \(j\); this gives a homogeneous term in the same marginal on a
shorter segment, a boundary term supplying the count mass \(p_{k-1}\) of the left sub-segment, and a convolution term
in the lower-rank marginals \(f_{X_{(k-j)}}\). Together these form a second-kind linear Volterra equation
forced by strictly lower ranks. Triangularity gives the rank-by-rank solution; the weight bound follows from
the coproduct raising the hyperlogarithm weight by one per integration (Chen series over the integer delay alphabet). \(\qquad\square\)

The count mass in the forcing is the \([z^{k-1}]\)-coefficient of the count generating function \(g(s,z)\) of
§3, so the hierarchy is the fixed-rank extraction of the group-like equation, as Proposition \ref{prop:coproduct-and-coupling}
requires. The kernel is the locally integrable kernel of §2, so a general rank \(X_{(k)}\) differs from the
boundary rank \(X_{(1)}\) only in its forcing, the count mass \(p_{k-1}\) and the lower ranks, not in its
kernel. The resolvent therefore solves each row and gives the unique continuous solution: every row is a
second-kind linear Volterra equation with a locally integrable kernel, so Theorem \ref{thm:existence-uniqueness-and-the-resolvent} of
\ref{sec:background-aggregate} applies verbatim, and the quasi-nilpotency of the Volterra operator
makes the resolvent series converge on every bounded interval. Existence and uniqueness for the whole
hierarchy follow rank by rank, since row \(k\) has the same kernel and a forcing already determined.

\textbf{Computation, and the boundary of the method.} The rows are solved numerically as written, and the
joint density they come from is evaluated by the dynamic programme of
\ref{sec:order-stats-cells}, at \(O(2^n n)\) by the last-car subset recursion. What makes
all of this available is the split, and the split is a property of a \emph{bounded} segment: the first car
is a distinguished car. In the thermodynamic limit no car is distinguished, the triangular structure is
lost, and the spatial gap sequence is not a Markov chain, the measured connected three-gap cumulant is
nonzero, so the correlation functions there obey a Kirkwood--Salsburg-type hierarchy that reaches
upward to \(g_{k+1}\) and does not truncate. Section \ref{sec:aggregate-boundary} develops that
contrast; the finite-\(s\) theory of this section is the regime in which the hierarchy closes.

\subsubsection{The leftmost order statistic}

The rank-one case is explicit. The leftmost order statistic \(X_{(1)}\) is the position of the first car from
the left, equivalently the boundary gap \(G_0\), and its density on \(s\in[\ell,\ell+1)\) is a piecewise
weight-\(\le(\ell-1)\) hyperlogarithm whose first two pieces are elementary,
\[\resizebox{\ifdim\width>\linewidth\linewidth\else\width\fi}{!}{$\displaystyle f_{X_{(1)}}(y;s)=\frac{1}{s-1}\ \text{on}\ y\in[0,s-1)\ \text{for}\ s\in[1,2),\qquad
  f_{X_{(1)}}(y;s)=\frac{1-\ln y+\ln(s-2)}{s-1}\ \text{on the next piece},$}\]
matching direct simulation. Higher weights are produced by
the same recursion and require a computer-algebra evaluation from weight three onward.

\textbf{Figure 1.} The order-statistic marginals \(f_{X_{(k)}}(y;s)\) at \(s=4.5\), from \(3\times10^5\) saturated
simulations. Each successive rank marches rightward into its feasible band, and the breaks at integer \(y\)
reflect the hard-core spacing; this is the aggregate rank density of Theorem \ref{thm:order-statistic-hierarchy} (unnormalised by
\(\Pr[N\ge k]\)). Generated.

\begin{figure}
\centering
\pandocbounded{\includegraphics[keepaspectratio,alt={order-statistic marginals}]{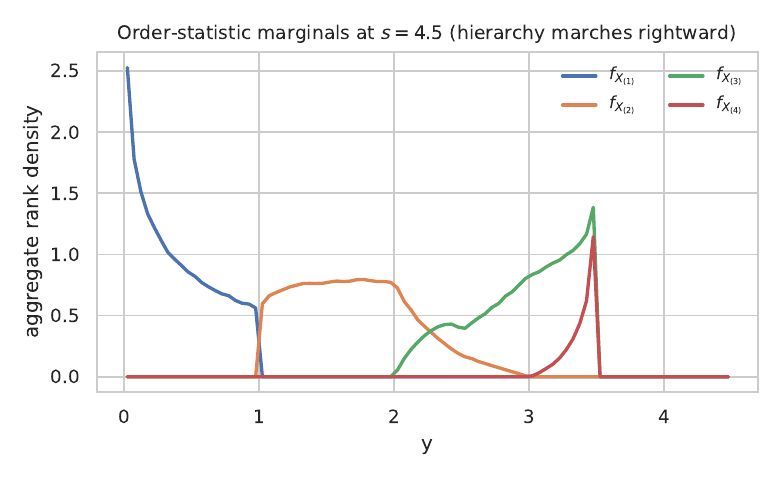}}
\caption{order-statistic marginals}
\end{figure}

\subsubsection{The aggregate mean sits between the count-conditioned extremes}

Averaging over the random saturated count \(N\) mixes the order statistic's law across every count with
positive jamming probability; a natural question is whether the resulting aggregate inherits a definite
position relative to the two extreme counts, \(N_{\min}(s)\) and \(N_{\max}(s)\).

\begin{quote}
\textbf{Observation (aggregate order statistics interpolate the count extremes).} At the running example
\(s=4.5\) (\(N\in\{2,3,4\}\)), the conditional mean \(\mathbb E[X_{(k)}\mid N=n]\) is monotone decreasing in
\(n\) across the whole window, for every rank \(k=1,2,3\); consequently the aggregate mean
\(\mathbb E[X_{(k)}]\) lies strictly between \(\mathbb E[X_{(k)}\mid N=N_{\min}]\) and
\(\mathbb E[X_{(k)}\mid N=N_{\max}]\). Checked numerically at this one example (nested Gauss--Legendre
quadrature over the exact jammed-cell polytopes; the count mass \(p_4\) cross-checked against an
independent closed form) --- not a proved statement for general \(s\) and \(k\).
\end{quote}

\textbf{Figure 2.} The conditional means \(\mathbb E[X_{(k)}\mid N=n]\) for \(k=1,2,3\) (lines) against \(n=2,3,4\) at
\(s=4.5\), with the aggregate mean marked (\(\star\)); each aggregate sits strictly inside the band spanned by
its two endpoint counts. Generated.

\begin{figure}
\centering
\pandocbounded{\includegraphics[keepaspectratio,alt={order-statistic monotonicity in the count}]{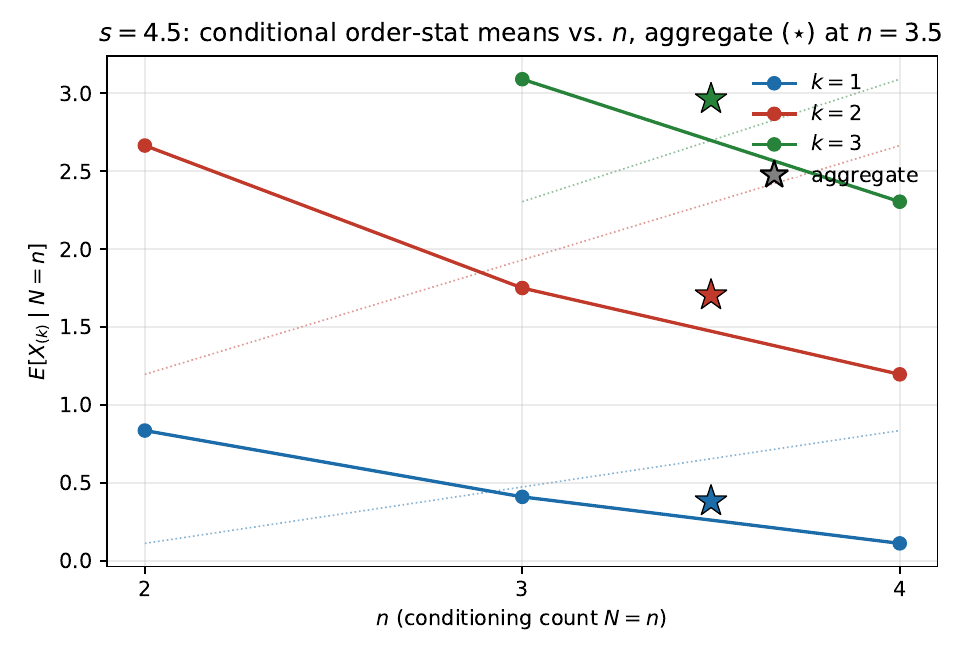}}
\caption{order-statistic monotonicity in the count}
\end{figure}

\subsubsection{The full order-statistic vector, the joint density}

Each single marginal \(f_{X_{(k)}}\) of Theorem \ref{thm:order-statistic-hierarchy} is a projection of the joint density of the whole vector
\(X_{(1)}<\dots<X_{(N)}\) at fixed count \(N=n\). That joint density is elementary before marginalisation and
stays in the hyperlogarithm class after it.

\begin{quote}
\refstepcounter{thmcnt}\label{prop:joint-order-statistic-density}\textbf{Proposition\nobreakspace{}\thethmcnt{} (joint order-statistic density).} Fix a count \(n\) and a non-integer \(s\). The joint
density of \((X_{(1)},\dots,X_{(n)})\) is piecewise rational: on each region of fixed temporal order it is the
product \(\prod_{k=1}^{n}\ell_k^{-1}\) of the free lengths \(\ell_k\) (the space available to the \(k\)-th car),
each affine in the gaps with coefficients in \(\{-1,0,+1\}\). Every marginal obtained by integrating out \(m\)
of the coordinates is therefore a hyperlogarithm of weight at most \(m\) (linear reducibility), with no
algebraic or elliptic singularity.
\end{quote}

\emph{Proof.} The strong-Markov split places each car uniformly on the current free space, contributing one
factor \(1/\ell_k\) per car; on a region of fixed temporal order the product is rational (\ref{sec:joint-order-stats},
\ref{thm:jamming-polytope-density}). The free lengths are \(\pm1\)-affine in the gaps (\ref{sec:joint-order-stats}), so the
singular locus of any marginalisation is a hyperplane arrangement, and Brown's reduction keeps the integral
linearly reducible, hence a hyperlogarithm (Brown 2009). \(\qquad\square\)

\textbf{Example (\(n=4\)).} On an \(n=4\) region the joint density is the single rational function
\(\prod_{k=1}^4\ell_k^{-1}\) (weight \(0\)). Keeping two gaps and Brown-eliminating the other two, on the actual
\(n=4\) and \(n=5\) free-length forms, never raises the degree of the kept variables: the singular locus stays a
hyperplane arrangement, so the two-variable marginal is a hyperlogarithm, no square root, no elliptic curve (a bilinear control term does produce quadrics, confirming the test
discriminates). This is the fixed-\(n\) input to the aggregate density of §5.

\subsubsection{From rank to count-weighting}

Theorem \ref{thm:order-statistic-hierarchy} resolves the coupled branch, in which conditioning on a rank ties the two sub-segments. The
remaining branch of Proposition \ref{prop:coproduct-and-coupling} weights the count alone, leaving the sub-segments decoupled. Whereas
a rank couples the sub-segments, a count-weighting leaves them decoupled; this character branch reduces to
values of the count generating function of §3 for a restricted weight class, made precise in Theorem \ref{thm:representability-character-case}.

\subsubsection{Representability of count-weightings}

The count-weighting \(\mathbb E[a_N]=\sum_n a_n\,p_n(s)\) is a character evaluation: it pairs the count law
\((p_n)\) against the weight sequence \((a_n)\).

\begin{quote}
\refstepcounter{thmcnt}\label{thm:representability-character-case}\textbf{Theorem\nobreakspace{}\thethmcnt{} (representability, character case).} If \((a_n)=\sum_j c_j\,\zeta_j^{\,n}\) is a finite
combination of geometric sequences, then \(\mathbb E[a_N]=\sum_j c_j\,g(s,\zeta_j)\) is a finite combination
of values of the count generating function, and satisfies a finite linear equation in \(s\).
\end{quote}

\emph{Proof.} By linearity \(\mathbb E[a_N]=\sum_j c_j\sum_n \zeta_j^{\,n}p_n(s)=\sum_j c_j\,g(s,\zeta_j)\), and
each \(g(s,\zeta_j)\) satisfies the count generating function's equation of §3 at \(z=\zeta_j\); a finite sum of
its solutions satisfies a finite linear system. \(\qquad\square\)

The conjectured exact class is the P-recursive sequences, those satisfying a linear recurrence with
polynomial coefficients, equivalently with a holonomic generating function.

\begin{quote}
\refstepcounter{thmcnt}\label{conj:representability-dichotomy}\textbf{Conjecture\nobreakspace{}\thethmcnt{} (representability dichotomy).} \(\mathbb E[a_N]\) satisfies a finite linear Volterra system
if and only if \((a_n)\) is P-recursive.
\end{quote}

The forward direction extends the character case to polynomial-times-character weights; the converse rests
on an order-bound lemma (an \(\ell\)-independent degree bound on the annihilating operator), which remains open.

The parity weighting \(a_n=(-1)^n\) is
the character contrast: \(\Pr[N\text{ even}]=\tfrac12(1+g(s,-1))\) is covered by Theorem \ref{thm:representability-character-case}. At \(s=3.5\), where
\(N\in\{2,3\}\), this gives \(\Pr[N\text{ even}]=p_2(3.5)=0.647\), with \(g(3.5,-1)=p_2-p_3=0.295\).

\subsection{The group-like corner: the count and max-gap Dyson--Schwinger equations}

\label{sec:aggregate-special-cases}

\emph{Two aggregates are group-like
elements of the parking Hopf algebra --- the count generating function and the max-gap distribution
function --- and each satisfies a quadratic Volterra equation. This section states both, records the
moment cascade as the \(z\)-expansion of the count equation, and ends with the two refinements (Tauberian constants, fractional
moments).}

\subsubsection{Two group-like aggregates}

Both aggregates of this section are split-compatible with a multiplicative composition law, the second regime of the Principle (§1); the equations are therefore quadratic rather than linear.

Both factorisations rest on one fact: given the first car's position \(U\), the two sub-segments are
independent under the strong-Markov split, and neither the count nor the maximal gap couples across the
first car. The two aggregates differ only in the algebra of the observable. The count is additive,
\(N=1+N_L+N_R\), so \(z^N=z\,z^{N_L}z^{N_R}\) holds pointwise and the count generating function
\(g(s,z)=\mathbb E[z^N]\) is multiplicative over the split. The maximal gap is extremal, so the distribution
function \(Q(s,x)=\Pr[\max_i G_i\le x]\) factors through the event identity
\(\{\max_i G_i\le x\}=\{\max_{i\in L}G_i\le x\}\cap\{\max_{i\in R}G_i\le x\}\), valid because the gaps
partition across the split with none straddling the first car. Both are the group-like branch of
Proposition \ref{prop:coproduct-and-coupling}: the count and its extremal companion.

\subsubsection{The count equation}

\begin{quote}
\refstepcounter{thmcnt}\label{thm:count-generating-function}\textbf{Theorem\nobreakspace{}\thethmcnt{} (count generating function).} \(g(s,z)=\mathbb E[z^N]\) satisfies the quadratic Volterra
equation
\[\resizebox{\ifdim\width>\linewidth\linewidth\else\width\fi}{!}{$\displaystyle (s-1)\,g(s,z)=z\int_0^{s-1} g(u,z)\,g(s-1-u,z)\,du,\qquad g(s,z)=1\ (s<1),\quad g(s,z)=z\ (1\le s<2),$}\]
the parking Dyson--Schwinger equation.
\end{quote}

\emph{Proof.} The count splits as \(N=1+N_L+N_R\) with \(N_L,N_R\) independent given the first car's position \(U\),
so \(z^N=z\,z^{N_L}z^{N_R}\) and \(\mathbb E[z^N\mid U]=z\,g(U,z)\,g(s-1-U,z)\). Averaging over \(U\) uniform on
\([0,s-1]\) and clearing the density gives the equation; a segment shorter than a car holds no car
(\(g=1\)), and one car fits but not two on \([1,2)\) (\(g=z\)). \(\qquad\square\)

Marking Rényi's split by the count is immediate, and the lattice analogue of this equation is solved:
Krapivsky (2020) derives the bivariate generating function of the jammed occupation number for lattice
random sequential adsorption in closed form, together with the exact support window of each count. The
continuum solution below is not rational but piecewise hyperlogarithmic.

The equation is group-like, \(\Delta g=g\otimes g\) --- the defining coproduct of the count character
(\ref{sec:background-hopf}). That is not decoration: it is what makes Rényi's
equation a consequence rather than a companion.

\begin{quote}
\refstepcounter{thmcnt}\label{thm:moment-cascade-is-the-tangent-tower}\textbf{Theorem\nobreakspace{}\thethmcnt{} (the moment cascade is the tangent tower).} Write \(g_k(s)=\partial_z^kg(s,z)\big|_{z=1}\)
for the \(k\)-th factorial moment of \(N(s)\). Differentiating Theorem \ref{thm:count-generating-function} \(k\) times at \(z=1\) gives a
\textbf{linear} Volterra equation for \(g_k\), forced by \(g_0,\dots,g_{k-1}\). At \(k=1\) it is
\[\resizebox{\ifdim\width>\linewidth\linewidth\else\width\fi}{!}{$\displaystyle (s-1)\,g_1(s)=(s-1)+2\!\int_0^{s-1}\!g_1(u)\,du,$}\]
which is \textbf{Rényi's mean equation} (Theorem \ref{thm:parking-integral-equation-renyi-1958}, \ref{sec:background-aggregate}); at \(k=2\),
\((s-1)g_2=4\!\int_0^{s-1}\!g_1+2\!\int_0^{s-1}\!g_2+2\!\int_0^{s-1}\!g_1(u)g_1(s-1-u)\,du\).
\end{quote}

\emph{Proof.} Write the right-hand side of Theorem \ref{thm:count-generating-function} as \(z\,C(z)\) with \(C(z)=\int_0^{s-1}g(u,z)g(s-1-u,z)\,du\).
Then \(\partial_z^k(zC)=k\,C^{(k-1)}+z\,C^{(k)}\), and \(C^{(k)}(1)\) is, by the Leibniz rule, a sum of
\(\int g_i(u)g_{k-i}(s-1-u)\,du\) over \(i\le k\). The terms with \(i\in\{0,k\}\) carry \(g_k\) against
\(g_0\equiv1\) and give the two linear occurrences; the rest involve only lower orders. Setting \(k=1\) and
\(k=2\) and using \(g_0\equiv1\) gives the two displayed equations. \(\square\)

So the additive mean equation of §2 is the first tangent of the group-like equation at the identity, and
the whole moment cascade is its higher tangents. Verified against direct solutions: the first tangent matches Rényi's
equation to \(1.3\times10^{-6}\), the second to \(2\times10^{-5}\) relative, and the solved mean returns
Rényi's constant as \(0.7484\) against \(m=0.7476\).

\textbf{The count law as a sum over trees.} Unrolling Theorem \ref{thm:count-generating-function} indexes its terms by the shape of the split.
A saturated configuration is an ordered binary tree --- the first car is the root, the two sub-segments
its subtrees --- and assigning to each tree the function
\[\resizebox{\ifdim\width>\linewidth\linewidth\else\width\fi}{!}{$\displaystyle \Phi(\varnothing)(s)=\mathbf 1\{s<1\},\qquad
  \Phi\bigl(\mathrm{node}(L,R)\bigr)(s)=\frac1{s-1}\int_0^{s-1}\!\Phi(L)(u)\,\Phi(R)(s-1-u)\,du$}\]
gives \(p_n(s)=\Pr[N(s)=n]=\sum_{T}\Phi(T)(s)\), the sum over the \(\mathrm{Cat}(n)\) trees with \(n\) nodes.
The target here is the algebra of functions under the \textbf{split-convolution}
\((f*g)(s)=\int_0^{s-1}f(u)g(s-1-u)\,du\), which is commutative and associative, not functions under
pointwise product; \(\Phi\) is a character into it. Verified, including that \(p_n\) is
supported exactly off \([n,2n+1)\).

\textbf{What the tree algebra does and does not give.} In the Connes--Kreimer algebra the same equation reads
\(X=B_+\bigl((1+X)^2\bigr)\) with \(X=g-1\), and \(P(h)=(1+h)^2\) is the \((\alpha,\beta)=(2,-\tfrac12)\) member
of Foissy's family \((1-\alpha\beta h)^{-1/\beta}\); since \(\beta\ne-1\) the subalgebra generated by the
graded pieces is of Faà di Bruno rather than ladder type (Foissy 2008). That the subalgebra is Hopf is
checked directly, not quoted,,
with four series outside the family failing at order four. \textbf{What is not established is the transport.}
No coproduct on the split-convolution algebra has been constructed for \(\Phi\) to intertwine with, so
nothing is carried from the tree algebra to \(g(s,z)\); in the renormalisation setting the coproduct earns
its keep through the Birkhoff decomposition of divergences, and the parking integrals converge. Whether
the Faà di Bruno structure has any consequence downstairs is \textbf{open}, and is the same question as the
group-flow generator conjecture of \ref{sec:future-work}.

\textbf{Example.} At \(s=2.5\) the segment holds one or two cars, and the equation gives \(g(2.5,z)=\tfrac13 z+\tfrac23 z^2\), so \(\Pr[N=2]=\tfrac23\).

The count depends only on how many cars attach, a number the split determines recursively: each
sub-segment is an independent smaller copy of the whole. The car positions, and with them the order
statistics, therefore never enter. The count generating function, its moments, and their \(s\to\infty\) asymptotics all follow from this
single equation.

\subsubsection{The max-gap equation}

The max gap is group-like as an extremal observable: the event that every gap is at most \(x\) factors over
the two sub-segments, because the gaps of the saturated configuration partition into those of the left and
right sub-segments with none straddling the first car. As with the count, the two sides are independent
given \(U\); the difference is the observable, additive for the count (\(z^N=z\,z^{N_L}z^{N_R}\)) and extremal
for the max gap (\(\{\max_i G_i\le x\}=\{\max_{i\in L}G_i\le x\}\cap\{\max_{i\in R}G_i\le x\}\)).

\begin{quote}
\refstepcounter{thmcnt}\label{thm:max-gap-distribution-function}\textbf{Theorem\nobreakspace{}\thethmcnt{} (max-gap distribution function).} \(Q(s,x)=\Pr[\max_i G_i\le x]\) satisfies the quadratic
Volterra equation
\[\resizebox{\ifdim\width>\linewidth\linewidth\else\width\fi}{!}{$\displaystyle (s-1)\,Q(s,x)=\int_0^{s-1} Q(u,x)\,Q(s-1-u,x)\,du,\qquad Q(s,x)=\mathbf 1\{s\le x\}\ (s<1).$}\]
\end{quote}

\emph{Proof.} The gaps of the full configuration partition into those of the two saturated sub-segments, so
\(\{\max_i G_i\le x\}=\{\max_{i\in L}G_i\le x\}\cap\{\max_{i\in R}G_i\le x\}\), and these events are independent given the first
car's position \(U\); hence \(\Pr[\max_i G_i\le x\mid U]=Q(U,x)\,Q(s-1-U,x)\), and averaging over \(U\) gives the
equation. On a segment shorter than a car the whole segment is one gap of length \(s\), so \(\max_i G_i=s\)
and \(Q(s,x)=\mathbf 1\{s\le x\}\). \(\qquad\square\)

The count-marked function \(Q(s,x,z)=\mathbb E[z^N\mathbf 1\{\max_i G_i\le x\}]\) satisfies the same equation
with a factor \(z\), and reduces to the count generating function \(g(s,z)\) at \(x\ge1\), where every saturated
gap lies below the threshold.

Both factorisations are properties of the strong-Markov split, not of saturation: the split applies equally
to the partial process, carrying a coverage variable, an extension not developed here.

\textbf{Figure 1.} The max-gap distribution \(Q(s,x)\) at \(s=4.5\): the Volterra march of Theorem \ref{thm:max-gap-distribution-function} (line) against
direct simulation (\(1.2\times10^5\) runs, points). \(Q\) rises from \(0\) and reaches \(1\) at \(x=1\), since jamming
forces every gap below one car length. Generated.

\begin{figure}
\centering
\pandocbounded{\includegraphics[keepaspectratio,alt={max-gap distribution}]{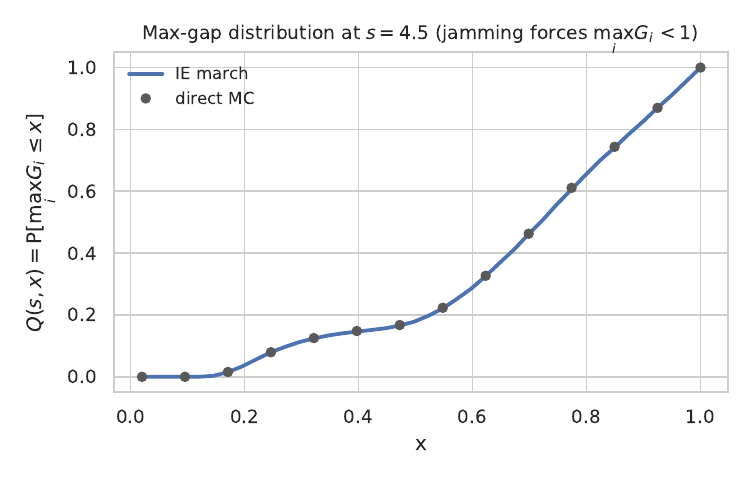}}
\caption{max-gap distribution}
\end{figure}

\subsubsection{Moments and two refinements}

The moment cascade is the \(z\)-expansion of Theorem \ref{thm:count-generating-function}. Writing \(M_k(s)=\partial_z^k g(s,z)\vert_{z=1}\) for
the \(k\)-th factorial moment and differentiating the Dyson--Schwinger equation \(k\) times gives a linear system
for \(M_1,\dots,M_k\) that is lower-triangular in the moment order. This system is the
Faà-di-Bruno graded-coordinate expansion: the parking equation \(X=1+B_+(X^2)\) belongs to Foissy's Faà-di-Bruno subalgebra (Foissy 2008; the cocycle \(B_+\) and the placement at
\((\alpha,\beta)=(2,-\tfrac12)\) are in \ref{sec:background-hopf}).

Two refinements are read off the same expansion.

\begin{itemize}
\tightlist
\item
  \textbf{Tauberian constants.} The cumulants are extensive, \(\kappa_k(s)=C_k\,s+O(1)\), with \(C_1=m\approx0.7476\)
  the Rényi parking constant and \(C_2\approx0.0382\) the Dvoretzky--Robbins variance rate; the constants
  \(C_3,\dots,C_{10}\) are computed from the cascade.
\item
  \textbf{Fractional moments.} The fractional moment \(\mathbb E[N^\nu]=\sum_n n^\nu p_n(s)\) weights the count by \(a_n=n^\nu\); this
  admits a finite equation for \(\nu\in\mathbb Z_{\ge0}\); whether it fails otherwise is the conjectural
  representability boundary of §4.
\end{itemize}

\subsubsection{One equation for every split-tree functional}

The count equation is usually read as a statement about the count. It is not. The equation depends only on
how an observable \emph{composes} under the split, and the count is one instance of a class it covers whole.

Place the first car at \(U\). The car occupies \([U,U+1]\) and separates the segment completely, so the
saturated configuration on \([0,s]\) is the union of independent saturated configurations on \([0,U]\) and
\([U+1,s]\), and its gap multiset is the \textbf{disjoint union} of theirs. Cars add the same way. Call

\[\resizebox{\ifdim\width>\linewidth\linewidth\else\width\fi}{!}{$\displaystyle T \;=\; c\,N \;+\; \sum_i h(G_i)$}\]

a \textbf{split-tree functional}, with \(c\in\mathbb R\) a per-car weight and \(h\) a function of a gap length.

\begin{quote}
\refstepcounter{thmcnt}\label{thm:one-equation-for-the-class}\textbf{Theorem\nobreakspace{}\thethmcnt{} (one equation for the class).} Write \(G(s,\theta)=\mathbb E\bigl[e^{\theta T}\bigr]\). Then
\[\resizebox{\ifdim\width>\linewidth\linewidth\else\width\fi}{!}{$\displaystyle (s-1)\,G(s,\theta)\;=\;e^{\theta c}\int_0^{s-1}G(u,\theta)\,G(s-1-u,\theta)\,\mathrm du,
\qquad G(s,\theta)=e^{\theta h(s)}\ \ (s<1).$}\]
Theorem \ref{thm:count-generating-function} is the case \(c=1\), \(h\equiv0\), with \(\theta=\log z\); a pure gap aggregate is the case \(c=0\).
\end{quote}

\emph{Proof.} Conditionally on \(U\) the two sub-segments fill independently, and by the disjoint-union
property \(T=c+T_L+T_R\) with \(T_L\) and \(T_R\) independent given \(U\). Hence
\(e^{\theta T}=e^{\theta c}e^{\theta T_L}e^{\theta T_R}\), and
\(\mathbb E[e^{\theta T}\mid U]=e^{\theta c}G(U,\theta)G(s-1-U,\theta)\). Averaging over \(U\) uniform on
\([0,s-1]\) and clearing the density gives the equation. A segment shorter than a car holds no car and is
a single gap of length \(s\), which fixes the boundary datum. \(\square\)

\begin{quote}
\refstepcounter{thmcnt}\label{cor:cascade-is-the-class-s-not-the-count-s}\textbf{Corollary\nobreakspace{}\thethmcnt{} (the cascade is the class's, not the count's).} Since \(G(s,0)=1\), differentiating
Theorem \ref{thm:one-equation-for-the-class} \(k\) times in \(\theta\) at \(0\) gives a linear Volterra equation for the \(k\)-th moment of \(T\),
forced by the lower ones. Theorem \ref{thm:moment-cascade-is-the-tangent-tower} is the case \(c=1\), \(h\equiv0\).
\end{quote}

\textbf{Where the order statistics enter.} Each gap is a difference of consecutive order statistics,
\(G_i=X_{(i+1)}-X_{(i)}-1\) in the interior, so every aggregate of the displayed shape is a functional of
the order statistics, and conversely every order-statistic functional of that shape is an aggregate.
The class is exactly where the two descriptions meet, and Corollary \ref{cor:cascade-is-the-class-s-not-the-count-s} covers it entirely. What the
aggregate machinery cannot reach is a functional that fails to decompose over the split --- an order
statistic \emph{individually}, for instance, whose rank depends on both sides at once and which therefore
needs the coupled hierarchy of the next section rather than a single equation.

\textbf{Example (four functionals, one equation).} At \(s=5\), with the mean and variance read off the first
and second \(\theta\)-tangents and checked against direct simulation of the process:

{\footnotesize\begin{longtable}[]{@{}
  >{\raggedright\arraybackslash}p{\dimexpr 0.1154\linewidth-2\tabcolsep\relax}
  >{\centering\arraybackslash}p{\dimexpr 0.1538\linewidth-2\tabcolsep\relax}
  >{\raggedright\arraybackslash}p{\dimexpr 0.1154\linewidth-2\tabcolsep\relax}
  >{\raggedleft\arraybackslash}p{\dimexpr 0.1538\linewidth-2\tabcolsep\relax}
  >{\raggedleft\arraybackslash}p{\dimexpr 0.1538\linewidth-2\tabcolsep\relax}
  >{\raggedleft\arraybackslash}p{\dimexpr 0.1538\linewidth-2\tabcolsep\relax}
  >{\raggedleft\arraybackslash}p{\dimexpr 0.1538\linewidth-2\tabcolsep\relax}@{}}
\toprule\noalign{}
\begin{minipage}[b]{\linewidth}\raggedright
\(T\)
\end{minipage} & \begin{minipage}[b]{\linewidth}\centering
\(c\)
\end{minipage} & \begin{minipage}[b]{\linewidth}\raggedright
\(h(g)\)
\end{minipage} & \begin{minipage}[b]{\linewidth}\raggedleft
\(\mathbb E[T]\) (tangent)
\end{minipage} & \begin{minipage}[b]{\linewidth}\raggedleft
\(\mathbb E[T]\) (simulated)
\end{minipage} & \begin{minipage}[b]{\linewidth}\raggedleft
\(\operatorname{Var}T\) (tangent)
\end{minipage} & \begin{minipage}[b]{\linewidth}\raggedleft
\(\operatorname{Var}T\) (simulated)
\end{minipage} \\
\midrule\noalign{}
\endhead
\bottomrule\noalign{}
\endlastfoot
count \(N\) & \(1\) & \(0\) & \(3.4889\) & \(3.4952\pm0.0079\) & \(0.2497\) & \(0.2501\pm0.0056\) \\
\(\sum_i G_i^2\) & \(0\) & \(g^2\) & \(0.8644\) & \(0.8588\pm0.0078\) & \(0.2407\) & \(0.2422\pm0.0054\) \\
\(\sum_i \sqrt{G_i}\) & \(0\) & \(\sqrt g\) & \(2.3214\) & \(2.3190\pm0.0061\) & \(0.1485\) & \(0.1450\pm0.0032\) \\
\(N+\sum_i G_i^2\) & \(1\) & \(g^2\) & \(4.3533\) & \(4.3540\pm0.0025\) & \(0.0233\) & \(0.0227\pm0.0005\) \\
\end{longtable}}

Nothing changes between the rows except \(c\) and \(h\). The third row is not smooth at the origin and is
covered all the same, since the proof uses only the disjoint-union property.

\refstepcounter{thmcnt}\label{rem:what-the-joint-transform-carries}\textbf{Remark\nobreakspace{}\thethmcnt{} (what the joint transform carries).} The fourth row has variance \(0.0233\), far below either
\(0.2497\) or \(0.2407\). Reading the covariance off the three values gives \(-0.2335\), a correlation of
\(-0.953\): cars and gap length compete for the same segment, so an extra car removes gap length almost
deterministically. Neither marginal equation shows this, and the joint transform of Theorem \ref{thm:one-equation-for-the-class} does.

\subsection{The aggregate multi-point density is a hyperlogarithm}

\label{sec:aggregate-boundary}

\emph{For every finite \(s\) the aggregate
\(k\)-point density of the saturated process, averaged over the random count \(N\), solves a triangular Volterra
hierarchy and is a piecewise hyperlogarithm. In the thermodynamic limit the limiting pair correlation is the
classical Bonnier--Boyer--Viot exponential period, expected to lie outside the finite-\(s\) hyperlogarithm
family; that separation is stated as expected, not proved.}

\subsubsection{\texorpdfstring{Regularity is a finite-\(s\) property}{Regularity is a finite-s property}}

The regularity of an aggregate density, meaning the transcendence class of its solution (here the
hyperlogarithms; §2), is a finite-\(s\) property. The aggregate density is representable --- it solves a
Volterra hierarchy --- and is a piecewise hyperlogarithm at every finite \(s\); the limit \(s\to\infty\) is
expected to leave that class, though the separation is not established here (Remark). In this section we
prove the finite-\(s\) structure and locate the limit.

\subsubsection{The aggregate multi-point density is a hyperlogarithm}

Averaging over the random count is what makes this density regenerate: the fixed-\((s,n)\) joint of \ref{sec:joint-order-stats} is severed from its own values at smaller lengths, whereas \(\rho_k\) is built from them. The aggregate density is therefore the third regime of the Principle (§1), augmented by the point count, and the limit treated below is where regeneration fails outright.

For fixed \(s\) the saturated process has a random count \(N\), over which the aggregate \(k\)-point density
averages. Unlike the fixed-\((s,n)\) joint of \ref{sec:joint-order-stats}, the aggregate density satisfies a recursion in
\(s\), and that recursion is its integral equation.

\begin{quote}
\refstepcounter{thmcnt}\label{thm:aggregate-multi-point-hierarchy}\textbf{Theorem\nobreakspace{}\thethmcnt{} (aggregate multi-point hierarchy).} Write \(X_1,\dots,X_N\) for the car positions. The
aggregate \(k\)-point density
\[\resizebox{\ifdim\width>\linewidth\linewidth\else\width\fi}{!}{$\displaystyle \rho_k(y_1,\dots,y_k;s)=\mathbb E\Big[\sum_{i_1\neq\cdots\neq i_k}\ \prod_{l=1}^{k}\delta(X_{i_l}-y_l)\Big]$}\]
satisfies (i) a second-kind linear Volterra equation in \(s\) (the first-car recursion), forced by the
lower orders \(\rho_1,\dots,\rho_{k-1}\), the equations for \(k=1,2,\dots\) forming a triangular system;
(ii) has a unique continuous solution, obtained order by order; and (iii) restricts on each
\(s\in[\ell,\ell+1)\) to a hyperlogarithm.
\end{quote}

\emph{Proof.} Condition on the first car's position \(U\), uniform on \([0,s-1]\); by translation invariance a
sub-configuration on \([a,b]\) is the process on \([0,b-a]\). Each of the \(k\) points is the first car, or lies
in the left sub-configuration \([0,U]\), or in the right one \([U+1,s]\). The terms in which a point is the
first car, or in which two points fall in different sub-configurations, involve only the lower orders
\(\rho_{<k}\); the terms with all \(k\) points in one sub-configuration reproduce \(\rho_k\) on a shorter segment.
Averaging over \(U\) gives a second-kind linear Volterra equation in \(s\) with the unit-delay kernel
\(K(s,u)=\tfrac{2}{s-1}\mathbf 1\{u\le s-1\}\) of §2, forced by \(\rho_{<k}\). The kernel is locally integrable,
so the Volterra existence--uniqueness theorem (Gripenberg--Londen--Staffans 1990; Brunner 2004) gives a unique
continuous solution, built order by order through the resolvent of §2. On \([\ell,\ell+1)\) only finitely many
counts occur, so \(\rho_k=\sum_n p_n(s)\,\rho_k^{(n)}\). Each fixed-\(n\) density \(\rho_k^{(n)}\) is a
hyperlogarithm by linear reducibility (\ref{sec:joint-order-stats}); each count mass \(p_n(s)=[z^n]g(s,z)\) is a
piecewise hyperlogarithm by Theorem \ref{thm:solution-space-by-count} with \(h\equiv1\); and the hyperlogarithms form a shuffle
algebra under finite \(\mathbb Q\)-linear combination (Brown 2009). Hence \(\rho_k\) is a hyperlogarithm on each
\([\ell,\ell+1)\), and a piecewise hyperlogarithm in \(s\). \(\qquad\square\)

\textbf{Example (the pair correlation).} The rows \(k=1,2\) are the intensity \(\rho_1\) and the pair correlation
\(\rho_2\). The intensity solves the scalar first-car equation
\[\resizebox{\ifdim\width>\linewidth\linewidth\else\width\fi}{!}{$\displaystyle (s-1)\,\rho_1(y;s)=\mathbf 1_{[0,s-1]}(y)+\int_y^{s-1}\rho_1(y;U)\,\mathrm dU
   +\int_0^{y-1}\rho_1(y-U-1;s-1-U)\,\mathrm dU,$}\]
and the pair density is forced by it: for \(y_1<y_2\) with \(y_2\ge y_1+1\),
\[\resizebox{\ifdim\width>\linewidth\linewidth\else\width\fi}{!}{$\displaystyle (s-1)\,\rho_2(y_1,y_2;s)=\rho_1(y_2-y_1-1;s-1-y_1)+\rho_1(y_1;y_2)
   +\int_{y_2}^{s-1}\!\rho_2(y_1,y_2;U)\,\mathrm dU
   +\int_0^{y_1-1}\!\rho_2(y_1-U-1,y_2-U-1;s-1-U)\,\mathrm dU
   +\int_{y_1}^{y_2-1}\!\rho_1(y_1;U)\,\rho_1(y_2-U-1;s-1-U)\,\mathrm dU,$}\]
the five terms being: the first car is the left point, the first car is the right point, both points in the
left sub-segment, both in the right, and the two points split across the first car (the first two present
only where the marked point can be the first car). This is the \(k=2\) instance of Theorem \ref{thm:aggregate-multi-point-hierarchy} --- a second-kind
linear Volterra equation forced by \(\rho_1\), hence a piecewise hyperlogarithm. The order-3 density
follows as the next row, forced by \(\rho_1\) and \(\rho_2\). The script marches \(\rho_1\) and \(\rho_2\) and
matches direct simulation (, \(\max|\mathrm{IE}-\mathrm{MC}|<0.12\) at step \(h=0.06\)).

\textbf{Figure 1.} The aggregate pair density \(\rho_2(y_1,y_2;s)\) at \(s=5\), from \(4\times10^5\) saturated
simulations. The black diagonal band is the hard-core exclusion \(|y_1-y_2|<1\); the bright ridge beside it is
the nearest-neighbour peak and the corner peaks are the boundary pairs. This is the \(k=2\) density of
Theorem \ref{thm:aggregate-multi-point-hierarchy}.

\begin{figure}
\centering
\pandocbounded{\includegraphics[keepaspectratio,alt={pair density heatmap}]{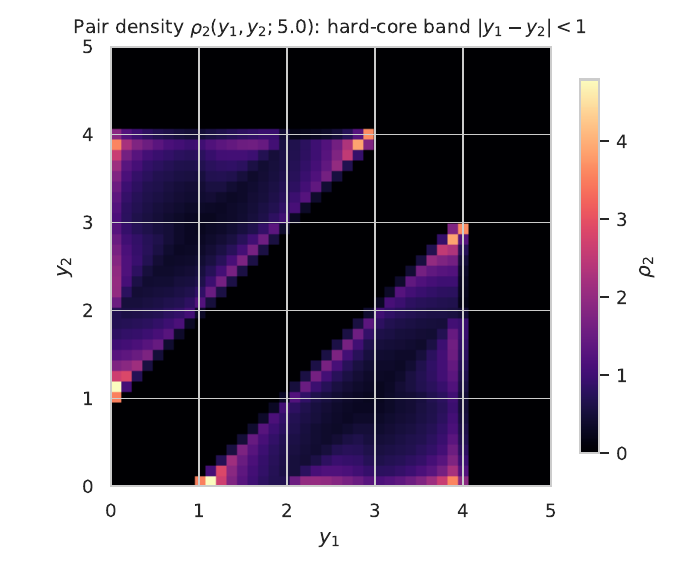}}
\caption{pair density heatmap}
\end{figure}

\textbf{Figure 2.} The 3-point density \(\rho_3(y_1,y_2,y_3;s)\) at \(s=6\) (denser cells shown, coloured by
\(\rho_3\)). The empty slabs are the triple hard-core exclusion --- any two coordinates within one car length ---
and the density concentrates on ordered triples at unit spacing. This is the \(k=3\) density of Theorem \ref{thm:aggregate-multi-point-hierarchy}, the
higher-order companion of Figure 1; its \(s\to\infty\) limit has no known closed form (§4).

\begin{figure}
\centering
\pandocbounded{\includegraphics[keepaspectratio,alt={three-point density}]{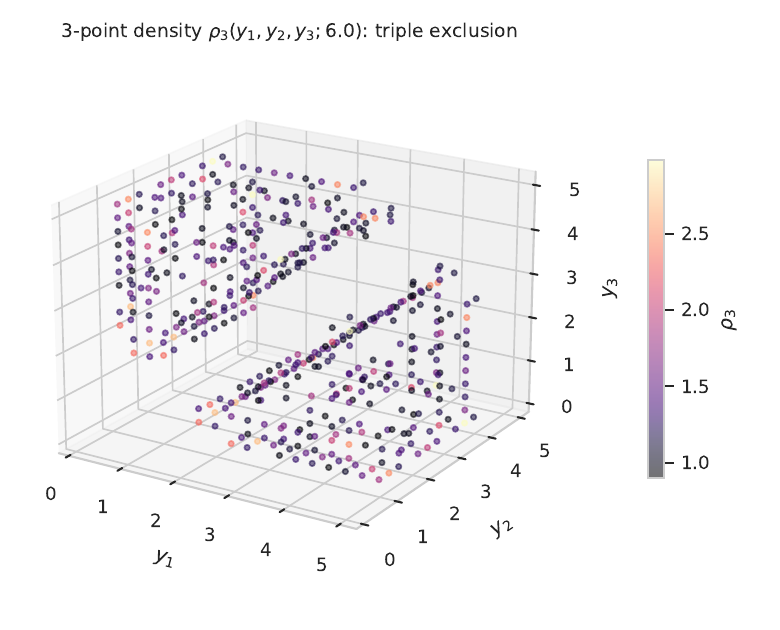}}
\caption{three-point density}
\end{figure}

\subsubsection{The asymptotic limit is classical}

The hyperlogarithm structure of Theorem \ref{thm:aggregate-multi-point-hierarchy} is a finite-\(s\) phenomenon and is the result of this section. Its
\(s\to\infty\) limit is not ours; it is a chain of classical facts, stated here as an observation and
attributed, not proved.

\begin{quote}
\textbf{Observation (thermodynamic limit).} As \(s\to\infty\) with \(N/s\to m\):
1. the aggregate intensity tends in the bulk to the constant \(m\), the Rényi parking constant (Rényi 1958);
2. the aggregate pair density \(\rho_2(y_1,y_2;s)\) tends to a translation-invariant stationary pair
correlation \(g_2(|y_1-y_2|)\), the infinite-volume limit of one-dimensional random sequential adsorption
(Penrose 2001);
3. \(g_2\) has the closed form computed by Bonnier--Boyer--Viot (1994), an \emph{exponential period} in the sense
of Kontsevich--Zagier (2001): an absolutely convergent integral \(\int e^{-f}\,\omega\) with \(f\) and the
algebraic differential form \(\omega\) defined over \(\overline{\mathbb Q}\) on a domain cut out by
polynomial inequalities over \(\overline{\mathbb Q}\);
4. the full joint law exists as a stationary point process (Penrose 2001) but is \textbf{non-renewal} --- the
gaps are correlated, so it does not reduce to independent gaps;
5. the higher correlation densities \(g_k\) \((k\ge3)\) satisfy the stationary Kirkwood--Salsburg (empty-interval)
hierarchy, each coupled to \(g_{k+1}\) (Evans 1993); the hierarchy does not truncate, so no \(g_k\) beyond
the pair is representable by a finite integral equation.
\end{quote}

\textbf{The Kirkwood--Salsburg hierarchy: the pair does not determine the rest.} In the thermodynamic limit the first-car
split is unavailable --- an infinite translation-invariant configuration has no distinguished first car --- so
the stationary correlations no longer obey the downward recursion of Theorem \ref{thm:aggregate-multi-point-hierarchy}. They obey instead the
sequential-adsorption counterpart of the Kirkwood--Salsburg hierarchy (Talbot, Tarjus, Van Tassel and Viot
2000, Eqs. (24)--(30)), the standard framework for the correlation functions of an interacting particle
system: the equation for the \(k\)-point density \(g_k\) also involves \(g_{k+1}\)
(Evans 1993 gives the empty-interval form for random sequential adsorption). The same upward-reaching structure is known in kinetic theory as the
Bogoliubov--Born--Green--Kirkwood--Yvon hierarchy; for sequential adsorption the Kirkwood--Salsburg name is
the one in use. Each order reaches \emph{upward}, and
no finite subsystem \(g_{\le k}\) is known to determine the rest. The pair \(g_2\) is the one order with a known
closed form (Bonnier--Boyer--Viot); \(g_3\) requires \(g_4\), and so on. This is the general case: not a single
integral equation but an infinite non-truncating system, of which only the first rung is solved.

\textbf{Illustration --- the connected correlations do not stop at the pair.} Were the hierarchy to reduce to the
pair --- the Kirkwood superposition approximation \(g_3\approx g_2 g_2 g_2\) --- every connected (irreducible) correlation beyond
order two would vanish. They do not. The normalised connected \(k\)-gap cumulant \(\kappa_k/\sigma^k\) of \(k\)
consecutive bulk gaps, measured at \(L=600\), is
\[\resizebox{\ifdim\width>\linewidth\linewidth\else\width\fi}{!}{$\displaystyle \kappa_2/\sigma^2=+0.089\ (34\ \text{s.e.}),\qquad
  \kappa_3/\sigma^3=+0.018\ (7\ \text{s.e.}),\qquad
  \kappa_4/\sigma^4=+0.000\ (\text{below resolution}).$}\]
The three-gap cumulant is resolvably nonzero, so \(g_3\) carries genuine three-body correlation and the
hierarchy does not truncate at the pair. The connected correlations then fall rapidly: the four-gap cumulant
already lies within its Monte-Carlo error of zero (Figure 3). The hierarchy therefore has content beyond the
pair, but that content decays fast --- the higher orders are increasingly negligible, though not proved to
vanish.

\textbf{Figure 3.} The connected \(k\)-gap cumulants \(\kappa_k/\sigma^k\) at \(L=600\) (error bars \(\pm1\) s.e.): the
pair \((k=2)\) and the triple \((k=3)\) are resolvably nonzero --- the hierarchy carries correlation beyond the
pair --- while the four-gap cumulant sits at the noise floor.

\begin{figure}
\centering
\pandocbounded{\includegraphics[keepaspectratio,alt={connected k-gap correlation ladder}]{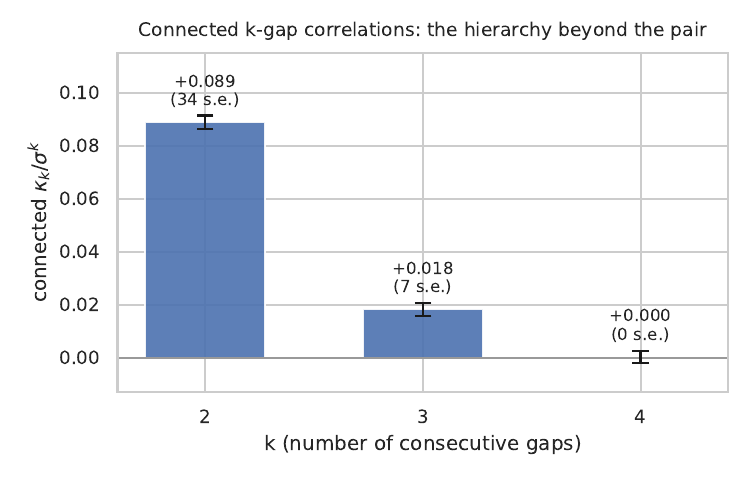}}
\caption{connected k-gap correlation ladder}
\end{figure}

\begin{quote}
\refstepcounter{thmcnt}\label{rem:non-truncation-is-the-loss-of-the-spatial}\textbf{Remark\nobreakspace{}\thethmcnt{} (non-truncation is the loss of the spatial Markov property).} The hierarchy's failure to truncate
and the correlation of the gaps are one statement, not two. Suppose that the stationary gap sequence
\((G_i)_{i}\) were a Markov chain in the spatial index; every finite-dimensional law of \((G_i)_i\) would then
be determined by the two-dimensional one, so no order beyond the pair would carry independent information
and the hierarchy would close at \(g_2\). Non-truncation, cited to Evans 1993 rather than proved here, is
therefore the analytic form of the statement that \((G_i)_i\) is not a Markov chain. What fails is the Markov
property of the gap sequence in the spatial index, not the existence of the limit; the limiting point
process exists and is stationary (Penrose 2001).
\end{quote}

The measured cumulants are evidence for the premise rather than for the conclusion, and the distinction is
worth keeping. A nonzero \(\kappa_2\) refutes the renewal property, and a nonzero \(\kappa_3\) refutes the
Kirkwood closure \(g_3\approx g_2g_2g_2\); neither refutes the Markov closure \(f_{123}=f_{12}f_{23}/f_2\), which
a Markov chain with correlated gaps satisfies while carrying a nonzero \(\kappa_3\). Testing that closure
directly is not part of the present evidence base. The connected correlations also decay rapidly, the
four-gap cumulant already lying within its Monte-Carlo error of zero, so the higher orders are not proved to
vanish.

The contrast with finite length is exact on one object. The finite-\(s\) three-point density \(\rho_3\) is
representable by an integral equation, forced by \(\rho_1,\rho_2\) through the first-car split (Theorem \ref{thm:aggregate-multi-point-hierarchy},
\(k=3\); Figure 2); its thermodynamic limit \(g_3\) is not --- the same density is representable at finite length
and governed by the non-truncating Kirkwood--Salsburg hierarchy in the limit. The split, a boundary operation, is what
the limit gives up.

\begin{quote}
\refstepcounter{thmcnt}\label{rem:regularity-boundary-and-the-open-joint}\textbf{Remark\nobreakspace{}\thethmcnt{} (regularity boundary and the open joint).} For every finite \(s\) the aggregate density
\(\rho_k(\,\cdot\,;s)\) is a piecewise hyperlogarithm (Theorem \ref{thm:aggregate-multi-point-hierarchy}); the limiting pair correlation \(g_2\) is an
exponential period. That \(g_2\) lies in a strictly larger transcendence class than the finite-\(s\) solutions
is expected, but rests on a transcendence separation not established here. Beyond the pair, the \(k\ge3\)
correlation functions have no known closed form, and the full joint is open (P5).
\end{quote}

\subsection{Illustrations: the count moments, the order-statistic correspondence, and the two ensembles}

\label{sec:aggregate-examples}

\emph{Worked illustrations for §§2--3; no theory or proofs. Each example confirms a piece of the finite-\(s\) count
theory and states how it extends Rényi's asymptotic account. Rényi (1958) and Dvoretzky--Robbins (1964)
gave the \(s\to\infty\) rate constants; the count Dyson--Schwinger equation of §3 gives the moments at every
finite \(s\), ties the count to the order statistics, and separates the jammed and unjammed ensembles.}

\subsubsection{Example 1 --- the count moments, interval by interval}

\textbf{Confirms.} The count Dyson--Schwinger equation of §3 reproduces the closed-form per-interval mean; the
mean slope \(\mathrm dM_1/\mathrm ds\) reaches \(0.7479\) at \(s=6.5\), against Rényi's \(m=0.7476\).

\textbf{Extends.} Rényi and Dvoretzky--Robbins gave only the \(s\to\infty\) rate constants. The equation gives the
raw moments \(\mathbb E[N^k]\) in closed form at every finite \(s\), interval by interval. The forms below sit on
one \emph{weight ladder}: on \([a,a{+}1)\) every moment has transcendental weight at most \(a-2\), independent of \(k\)
(rational on \([2,3)\), one logarithm on \([3,4)\), one dilogarithm on \([4,5)\)); a higher moment enriches the
basis at each weight but does not raise the weight. The Sage cascade
forms are march-verified over \([2,5)\) (max relative error \(1.9\times10^{-3}\),):

{\footnotesize\begin{longtable}[]{@{}
  >{\raggedright\arraybackslash}p{\dimexpr 0.2000\linewidth-2\tabcolsep\relax}
  >{\raggedright\arraybackslash}p{\dimexpr 0.2000\linewidth-2\tabcolsep\relax}
  >{\raggedright\arraybackslash}p{\dimexpr 0.2000\linewidth-2\tabcolsep\relax}
  >{\raggedright\arraybackslash}p{\dimexpr 0.2000\linewidth-2\tabcolsep\relax}
  >{\raggedright\arraybackslash}p{\dimexpr 0.2000\linewidth-2\tabcolsep\relax}@{}}
\toprule\noalign{}
\begin{minipage}[b]{\linewidth}\raggedright
interval
\end{minipage} & \begin{minipage}[b]{\linewidth}\raggedright
\(\mathbb E[N]\)
\end{minipage} & \begin{minipage}[b]{\linewidth}\raggedright
\(\mathbb E[N^2]\)
\end{minipage} & \begin{minipage}[b]{\linewidth}\raggedright
\(\mathbb E[N^3]\)
\end{minipage} & \begin{minipage}[b]{\linewidth}\raggedright
weight
\end{minipage} \\
\midrule\noalign{}
\endhead
\bottomrule\noalign{}
\endlastfoot
\([1,2)\) & \(1\) & \(1\) & \(1\) & 0 \\
\([2,3)\) & \(\dfrac{3s-5}{s-1}\) & \(\dfrac{7s-13}{s-1}\) & \(\dfrac{15s-29}{s-1}\) & 0 \\
\([3,4)\) & weight-1 (log) & weight-1 (log) & weight-1 (log) & 1 \\
\([4,5)\) & weight-2 (dilog) & weight-2 (dilog) & weight-2 (dilog) & 2 \\
\end{longtable}}

On \([3,4)\) each moment gains a single logarithm \(\log(s-2)\):
\[\resizebox{\ifdim\width>\linewidth\linewidth\else\width\fi}{!}{$\displaystyle \mathbb E[N]=\frac{7s-4\log(s{-}2)-17}{s-1},\quad \mathbb E[N^2]=\frac{29s-20\log(s{-}2)-79}{s-1},\quad \mathbb E[N^3]=\frac{103s-76\log(s{-}2)-293}{s-1}.$}\]

On \([4,5)\) each moment carries the same dilogarithm \(\mathrm{Li}_2(3-s)\); the explicit forms (e.g.
\(\mathbb E[N]=1-\tfrac{2}{3(s-1)}(72+\pi^2-21s+6\log2+30\log\tfrac{s-2}{2}+12\log(s{-}3)\log(s{-}2)+12\,\mathrm{Li}_2(3{-}s))\))
are. The third cumulant rate is \(C_3\approx-2.4\times10^{-5}\) (\texttt{G14}), so
\(N(s)\) is near-normal to third order.

The piecewise moments converge to their linear asymptotes as \(s\to\infty\): the occupancy
\(M_1(s)/s\to m=0.7476\) (from below, deviation \(\approx(1-m)/s\), the wasted end-room) and the variance rate
\(\mathrm{Var}(N)/s\to0.0382\) (Dvoretzky--Robbins, from above). The small-\(s\) wiggles are the
interval-by-interval polylogarithm structure settling onto the line; the variance rate, still rising near
\(s=7\), turns over and reaches its limit by \(s\approx15\).

\textbf{Figure 1.} Left: the exact piecewise \(M_1(s)\) and \(10\times\mathrm{Var}(s)\) hug their linear asymptotes.
Right: the occupancy \(M_1(s)/s\to m=0.7476\) (Rényi 1958) and the variance rate \(\mathrm{Var}(s)/s\to0.0382\)
(Dvoretzky--Robbins 1964); the oscillations settle by \(s\approx15\). Deterministic cascade march to \(s=30\),.

\begin{figure}
\centering
\pandocbounded{\includegraphics[keepaspectratio,alt={M1 and variance-rate convergence to their asymptotic constants}]{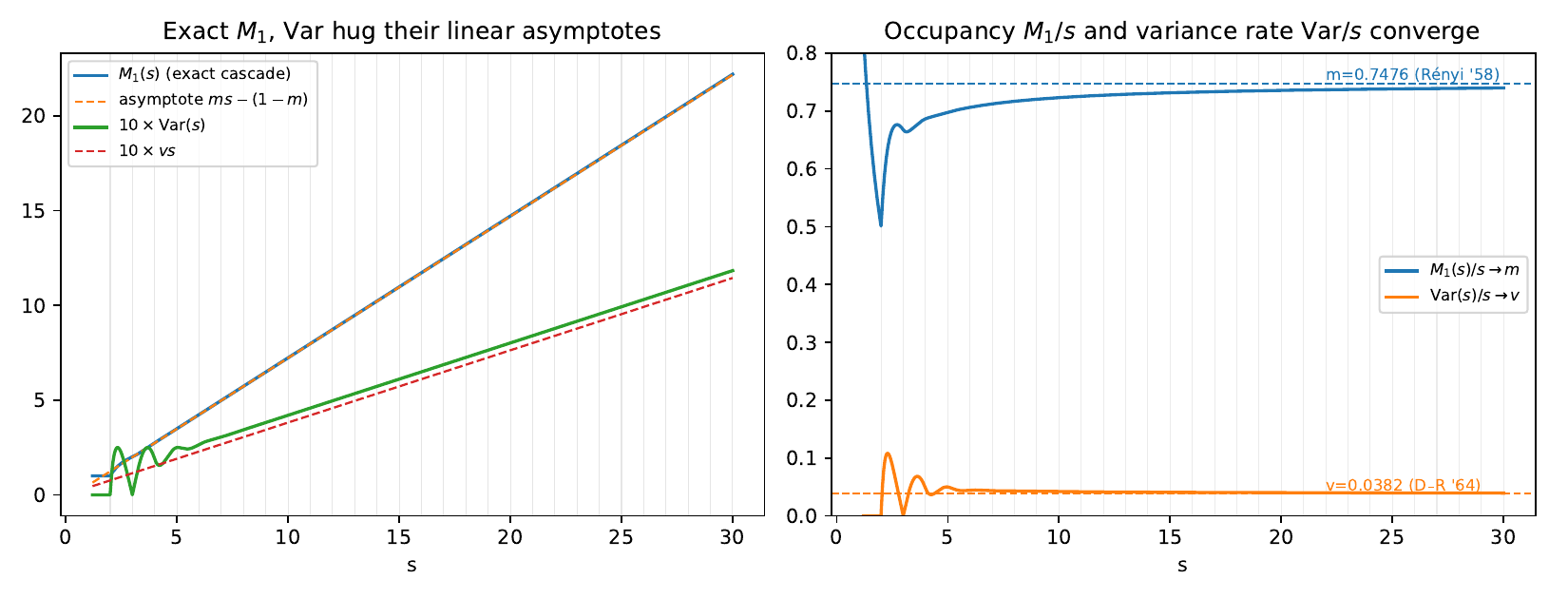}}
\caption{M1 and variance-rate convergence to their asymptotic constants}
\end{figure}

\subsubsection{\texorpdfstring{Example 2 --- the order-statistic \(\leftrightarrow\) Rényi correspondence}{Example 2 --- the order-statistic \textbackslash leftrightarrow Rényi correspondence}}

\textbf{Confirms.} The mean count \(M_1(s)\) from the Dyson--Schwinger equation equals \(\mathbb E[N]\) computed from
a direct simulation of the saturated configuration through its spatial order statistics.

\textbf{Extends.} The equation lives in the temporal (arrival-time) filtration, whereas the order statistics are
the sorted positions. The agreement below shows that the count is the \(n\)-marginal of the fixed-\(n\)
order-statistic density, linking the count to the positional description at finite \(s\), not only in the
limit. The DSE column is the grid march, biased high by \(O(h)\), up to \(1.7\times10^{-3}\) on this range; the simulation matches the exact mean
to that precision.

{\footnotesize\begin{longtable}[]{@{}
  >{\raggedright\arraybackslash}p{\dimexpr 0.3333\linewidth-2\tabcolsep\relax}
  >{\raggedright\arraybackslash}p{\dimexpr 0.3333\linewidth-2\tabcolsep\relax}
  >{\raggedright\arraybackslash}p{\dimexpr 0.3333\linewidth-2\tabcolsep\relax}@{}}
\toprule\noalign{}
\begin{minipage}[b]{\linewidth}\raggedright
\(s\)
\end{minipage} & \begin{minipage}[b]{\linewidth}\raggedright
\(M_1\) (DSE march)
\end{minipage} & \begin{minipage}[b]{\linewidth}\raggedright
\(\mathbb E[N]\) (order-statistic MC, \(\pm1\) s.e.)
\end{minipage} \\
\midrule\noalign{}
\endhead
\bottomrule\noalign{}
\endlastfoot
\(2.5\) & \(1.6680\) & \(1.6668\pm0.0014\) \\
\(3.5\) & \(2.3527\) & \(2.3513\pm0.0014\) \\
\(4.5\) & \(3.1136\) & \(3.1107\pm0.0012\) \\
\(5.5\) & \(3.8616\) & \(3.8585\pm0.0014\) \\
\end{longtable}}

\subsubsection{Example 3 --- the jammed and unjammed count ensembles}

\textbf{Confirms.} The jammed count mass \(p_n(s)=[z^n]g(s,z)\) from the Dyson--Schwinger equation equals the
histogram of the saturated count from a direct run of the process.

\textbf{Extends.} The saturated (jammed) count is a marginal of the partial (unjammed) ensemble: \(p_n(s)\) is the
mass of the fixed-\(n\) partial density \(f_{\mathrm{UCP}}(\,\cdot\,;s,n)\) on the region where all \(n{+}1\)
gaps are shorter than one car length,
\[\resizebox{\ifdim\width>\linewidth\linewidth\else\width\fi}{!}{$\displaystyle p_n(s)=\int_{\text{all gaps}<1} f_{\mathrm{UCP}}(\bm x;s,n)\,\mathrm d\bm x.$}\]
The jammed equation of §3 and the unjammed fixed-\(n\) ensemble are two views of one object; the table below
confirms the count marginal numerically at \(s=3.5\).

{\footnotesize\begin{longtable}[]{@{}lll@{}}
\toprule\noalign{}
\(n\) & \(p_n(s)\) (DSE) & histogram (process MC, \(\pm1\) s.e.) \\
\midrule\noalign{}
\endhead
\bottomrule\noalign{}
\endlastfoot
\(2\) & \(0.6473\) & \(0.6489\pm0.0011\) \\
\(3\) & \(0.3527\) & \(0.3511\pm0.0011\) \\
\end{longtable}}

\subsubsection{Example 4 --- the unjammed process has an equation of its own}

Rényi's equation describes the saturated segment, and the constant \(m\) belongs to that limit. The
unjammed process --- arrivals still arriving, some gap still admitting a car --- is the same attachment
chain stopped before absorption, and the same first-car split gives it an equation in which time
appears alongside length.

Let arrivals form a rate-one Poisson field in position and time on \([0,s-1]\times[0,\infty)\), an
arrival being accepted when it fits. On an empty segment the free measure is \(s-1\), so the first
acceptance occurs at a time \(\tau\sim\mathrm{Exp}(s-1)\) and a position \(U\) uniform on \([0,s-1]\),
independent of \(\tau\). The strong Markov property restarts both sides for the \textbf{remaining} time
\(t-\tau\). Writing \(M(s,t)=\mathbb E[N(s,t)]\) and \(I(s,t)=\int_0^{s-1}M(u,t)\,\mathrm du\),

\[\resizebox{\ifdim\width>\linewidth\linewidth\else\width\fi}{!}{$\displaystyle M(s,t)\;=\;\int_0^{t} e^{-(s-1)r}\Bigl[(s-1)+2\,I(s,t-r)\Bigr]\mathrm dr ,
\qquad M(s,t)=0\ \ (s<1).$}\]

One Volterra equation in two variables, from the split that gives Rényi's. Once no gap admits a car the
process stops, so \(M(s,\cdot)\) increases to the saturated mean and Rényi's equation is the \(t\to\infty\)
limit of this one.

{\footnotesize\begin{longtable}[]{@{}
  >{\raggedright\arraybackslash}p{\dimexpr 0.1304\linewidth-2\tabcolsep\relax}
  >{\raggedleft\arraybackslash}p{\dimexpr 0.1739\linewidth-2\tabcolsep\relax}
  >{\raggedleft\arraybackslash}p{\dimexpr 0.1739\linewidth-2\tabcolsep\relax}
  >{\raggedleft\arraybackslash}p{\dimexpr 0.1739\linewidth-2\tabcolsep\relax}
  >{\raggedleft\arraybackslash}p{\dimexpr 0.1739\linewidth-2\tabcolsep\relax}
  >{\raggedleft\arraybackslash}p{\dimexpr 0.1739\linewidth-2\tabcolsep\relax}@{}}
\toprule\noalign{}
\begin{minipage}[b]{\linewidth}\raggedright
\(s\)
\end{minipage} & \begin{minipage}[b]{\linewidth}\raggedleft
\(M(s,2)\)
\end{minipage} & \begin{minipage}[b]{\linewidth}\raggedleft
\(M(s,4)\)
\end{minipage} & \begin{minipage}[b]{\linewidth}\raggedleft
\(M(s,8)\)
\end{minipage} & \begin{minipage}[b]{\linewidth}\raggedleft
Rényi \(M(s)\)
\end{minipage} & \begin{minipage}[b]{\linewidth}\raggedleft
residual at \(t=8\)
\end{minipage} \\
\midrule\noalign{}
\endhead
\bottomrule\noalign{}
\endlastfoot
\(2\) & \(0.8647\) & \(0.9817\) & \(0.9997\) & \(1.0020\) & \(0.0023\) \\
\(3\) & \(1.4513\) & \(1.7176\) & \(1.8661\) & \(2.0010\) & \(0.1349\) \\
\(4\) & \(2.0386\) & \(2.3632\) & \(2.5473\) & \(2.7441\) & \(0.1967\) \\
\(5\) & \(2.6340\) & \(3.0345\) & \(3.2553\) & \(3.4870\) & \(0.2317\) \\
\end{longtable}}

Two things the table shows. A segment of length \(2\) holds at most one car and has all but reached its
saturated mean by \(t=8\), whereas at \(s=5\) a fifth of a car is still missing: the residual grows with the
length, because a longer segment has more small gaps left to fill and filling them is what takes the
time. And the approach is slow --- algebraic in \(t\), not exponential --- which is the finite-time face of the
jamming asymptotics. Solving the equation and simulating the timed process agree at every \((s,t)\)
tested.

The example is here for what it says about the machinery rather than about the numbers. Nothing in the
derivation used saturation: the split, the conditional independence of the two sides, and the resulting
Volterra equation are properties of the attachment chain, and absorption enters only through the
boundary datum. Rényi's equation is the absorbed case of a construction that does not require it.

\subsubsection{Example 5 --- the third cumulant, and where the cascade runs out of signal}

Corollary \ref{cor:cascade-is-the-class-s-not-the-count-s} gives a linear equation for every moment. Carrying it to \(k=3\) shows both what the cascade
delivers and where numerical resolution, rather than the theory, becomes the binding constraint.

Write \(g_k(s)=\partial_z^k g(s,z)\big|_{z=1}\) for the falling-factorial moments and
\(C(z)=\int_0^{s-1}g(u,z)g(s-1-u,z)\,\mathrm du\). From \(\partial_z^k(zC)=k\,C^{(k-1)}+z\,C^{(k)}\) and
\(C^{(k)}(1)=\sum_j\binom kj\int g_j(u)g_{k-j}(s-1-u)\,\mathrm du\), together with \(g_0\equiv1\):

\[\resizebox{\ifdim\width>\linewidth\linewidth\else\width\fi}{!}{$\displaystyle (s-1)g_3=6\!\int_0^{s-1}\!\!g_2+6\!\int_0^{s-1}\!\!g_1(u)g_1(s{-}1{-}u)+2\!\int_0^{s-1}\!\!g_3
+6\!\int_0^{s-1}\!\!g_1(u)g_2(s{-}1{-}u).$}\]

Linear in \(g_3\), forced by \(g_1\) and \(g_2\), exactly as Corollary \ref{cor:cascade-is-the-class-s-not-the-count-s} predicts. Converting to cumulants,
\(\kappa_3=\mathbb E N^3-3\mathbb E N^2\,\mathbb E N+2(\mathbb E N)^3\).

\textbf{The first two cumulants.} Figure 1 puts \(\kappa_1/s\) and \(\kappa_2/s\) against their limits. The mean
does not approach \(m\) from below at rate \(1/s\) by accident: Rényi's expansion is
\(M(s)=ms-(1-m)+o(1)\), so \(\kappa_1/s=m-(1-m)/s\), and the dotted curve is that line rather than a fit.
The variance rate approaches \(v=0.0382\) from above, reaching \(0.0391\) at \(s=40\).

\begin{figure}
\centering
\pandocbounded{\includegraphics[keepaspectratio,alt={mean and variance against their asymptotics}]{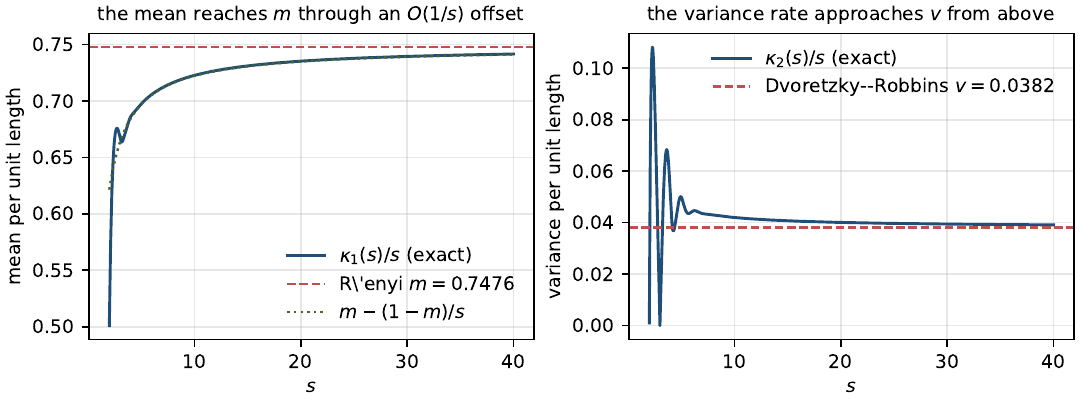}}
\caption{mean and variance against their asymptotics}
\end{figure}

\textbf{The third.} Figure 2 plots \(\kappa_3/\kappa_2^{3/2}\). It changes sign --- \(+0.052\) at \(s=5\),
\(-0.0003\) at \(s=10\) --- and then stays within \(10^{-3}\) of zero. Checked against direct simulation of
\(4\times10^5\) runs, the cascade agrees at \(s=7\) (\(-0.00144\) against \(-0.00172\pm0.00065\)) and sits
about two standard errors away at \(s=5\) (\(0.00651\) against \(0.00746\pm0.00048\)), so the equation is
right and the residual is discretisation.

\begin{figure}
\centering
\pandocbounded{\includegraphics[keepaspectratio,alt={the standardised third cumulant}]{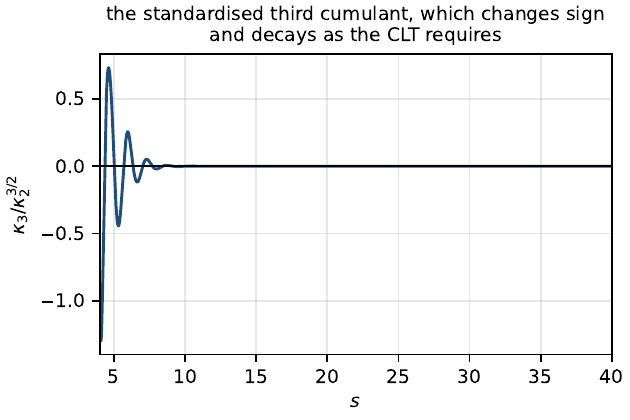}}
\caption{the standardised third cumulant}
\end{figure}

What the figure does \textbf{not} support is a value for the extensive rate. The count's third cumulant is
so small that \(\kappa_3/s\) at \(s=40\) is \(-2\times10^{-5}\), which is at the resolution of a grid solve
and cannot be separated from discretisation error; a rate constant is not claimed here. That the
standardised cumulant decays is the part the data carry, and it is the part the central limit theorem
requires. The contrast with the gap functionals of the asymptotics section, whose third cumulants are
extensive with resolved constants, is itself informative: the count is very nearly symmetric,
and its skewness is a weaker signal than a nonlinear gap score's.

The extensive rate \(C_3\) is not claimed here.

\subsubsection{Example 6 --- the same split one dimension up, and how it generalises}

Rényi's split cuts a segment in two. The two-dimensional analogue puts one uniform point on each side of
a triangle; the three points span an inscribed triangle, and removing it leaves exactly three corner
triangles. The corners are not similar to the parent, but affine maps act transitively on triangles,
preserve ratios of area, and carry a uniform point on a side to a uniform point on a side, so the
construction is affine-equivariant and \textbf{area is the whole state} --- the same collapse that Theorem \ref{thm:rigidity-of-the-attachment-kernel} of
\ref{sec:aggregate-representations} studies for length in one dimension.

Writing \(x,y,z\) for the three side fractions, the corner areas are \(z(1-y)\), \(x(1-z)\) and \(y(1-x)\) times
the parent's, each of mean \(\tfrac14\), and the removed triangle also has mean \(\tfrac14\). Each corner
fraction is a product of two independent uniform variables, so its density is \(-\log t\) on \((0,1)\), and
the count of pieces above a fixed area obeys

\[\resizebox{\ifdim\width>\linewidth\linewidth\else\width\fi}{!}{$\displaystyle M(a)\;=\;1+3\int_0^1(-\log\alpha)\,M(\alpha a)\,\mathrm d\alpha\qquad(a\ge1),\qquad M(a)=0\ \ (a<1).$}\]

\textbf{How it generalises.} In a \(d\)-simplex, place one uniform point on each of the \(\binom{d+1}2\) edges and
take the corner at each vertex \(V_i\) to be the simplex spanned by \(V_i\) and the \(d\) chosen points on the
edges meeting it. Those \(d+1\) corners are simplices in every dimension, the corner at \(V_i\) has volume
fraction \(\prod_{j\ne i}t_{ij}\) --- a product of \(d\) independent uniform variables --- and affine
equivariance again leaves volume as the only state. So the recursion is \((d+1)\)-way in every dimension,
with \(\mathbb E[\alpha^\theta]=(\theta+1)^{-d}\) and growth exponent
\[\resizebox{\ifdim\width>\linewidth\linewidth\else\width\fi}{!}{$\displaystyle M(a)\sim C\,a^{\theta_d},\qquad (d+1)(\theta_d+1)^{-d}=1,\qquad
  \theta_d=(d+1)^{1/d}-1 .$}\]
At \(d=1\) this gives \(\theta_1=1\): the one-dimensional member grows linearly, which is Rényi's scaling
recovered as a special case. At \(d=2\) it gives \(\sqrt3-1\), and \(\theta_d\) decreases to zero.

Two things are worth separating. What survives in every dimension is the \((d+1)\)-way corner recursion.
What does \textbf{not} is the picture of removing a single blocker: the corners leave a residual region whose
volume fraction is positive in every dimension, and only at \(d=2\) is that residual itself a simplex. For
a tetrahedron cut at edge midpoints it is an octahedron. The triangle picture is therefore a
two-dimensional accident, while the recursion behind it is not.

The contrast with the process this paper studies is the point of the example. These simplex splits are
self-similar in Bertoin's sense, since the law of the volume fractions does not depend on the parent's
volume; the parking process is not, because a car has an absolute length, and the departure is exactly
\(1/s\) (\ref{sec:background-aggregate}). The higher-dimensional relatives are the better-behaved
family, and the one-dimensional parking process is the anomaly.

The exponent \(\sqrt3-1=0.7321\) lies near Rényi's \(m=0.7476\) by coincidence and for no structural reason.

\section{Implementation details}

\label{sec:implementation-details}

The results in this paper rest on two kinds of computation: symbolic evaluation of the exact order-statistic and jamming densities, and a machine-checked formalization of the structural claims. This section records the tooling each uses and how far each reaches.

\subsection{Symbolic computation}

The joint density is a rational function on each gap-sign cell, and its marginals and count integrals are multiple polylogarithms. Three tools cover the pipeline. \textbf{SageMath} performs the rational-function and polynomial arithmetic, the Fourier--Motzkin cell enumeration, and the polytope-volume computations. The per-cell polylogarithmic integrations that a general-purpose computer-algebra system cannot close are carried out by a dedicated hyperlogarithm integrator, the sole integration step in the pipeline. Every resulting form is archived as plain text and re-evaluated independently, under the identification of the iterated integral \(\mathrm{Hlog}(z,[a_1,\dots,a_w])\) with the Goncharov function \(G(\{a_1,\dots,a_w\},z)\), together with an independent recursion evaluator; the symbolic forms therefore remain checkable independently of the program that produced them.

The reach of the exact symbolic computation is the all-jammed facet. The saturation probability \(\Pr[N=6]\) on the interval \((6,7)\) is obtained cell by cell, and \(719\) of the \(720\) gap-sign cells reduce to closed polylogarithmic form, the one remaining cell being evaluated numerically; single-gap marginal pieces are computed at \(n=4\). Each exact form is cross-validated against direct Monte-Carlo simulation of the parking process and agrees to within Monte-Carlo error.

\subsection{Machine-checked formalization (Lean 4 / Mathlib)}

A large part of this paper's structural claims is machine-checked in Lean 4 against Mathlib (73 modules, roughly 555 checked declarations). The development is \emph{axiom-honest}: every external theorem is carried as an explicit hypothesis on the results that use it, never as an \texttt{axiom}, and each declaration's axiom footprint is audited to lie within Lean's three foundational axioms (\texttt{propext}, \texttt{Classical.\allowbreak choice}, \texttt{Quot.\allowbreak sound}). Exactly one genuine axiom remains, the Rényi Dyson--Schwinger equation in its real-variable form, isolated in a single module that nothing else builds on. There are no \texttt{sorry}s and no \texttt{native\_\allowbreak decide}.

Everything is derived from the process measure \(M0\) (the sequential-uniform law, an iterated composition of uniform-on-the-free-set kernels) and the free-length affineness \(F1\), except where a named external theorem is quoted as a hypothesis.

\subsubsection{Dependency structure}

The proved results form a directed acyclic graph rooted at the process measure \(M0\) and the
free-length affineness \(F1\); the two conditional clusters (dashed) carry the explicit hypotheses of
the table below.

\begin{figure}
\centering
\pandocbounded{\includegraphics[keepaspectratio,alt={Dependency graph of the machine-checked results. Roots (shaded) are the process measure and the free-length affineness; dashed clusters are conditional on the hypotheses listed below.}]{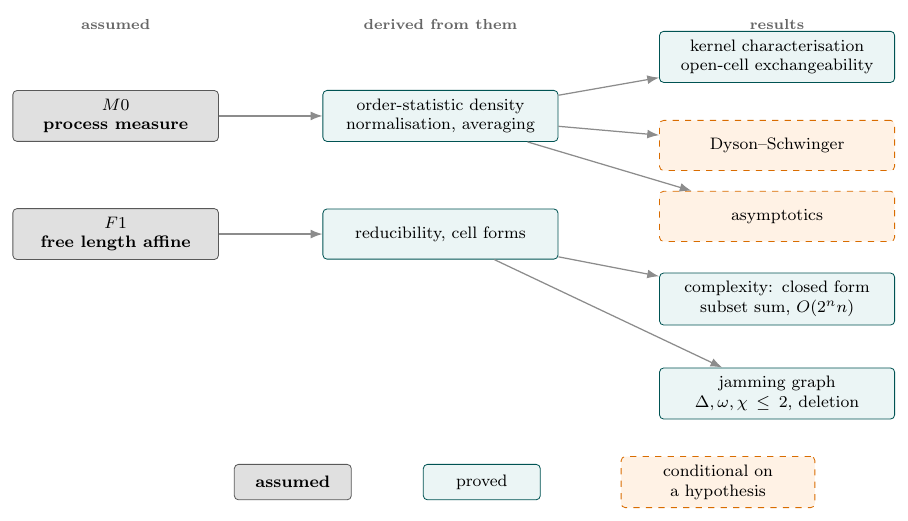}}
\caption{Dependency graph of the machine-checked results. Roots (shaded) are the process measure and the free-length affineness; dashed clusters are conditional on the hypotheses listed below.}
\end{figure}

\subsubsection{Discharged versus assumed, and why}

{\footnotesize\begin{longtable}[]{@{}
  >{\raggedright\arraybackslash}p{\dimexpr 0.2500\linewidth-2\tabcolsep\relax}
  >{\raggedright\arraybackslash}p{\dimexpr 0.2500\linewidth-2\tabcolsep\relax}
  >{\raggedright\arraybackslash}p{\dimexpr 0.2500\linewidth-2\tabcolsep\relax}
  >{\raggedright\arraybackslash}p{\dimexpr 0.2500\linewidth-2\tabcolsep\relax}@{}}
\toprule\noalign{}
\begin{minipage}[b]{\linewidth}\raggedright
result cluster
\end{minipage} & \begin{minipage}[b]{\linewidth}\raggedright
discharged (proved from \(M0\)/\(F1\))
\end{minipage} & \begin{minipage}[b]{\linewidth}\raggedright
assumed (explicit hypothesis)
\end{minipage} & \begin{minipage}[b]{\linewidth}\raggedright
why assumed
\end{minipage} \\
\midrule\noalign{}
\endhead
\bottomrule\noalign{}
\endlastfoot
process measure & \(M0\), \texttt{PM-\allowbreak norm}, total mass one, arrival kernel, first-arrival disintegration & --- & --- \\
order-statistic density & positional density, \(n!\)-symmetrisation, \(F1\), reducibility & --- & --- \\
log-convexity & per-cell and \(n=2\) on the density itself; exact global-failure counterexample & --- & --- \\
complexity & unjammed closed form, jammed subset-sum reduction, counting identity, the Held--Karp subset-recursion identity, refutation of the \(O(n^3)\)/FP claim & --- & --- \\
jamming graph & linear-forest structure, \(\Delta\le2\), \(\omega\le2\), \(\chi\le2\), independence / matching / domination, deletion law & --- & --- \\
weight ladder & the parking arrangement is reducible; the weight algebra & Aomoto--Goncharov period theorem, and a marginal-equals-integral identification & the period theorem is deep analysis absent from Mathlib; the marginal identification is separate work \\
Dyson--Schwinger & the averaging over the first car; the Cauchy-product algebra; coefficient and generating-function forms; base-coordinate non-vacuity & \texttt{CountSplit} (strong-Markov independence of the two sides) and a segment-length measurability hypothesis & \texttt{CountSplit} is the open probabilistic half of the sub-interval factorisation; the measurability needs joint variable--configuration measurability the library lacks \\
asymptotics & tightness arithmetic, extreme-value-class flip, concentration increments, median mean unconditionally & Chen--Stein, Penrose--Yukich, Bolthausen, Saulis--Statulevičius, the Rényi splitting law & no extreme-value, Poisson-approximation, central-limit-rate or Skorokhod-tightness theory in Mathlib \\
gap law & constructed as a pushforward; its mass equals \(\Pr[N=n]\) & --- & --- \\
sole axiom & --- & the real-variable Dyson--Schwinger equation & the nonnegative-valued conditional derivation does not extend to signed arguments; deleting the axiom without a real proof would remove the recorded assumption \\
\end{longtable}}

A hypothesis is only safer than an axiom if it is \emph{satisfiable}: the development includes a regression witness proving that one badly-encoded hypothesis was contradictory, and the Dyson--Schwinger hypothesis is machine-shown non-vacuous at its base coordinate.

\subsubsection{Obstacles}

\begin{enumerate}
\def\labelenumi{\arabic{enumi}.}
\tightlist
\item
  \textbf{Strong-Markov independence of the two sub-segments}: a two-sided-chain construction (a kernel on pairs of complementary histories, an interleaving intertwining, and a run-to-absorption argument). This turns the Dyson--Schwinger equation from an axiom into a theorem and unlocks the moment cascade. The one-step and terminal-event pieces are proved; only the independence remains.
\item
  \textbf{The real-variable Dyson--Schwinger form}: the proved generating-function equation is nonnegative-valued; the signed-argument form needs a parallel Bochner-integral assembly.
\item
  \textbf{Segment-length measurability}: measurability of \(u\mapsto\Pr[N(u)=i]\) needs joint measurability of the free length, density and jammed set in the length argument, none of which is built for that variable (the base case is proved).
\item
  \textbf{The position-to-slack change of coordinates}: a single construction, with its Jacobian, that would discharge the remaining partial results in the order-statistics sections at once.
\item
  \textbf{Theory absent from Mathlib}: no free quasi-symmetric functions, graph spectra, Palm calculus, extreme-value or central-limit-rate machinery, differential Galois theory, a log-convexity API (the last built by hand here), a cost model for the subset recursion, so the \(O(2^n n)\) bound is machine-checked, as is the identity it bounds.
\end{enumerate}

\section{Retrospect and future work}

\label{sec:future-work}

\emph{The exact finite-\((s,n)\) joint order-statistic law is elementary to state, yet it is essentially absent from roughly seventy years of car-parking work. We read that absence historically, and ask what it says about where the subject goes next. The open problems themselves, each with the obstruction blocking it and the step that would remove it, are stated in \ref{sec:summary-of-results}.}

\subsection{Why the finite joint law went unstudied; a short historical reading}

\emph{Interpretation, not established fact; a discussion section. The exact finite-\((s,n)\) joint order-statistic
density is elementary to \textbf{state}, yet it is essentially absent from \textasciitilde70 years of car-parking / RSA work.
The telling feature is the \textbf{silence}: no visible trail of papers attacking it and failing. That is the
signature of a question never posed, not a problem that resisted. Four forces, shallowest to deepest:}

\begin{enumerate}
\def\labelenumi{\arabic{enumi}.}
\tightlist
\item
  \textbf{The founding questions were asymptotic by construction.} Rényi (1958) introduced parking to extract a
  single thermodynamic number; the jamming constant \(\mathcal M_1\approx0.7476\). In that framing finite-\(s\)
  behaviour is a boundary transient to integrate away, not the object of study.
\item
  \textbf{The stat-mech lineage measured other things.} The irreversible-deposition tradition (Evans 1993;
  Krapivsky and collaborators) studies the \textbf{coverage fraction} \(\theta_{\mathrm{jam}}\), its \textbf{kinetics},
  and the \textbf{bulk pair-correlation}; never the finite joint order-statistic law.
\item
  \textbf{The moment program parked at the constant.} Rényi \(\to\) Dvoretzky--Robbins (1964: variance constant
  \(+\) count CLT) \(\to\) Mannion (higher cumulants, asymptotically) carried the moments \emph{precisely to their
  asymptotic constants and stopped}. What was never taken: the exact finite-\(L\) moment functions \(M_k(L)\)
  (piecewise-polylog solutions of the same Volterra equations) and any finite-\(L\) closure of the correlation
  hierarchy.
\item
  \textbf{Toolkit mismatch; the deep cause.} The finite law is a piecewise multiple polylogarithm; the machinery
  that makes it tractable; hyperlogarithm integration and the period / coaction calculus (Goncharov, Brown,
  Panzer); postdates the founding work and sits outside the RSA community's tools. A motivated classical
  attacker would have stalled exactly here.
\end{enumerate}

\emph{The point of the ordering is the argument: the shallow causes (1--3) were contingent and are now fixed; the
deep cause (4) is why the question stayed open; and is precisely the machinery this program imports.}

\subsection{Engineering limitations}

The model is deliberately abstract, and three of its idealisations bear on whether the results transfer to a deployed system.

\textbf{Sensing and dynamics.} The process carries no sensing or actuation error. A physical robot attaches under aleatoric uncertainty in its own pose and in the perceived extent of its neighbours, and the standard treatments of that uncertainty (Thrun, Burgard and Fox 2005; Barfoot 2017) replace an exact hard-core constraint by a probabilistic one. Whether the exact finite-\(s\) laws survive that replacement, or degrade in a controlled way, is not addressed here.

\textbf{Global information.} Each arrival is uniform on the free part of the \emph{whole} segment, so realising the process requires a map of the whole segment. A resource-limited robot obtains that map in one of two ways, and neither is realistic at scale: it traverses the segment to learn its length and its occupied intervals before choosing a position, or it holds the entire configuration in memory. Our earlier work used platforms for which both are out of reach.

\textbf{A locally random rule.} A robot can instead sample uniformly within a bounded window of its current position. Such a \emph{locally random} rule needs neither a full traversal nor global memory, and it recovers the present process as the window grows to the segment length. Its saturation density, and whether its finite-\(s\) order statistics remain exactly computable, are open.

\begin{quote}
\textbf{Goal.} Formulate locally random attachment with the window width as a parameter, and determine whether the splitting law survives it. Because the window interpolates between purely local placement and the present model, the question is how the exact theory degrades as the window shrinks.
\end{quote}

\section{References}

\begin{itemize}
\tightlist
\item
  \textbf{Aomoto, K. (1982).} Addition theorem of Abel type for hyper-logarithms. \emph{Nagoya Math. J.} \textbf{88}, 55--71.
\item
  \textbf{Arnold, B. C., Balakrishnan, N. \& Nagaraja, H. N. (1992).} \emph{A First Course in Order Statistics.} Wiley.
\item
  \textbf{Bapat, R. B. \& Beg, M. I. (1989).} Identities and recurrence relations for order statistics corresponding to nonidentically distributed variables. \emph{Comm. Statist. Theory Methods} \textbf{18}(5), 1993--2004.
\item
  \textbf{Bapat, R. B. \& Beg, M. I. (1989).} Order statistics for nonidentically distributed variables and permanents. \emph{Sankhyā A} \textbf{51}, 79--93.
\item
  \textbf{Barbour, A. D., Holst, L. \& Janson, S. (1992).} \emph{Poisson Approximation.} Oxford University Press.
\item
  \textbf{Barfoot, T. D. (2017).} \emph{State Estimation for Robotics.} Cambridge University Press.
\item
  \textbf{Beirlant, J. \& van Zuijlen, M. C. A. (1985).} The empirical distribution function and strong laws for functions of order statistics of uniform spacings. \emph{J. Multivariate Anal.} \textbf{16}, 300--317.
\item
  \textbf{Bertoin, J. (2006).} \emph{Random Fragmentation and Coagulation Processes.} Cambridge Univ. Press.
\item
  \textbf{Blaisdell, B. E. \& Solomon, H. (1970).} On random sequential packing in the plane and a conjecture of Palàsti. \emph{J. Appl. Probab.} \textbf{7}(3), 667--698.
\item
  \textbf{Bonnier, B., Boyer, D. \& Viot, P. (1994).} Pair correlation function in random sequential adsorption processes. \emph{J. Phys. A: Math. Gen.} \textbf{27}, 3671--3682.
\item
  \textbf{Boucheron, S., Lugosi, G. \& Massart, P. (2013).} \emph{Concentration Inequalities.} Oxford University Press.
\item
  \textbf{Brown, F. (2009).} Multiple zeta values and periods of moduli spaces \(\mathfrak{M}_{0,n}\). \emph{Ann. Sci. Éc. Norm. Supér. (4)} \textbf{42}, 371--489.
\item
  \textbf{Brown, F. (2009).} The massless higher-loop two-point function. \emph{Comm. Math. Phys.} \textbf{287}, 925--958.
\item
  \textbf{Brunner, H. (2004).} \emph{Collocation Methods for Volterra Integral and Related Functional Equations.} Cambridge Univ. Press.
\item
  \textbf{Błaszczyszyn, B., Yogeshwaran, D. \& Yukich, J. E. (2019).} Limit theory for geometric statistics of point processes having fast decay of correlations. \emph{Ann. Probab.} \textbf{47}(2), 835--895.
\item
  \textbf{Chatterjee, S. (2007).} Stein's method for concentration inequalities. \emph{Probab. Theory Related Fields} \textbf{138}, 305--321.
\item
  \textbf{Chen, K.-T. (1977).} Iterated path integrals. \emph{Bull. Amer. Math. Soc.} \textbf{83}, 831--879.
\item
  \textbf{Chiu, S. N., Stoyan, D., Kendall, W. S. \& Mecke, J. (2013).} \emph{Stochastic Geometry and its Applications} (3rd ed.). Wiley.
\item
  \textbf{Coffman, E. G., Flatto, L. \& Jelenković, P. (2000).} Interval packing: the vacant-interval distribution. \emph{Ann. Appl. Probab.} \textbf{10}(1), 240--257.
\item
  \textbf{Coletti, C. F., Gallo, S., Roldán-Correa, A. \& Valencia, L. (2024).} CLT/LIL/concentration for lattice parking occupation density. \emph{J. Stat. Phys.} \textbf{191}, 146.
\item
  \textbf{Cunden, F. D., Cuppone, D., Gramegna, F. \& Vivo, P. (2026).} Maximal minimal spacing for random points. arXiv:2606.04837.
\item
  \textbf{Daley, D. J. \& Vere-Jones, D. (2003).} \emph{An Introduction to the Theory of Point Processes}, Vol. I (2nd ed.). Springer.
\item
  \textbf{Daly, F. \& Wade, A. R. (2025).} Rates of convergence for extremal spacings in Kakutani's random interval-splitting process. arXiv:2508.20749.
\item
  \textbf{David, H. A. \& Nagaraja, H. N. (2003).} \emph{Order Statistics} (3rd ed.). Wiley.
\item
  \textbf{de Haan, L. \& Ferreira, A. (2006).} \emph{Extreme Value Theory: An Introduction.} Springer.
\item
  \textbf{Dubhashi, D. \& Panconesi, A. (2009).} \emph{Concentration of Measure for the Analysis of Randomized Algorithms.} Cambridge University Press.
\item
  \textbf{Dvoretzky, A. \& Robbins, H. (1964).} On the parking problem. \emph{Publ. Math. Inst. Hungar. Acad. Sci.} \textbf{9}, 209--225.
\item
  \textbf{Esary, J. D., Proschan, F. \& Walkup, D. W. (1967).} Association of random variables, with applications. \emph{Ann. Math. Statist.} \textbf{38}, 1466--1474.
\item
  \textbf{Evans, J. W. (1993).} Random and cooperative sequential adsorption. \emph{Rev.~Mod. Phys.} \textbf{65}, 1281--1329.
\item
  \textbf{Foissy, L. (2008).} Faà di Bruno subalgebras of the Hopf algebra of planar trees from combinatorial Dyson--Schwinger equations. \emph{Adv. Math.} \textbf{218}, 136--162.
\item
  \textbf{Gerin, L. (2015).} The Page--Rényi parking process. arXiv:1411.8002.
\item
  \textbf{Golumbic, M. C. (1980).} \emph{Algorithmic Graph Theory and Perfect Graphs.} Academic Press.
\item
  \textbf{Goncharov, A. B. (2001).} Multiple polylogarithms and mixed Tate motives. arXiv:math/0103059.
\item
  \textbf{Goncharov, A. B. (2005).} Galois symmetries of fundamental groupoids and noncommutative geometry. \emph{Duke Math. J.} \textbf{128}, 209--284.
\item
  \textbf{González, J. J., Hemmer, P. C. \& Høye, J. S. (1974).} Cooperative effects in random sequential polymer reactions. \emph{Chem. Phys.} \textbf{3}, 228--238.
\item
  \textbf{Gripenberg, G., Londen, S.-O. \& Staffans, O. (1990).} \emph{Volterra Integral and Functional Equations.} Cambridge Univ. Press.
\item
  \textbf{Haenggi, M. (2012).} \emph{Stochastic Geometry for Wireless Networks.} Cambridge Univ. Press.
\item
  \textbf{Held, M. \& Karp, R. M. (1962).} A dynamic programming approach to sequencing problems. \emph{J. SIAM} \textbf{10}, 196--210.
\item
  \textbf{Iwasa, Y. \& Fukuda, K. (2008).} arXiv:0810.5632.
\item
  \textbf{Joag-Dev, K. \& Proschan, F. (1983).} Negative association of random variables, with applications. \emph{Ann. Statist.} \textbf{11}, 286--295.
\item
  \textbf{Joni, S. A. \& Rota, G.-C. (1979).} Coalgebras and bialgebras in combinatorics. \emph{Stud. Appl. Math.} \textbf{61}, 93--139.
\item
  \textbf{Justicz, J., Scheinerman, E. R. \& Winkler, P. (1990).} Random intervals. \emph{Amer. Math. Monthly} \textbf{97}, 881--889.
\item
  \textbf{Kallenberg, O. (2021).} \emph{Foundations of Modern Probability} (3rd ed.). Springer.
\item
  \textbf{Karlin, S. \& Rinott, Y. (1980).} Classes of orderings of measures and related correlation inequalities. I. Multivariate totally positive distributions. \emph{J. Multivariate Anal.} \textbf{10}, 467--498.
\item
  \textbf{Kontsevich, M. \& Zagier, D. (2001).} Periods. In \emph{Mathematics Unlimited --- 2001 and Beyond}, 771--808. Springer.
\item
  \textbf{Korevaar, J. (2004).} \emph{Tauberian Theory: A Century of Developments.} Springer.
\item
  \textbf{Kornberg, R. D. \& Stryer, L. (1988).} Statistical distributions of nucleosomes: nonrandom locations by a stochastic mechanism. \emph{Nucleic Acids Res.} \textbf{16}, 6677--6690.
\item
  \textbf{Krapivsky, P. L. (2020).} Large deviations in one-dimensional random sequential adsorption. \emph{Phys. Rev.~E} \textbf{102}, 062108.
\item
  \textbf{Krapivsky, P. L., Redner, S. \& Ben-Naim, E. (2010).} \emph{A Kinetic View of Statistical Physics.} Cambridge University Press.
\item
  \textbf{Kumar, G. P. \& Berman, S. (2014).} Statistical analysis of stochastic multi-robot boundary coverage. \emph{Proc. IEEE Int. Conf. Robotics and Automation (ICRA)}, 74--81.
\item
  \textbf{Kumar, G. P. \& Berman, S. (2016).} The probabilistic analysis of the communication network created by dynamic boundary coverage. arXiv:1604.01452.
\item
  \textbf{Kumar, G. P. \& Berman, S. (2017).} Design of stochastic robotic swarms for target performance metrics in boundary coverage tasks. arXiv:1702.02511.
\item
  \textbf{Kumar, G. P. (2016).} \emph{Development and Analysis of Stochastic Boundary Coverage Strategies for Multi-Robot Systems.} PhD thesis, Arizona State University.
\item
  \textbf{Lachièze-Rey, R., Schulte, M. \& Yukich, J. E. (2019).} Normal approximation for stabilizing functionals. \emph{Ann. Appl. Probab.} \textbf{29}(2), 931--993.
\item
  \textbf{Last, G. \& Penrose, M. D. (2017).} \emph{Lectures on the Poisson Process.} Cambridge University Press.
\item
  \textbf{Last, G., Peccati, G. \& Schulte, M. (2016).} Normal approximation on Poisson spaces: Mehler's formula, second order Poincaré inequalities and stabilization. \emph{Probab. Theory Related Fields} \textbf{165}, 667--723.
\item
  \textbf{Leadbetter, M. R., Lindgren, G. \& Rootzén, H. (1983).} \emph{Extremes and Related Properties of Random Sequences and Processes.} Springer.
\item
  \textbf{Loday, J.-L. \& Ronco, M. (1998).} Hopf algebra of the planar binary trees. \emph{Adv. Math.} \textbf{139}, 293--309.
\item
  \textbf{Malvenuto, C. \& Reutenauer, C. (1995).} Duality between quasi-symmetric functions and the Solomon descent algebra. \emph{J. Algebra} \textbf{177}, 967--982.
\item
  \textbf{Mannion, D. (1964).} Random space-filling in one dimension. \emph{Publ. Math. Inst. Hungar. Acad. Sci.} \textbf{9}, 143--153.
\item
  \textbf{Mannion, D. (1976).} Random packing of an interval. \emph{Adv. Appl. Probab.} \textbf{8}(3), 477--501.
\item
  \textbf{Matérn, B. (1986).} \emph{Spatial Variation} (2nd ed.). Springer.
\item
  \textbf{McGhee, J. D. \& von Hippel, P. H. (1974).} Theoretical aspects of DNA--protein interactions: cooperative and non-cooperative binding of large ligands to a one-dimensional homogeneous lattice. \emph{J. Mol. Biol.} \textbf{86}, 469--489.
\item
  \textbf{Meyn, S. \& Tweedie, R. L. (2009).} \emph{Markov Chains and Stochastic Stability} (2nd ed.). Cambridge Univ. Press.
\item
  \textbf{Mourrat, J.-C. (2013).} Rate of convergence in the martingale CLT. \emph{Bernoulli} \textbf{19}(2), 633--645.
\item
  \textbf{Ney, P. E. (1962).} A random interval filling problem. \emph{Ann. Math. Statist.} \textbf{33}, 702--718.
\item
  \textbf{Panzer, E. (2015).} \emph{Feynman Integrals and Hyperlogarithms.} PhD thesis, Humboldt-Universität zu Berlin.
\item
  \textbf{Pavlic, T. P., Wilson, S., Kumar, G. P. \& Berman, S. (2015).} Control of stochastic boundary coverage by multirobot systems. \emph{J. Dyn. Syst. Meas. Control} \textbf{137}, 034504.
\item
  \textbf{Penrose, M. D. \& Yukich, J. E. (2002).} Limit theory for random sequential packing and deposition. \emph{Ann. Appl. Probab.} \textbf{12}(1), 272--301.
\item
  \textbf{Penrose, M. D. (2001).} Random parking, sequential adsorption, and the jamming limit. \emph{Comm. Math. Phys.} \textbf{218}, 153--176.
\item
  \textbf{Penrose, M. D. (2003).} \emph{Random Geometric Graphs.} Oxford University Press.
\item
  \textbf{Pyke, R. (1965).} Spacings. \emph{J. Roy. Statist. Soc. B} \textbf{27}(3), 395--449.
\item
  \textbf{Rényi, A. (1958).} On a one-dimensional problem concerning random space-filling. \emph{Publ. Math. Inst. Hungar. Acad. Sci.} \textbf{3}, 109--127.
\item
  \textbf{Schreiber, T., Penrose, M. D. \& Yukich, J. E. (2007).} Gaussian limits for multidimensional random sequential packing at saturation. \emph{Comm. Math. Phys.} \textbf{272}, 167--183.
\item
  \textbf{Shorack, G. R. \& Wellner, J. A. (1986).} \emph{Empirical Processes with Applications to Statistics.} Wiley.
\item
  \textbf{Stanley, R. P. (1980).} Differentiably finite power series. \emph{European J. Combin.} \textbf{1}, 175--188.
\item
  \textbf{Talbot, J., Tarjus, G., Van Tassel, P. \& Viot, P. (2000).} From car parking to protein adsorption: an overview of sequential adsorption processes. \emph{Colloids and Surfaces A} \textbf{165}, 287--324.
\item
  \textbf{Thrun, S., Burgard, W. \& Fox, D. (2005).} \emph{Probabilistic Robotics.} MIT Press.
\item
  \textbf{Tonks, L. (1936).} The complete equation of state of one, two and three-dimensional gases of hard elastic spheres. \emph{Phys. Rev.} \textbf{50}, 955--963.
\item
  \textbf{Torquato, S. (2002).} \emph{Random Heterogeneous Materials.} Springer.
\item
  \textbf{van Lieshout, M. N. M. \& Baddeley, A. J. (1996).} A nonparametric measure of spatial interaction in point patterns. \emph{Statist. Neerlandica} \textbf{50}, 344--361.
\item
  \textbf{Vaughan, R. J. \& Venables, W. N. (1972).} Permanent expressions for order statistic densities. \emph{J. Roy. Statist. Soc. B} \textbf{34}(2), 308--310.
\item
  \textbf{Widom, B. (1966).} Random sequential addition of hard spheres to a volume. \emph{J. Chem. Phys.} \textbf{44}, 3888--3894.
\item
  \textbf{Zeilberger, D. (1990).} A holonomic systems approach to special functions identities. \emph{J. Comput. Appl. Math.} \textbf{32}, 321--368.
\end{itemize}

\end{document}